\documentclass{article}
\usepackage[margin = 1in]{geometry}
\usepackage{amsmath}
\usepackage{amsfonts,amscd,amssymb,bm,bbm,mathrsfs}
\usepackage{amsthm}
\usepackage{mathtools}
\usepackage{algorithm}
\usepackage[noend]{algorithmic}
\usepackage{comment}

\usepackage{xcolor}
\usepackage{hyperref}

\usepackage{cleveref}
\usepackage{autonum}

\hypersetup{
    colorlinks,
    linkcolor={blue!50!black},
    citecolor={blue!50!black},
}
\colorlet{linkequation}{blue}

\def\N{{\mathbb N}}
\def\Var{{\rm Var}}

\def\proj{{\rm proj}}

\def\P{{\mathbb P}}
\def\Q{{\mathbb Q}}

\newtheorem{theorem}{Theorem}
\newtheorem*{theorem*}{Theorem}
\newtheorem{lemma}{Lemma}
\newtheorem{remark}{Remark}
\newtheorem*{remark*}{Remark}
\newtheorem*{lemma*}{Lemma}
\newtheorem{corollary}{Corollary}
\newtheorem{proposition}[theorem]{Proposition}
\newtheorem{definition}{Definition}

\newtheorem{assumption}{Assumption}
\newtheorem{example}{Example}

\def\argmin{\mathrm{argmin}}
\def\de{{\rm d}}
\def\R{{\mathbb R}}
\def\E{{\mathbb E}}
\def\bzero{{\boldsymbol 0}}
\def\ones{{\boldsymbol 1}}
\def\id{{\mathbf I}}
\def\cN{{\mathcal N}}
\def\sT{{\mathsf T}}

\def\op{{\rm op}}
\newcommand{\nrmp}[1]
{{|\!|\!|{#1}|\!|\!|}}
\newcommand{\nrmps}[1]
{{|\!|\!|{#1}|\!|\!|}}

\def\eps{{\varepsilon}}

\newcommand{\what}{\widehat}
\def\diag{{\rm diag}}

\def\NN{{\rm NN}}

\def\sP{{\mathsf P}}
\def\cP{{\mathcal P}}
\def\cF{{\mathcal F}}

\newcommand{\<}{\langle}
\renewcommand{\>}{\rangle}

\def\KL{\mathsf{KL}}
\def\TV{\mathsf{TV}}

\def\bgamma{{\boldsymbol \gamma}}

\def\bA{{\boldsymbol A}}

\def\bG{{\boldsymbol G}} 
 
\def\bI{{\boldsymbol I}}

\def\bP{{\boldsymbol P}}

\def\bW{{\boldsymbol W}}

\def\bg{{\boldsymbol g}}

\def\bj{{\boldsymbol j}}

\def\bbm{{\boldsymbol m}}

\def\bs{{\boldsymbol s}} 
 
\def\bu{{\boldsymbol u}} 
\def\bv{{\boldsymbol v}} 
\def\bw{{\boldsymbol w}} 
\def\bx{{\boldsymbol x}} 
\def\by{{\boldsymbol y}} 
\def\bz{{\boldsymbol z}} 

\def\btheta{{\boldsymbol \theta}}
\def\bmu{{\boldsymbol \mu}}

\def\cT{{\mathcal T}}
\def\cA{{\mathcal A}}
\def\pa{{\rm pa}}
\def\cC{{\mathcal C}}
\def\cV{{\mathcal V}}

\def\cX{{\mathcal X}}
\def\cY{{\mathcal Y}}
\def\cS{{\mathcal S}}

\def\normalize{{\rm normalize}}
\def\softmax{{\rm softmax}}
\def\ReLU{{\rm ReLU}}
\def\Ind{{\rm Ind}}

\def\qnetwork{{\mathsf q}}
\def\hnetwork{{\mathsf h}}
\def\bnetwork{{\mathsf b}}

\def\xnetwork{{\mathsf x}}

\def\fnetwork{{\mathsf f}}
\def\message{{\nu}}

\def\root{{\rm r}}
\def\nchild{{m}}
\def\dim{{d}}

\def\MP{{\rm MP}}

\def\tr{{\rm trace}}

\def\truemu{{\mu_\star}}

\newcommand{\indicator}{\mathbbm{1}}

\newcommand{\real}{\ensuremath{\mathbb{R}}}

\newcommand{\bigO}{\mathcal{O}}

\newcommand{\indep}{\perp\!\!\!\perp}

\newcommand{\mmax}{\overline{m}}

\newcommand{\defn}{\coloneqq}

\newcommand{\jth}{{(j)}}

\newcommand{\Term}{{T}}

\newcommand{\FunSpace}{{\mathcal{S}}}
\newcommand{\polyshort}{{C}}
\newcommand{\polyshortprime}{{C'}}

\newcommand{\myunder}[1]{\noindent{\underline{#1}}}

\newcommand{\lops}[1]{\|{#1}\|_{\mathrm{op}}}

\newcommand{\feature}{\mathrm{f}}
\newcommand{\positional}{\mathrm{p}}
\newcommand{\hmatrix}{{\mathsf H}}
\newcommand{\gmatrix}{{\mathsf G}}
\newcommand{\gnetwork}{{\mathsf g}}
\newcommand{\qmatrix}{{\mathsf Q}}

\newcommand{\pomatrix}{{\boldsymbol P}}
\newcommand{\boldzero}{{\boldsymbol 0}}
\newcommand{\mask}{{\boldsymbol M}}
\newcommand{\pnetwork}{{\mathsf p}}
\newcommand{\bbmatrix}{{\mathsf B}}

\newcommand{\bdelta}{{\boldsymbol \delta}}
\newcommand{\bWauxiliary}{\overline{\bW}}

\newcommand{\perturb}{{h}}
\newcommand{\diffperturb}{\Delta^{\perturb}}

\newcommand{\softmaxscore}{{p}}
\newcommand{\softmaxscoreinf}[1]{\what\P_{#1}}
\newcommand{\clipfntprob}[2]{\what\P^{(#1)}_{#2}}

\newcommand{\scorefun}{{\sf S}}
\newcommand{\revdef}{\eqqcolon}
\newcommand{\scorediffnorm}{r_0}
\newcommand{\boundscoreconst}{c_1}

\newcommand{\orlicz}[2]{\|#1\|_{\psi_{#2}}}
\newcommand{\orliczbig}[2]{\Big\|#1\Big\|_{\psi_{#2}}}

\newcommand{\Varterm}{V}
\newcommand{\scorejoint}{\P^{(K,r)}_{\txt|\image}}
\newcommand{\scorejointloo}{\P^{(K,r,-1)}_{\txt|\image}}
\newcommand{\scorejointinf}{\P^{(r)}_{\txt|\image}}
\newcommand{\tscorejoint}{\Term^{(K,r)}_{\txt|\image}}
\newcommand{\tscorejointloo}{\Term^{(K,r,-1)}_{\txt|\image}}
\newcommand{\tscorejointinf}{\Term^{(r)}_{\txt|\image}}

\newcommand{\shortscoreratio}{s}
\newcommand{\shortempprob}{\mathcal{R}}
\newcommand{\residual}{\mathcal{R}}

\def\adaweighta{{W^{(1)}_{{\sf ada}}}}
\def\adaweightb{{W^{(2)}_{{\sf ada}}}}

\newcommand{\querymatrix}{W_Q}
\newcommand{\keymatrix}{W_K}
\newcommand{\valuematrix}{W_V}
\newcommand{\numfflayer}{J}

\newcommand{\state}{S}
\newcommand{\staten}{S}
\newcommand{\readbd}{B_{\read}}
\newcommand{\Sterm}{\widetilde{T}}

\newcommand{\boundadapt}{B_{\mathrm{Adap}}}

\newcommand{\diminadapt}{{M}}

\newcommand{\EstPar}{{\widehat\btheta}}

\newcommand{\mmaxone}{\underline{m}}
\newcommand{\mmaxb}{\overline{m}}

\newcommand{\Parspaceclip}{\Theta_{L,J,D,D',B}}
\newcommand{\Parspacecdm}{\Theta_{L,J,D,D',B,M}}
\newcommand{\Parspacecdmone}{\Theta^{\cdm}_{L,J,D,D',B}}
\newcommand{\Parspacevlm}{\Theta_{L,J,D,D',B,M}}
\newcommand{\Parspacevlmone}{\Theta^{\vlm}_{L,J,D,D',B}}

\newcommand{\lone}[1]{\|#1\|_1}

\newcommand{\bigOtil}{\widetilde{\mathcal{ O}}}
\newcommand{\truemucdm}{\mu_{\star,t}}
\newcommand{\truebbmcdm}{\bbm_{\star,t}}
\newcommand{\simiid}{\sim_{iid}}

\newcommand{\readbdvlmenc}{B^\vlm_\read}

\newcommand{\readbdcdmenc}{B^\txt_\read}
\newcommand{\readsimscore}{{\sf trun}}
\newcommand{\readtxt}{{\sf trun}_{\txt}}
\newcommand{\readimage}{{\sf trun}_{\image}}

\newcommand{\readclip}{{\sf read}_{\clip}}
\newcommand{\readvlm}{{\sf read}_{\vlm}}
\newcommand{\encodeclip}{{\sf Emb}_{\clip}}
\newcommand{\mmaintextone}{\underline{m}}
\newcommand{\mmaintextmax}{\overline{m}}
\newcommand{\numbatch}{{n}}

\newcommand{\ATF}{{\tt Standard\,\,TF}}
\newcommand{\GTF}{{\tt Guided\,\,TF}}
\newcommand{\STF}{{\tt Shallow\,\,TF}}
\newcommand{\Bayes}{{\tt Bayes}}
\newcommand{\MBP}{{\tt Mis-spec.\,\,BP}}
\newcommand{\JT}{{\tt Joint\,\,Training}}
\newcommand{\Hmessage}{\mathcal{H}}
\newcommand{\Qmessage}{\mathcal{Q}}
\newcommand{\Bmessage}{\mathcal{B}}

\usepackage{enumitem}
\usepackage{mathabx}
\usepackage{cleveref}
\usepackage{subcaption}
\usepackage{booktabs}
\usepackage{tocbibind}
\addtocontents{toc}{\protect\setcounter{tocdepth}{-1}}

\def\showinterpretremark{1}
\def\showZSCprooftech{0}

\def\showappendix{1}
\def\showacknowledgement{0}
\def\showprelim{1}
\def\showfigure{1}

\def\image{{\rm im}}
\def\txt{{\rm tx}}
\def\root{{\rm r}}
\def\xi{{\bx_{\image}}}
\def\xt{{\bx_{\txt}}}

\def\depth{{J}}
\def\width{{j}}
\def\widthvector{{\bj}}

\def\NNs{{\tau}}

\def\xib{{\overline{\bx}_{\image}}}
\def\xtb{{\overline{\bx}_{\txt}}}
\def\xiu{{\underline{\bx}_{\image}}}
\def\xtu{{\underline{\bx}_{\txt}}}
\def\cXi{{\cX_{\image}}}
\def\cXt{{\cX_{\txt}}}

\def\kb{{\overline{k}}}
\def\ku{{\underline{k}}}

\def\clip{{\sf clip}}

\def\cdm{{\sf cdm}}
\def\vlm{{\sf vlm}}

\def\Ei{{\sf E_{\image}}}
\def\Et{{\sf E_{\txt}}}
\def\trueEi{{\sf E_{\image, \star}}}
\def\trueEt{{\sf E_{\txt, \star}}}
\def\KL{{\mathsf D}_{\rm KL}}
\def\TV{{\mathsf D}_{\rm TV}}
\def\Renyi{{\mathsf D}_{\rm \alpha}}
\def\Renyitwo{{\mathsf D}_{\rm 2}}
\def\scorelink{{\Upsilon}}
\def\truescorelink{\Upsilon_\star}
\def\dimclip{{p}}
\def\truedimclip{{p_\star}}
\def\Mt{{\sf M}_t}
\def\Mthat{\widehat{\mathsf{M}}_t}

\def\Ematrix{{\boldsymbol E}}
\def\trueEmatrix{{\boldsymbol E}}

\def\Binvbd{L_B}
\def\tslinvLip{L_\Gamma}
\def\Adapter{{\rm Adap}}

\def\risk{{\sf R}}
\def\orisk{\overline{\sf R}}
\def\simscore{{\sf S}}
\def\truesimscore{{\sf S_\star}}
\def\mutinfor{{\rm MI}}
\def\const{{\rm const}}
\def\suffmeasure{{\rm Suff}}
\def\cls{{\rm cls}}
\def\Mspace{{\mathcal M}}

\def\sQ{{\sf Q}}
\def\sE{{\sf E}}
\def\sep{{\rm sp}}
\def\col{{\rm col}}
\def\row{{\rm row}}
\def\linkweight{{w}}
\def\encode{{\sf Emb}}
\def\read{{\sf read}}

\def\excess{{\sf Excess}}
\def\transbd{B_{\psi}}
\def\cS{{\mathcal S}}
\def\Perm{{\boldsymbol \Pi}}

\def\pflip{{p_{\rm flip}}}
\def\tO{{\tilde O}}

\newcommand{\boundimage}{B_{x_\image}}

\newcommand{\tokenmatrix}{\mathsf{H}}
\newcommand{\outtokenmatrix}{\widetilde{\mathsf{H}}}
\newcommand{\outtokenmatrixb}{\widebar{\mathsf{H}}}
\newcommand{\tokenmatrixb}{\widebar{\mathsf{H}}}

\newcommand{\radius}{{\mathsf{R}}}
\newcommand{\trunspace}{{\mathcal{H}}}
\newcommand{\feedforward}{{\mathrm{FF}}}
\newcommand{\Par}{{\boldsymbol{\theta}}}
\newcommand{\ParSpace}{{{\Theta}}}
\newcommand{\ffnweight}{W}

\newcommand{\ff}{\mathsf{ff}}

\newcommand{\bound}{B}

\newcommand{\attention}{{\mathrm{Attn}}}
\newcommand{\FF}{\feedforward}
\newcommand{\AT}{\attention}
\newcommand{\attn}{\attention}
\newcommand{\Attn}{\attention}
\newcommand{\MAttn}{{\mathrm{MAttn}}}
\newcommand{\horizon}{{N}}
\newcommand{\softmaxshort}{{\sigma}}
\newcommand{\termattn}{{U}}
\newcommand{\TF}{\mathrm{TF}}
\newcommand{\tf}{\mathsf{tf}}
\newcommand{\normalizeshort}{\mathrm{norm}}
\newcommand{\layer}{L}

\newcommand{\statesize}{{S}}

\newcommand{\abwimage}{{\boldsymbol{Y}}}
\newcommand{\abwimagedist}[1]{\widebar{\P}_{\image|\txt}^{(#1)}}
\newcommand{\bwimage}{\widetilde{\boldsymbol{Y}}}
\newcommand{\bwimagedist}[1]{\widetilde{\P}_{\image|\txt}^{(#1)}}

\newcommand{\bnoise}{\boldsymbol{W}}

\newcommand{\gnoise}{{\boldsymbol{g}}}
\newcommand{\IdMat}{\id}
\newcommand{\truepostmean}{\boldsymbol{m}}
\newcommand{\dimimage}{\dim_\image}
\newcommand{\dimtxt}{\dim_\txt}
\newcommand{\probngauss}[1]{\P_{\image|\txt}^{\square{\tiny#1}}}

\newcommand{\trueprobnext}{{\mu_\star}}
\newcommand{\estprobnext}{{\widehat\mu}}
\newcommand{\probnext}{{\mu}}
\newcommand{\probnextspace}{\mathcal{U}}
\newcommand{\xtsub}[1]{x_{\txt,#1}}
\newcommand{\cXtsub}[1]{\cX_{\txt,#1}}

\newcommand{\distance}{{\mathsf{D}}}

\newcommand{\embeddim}{D}
\newcommand{\hiddendim}{D'}
\newcommand{\linf}[1]{\|#1\|_\infty}

\usepackage{lipsum}

\newcommand\blfootnote[1]{%
  \begingroup
  \renewcommand\thefootnote{}\footnote{#1}%
  \addtocounter{footnote}{-1}%
  \endgroup
}

\iftrue
\makeatletter
\renewcommand{\paragraph}{%
  \@startsection{paragraph}{4}%
  {\z@}{0.8ex \@plus 1ex \@minus .2ex}{-1em}%
  {\normalfont\normalsize\bfseries}%
}
\makeatother
\fi

\def\spacingset#1{\renewcommand{\baselinestretch}%
{#1}\small\normalsize} \spacingset{1}

\title{A Statistical Theory of Contrastive Pre-training and \\ Multimodal Generative AI}

\date{}

\begin{document}

\author{Kazusato Oko\thanks{These authors contributed equally to this work; more junior authors listed first. } \thanks{Department of EECS, UC Berkeley. Email: \texttt{oko@berkeley.edu}.} \and Licong Lin\footnotemark[1] \thanks{Department of Statistics, UC Berkeley. Email: \texttt{liconglin@berkeley.edu}.} \and Yuhang Cai\footnotemark[1]  \thanks{Department of Mathematics, UC Berkeley. Email: \texttt{willcai@berkeley.edu}.} \and Song Mei\footnotemark[1] \thanks{Department of Statistics and Department of EECS, UC Berkeley. Email: \texttt{songmei@berkeley.edu}. Corresponding author.} }

\maketitle

\begin{abstract}
    Multi-modal generative AI systems, such as those combining vision and language, rely on \textit{contrastive pre-training} to learn representations across different modalities. While their practical benefits are widely acknowledged, a rigorous theoretical understanding of the contrastive pre-training framework remains limited. This paper develops a theoretical framework to explain the success of contrastive pre-training in downstream tasks, such as zero-shot classification, conditional diffusion models, and vision-language models. We introduce the concept of \textit{approximate sufficient statistics}, a generalization of the classical sufficient statistics, and show that near-minimizers of the contrastive pre-training loss are approximately sufficient, making them adaptable to diverse downstream tasks. We further propose the \textit{Joint Generative Hierarchical Model} for the joint distribution of images and text, showing that transformers can efficiently approximate relevant functions within this model via belief propagation. Building on this framework, we derive sample complexity guarantees for multi-modal learning based on contrastive pre-trained representations. Numerical simulations validate these theoretical findings, demonstrating the strong generalization performance of contrastively pre-trained transformers in various multi-modal tasks. \blfootnote{Code for our experiments is available at~\url{https://github.com/willcai7/Multimodal-GHM}. }
\end{abstract}

\section{Introduction}
Multi-modal generative AI systems, such as DALL-E \cite{dalle2} for generating images from text prompts and GPT-4V \cite{gpt4v} for generating text based on both image and text inputs, have achieved remarkable empirical success. The training process for such systems often begins with contrastive pre-training \cite{radford2021learning, jia2021scaling}, which learns lower-dimensional neural network representations for each modality using large-scale pretraining datasets. Subsequently, the contrastively pre-trained representations of one modality are fixed and used to guide the training of a generative model for the other modality.

To elaborate, we focus on multi-modal learning in the image-text domain\footnote{We use ``image-text domain'' as convenient terminology, but the theory applies universally and is not restricted to this setting. Our analysis focuses on two domains for simplicity but extends to multi-domain scenarios. Notably, in the machine learning literature, ``multi-modal learning'' does not necessarily require three or more domains.}, 
where the contrastive pre-training process is known as Contrastive Language-Image Pretraining (CLIP) \cite{radford2021learning}. Given a dataset of paired image-text samples $(\xi, \xt) \in \cXi \times \cXt$, CLIP trains a pair of neural network encoders, $(\Ei: \cXi \to \R^\dimclip, \Et: \cXt \to \R^\dimclip)$, by aligning paired image-texts while simultaneously pushing apart non-paired ones. This alignment is achieved by minimizing the contrastive loss defined in Eq.~(\ref{eqn:CLIP_risk}). The pre-trained CLIP encoders have shown exceptional performance in various downstream tasks, including:
\begin{itemize}[topsep=5pt]
\item \textit{Zero-shot classification} \cite{radford2021learning, jia2021scaling}. The goal is to predict the label $y \in \cY$ for a new image $\xi \in \cXi$. Using the pre-trained encoders $(\Ei, \Et)$, a good classifier can be constructed without the need for fine-tuning on task-specific data. 
\item
\textit{Conditional diffusion models} \cite{dalle2, esser2024scaling}: The task is to generate an image $\xi \in \cXi$ from a text prompt $\xt \in \cXt$. In these models, the text embedding $\Et(\xt)$ is used in the conditional denoising function, without directly referencing the original text prompt during training.
\item 
\textit{Vision-language models} \cite{li2022blip, liu2024improved}. The task is to generate text $\xt \in \cXt$ from an image prompt $\xi \in \cXi$. In such models, the image embedding $\Ei(\xi)$ is used in the auto-regressive transformer, without directly referencing the original image prompt during training.
\end{itemize}

The empirical success of multimodal learning underscores the need for a theoretical framework to better understand this paradigm, ideally within the context of statistical learning theory. To achieve this, two key theoretical questions need to be addressed: 
\begin{itemize}[topsep=5pt, leftmargin=17pt]
\item[1)] \textit{Why are CLIP encoders effective representations for downstream tasks?} The statistical properties of contrastive loss minimizers have been extensively studied in the literature \cite{saunshi2019theoretical, tosh2021contrastive1, tosh2021contrastive2, haochen2021provable}. Existing works often leverage the structure of the contrastive loss and its connection to downstream tasks to show that linear functions of learned representations perform well in these settings. However, such analyses fall short in explaining tasks like zero-shot classification, where no fine-tuning is required, as well as tasks involving conditional diffusion models and vision-language models, where linear functions of learned representations are insufficient to capture relevant functions. 
\item[2)]
\textit{Why do the encoders and downstream functions admit efficient neural network approximations? } This question has received relatively less attention. While neural networks are universal function approximators \cite{barron1993universal}, they can suffer from the curse of dimensionality \cite{bach2017breaking} when dealing with general high-dimensional target functions. The primary theoretical challenge lies in constructing a tractable yet realistic statistical model for the joint image-text distribution. A Gaussian assumption, though mathematically convenient, is often overly restrictive and unrealistic, whereas a fully non-parametric approach could lead to the curse of dimensionality. 
\end{itemize}

This paper addresses the two theoretical questions outlined above. In \Cref{sec:CLIP_minimizer}, we reveal a surprisingly simple property of the near-minimizers of the CLIP loss: they are pairs of \textbf{approximate sufficient statistics}, a generalization of the classical concept of sufficient statistics. Due to their approximate sufficiency and the straightforward implications of data processing inequalities, these representations can adapt to a variety of downstream tasks, including zero-shot classification, conditional diffusion models, and vision-language models. Furthermore, when a simple ``canonical representation” of the data exists, we show that it can be recovered from any near-minimizer of the CLIP loss through a simple two-layer network. This enables CLIP representations to effectively adapt to downstream tasks where the canonical representations serve as sufficient statistics. 

In \Cref{sec:GHMs}, we apply our general framework to a statistical model for the joint distribution of images and text, which we call the \textbf{Joint Generative Hierarchical Model} (JGHM). The JGHM is a graphical model consisting of two trees with a shared root, where the root node captures high-level features, and the leaf nodes represent observed images or text. We demonstrate that transformers \cite{vaswani2017attention} can efficiently approximate the relevant functions within JGHMs by approximating the belief propagation algorithm, thus \textbf{breaking the curse of dimensionality}. Building on this insight, we derive end-to-end sample complexity results for tasks such as zero-shot classification, conditional diffusion models, and vision-language models, all utilizing the pre-trained CLIP representations. 

Numerical simulations are presented in \Cref{sec:simulations} within the simulated JGHM framework. The experimental results demonstrate that transformers trained using the Adam algorithm~\cite{kingma2014adam} can achieve near-optimal minimizers, exhibiting strong generalization performance. Additionally, out-of-distribution tests show that the  minimizers obtained by Adam closely emulate the behavior of belief propagation, a result of independent interest. 


\section{Related literature}\label{sec:rl_main}

\paragraph{Contrastive learning and multi-modal learning.} CLIP \cite{radford2021learning} and ALIGN \cite{jia2021scaling} leverage large-scale contrastive pretraining to extract visual and textual embeddings, relying on loss functions such as NCE \cite{gutmann2010noise}, InfoNCE \cite{oord2018representation}, and Multi-class N-pair loss \cite{sohn2016improved} to distinguish paired from non-paired samples. Conditional Diffusion Models, exemplified by DALL-E \cite{dalle2} and Stable Diffusion \cite{esser2024scaling}, generate realistic images from text prompts, while Vision-Language Models like Flamingo \cite{alayrac2022flamingo}, BLIP \cite{li2022blip}, and Llava \cite{liu2024visual, liu2024improved} interpret and describe images based on textual inputs. These frameworks highlight the versatility of contrastive learning in advancing multimodal understanding and generation.

\paragraph{Theories of Contrastive Learning and CLIP.} Numerous studies have shown that InfoNCE loss (derived from the InfoMax principle \cite{linsker1988self}) maximizes a lower bound on mutual information between positive sample pairs \cite{oord2018representation, poole2019variational, hjelm2018learning, bachman2019learning, tian2020contrastive, zimmermann2021contrastive, lu2024f}, which aligns with \Cref{lem:global_min_contrastive_loss} and \Cref{prop:approximate_global_min_contrastive_loss}. Theoretical analysis of the adaptation properties of contrastive learning has been investigated in a series of work \cite{wang2020understanding,saunshi2019theoretical,tosh2021contrastive1,tosh2021contrastive2,haochen2021provable}. Our work diverges from these existing theories of contrastive learning in three key ways: (1) While many studies provide “absolute risk bounds” for downstream tasks under structural conditions, our work offers “excess risk bounds,” which require more refined statistical analysis; (2) We analyze the multimodal learning, including zero-shot prediction task, conditional diffusion models, and vision-language models, which have not been addressed in these work; and (3) We proposed a data distribution for image and text pairs and provided end-to-end statistical efficiency guarantees for multimodal learning through neural networks. 

Closest to our approach are \cite{uesaka2024understanding} and \cite{chen2023understanding}. The former uses point-wise mutual information to bound excess risk in downstream classification, while the latter examines CLIP's minimizer under completeness conditions, demonstrating its strong zero-shot classification capabilities. In contrast, our work (1) adopts a sufficient statistics framework to interpret CLIP, (2) uncovers additional properties of CLIP representations, and (3) provides a unified theory for multimodal learning, including vision-language models and conditional diffusion frameworks.

\paragraph{Approximate sufficient statistics.} The concept of approximate sufficient statistics was mentioned in \cite{chen2020neural}, which proposed an approach to find them. However, this work did not provide a formal definition of approximate sufficient statistics or explore its theoretical properties. The relationship between contrastive loss minimizers and sufficient statistics was examined in \cite{xu2024dependence}, but the notion of approximate sufficient statistics was not considered. After an extensive review of the literature, we conclude that the definition of approximate sufficient statistics and its connection to the approximate minimizer of CLIP loss, to the best of the authors’ knowledge, is novel.  

Further related works are summarized in \Cref{sec:related-literature}.

\section{Statistical properties of contrastive pre-training}\label{sec:CLIP_minimizer}

In this section, we demonstrate that CLIP provides effective representations that can adapt to downstream tasks. In \Cref{sec:CLIP_sufficient_stats}, we show that any near-minimizer of the CLIP risk yields a pair of near-sufficient statistics. In \Cref{sec:CLIP_ZSC_VLM_CDM}, we demonstrate that this near-sufficiency facilitates the adaptability of CLIP representations to various downstream tasks. Furthermore, in \Cref{sec:near_optimal_near_equivalence}, we show that if the joint distribution allows for a canonical representation with certain well-posedness properties, a simple adapter (a small network) enables efficient neural network approximations for downstream tasks where the canonical representations serve as sufficient statistics.

\subsection{Near-sufficiency of CLIP minimizers}\label{sec:CLIP_sufficient_stats}

To simplify the discussion and avoid measure-theoretic complications, we assume that both $\cXi$ and $\cXt$ are discrete spaces. Let the image-text pair $(\xi, \xt) \in \cXi \times \cXt$ follow a joint distribution $\P_{\image, \txt} \in \cP(\cXi \times \cXt)$. We denote the marginal distributions of $\xi$ and $\xt$ as $\P_{\image}$ and $\P_{\txt}$, respectively, and the conditional distributions of $\xi$ given $\xt$ and $\xt$ given $\xi$ as $\P_{\image | \txt}$ and $\P_{\txt | \image}$, respectively. For clarity, we will omit subscripts in probability expressions when the context is clear.

{
In the CLIP framework, paired image-texts are used as positive samples, while unpaired image-texts are used as negative samples. Specifically, within each batch we  have 
 $K$ i.i.d. samples $(\xi_{,i}, \xt_{,i})_{i=1}^K$ from $\P_{\image, \txt}$, and we use $(\xi_{,i}, \xt_{,i})_{i=1}^K$ as the paired 
 image-text samples and  $(\xi_{,i}, \xt_{,j})_{i\neq j;i, j=1}^K$ as the non-paired samples.  
}


Let $\Ei: \cX_i \to \R^\dimclip$ and $\Et: \cXt \to \R^\dimclip$ represent the image and text encoders, respectively, both parameterized by neural networks. Using a user-defined similarity link function $\scorelink: \R^\dimclip \times \R^\dimclip \to \R$, the similarity score for an image-text couple $(\xi, \xt)$ is given by $\simscore_{\Ei, \Et}(\xi, \xt) \defn\scorelink(\Ei(\xi), \Et(\xt))$. The CLIP risk function is the expected InfoNCE loss over paired and non-paired samples:
\begin{align}
&~\orisk_{\clip,K}(\simscore) \defn  \E \Big[ - \log \frac{\exp(\simscore(\xi_{,1}, \xt_{,1})  ) }{\sum_{j \in [K]} \exp(\simscore(\xi_{,1}, \xt_{, j}))} \Big] 
+ 
\E \Big[ - \log \frac{\exp(\simscore(\xi_{,1}, \xt_{,1})  ) }{\sum_{j \in [K]} \exp(\simscore(\xi_{, j}, \xt_{,1}))} \Big], \\
&~\risk_{\clip,K}(\Ei, \Et) \defn \orisk_{\clip,K}(\simscore_{\Ei, \Et}), \label{eqn:CLIP_risk}
\end{align}
where the expectation is over $(\xi_{,i}, \xt_{,i})_{i=1}^K\simiid \P_{\image, \txt}$. This risk comprises the cross-entropy losses for classifying paired and non-paired samples, based on $\softmax((\xi_{,1}, \xt_{,j})_{j \in [K]})$ and $\softmax((\xi_{,j}, \xt_{,1})_{j \in [K]})$, respectively. The function $\orisk_{\clip, K}$ is defined over all possible similarity scores $\simscore: \cXi \times \cXt \to \R$, while $\risk_{\clip, K}$ is defined over all couples of encoders $(\Ei: \cXi \to \R^{\dimclip}, \Et: \cXt \to \R^{\dimclip})$. 

\paragraph{Global minimizers of CLIP as sufficient statistics.} The InfoNCE loss, first introduced by \cite{oord2018representation}, underpins the CLIP framework and leads to the following characterization of its global minimizers. For completeness, the proof is provided in \Cref{app:proof_global_min_contrastive_loss}. 
\begin{lemma}[Global CLIP minimizer \cite{oord2018representation}]\label{lem:global_min_contrastive_loss}
Consider minimizing $\orisk_{\clip,K}$ over all possible similarity scores $\simscore: \cXi \times \cXt \to \R$. For all $K\geq3$, the set of global minimizers of $\orisk_{\clip,K}$, denoted by $\mathcal{M}_\mathcal{S}$, is given by 
\begin{equation}\label{eqn:global_min_clip_risk}
\mathcal{M}_\mathcal{S}= \Big\{\truesimscore:\truesimscore(\xi,\xt)=\log \Big[ \frac{\P_{\image, \txt}( \xi, \xt) }{ \P_{\image}(\xi) \cdot \P_{\txt}(\xt)} \Big]+\const, \text{  for some } \const\in\R \Big\}.
\end{equation}
Moreover, in the limit as $K \to \infty$, the minimum CLIP risk yields the negative mutual information of $(\xi, \xt)$ under the joint distribution $\P_{\image, \txt}$: 
\[
\lim_{K \to \infty} \big[- \frac{1}{2} 
 \inf_{\simscore} \orisk_{\clip, K}(\simscore) + \log K\big] =  \mutinfor(\xi, \xt) \defn\E_{\P_{\image, \txt}}\Big[\log \big[ \P( \xi, \xt) / [ \P(\xi) \cdot \P(\xt)] \big]\Big]. 
\]
\end{lemma}

As a corollary, using the Fisher-Neyman factorization theorem \cite{fisher1922mathematical, neyman1936teorema}, any pair of encoders that achieve the minimum CLIP risk serves as sufficient statistics.

\begin{corollary}[CLIP minimizers as sufficient statistics]\label{cor:sufficient_stats_FNF}
Suppose there exists a pair of encoders $(\trueEi, \trueEt)$ such that $\risk_{\clip, K}(\trueEi, \trueEt) = \inf_{\simscore} \orisk_{\clip,K}(\simscore)$. Then, $\trueEi(\xi)$ and $\trueEt(\xt)$ are sufficient statistics for the statistical models $\P_{\image | \txt}(\xi | \xt)$ and $\P_{\txt | \image}(\xi | \xt)$, respectively. Specifically, the mutual information satisfies:
\[
\mutinfor(\xi, \xt) = \mutinfor(\trueEi(\xi), \xt) = \mutinfor(\xi, \trueEt(\xt)). 
\]
\end{corollary}

\begin{proof}[Proof of \Cref{cor:sufficient_stats_FNF}] By \Cref{lem:global_min_contrastive_loss} and the condition that $\risk_{\clip, K}(\trueEi, \trueEt) = \inf_{\simscore} \orisk_{\clip,K}(\simscore)$,  the conditional distribution can be expressed as:
\[
\P_{\image| \txt}(\xi | \xt) = \exp\{- \const\} \cdot  \P_{\image}(\xi) \cdot \exp\{ \scorelink(\trueEi(\xi), \trueEt(\xt)) \}. 
\]
By the Fisher-Neyman factorization theorem (see e.g., Theorem~3.6~in~\cite{keener2010theoretical}), $\trueEi(\xi)$ is a sufficient statistic for the model $\P_{\image | \txt}(\xi | \xt)$. Similarly, by symmetry, $\trueEt(\xt)$ is a sufficient statistic for the model $\P_{\txt | \image}(\xt | \xi)$. 
\end{proof}

\Cref{lem:global_min_contrastive_loss} has appeared in various forms across the literature \cite{oord2018representation, poole2019variational, hjelm2018learning, bachman2019learning, tian2020contrastive, zimmermann2021contrastive}. Similarly, the interpretation of CLIP minimizers as sufficient statistics in \Cref{cor:sufficient_stats_FNF} aligns with the InfoMax principle introduced in earlier works \cite{linsker1988self, chen2020neural}. While we do not claim originality for either result, to the best of our knowledge, the use of the Fisher-Neyman factorization theorem to establish \Cref{cor:sufficient_stats_FNF} is novel and may be of particular interest to the statistics community. 

\Cref{cor:sufficient_stats_FNF} implies that $\trueEi(\xi)$ captures all the information necessary to predict $\xt$, making it an effective representation of the image (similar argument applies to $\trueEt(\xt)$). It's worth noting that there may be infinitely many minimizers of $\risk_{\clip, K}$, and \Cref{cor:sufficient_stats_FNF} holds for all of them. In practice, finding the exact minimizer of the CLIP risk is not feasible; however, an approximate version of the results still holds, which we discuss next.


\paragraph{Near-minimizers of CLIP as near-sufficient statistics.} We now demonstrate that approximate minimizers of the CLIP risk serve as approximate sufficient statistics. To this end, we extend the classical notion of sufficiency to encoders and similarity scores, formalizing the concept as follows:
\begin{definition}[Approximate sufficiency]\label{def:sufficiency} For an image encoder $\Ei : \cXi \to \R^m$, its sufficiency is measured as
\begin{equation}\label{eqn:sufficiency_encoder}
\suffmeasure(\Ei) = \E_{\xi\sim\P_\image}\Big[\KL\Big( \P_{\txt | \image}(\cdot | \xi) \Big|\Big|  \P_{\txt|\image}(\cdot  | \Ei(\xi)) \Big) \Big],
\end{equation}
where $\P_{\txt | \image}(\xt| \Ei(\xi))$ denotes the conditional distribution of $\xt$ given $\Ei(\xi)$ under $\P_{\image, \txt}$. The sufficiency measure for a text encoder $\Et: \cXt \to \R^m$ is defined symmetrically. 

A similarity score $\simscore: \cXi \times \cXt \to \R$ induces a probability distribution $\what \P_\simscore$ over $\cXi \times \cXt$:
\begin{equation}
\what \P_\simscore(\xi, \xt) \defn \frac{\exp(\simscore(\xi, \xt) ) \P_{\image}(\xi) \P_{\txt}(\xt)}{\sum_{\xi', \xt'} \exp(\simscore(\xi', \xt')) \P_{\image}(\xi') \P_{\txt}(\xt')}.
\end{equation}
Its sufficiency measure, $\suffmeasure(\simscore)$, is defined as
{
\begin{equation}\label{eqn:sufficiency_score}
\begin{aligned}
\suffmeasure(\simscore) =&~ \E_{\xi\sim\P_\image}\Big[\KL\Big( \P_{\txt | \image}(\cdot | \xi) \Big|\Big|  \what \P_\simscore(\cdot  | \xi) \Big) \Big] + \E_{\xt\sim\P_\txt}\Big[\KL\Big( \P_{\image | \txt}(\cdot | \xt) \Big|\Big|  \what \P_\simscore(\cdot  | \xt) \Big)\Big],
\end{aligned}
\end{equation}
}
where $\what \P_\simscore(\xt | \xi)$ and $\what \P_\simscore(\xi | \xt)$ are the conditional distributions induced by $\what \P_\simscore$. By this definition, it follows that $\suffmeasure(\Ei) + \suffmeasure(\Et) \le \suffmeasure(\Upsilon(\Ei(\cdot), \Et(\cdot)))$. 

We say that $\star \in \{ \Ei, \Et, \simscore \}$ is $\eps$-sufficient if $\suffmeasure(\star) \leq \eps$. Statistics with small sufficiency measures are called ``approximate sufficient statistics'' or ``near-sufficient statistics''. 
\end{definition}

Approximate sufficiency has a more intuitive form via the information loss: 
\[
\suffmeasure(\Ei) = \mutinfor(\xi, \xt) - \mutinfor(\Ei(\xi), \xt),
\]
with the proof provided in~\Cref{sec:properties_approx_suff}. This implies that when $\suffmeasure(\Ei) = 0$, we have $\mutinfor(\xi, \xt) = \mutinfor(\Ei(\xi), \xt)$, aligning $0$-sufficiency with the classical notion of sufficiency. Although the concept has been mentioned in the literature \cite{chen2020neural}, we are unaware of any formal or rigorous definition of approximate sufficient statistics in prior work. In \Cref{sec:CLIP_ZSC_VLM_CDM}, we illustrate that near-sufficient encoders achieve strong performance on downstream tasks, including zero-shot classification and conditional diffusion models

We introduce an assumption on the boundedness of the score function, which allows us to show that near-minimizers of the CLIP risk function serve as near-sufficient statistics.

\begin{assumption}[Bounded score]\label{ass:bounded_score}
Let $\FunSpace$ denote the set of score functions over which the minimization is performed. There exists a constant $c_1>0$ such that for all pairs $(\xi,\xt)$, we have $\frac{\P_{\image, \txt}( \xi, \xt) }{ \P_{\image}(\xi) \cdot \P_{\txt}(\xt)}\in[1/c_1,c_1]$ and  $\exp(\scorefun(\xi,\xt))\in[1/c_1,c_1]$ for all $\scorefun\in\FunSpace$. 
\end{assumption}
{{We refer to Appendix~\ref{sec:ablation_bounded}} for more discussions on Assumption~\ref{ass:bounded_score}.}
 Building on this assumption, we establish the following result.
\begin{proposition}[Near-minimizer of CLIP as near-sufficient statistics]\label{prop:approximate_global_min_contrastive_loss}
Assume \Cref{ass:bounded_score} holds, and let $\orisk_{\clip, K}$ denote the CLIP risk as defined in Eq.~(\ref{eqn:CLIP_risk}). Suppose $\truesimscore$ is a global minimizer of $\orisk_{\clip, K}(\simscore)$ as defined in Eq.~(\ref{eqn:global_min_clip_risk}). Then, there exists a constant $\polyshort>0$, which depends polynomially on $c_1$, such that for any $\simscore\in\FunSpace$, its sufficiency can be bounded in terms of its CLIP excess risk. Specifically, for any $K \geq 3$, we have:
\begin{align}\label{eqn:sufficiency_excess_risk}
\lim_{K' \to \infty}\Big[ \orisk_{\clip,  K'}(\simscore) - \orisk_{\clip,  K'}(\truesimscore)\Big] = \suffmeasure(\simscore) \leq \underbrace{\Big[ \orisk_{\clip,  K}(\simscore) - \orisk_{\clip,  K}(\truesimscore) \Big]}_{\textup{CLIP excess risk}}\cdot\Big(1 +\frac{\polyshort}{K}\Big). 
\end{align} 
\end{proposition}
The proof of \Cref{prop:approximate_global_min_contrastive_loss} is provided in \Cref{app:approximate_global_min_contrastive_loss}. The first equality in Eq.~(\ref{eqn:sufficiency_excess_risk}) follows established results in prior literature, such as \cite{wang2020understanding, zimmermann2021contrastive}. The primary contribution of \Cref{prop:approximate_global_min_contrastive_loss} lies in the non-asymptotic sufficiency bound in Eq.~(\ref{eqn:sufficiency_excess_risk}), which improves upon prior results, e.g., \cite[Theorem 1]{wang2020understanding}. Compared to \cite{wang2020understanding}, the bound presented here offers two significant improvements: (1) the error decays at a faster rate of $K^{-1}$ rather than $K^{-1/2}$, and (2) the error is multiplicative rather than additive. The multiplicative error bound ensures that, for any finite K, the exact minimizer of the CLIP risk is 0-sufficient, whereas an additive error bound does not provide this guarantee. 
{
On the other hand, one can establish an additive error bound without requiring the boundedness conditions in Assumption~\ref{ass:bounded_score}. We refer the readers to Appendix~\ref{sec:alternative_bounded_score} for more details.
}

\subsection{Adaptation to various downstream tasks}\label{sec:CLIP_ZSC_VLM_CDM}

Consider the couple of encoders $(\Ei: \cXi \to \R^{\dimclip}, \Et: \cXt \to \R^{\dimclip})$ and a link function $\scorelink: \R^{\dimclip} \times \R^{\dimclip} \to \R$, such that $\simscore(\xi, \xt) \defn \scorelink(\Ei(\xi), \Et(\xt))$ is a near-minimizer of the CLIP risk. By~\Cref{prop:approximate_global_min_contrastive_loss}, $\Ei$ and $\Et$ are near-sufficient statistics of the conditional models. In this section, we show that the error in downstream tasks is bounded by their sufficiency through direct applications of data-processing inequalities.

\paragraph{Zero-shot classification (ZSC).} In the zero-shot classification task \cite{radford2021learning, jia2021scaling}, the goal is to predict the label $y$ for a new image $\xi$ without having trained on a task-specific dataset. The ZSC approach starts by sampling $(\xt(y))_{y \in \cY}$ from a chosen distribution, computing the similarity score functions $(\simscore(\xi, \xt(y)) )_{y \in \cY}$, and then selecting the predicted label for $\xi$ using the formula $\arg\max_{y \in \cY} \simscore(\xi,\xt(y))$. 
To provide a theoretical foundation for this method, we assume  the data distribution satisfies a conditional independence criterion: 
\begin{assumption}[Conditional independence]\label{ass:conditional_independence}
For the joint distribution $(\xi, \xt, y) \sim \P_{\image, \txt, \cls}$, the image $\xi$ and the label $y$ are conditionally independent given $\xt$. Notably, a special case of this assumption arises when $y$ is a deterministic function of $\xt$. 
\end{assumption}

We propose a modified zero-shot classification procedure and establish a theoretical guarantee for its performance. For each $y \in \cY$, we generate $M$ independent samples $(\xt^{(j)}(y) )_{j \in [M]} \sim_{iid} \P_{\txt | \cls}(\xt|y)$. The classifier’s predicted distribution is then defined as the softmax over aggregated score functions $L(\xi, (\xt^\jth(y) )_{j \in [M]} )$: 
\begin{align}\label{eqn:ZSC_classifier}
\clipfntprob{M}{\simscore}(\cdot | \xi) \defn&~ \softmax((L(\xi, (\xt^\jth(y) )_{j \in [M]}) )_{y \in \cY}), \\
L(\xi, (\xt^\jth(y) )_{j \in [M]} ) =&~ \textstyle \log\Big[M^{-1} \sum_{j = 1}^M \exp(\simscore(\xi,\xt^\jth(y) ) )\Big] + \log \P(y). 
\end{align}
\Cref{prop:validity_zero_shot_classification} below shows that the error rate of this classifier $\clipfntprob{M}{\simscore}(\cdot | \xi)$ is bounded by the sufficiency of the similarity score, and hence bounded by the CLIP excess risk. 
\begin{proposition}[Zero-shot classification error bound]\label{prop:validity_zero_shot_classification}
Assume \Cref{ass:bounded_score}~and~\ref{ass:conditional_independence} hold. Let $\clipfntprob{M}{\scorefun}(\cdot | \xi)$ be as defined in Eq.~(\ref{eqn:ZSC_classifier}), and let $\P_{\cls| \image}(\cdot | \xi) \in \cP(\cY)$ denote the conditional distribution of $y$ given $\xi$ under $\P_{\image, \txt, \cls}$. Then there exists a constant $\polyshort>0$, which depends polynomially on $c_1$, such that for any $\simscore \in\FunSpace$, with probability at least $1-\delta$, 
\begin{align}
\E_{\xi\sim\P_\image}\Big[\KL\Big( \P_{\cls| \image}(y | \xi) \Big|\Big| & \clipfntprob{M}{\simscore}(y | \xi) \Big)\Big]
\leq 2\suffmeasure(\simscore) +\polyshort\cdot{\frac{\log(2/\delta)}{M}} \\
\leq&~ \underbrace{\Big[ \orisk_{\clip,  K}(\simscore) - \orisk_{\clip,  K}(\truesimscore) \Big]}_{\textup{CLIP excess risk}} \Big(2+\frac{\polyshort}{K}\Big) +\polyshort\cdot{\frac{\log(2/\delta)}{M}}.
\end{align} 
\end{proposition}
The proof of \Cref{prop:validity_zero_shot_classification} is provided in \Cref{app:proof_validity_zero_shot_classification}.
\if\showZSCprooftech1
Briefly, as $M \to \infty$, the distribution $\clipfntprob{M}{\simscore}(y | \xi) \to \clipfntprob{\infty}{\simscore}(y | \xi) = \E_{\xt \sim \what \P_{\simscore}(\xt| \xi)} [\P_{\cls| \txt}(y | \xt )]$. Notice that $ \P_{\cls| \image}(y | \xi) = \E_{\xt \sim \P_{\txt|\image}(\cdot| \xi)} [\P_{\cls| \txt}(y | \xt )]$. By applying \Cref{ass:conditional_independence} and the data-processing inequality, we derive the following bound:
\[
\E_{\xi}\KL( \P_{\cls| \image}(y | \xi) ||  \clipfntprob{\infty}{\simscore}(y | \xi)) \le \E_{\xi}\KL( \P_{\txt | \image}(\xt | \xi) ||  \what \P_{\simscore}(\xt | \xi)) \le \suffmeasure(\simscore).
\]
\fi
\Cref{prop:validity_zero_shot_classification} establishes that the zero-shot classification (ZSC) approach performs well when the similarity score is near-sufficient and $M$ is large. Notably, the original ZSC method in CLIP corresponds to the argmax decision rule applied on $\clipfntprob{M}{\simscore}(\cdot | \xi)$ with $M = 1$. This method performs well using only a single text sample, likely because $\exp(\simscore(\xi,\xt^\jth(y)))$ exhibits strong concentration around its expectation for fixed pairs of $(\xi, y)$, thus reducing the need for averaging over multiple samples of $\xt^\jth(y)$.
{
Our simulations on both synthetic and real data further show that increasing 
$M$ improves ZSC performance, with the gain scaling as $1/M$, consistent with Proposition~\ref{prop:validity_zero_shot_classification}; see Appendix~\ref{sec:ablation_vary_m} for more details. 
}

\paragraph{Conditional Diffusion Models (CDMs).} Text-to-image CDMs take text prompts as input and generate natural images by solving a stochastic differential equation (SDE).  We consider the stochastic localization formulation \cite{eldan2013thin, el2022sampling} of CDMs, where the drift term of the SDE is determined by a neural network trained to approximate the conditional denoising function $\bbm_t: \R^{\dim_{\image}} \times \cXt \to \R^{\dim_{\image}}$, defined as
\begin{equation}\label{eqn:CDM_true_m}
\bbm_t(\bz, \xt) = \E_{(\xi, \bg) \sim \P_{\image | \txt}(\cdot|\xt) \times \cN(\bzero, \id_{d_{\image}})}[\xi|\bz = t \cdot \xi+\sqrt{t}\cdot\bg,\xt].
\end{equation}
This neural network approximates the conditional denoising function by minimizing risk over $\Mspace_t \subseteq \{ \Mt: \R^{\dim_{\image}} \times \R^\dimclip \to \R^{\dim_{\image}} \}$, a function class where the inputs are a noisy image and the CLIP text representation. The population risk minimization formulation gives
{
\begin{align}\label{eqn:conditional_denoising_with_CLIP}
 \Mthat = \arg\min_{\Mt \in \Mspace_t} \Big\{ \risk_{\cdm, t}(\Mt, \Et) \defn \E_{(\xi, \xt, \bg) \sim \P_{\image, \txt} \times \cN(\bzero, \id_{\dim_{\image}})}\Big[ \big\| \xi - \Mt(t \xi + \sqrt{t} \bg, \Et(\xt)) \big\|_2^2 \Big] \Big\}.
\end{align}
}
Notice that the global minimizer of this formulation, when $\Mspace_t$ includes all measurable functions, yields $\Mthat(\bz, \sE) = \E[\xi| \bz = t \xi + \sqrt{t} \bg, \Et(\xt) = \sE]$, which differs from the true conditional denoising function $\bbm_t(\bz, \xt)$, as defined in Eq.~(\ref{eqn:CDM_true_m}). Nevertheless, \Cref{prop:CDM_CLIP_error} below shows that the estimation error  of $ \Mthat$ is bounded by the sufficiency of the text encoder $\Et$, and hence bounded by the CLIP excess risk. 
\begin{proposition}[Estimation error bound for CDMs]\label{prop:CDM_CLIP_error}
Assume $\sup_{\xi \in \cXi} \| \xi \|_\infty \le \boundimage$, and let $\Mspace_t$ include all measurable functions. Let the joint distribution of $(\xi, \xt, \bz_t)$ be given by $(\xi, \xt, \bg) \sim \P_{\image, \txt} \times \cN(\bzero, \id_{\dim_{\image}})$ with $\bz_t = t \cdot \xi + \sqrt{t} \cdot \bg$. Then for any $t\geq0$, the error rate of $ \Mthat$, as defined in Eq.~(\ref{eqn:conditional_denoising_with_CLIP}), is bounded by the sufficiency of the encoder $\Et$: 
\begin{equation}\label{eqn:denoising_generalization_CLIP}
\begin{aligned}
&~\E_{(\xi, \xt, \bz_t)}
\Big[ \frac{1}{\dim_\image}\cdot\big\| \bbm_t(\bz_t, \xt) - \Mthat(\bz_t, \Et(\xt)) \big\|_2^2 \Big]
\le 2 \boundimage^2 \cdot \suffmeasure(\Et). 
\end{aligned}
\end{equation} 
\end{proposition}
The proof of \Cref{prop:CDM_CLIP_error} is provided in \Cref{app:proof_CDM_CLIP_error}. Briefly, the left-hand-side of (\ref{eqn:denoising_generalization_CLIP}), scaled by a factor of $2 \boundimage^2\dim_\image$, can be bounded as follows:
{
\begin{align}
\E_{(\xt,\bz_t)}[\KL(\P(\xi | \xt, \bz_t)|| \P(\xi | \Et(\xt), \bz_t))] \le \E_{\xt}[\KL(\P(\xi | \xt)||\P(\xi | \Et(\xt)))] \le \suffmeasure(\Et),
\end{align}
}
where the first inequality follows from the data-processing inequality, and the second inequality is due to the definition of $\suffmeasure(\Et)$. 

Using \Cref{prop:CDM_CLIP_error} along with a standard Girsanov theorem analysis of diffusion models, we derive \Cref{cor:diffusion_sampling_error}, which provides a sampling error bound of CDMs. 
\begin{corollary}[Sampling Error Bound for CDMs]\label{cor:diffusion_sampling_error}
Under the setting and assumptions of \Cref{prop:CDM_CLIP_error}, let $\abwimagedist{T}(\cdot | \xt)$ denote the distribution of $\abwimage_T / T$, where $\abwimage_t$ is the solution to the SDE with drift term given by the risk minimizer $ \Mthat$ in Eq.~(\ref{eqn:conditional_denoising_with_CLIP}): 
\[
\de \abwimage_t =  \Mthat(\abwimage_t, \Et(\xt)) \de t + \de \bnoise_t,~~~~ \bz_0 = \bzero,~~~~ \bnoise_t \text{ is Brownion motion. }
\] 
Let $\probngauss{\delta}(\cdot|\xt)$ denote the distribution of $\xi+\delta\gnoise$, where $(\xi,\gnoise)\sim \P_{\image|\txt}(\cdot|\xt)\times\cN(0,\IdMat_{\dimimage}).$
Then we have the following bound on the sampling error
\[
\E_{\xt \sim \P_{\txt}}[\KL(\probngauss{\frac{1}{\sqrt{T}}}(\cdot | \xt)||\abwimagedist{T}(\cdot | \xt))] \le  \dim_\image \boundimage^2T\cdot\suffmeasure(\Et).  
\]
\end{corollary}
The proof of \Cref{cor:diffusion_sampling_error} is provided in \Cref{sec:proof_diffusion_sampling_error}. 
Additionally, we perform similar analyses for the sampling error bound for vision-language models
in \Cref{sec:error_bound_VLM}. 

\subsection{Adaptation to tasks with canonical representation}\label{sec:near_optimal_near_equivalence}


In certain cases, the joint distribution of images and text admits canonical representations $(\trueEi, \trueEt)$, which serve as sufficient statistics and are also sufficient for downstream tasks. We show that under certain conditions on these canonical representations, a simple adapter $\Adapter$—a small neural network—can transform any near-minimizer $(\Ei, \Et)$ of the CLIP risk into the canonical representations $(\trueEi, \trueEt)$. Consequently, near-minimizers of the CLIP risk can effectively adapt to downstream tasks using these canonical representations as sufficient statistics. To formalize this idea, we impose the following assumption on the canonical representations of the joint distribution $\P_{\image, \txt}$. Specifically, we require that the representation functions are linearly independent and that the inverse of the true link function is Lipschitz.
\begin{assumption}[Well-posed canonical representation]\label{ass:well_posed_representation}
Assume there exist canonical representations $\trueEi: \cXi \to \R^\truedimclip$ and $\trueEt: \cXt \to \R^\truedimclip$, along with a univariate, monotone, and invertible link function $\truescorelink$, such that
\[
\truesimscore(\xi, \xt) \defn \log \frac{\P_{\image, \txt}(\xi, \xt) }{ \P_{\image}(\xi) \P_{\txt}(\xt) } \revdef \truescorelink( \< \trueEi(\xi), \trueEt(\xt) \> ). 
\]
We further assume the following conditions on  $\trueEi$ and $\truescorelink$:
\begin{itemize}
\item[(a)] $ \E_{\xi \sim \P_{\image}} [\trueEi(\xi) \trueEi(\xi)^\sT]  \succeq  \IdMat_{\truedimclip}/\Binvbd^2$ for some $\Binvbd>0$. 
\item[(b)] The true link function $\truescorelink$ is invertible over the feasible range of $\< \trueEi(\xi), \trueEt(\xt) \>$, and its inverse function $\truescorelink^{-1}$ is $\tslinvLip$-Lipschitz.     
\end{itemize}
\end{assumption}

We provide two examples where these assumptions are satisfied.
\begin{example}[Separator representation]\label{example:separator}
Let $\bs\in\cS$ with $|\cS| = \truedimclip$ be a separator of $(\xi, \xt)$, meaning that under the joint distribution $\P_{\image,\txt,\sep}(\xi, \xt, \bs)$, $(\xi, \xt)$ are conditionally independent given $\bs$. In this case, the canonical representations are given by $\trueEi(\xi) = [ \P(\bs | \xi)/\P(\bs)]_{\bs \in \cS} \in \R^{\truedimclip}$ and $\trueEt(\xt) = [ \P(\bs | \xt)]_{\bs \in \cS} \in \R^{|\cS|}$, with the link function defined as $\truescorelink(t) = \log(t)$. This setup leads to
\[
\begin{aligned}
\truescorelink(\langle \trueEt(\xt), \trueEi(\xi)\rangle ) =&~ \log \sum_{\bs \in \cS} \P(\bs | \xi) \P(\bs | \xt) / \P(\bs) = \log \{ \P(\xi, \xt) / [\P(\xi) \P(\xt)]\}.
\end{aligned}
\]
In this example,
\Cref{ass:well_posed_representation}(a) holds if the matrix $(\E_{\xi \sim \P_{\image}} [\frac{\P(\bs_1 | \xi) \P(\bs_2 | \xi) }{ \P(\bs_1) \P(\bs_2)} ])_{\bs_1, \bs_2 \in \cS}  \succeq  \frac{\IdMat_{\truedimclip}}{\Binvbd^2}$; 
\Cref{ass:well_posed_representation}(b) holds if we impose a uniform upper bound on $\frac{\P_{\image, \txt}(\xi, \xt)}{ \P_{\image}(\xi) \P_{\txt}(\xt)}$. 
\end{example}

\begin{example}[Exponential family representation]\label{example:exponential_family}
Take $\truescorelink(t) = t$ as the identity function. Then, $\P_{\image|\txt}(\xi | \xt) \\= \P_{\image}(\xi)\exp\{ \langle \trueEi(\xi), \trueEt(\xt)\rangle \}$ defines an exponential family, with $\trueEt(\xt)$ as the natural parameter and $\trueEi(\xi)$ as the sufficient statistic (a similar formulation holds for the reverse conditional distribution). In this case, \Cref{ass:well_posed_representation}(b) is automatically satisfied, as $\truescorelink^{-1}$ is simply the identity function. 
\end{example}

Recall that we assumed representations $\Ei: \cXi \to \R^{\dimclip}$ and $\Et: \cXt \to \R^{\dimclip}$, along with a link function $\scorelink: \R^{\dimclip} \times \R^{\dimclip} \to \R$, such that $\simscore(\xi, \xt) \defn \scorelink(\Ei(\xi), \Et(\xt))$ is a near-minimizer of the CLIP risk $\orisk_{\clip, K}$, rendering $\Ei$ and $\Et$ near-sufficient. The following result shows that a simple adapter exists that can transform these near-sufficient representations $(\Ei, \Et)$ into the canonical representations $(\trueEi, \trueEt)$. 

\begin{proposition}[Near-equivalence to the canonical representations]\label{prop:representation_equivalence}
Suppose \Cref{ass:bounded_score} and \Cref{ass:well_posed_representation} hold. Let $\diminadapt\geq1$ be some integer, and define $\boundadapt \defn (\diminadapt \cdot \E_{\xi\sim\P_\xi}\|\Ei(\xi)\|_2^2 )^{1/2}$. Then, there exists a constant $\polyshort>0$, which depends polynomially on $\boundscoreconst$, and a parameter $\btheta = (\adaweighta \in \R^{\truedimclip \times \diminadapt}, \adaweightb \in \R^{\diminadapt \times \dimclip} )$ with $\lops{\adaweighta}\leq \polyshort\Binvbd/\sqrt{\diminadapt},\lops{\adaweightb}\leq \polyshort\boundadapt$, such that defining a simple adapter
\[
\Adapter_\btheta(\Et) \defn \adaweighta \Big( \truescorelink^{-1}\Big( \log \Big[ \diminadapt \cdot \softmax\big(\scorelink(\adaweightb_{, j:}, \Et) \big) \Big] \Big) \Big) \Big)_{j \in [\diminadapt]}, 
\]
the transformed embedding $\what \Et(\xt)\defn \Adapter_\btheta(\Et(\xt))$ satisfies
\begin{align}
\E_{\xt}[\| \what \Et(\xt) - \trueEt(\xt) \|_2^2] \le \polyshort \cdot \Binvbd^2 \cdot \tslinvLip^2 \cdot \truedimclip\cdot ( \suffmeasure(\scorefun) + \diminadapt^{-1} ). \label{eq:bound_adaptation}
\end{align}
\end{proposition}
The proof of \Cref{prop:representation_equivalence} is provided in \Cref{app:proof_representation_equivalence}. In short, we exploit the fact that $\truescorelink( \langle \trueEi, \trueEt \rangle ) \\\approx \scorelink(\Ei, \Et)$, which leads to the heuristic approximation $\trueEt \approx \trueEi^\dagger \truescorelink^{-1} \scorelink(\Ei, \Et)$. Here, $\trueEi^\dagger$ is interpreted as a high-dimensional matrix. To reduce the dimensionality, we introduce a sampling approach to approximate $\trueEi^\dagger \truescorelink^{-1} \scorelink(\Ei, \Et)$ with lower-dimensional operators. A similar approach was used in \cite{tosh2021contrastive1} in a more restricted setting: when the link functions $(\truescorelink, \scorelink)$ are the logarithm, the canonical representation $\trueEt$ can be efficiently recovered via a linear transformation of $\Et$.

\begin{remark}[Adaptation to downstream tasks with canonical representation]\label{rmk:adaptation_canonical}
When $\scorelink$ is a simple function, the adapter $\Adapter_\btheta$ can be efficiently approximated by a shallow neural network. Consequently, consider a target function $f_\star(\trueEt(\xt))$ that depends on $\xt$ through the canonical representation $\trueEt$, and assume that $f_\star$ can be efficiently approximated by a neural network. Under the conditions of \Cref{prop:representation_equivalence}, where $\simscore(\xi, \xt) \defn \scorelink(\Ei(\xi), \Et(\xt))$ is a near-minimizer of the CLIP risk $\orisk_{\clip, K}$, it follows that $f_\star(\trueEt(\xt))$ can be efficiently approximated by a neural network applied to $\Et(\xt)$. 

This strategy will be applied in \Cref{sec:CDM_GHM}~and Appendix~\ref{sec:VLM_GHM} to conditional diffusion models (CDMs) and vision-language models (VLMs), where we construct efficient neural network approximations for prediction functions based on pre-trained CLIP encoders.
\end{remark}

{
\begin{remark}[An improved bound]
The error bound in Eq.~\eqref{eq:bound_adaptation} depends on the embedding $\Et(\xt)$ through the term $\suffmeasure(\scorefun)$. This term  can be replaced by $\suffmeasure(\Et)$ if the link function $\scorelink$ and the image embedding $\Ei(\xi)$ are chosen such that $\scorelink(\Ei(\xi),\Et(\xt))=\log \frac{\P_{\image|\txt}(\xi|\Et(\xt))}{\P_{\image}(\xi)}$.
However, while such a link function and embedding exist in principle, there is no guarantee that the link function is ``simple'' and can be efficiently approximated by a shallow neural network. We refer readers to the end of~\Cref{app:proof_representation_equivalence} for more details. 
\end{remark}
}

\section{Sample-efficient learning in hierarchical models}\label{sec:GHMs}

In the previous section, we showed that near-minimizers of the CLIP risk are near-sufficient and adaptable to downstream tasks, including zero-shot classification (ZSC), conditional diffusion models (CDMs), and vision-language models (VLMs). Despite these findings, it remains unclear why certain neural networks can efficiently learn these near-minimizers and the associated functions within CDMs and VLMs. In this section, we address this question by introducing a concrete data generation model for image-text pairs.

Specifically, we assume that the image-text pairs are generated according to a \textit{joint generative hierarchical model} (JGHM), which integrates two generative hierarchical models (GHMs) with a shared root. A GHM is a tree-structured graphical model in which the root node represents the highest-level features; these features hierarchically generate lower-level features based on a transition kernel, eventually reaching the leaf nodes that represent observed images or text. GHMs have been widely used in theoretical modeling for images and language independently \cite{mossel2016deep, petrini2023deep, sclocchi2024phase, tomasini2024deep, cagnetta2024towards, garnier2024transformers, kadkhodaie2023learning, kadkhodaie2023generalization, mei2024u}. The JGHM framework extends GHMs to jointly model paired image and text data\footnote{We use the JGHM as a working model and acknowledge that it may not fully capture the complexity of the image-text distribution. Developing a more realistic model for image-text data is left for future work. In this paper, we focus on a model that captures the hierarchical structure of image-text pairs and provides an efficient sample complexity bound.}. In the following, we formally define the JGHM, building on the GHM framework presented in \cite{mei2024u}. 

\paragraph{The joint tree structure.}
Consider a joint tree structure $\cT = { \cT_\image, \cT_\txt }$, consisting of two trees, $\cT_\image$ and $\cT_\txt$, each of height $L$. These trees generate images and text, respectively, and share a common root node, $\root$, which represents shared information across the image and text domains. 
Let the sets of nodes in the image and text trees be $\cV_\image$ and $\cV_\txt$, respectively. The root is defined as level $0$, and the set of nodes at a distance $\ell$ from the root is referred to as level $\ell$. These nodes are denoted by $\cV_{\image}^{(\ell)}$ in the image tree and $\cV_\txt^{(\ell)}$ in the text tree.
Let \(\cC(v)\) represent the set of its children defined within either \(\cT_\image\) or \(\cT_\txt\), as appropriate. 
We assume that for any $v \in \cV^{(\ell-1)}_\image$ (or $\cV^{(\ell-1)}_\txt$), the number of children is fixed at $\nchild_\image^{(\ell)}$ (or $\nchild_\txt^{(\ell)}$) for $\ell \in [L]$, except for leaf nodes $v \in \cV^{(L)}_\image$ (or $\cV^{(L)}_\txt$), which have no children. 
The number of nodes at each layer is denoted by $\dim^{(\ell)}_\image=|\cV^{(\ell)}_\image|$ and $\dim^{(\ell)}_\txt=|\cV^{(\ell)}_\txt|$.
In particular, the total number of leaf nodes is represented by $\dim_\image = \dim^{(L)}_\image=|\cV^{(L)}_\image|$ and $\dim_\txt = \dim^{(L)}_\txt=|\cV^{(L)}_\txt|$.   Additionally, we define $\mmaintextone\defn \max\{\nchild_\image^{(1)},~\nchild_\txt^{(1)}\}, \mmaintextmax \defn \max_{\ell\in[L]}\max\{\nchild_\image^{(\ell)},\nchild_\txt^{(\ell)}\}$. 

\paragraph{Joint generative hierarchical models (JGHMs).}
Building on the joint tree structure, we define the joint generative model for the image $\xi$ and text $\xt$. Each node in the tree is associated with a variable: the root node is represented by $x^{(0)}_{\root}=x^{(0)}_{\image,\root}=x^{(0)}_{\txt,\root}\in \cS_\root$; for nodes $v \in \cV^{(\ell)}_\image$ at levels $1\leq \ell\leq L$, the variables are $x^{(\ell)}_{\image, v} \in \cS_\image$; and for nodes $v \in \cV^{(\ell)}_\txt$ at levels $1\leq \ell\leq L$, the variables are $x^{(\ell)}_{\txt, v} \in \cS_\txt$. Here, $\cS_\root$, $\cS_\image$, and $\cS_\txt$ denote the spaces of root, image, and text variables, respectively. For simplicity, we set $\cS_\root = \cS_\image = \cS_\txt = [\state]$ for some $\state \in \N_{ > 0}$; however, our theoretical results extend naturally to the more general case where these spaces differ. We collectively denote the variables associated with $\cV^{(\ell)}_{\image}$ and $\cV^{(\ell)}_{\txt}$ as $\bx^{(\ell)}_{\image} = (x_{{\image},v}^{(\ell)})_{v \in \cV^{(\ell)}_{\image}}$ and $\bx^{(\ell)}_{\txt} = (x_{{\txt},v}^{(\ell)})_{v \in \cV^{(\ell)}_{\txt}}$, respectively. For the leaf level $\ell=L$, we sometimes omit the superscript $(L)$ for brevity.

The joint distribution $\truemu(x^{(0)}_\root, \bx^{(1)}_\image, \ldots, \bx^{(L-1)}_\image, \bx_\image, \bx^{(1)}_\txt, \ldots, \bx^{(L-1)}_\txt, \bx_\txt)$ is defined as
\begin{align}\label{eqn:hierarchical_bayes}
&~\truemu(x^{(0)}_\root, \bx^{(1)}_\image, \ldots, \bx^{(L-1)}_\image, \bx_\image, \bx^{(1)}_\txt, \ldots, \bx^{(L-1)}_\txt, \bx_\txt)\\ \propto&~ \textstyle \psi_\root^{(0)} (x^{(0)}_\root) \cdot \psi_{\image}^{(1)}(x^{(0)}_\root, \bx^{(1)}_\image) \cdot \Big(\prod_{v \in \cV^{(1)}_{\image}} \psi_{\image}^{(2)}(x_{\image, v}^{(1)}, x_{\image, \cC(v)}^{(2)}) \Big) \cdots \Big( \prod_{v \in \cV^{(L - 1)}_{\image}}\psi_{\image}^{(L)}(x_{\image, v}^{(L-1)}, x_{\image, \cC(v)})\Big) \cdot
\\ 
&\quad\quad\quad\quad\ \textstyle \psi_{\txt}^{(1)}(x^{(0)}_\root, \bx_{\txt}^{(1)}) \cdot \Big(\prod_{v \in \cV^{(1)}_{\txt}} \psi_{\txt}^{(2)}(x_{\txt, v}^{(1)}, x_{\txt, \cC(v)}^{(2)}) \Big) \cdots \Big( \prod_{v \in \cV^{(L - 1)}_{\txt}}\psi_{\txt}^{(L)}(x_{\txt, v}^{(L-1)}, x_{\txt, \cC(v)})\Big), 
\end{align}
where $\psi_\root^{(0)}\colon [\state]\to \R_{\ge 0}$, $\psi_\image^{(\ell)}\colon  [\state]\times [\state]^{m_\image^{(\ell)}}\to \R_{\ge 0}$, and $\psi_\txt^{(\ell)}:[\state] \times [\state]^{m_\txt^{(\ell)}}\to \R_{\ge 0}$ define the transition probabilities of child nodes conditioned on their parent node. This joint distribution models the image-text generation process. Starting at the shared root of the image-text tree, initialized according to $\psi_\root^{(0)}$, values are sampled level-by-level through the transition probabilities $\psi_{\{\image,\txt\}}^{(\ell)}$. This process continues until the leaf variables, $\xi$ and $\xt$, are generated. The observed data consist of the leaf variables $\xi$ (image) and $\xt$ (text), while the intermediate variables are typically unobserved.

\if\showfigure1
\begin{figure}[t]
\centering 
\includegraphics[width=0.46\linewidth]{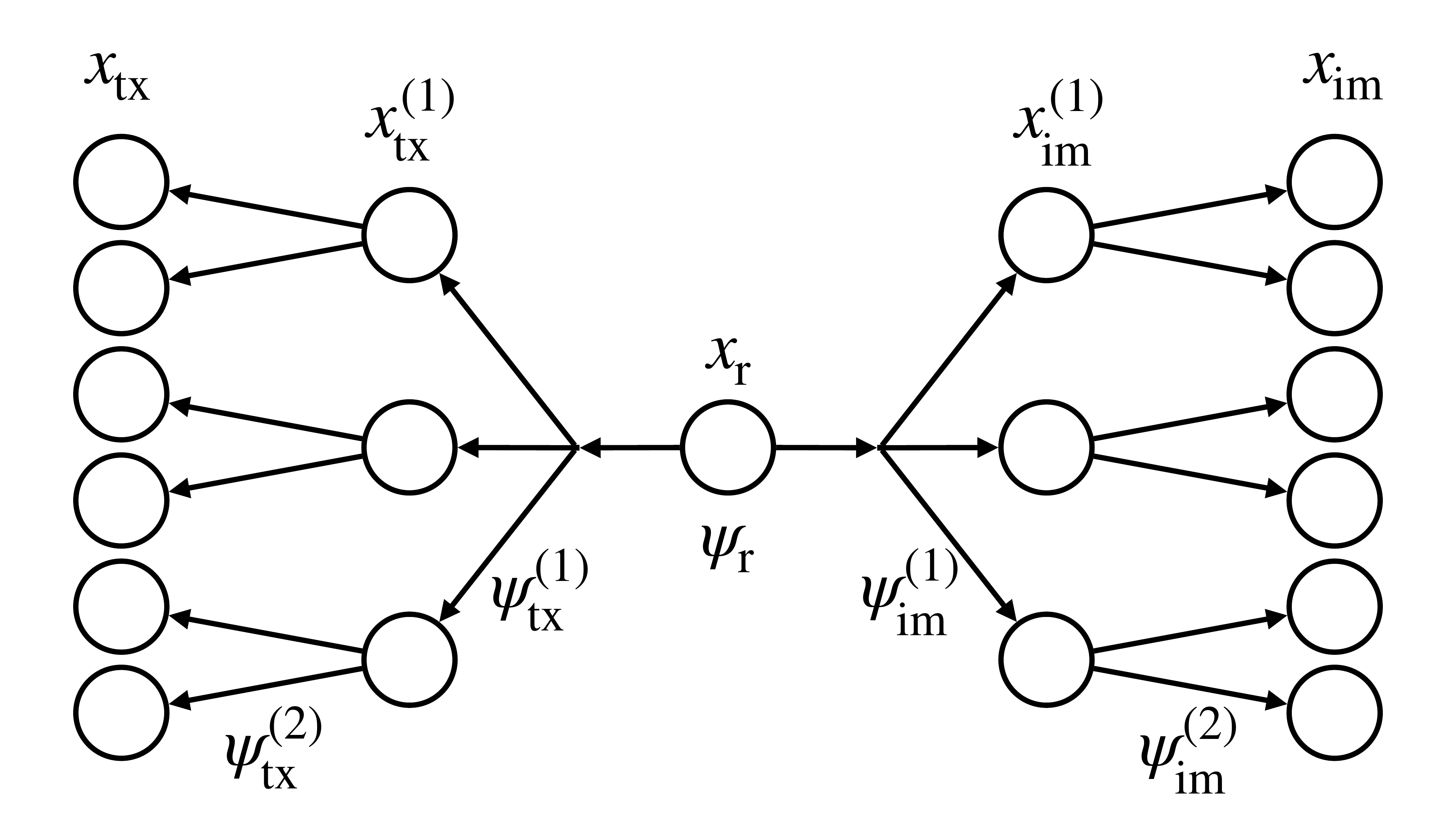}
\hspace{10pt}
\includegraphics[width=0.46\linewidth]{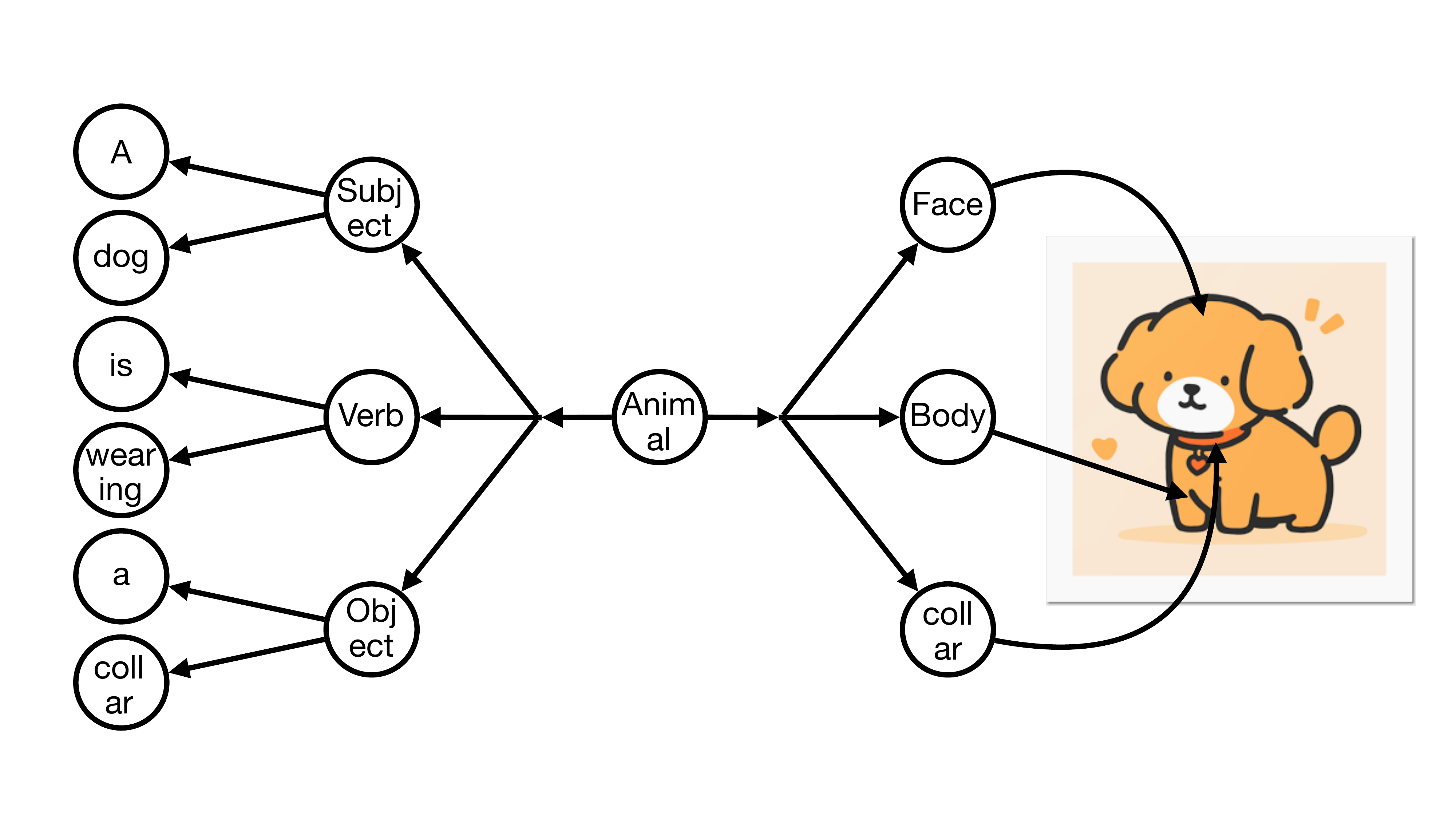}
\caption{{ Left: the JGHM used to generate the joint distribution of text and images. Right: an illustrative example of a generated text-image pair.} }\label{fig:JGHM} 
\end{figure}
\fi

We impose the following factorization assumption on the $\psi$ functions in the JGHM model. This assumption implies that the child nodes are conditionally independent of the parents, and that the transition is homogeneous across all nodes within a particular layer. Although this assumption, inherited from \cite{mei2024u}, is not strictly necessary for the theoretical framework and could be relaxed with additional technical work, it significantly simplifies the presentation and proof.  Therefore, we retain it here for convenience. 
\begin{assumption}[Factorization of $\psi$]\label{assumption:factorization}
For each $v\in \cV^{(\ell)}_{\image}$, let there be a known ordering function $\iota: \cC_{\image}(v) \to [\nchild^{(\ell)}_{\image}]$ that is bijective. A similar ordering function $\iota$ is defined for each $v\in \cV^{(\ell)}_{\txt}$ as well. 
For $\square \in \{ \image, \txt\}$, each layer $\ell \in [L]$ and node $v \in \cV^{(\ell-1)}_\square$, we assume
\begin{equation}
\begin{aligned}
\label{eqn:factorization_ass}
\psi^{(\ell)}_\square(x^{(\ell - 1)}_{\square,v}, x^{(\ell)}_{\square,\cC(v)}) = \prod_{v' \in \cC(v)} \psi^{(\ell)}_{\square,\iota(v')}(x_{\square,v}^{(\ell - 1)}, x_{\square,v'}^{(\ell)}). 
\end{aligned}
\end{equation}
\end{assumption}
In addition, we assume boundedness for the $\psi$ functions.
\begin{assumption}[Boundedness of $\psi$]\label{ass:bounded_transition}
There exists some $\transbd>0$ such that for any  $x,x'\in [\state]$,
\begin{equation}
1/\transbd \le \psi_\root^{(0)}(x),\psi_{\image,\iota}^{(\ell)}(x, x') ,\psi_{\txt,\iota}^{(\ell)}(x, x')  \le \transbd.
\end{equation}
\end{assumption}
A schematic illustration of the JGHM with two layers is shown in \Cref{fig:JGHM}.

\subsection{Sample-efficient learning of CLIP encoders and ZSC}\label{sec:clip_JGHM}

{
Consider a set of $\numbatch K$ i.i.d. samples $\{ (\xi^{(i)}, \xt^{(i)}) \}_{i \ge 1}$ drawn from the distribution $\truemu$ under the JGHM. They can be  reorganized into the form 
$\{  (\xi_{,j}^{(i)}, \xt_{, j}^{(i)})_{j\in[K]}\}_{i \in [\numbatch ]}$, 
where $K$ is the batch size and $\numbatch $ is the number of batches.}
Our goal is to learn encoders for both the image and text components by minimizing the CLIP loss. The optimal similarity score under the CLIP loss is given by the logarithmic probability ratio 
$\truesimscore(\xi,\xt)= \log\frac{\truemu(\xi, \xt) }{\truemu(\xi) \truemu(\xt)}
$
. 
We seek to analyze the sample complexity required to learn this optimal similarity score using empirical risk minimization over the class of transformers.

\paragraph{The neural network architecture.} 
The similarity score consists of three main components: a transformer encoder\footnote{While we use a transformer architecture to align with practical implementations, the theoretical framework does not require the use of transformers to avoid the curse of dimensionality. Any network capable of approximating the belief propagation algorithm can be utilized. We do not claim that transformers are the optimal architecture for this purpose.} for images, $\NN_\image^{\bW_{\image}} \colon [\state]^{\dim_\image} \to \R^\state$; a transformer encoder for text, $\NN_\txt^{\bW_{\txt}}\colon [\state]^{\dim_\txt} \to \R^\state$; and a parameterized similarity link function,  
$\NNs^{\linkweight}(h,h')=\log \readsimscore(\sum_{s\in [\state]}h_s h_s' w_s)$, where $\linkweight, h, h' \in \R^\state$, and $\readsimscore(\cdot):\R\mapsto\R_{>0}$ is a truncation function. The similarity score, $\simscore_\NN^\btheta$, with parameters $\btheta = (\bW_{\image}, \bW_{\txt}, \linkweight)$, is defined as
\begin{align}
    \simscore_\NN^\btheta(\xi, \xt) \defn \NNs^{\linkweight}\big(\softmax(\NN_\image^{\bW_{\image}}(\xi)),\softmax(\NN_\txt^{\bW_{\txt}}(\xt))\big).\label{eq:simscore_architecture}
\end{align}
The same network architecture is used for the vision transformer $\NN_\image^{\bW_{\image}}$ and the text transformer $\NN_\txt^{\bW_{\txt}}$, but with different weights. For simplicity, we describe the architecture generically and omit subscripts. The neural network output is given by $\NN^{\bW}(\bx) = \readclip(\TF^{\bW}(\encodeclip(\bx)))$. 
The only trainable part, $\TF^{\bW}$, is a repetition of transformer blocks as described below. 
The fixed embedding function, $\encodeclip\colon \R^d \to \R^{D \times d}$, maps the input $\bx \in \R^d$ (including positional encoding) to a matrix $\hmatrix^{(L)} = \encodeclip(\bx) \in \R^{D\times d}$, and the fixed readout function, $\readclip\colon \R^{D \times d}\to \R^S$, extracts an $(S\times 1)$ 
submatrix from the output of the last transformer block $\hmatrix^{(0)}=\TF^{\bW}(\encodeclip(\bx))$. Definitions of the functions $\encodeclip$ and $\readclip$ are provided in Appendix~\ref{subsubsection:CLIP-Overview-Transformer}.

\begin{definition}[The transformer architecture]\label{def:transformer_CLIP}
The transformer, $\TF^{\bW}: \R^{D \times d} \to \R^{D \times d}$, consists of $L$-blocks of the $(\depth+1)$-layer fully-connected ReLU network $\FF^{(\ell)}: \R^{D\times d} \to \R^{D \times d}$, applied column-wise, and the self-attention layer $\Attn^{(\ell)}: \R^{D\times d} \to \R^{D \times d}$, defined as: 
\begin{align}
  \FF^{(\ell)}(\xnetwork) =&~  \ffnweight_{\depth+1}^{(\ell)} \,[\ \cdot\ ;1] \, \circ \, \ReLU\, (\ffnweight_{\depth}^{(\ell)}\,[\ \cdot\ ;1]) \,
  \circ \cdots \circ \ReLU(\ffnweight_{1}^{(\ell)} \,[\xnetwork;1]\,),\quad (\xnetwork \in \R^{D})
    \\
 \Attn^{(\ell)}(\qmatrix) =&~  \valuematrix^{(\ell)} \qmatrix \cdot  \softmax_{\col}\big(\qmatrix^\sT (\keymatrix^{(\ell)})^\sT \querymatrix^{(\ell)} \qmatrix \big).
\end{align}
Here, $[\ \cdot\ ;1]$ appends a constant 1 to the end of a vector, introducing an intercept term. 
Starting from $\hmatrix^{(L)}$, the $\ell$-th block computes intermediate representations $\hmatrix^{(\ell)}\in \R^{D\times d}$ and $\qmatrix^{(\ell)}\in \R^{D\times d}$ as follows: 
\begin{align}
    \qmatrix^{(\ell)} &= \hmatrix^{(\ell)} + \FF^{(\ell)}(\hmatrix^{(\ell)}),
    \\
    \hmatrix^{(\ell-1)} &= \normalize(\qmatrix^{(\ell)} + \Attn^{(\ell)}(\qmatrix^{(\ell)})).
\end{align}
For simplicity,  $\FF^{(\ell)}$ is treated as a function from $\R^{D\times d}$ to $\R^{D\times d}$, though applied column-wise.
The transformer weights, denoted by $\bW$ (subscripts ${\image, \txt}$ correspond to specific transformers), are given by:  
\begin{align}\label{eqn:W_matrix_CLIP_encoder}
\bW = \Big\{  \querymatrix^{(\ell)}, \keymatrix^{(\ell)}, \valuematrix^{(\ell)} \in \R^{D \times D},  \ffnweight_{1}^{(\ell)}\in\R^{D' \times (D+1)},\{ \ffnweight_{i}^{(\ell)} \in \R^{D' \times (D'+1)} \}_{i = 2}^{\depth},\ffnweight_{\depth+1}^{(\ell)} \in \R^{D\times (D'+1)} \Big\}_{\ell \in [\layer]}. 
\end{align}
Here, $\softmax_{\col}$ denotes a column-wise softmax operation, where for any matrix $\bA \in \R^{d \times d}$, each column of $\softmax_{\col}(\bA) \in \R^{d \times d}$ is the softmax of the corresponding column in $\bA$. The function $\normalize: \R^{D \times d} \to \R^{D \times d}$ performs column-wise normalization, where each column of $\normalize(\hmatrix) \in \R^{D \times d}$ is the normalized version of the corresponding column in $\hmatrix$, with its formal definition provided in Appendix~\ref{subsubsection:CLIP-Overview-Transformer}.
\end{definition}

Intuitively, each column vector of $\hmatrix^{(\ell)}$ corresponds to a leaf node $v$. As we will show in Appendix~\ref{sec:clip_pf_overview}, each transformer block approximates one step of belief propagation. Consequently, the blocks are indexed in decreasing order $(\ell = \layer, \dots, 1)$ to align with the belief propagation process. Some modifications are also incorporated, such as placing the feedforward layer first and using a multi-layer network for the feedforward component. However, these changes are not essential and can be effectively simulated within the original transformer architecture~\cite{vaswani2017attention}.

\paragraph{The ERM estimator.} To find the optimal similarity score, we solve the empirical risk minimization problem defined by the following objective: 
\begin{align}
    \label{eq:clip-empirical-risk-minimization}
    \textstyle \EstPar
    =
    \underset{\Par \in \Parspaceclip }{\argmin} \Big\{ \what \risk_{\clip,K}(\simscore_\NN^\Par) \defn&~ \textstyle  \frac{1}{{ \numbatch }}\sum_{i = 1}^{{ \numbatch }} \Big[ - { \frac1K\sum_{k=1}^K}\log \frac{\exp(\simscore_\NN^\Par(\xi_{,{k}}^{(i)}, \xt_{,{k}}^{(i)})  ) }{\sum_{j \in [K]} \exp(\simscore_\NN^\Par(\xi_{,{k}}^{(i)}, \xt_{, j}^{(i)}))} \Big] \\
    &~\textstyle + \frac{1}{{ \numbatch }}\sum_{i = 1}^{{ \numbatch }} \Big[ - { \frac1K\sum_{k=1}^K}\log \frac{\exp(\simscore_\NN^\Par(\xi_{,{k}}^{(i)}, \xt_{,{k}}^{(i)})  ) }{\sum_{j \in [K]} \exp(\simscore_\NN^\Par(\xi_{, j}^{(i)}, \xt_{,{k}}^{(i)}))} \Big]\Big\}, 
\end{align}
where the parameter space is defined as:
\begin{align}\label{eqn:CLIP_param_space}
&~\Parspaceclip \defn\Big\{ \bW_{\image}, \bW_{\txt} \text{ as defined in Eq.~(\ref{eqn:W_matrix_CLIP_encoder})}, \linkweight  \text{ as defined in Eq.~(\ref{eq:simscore_architecture})};
\\
&~~~~~~~~~~\nrmps{\Par} \defn \| \linkweight \|_\infty \vee \max_{\square \in \{\image, \txt\} }\max_{i \in [\depth+1], \ell \in [L]}\{ \| \ffnweight_{i, \square}^{(\ell)} \|_{\op}, \| {\querymatrix}_{,\square}^{(\ell)}\|_{\op}, \| {\keymatrix}_{,\square}^{(\ell)} \|_{\op}, \| {\valuematrix}_{,\square}^{(\ell)} \|_{\op} \}  \le B \Big\}.
\end{align}
We expect the empirical risk minimizer, $\simscore_\NN^{\EstPar}$, to closely approximate the optimal similarity score $\truesimscore$, which minimizes the population risk $\orisk_{\clip, K}$ over all functions as defined in Eq.~(\ref{eqn:CLIP_risk}). This is quantified through the excess risk
$\excess_K(\simscore_\NN^{\EstPar}, \truesimscore) \defn \orisk_{\clip, K}(\simscore_\NN^{\EstPar}) - \orisk_{\clip, K}(\truesimscore)$. 
The following theorem provides a bound on the excess risk of the estimator $\simscore_\NN^{\EstPar}$.

\begin{theorem}[Sufficiency and excess risk bound of CLIP]\label{theorem:CLIP-Generalization}
Suppose \Cref{assumption:factorization} and \Cref{ass:bounded_transition} hold. Let $\Parspaceclip$ denote the parameter space defined in Eq.~(\ref{eqn:CLIP_param_space}), with $J = \bigOtil(L)$, $D = \bigO(\state \layer)$, $D' =\bigOtil(\mmaintextmax \state \layer^3)$, and $\bound = \bigOtil( \state \layer+\mmaintextmax^2)$. Let $\EstPar$ be the empirical risk minimizer as defined in Eq.~(\ref{eq:clip-empirical-risk-minimization}). Then, with  probability at least $1-1/n$, we have 
\begin{align}\label{eqn:excess_risk_bound_CLIP}
\textstyle \excess_K(\simscore_\NN^{\EstPar}, \truesimscore) = \bigOtil\Big( \sqrt{\frac{ \state^2 \layer^{11}\mmaintextmax^2}{n}} \Big), 
\end{align}
where $\bigOtil$ hides polynomial factors in $\log(\mmaintextmax \state \layer n \transbd)$. 

Moreover, combined with \Cref{prop:approximate_global_min_contrastive_loss}, this excess risk bound also provides an upper bound on the sufficiency of the learned encoders and the similarity score, $\suffmeasure(\NN_\txt^{\bW_{\txt}})$, $\suffmeasure(\NN_\image^{\bW_{\image}})$, and $\suffmeasure(\simscore_\NN^{\EstPar})$. 
\end{theorem}

The proof of \Cref{theorem:CLIP-Generalization} is provided in \Cref{app:proof_CLIP-Generalization}. We note that the sample complexity bound in this theorem is not intended to be the tightest possible, and refining it remains an intriguing direction for future research. Theorem~\ref{theorem:CLIP-Generalization} establishes that the excess risk vanishes whenever $n \gg \state^2 \layer^{11} \mmaintextmax^2$, with the required sample size being sub-linear in $\dim$. Crucially, this result avoids the curse of dimensionality, demonstrating that the JGHM can be efficiently learned via ERM over transformers. While simpler two-layer neural networks could be used as encoders, their approximation error and sample complexity would likely scale exponentially with the dimension $\dim$, leading to a curse of dimensionality. In contrast, transformers circumvent this issue by efficiently approximating belief-propagation. 
{
In Remark~\ref{remark:exponential_dependency_on_B}, we further show that the $1/\sqrt{n}$ rate can be improved to $1/\sqrt{nK}$, so the bound scales with the total sample size rather than the number of batches. This improvement, however, comes at the cost of an exponential dependence on $\mmaxb$, which is unavoidable under the current assumptions (see Remark~\ref{remark:exponential_dependency_on_B} for more details).
}

\paragraph{Proof strategy of \Cref{theorem:CLIP-Generalization}: Transformers efficiently approximate belief propagation.} 
The excess risk $\excess_K(\simscore_\NN^{\hat \btheta}, \truesimscore)$ can be decomposed into two components: approximation error and generalization error:
\[
\begin{aligned}
\textstyle \excess_K(\simscore_\NN^{\hat \btheta}, \truesimscore) \le \underbrace{\inf_{\btheta \in \Theta} \orisk_{\clip, K}(\simscore_{\NN}^\btheta) - \orisk_{\clip, K}(\truesimscore)}_{\text{approximation error}} + 2 \cdot \underbrace{\sup_{\btheta \in \Theta} \Big| \what \risk_{\clip, K}(\simscore_\NN^{\hat \btheta}) - \orisk_{\clip, K}(\simscore_\NN^{\hat \btheta}) \Big|}_{\text{generalization error}}. 
\end{aligned}
\]
The generalization error is controlled using standard parameter counting arguments and the chaining approach. The main focus, therefore, lies in bounding the approximation error. This is achieved by first introducing the belief propagation (BP) algorithm, which computes the conditional probabilities $(\P(x_{\root}| \xi), \P(x_{\root} | \xt))$, as shown in Eq.~(\ref{eq:Appendix-CLIP-ErrorPropagation-15}), and then showing that transformers can effectively approximate BP. See Appendix~\ref{sec:clip_pf_overview} for more detail.


\if\showinterpretremark1
\begin{remark}
We note that while the BP algorithm serves as a theoretical proof technique, we cannot conclude that the pre-trained CLIP encoders implement BP in JGHM. Investigating whether the trained CLIP encoders approximate BP remains an intriguing direction for future interpretability research. Our simulation studies in out-of-distribution settings, as shown in \Cref{fig:OOD}, provide partial evidence relevant to this question. This remark also applies to the CDM and VLM tasks. 
\end{remark}
\fi

\begin{remark}
    While classical algorithms such as maximum likelihood estimation can also efficiently learn the similarity score from JGHM, our theory shows that a neural network (NN)-based approach with contrastive pre-training can achieve the same result. A key advantage of NN-based approaches is their flexibility: they rely less on the precise specification of the underlying graphical model, whereas classical methods  struggle if the model is misspecified. This makes NN-based approaches especially useful when the data-generating process is unknown or difficult to model. This same remark applies to the CDM and VLM tasks. 
\end{remark}

\paragraph{Sample-efficient zero-shot classification.} 
Combining \Cref{theorem:CLIP-Generalization} with \Cref{prop:validity_zero_shot_classification} provides an end-to-end theory for the performance of zero-shot classification using the classifier $\clipfntprob{M}{\simscore}(\cdot | \xi)$, as defined in Eq.~(\ref{eqn:ZSC_classifier}). Here $\simscore = \simscore_\NN^{\EstPar}$ is the similarity score corresponding to the empirical risk minimizer given in Eq.~(\ref{eq:clip-empirical-risk-minimization}). 
\begin{corollary}\label{cor:ZSC_GHM}
Suppose that \Cref{ass:conditional_independence},~\ref{assumption:factorization}~and~\ref{ass:bounded_transition} hold. Let $\EstPar$ be the empirical risk minimizer defined in Eq.~(\ref{eq:clip-empirical-risk-minimization}), and let $\clipfntprob{M}{\simscore}(\cdot | \xi)$ be the zero-shot classifier as defined in Eq.~(\ref{eqn:ZSC_classifier}) with $\simscore = \simscore_\NN^{\EstPar}$. Then, with probability at least $1 - \eta$,
\begin{align}
\textstyle \E_{\xi\sim\P_\image}\Big[\KL\Big( \P_{\cls| \image}(y | \xi) \Big|\Big| \clipfntprob{M}{\simscore}(y | \xi) \Big)\Big]
\leq \bigOtil\Big( \sqrt{\frac{ \state^2 \layer^{11}\mmaintextmax^2}{n}} + {\frac{\log(2/\eta)}{M}}\Big),
\end{align} 
where $\bigOtil$ hides polynomial factors in $(\log(\mmaintextmax \state \layer n \transbd), (\transbd)^{\mmaintextone})$.
\end{corollary}
The proof of Corollary~\ref{cor:ZSC_GHM} is provided in Appendix~\ref{sec:clip_cor_pf}. 

\subsection{Sample-efficient learning of CDMs}\label{sec:CDM_GHM}

In this section, we investigate the conditional denoising models (CDMs) within the JGHM. Consider the joint distribution of noisy image, clean image, and text $(\bz_t, \xi, \xt)$, generated as follows: $(\xi, \xt)\sim\truemu$, and $\bz_t = t \cdot \xi + \sqrt{t} \cdot \bg$, where $\bg \sim \cN(\bzero, \id_{\dim_\image})$ represents independent Gaussian noise. We denote the joint distribution of $(\bz_t, \xi, \xt)$ by $\truemucdm$. 

Suppose we are given a dataset of iid samples $\{(\bz_t^{(i)}, \xi^{(i)}, \xt^{(i)})\}_{i \in [n]} \sim_{iid} \truemucdm$. With a text representation $\Et(\xt)\in\R^\dimclip$ (e.g., a CLIP-based embedding), the goal is to learn a conditional denoiser $\Mt(\bz_t, \Et(\xt))$ that closely approximates the clean image $\xi$. Under an appropriate loss function, the optimal denoiser is the Bayes denoiser $\truebbmcdm(\bz_t, \xt) = \E_{(\bz_t, \xi, \xt) \sim \truemucdm}[\xi | \bz_t, \xt]$, which computes the posterior expectation of $\xi$ given $(\bz_t, \xt)$. This section aims to analyze the sample complexity of learning this conditional denoiser using empirical risk minimization over the class of transformers.

\paragraph{The neural network architecture.}
The conditional denoiser is modeled as
\[
\Mt^\Par(\bz_t, \Et(\xt)) = \read_{\cdm} \circ \TF_{\cdm} \circ \encode_{\cdm}(\bz_t, \Adapter(\Et(\xt))), 
\] 
where each component is defined as follows. The function $\read_{\cdm}: \R^{D \times \dimimage} \to \R^{\dimimage}$ extracts the final denoised image, whereas the embedding function $\encode_{\cdm}: \R^{\dimimage} \times \R^{\state} \to \R^{D \times \dimimage}$ maps the input features into a transformer-compatible embedding, with specific details provided in Appendix~\ref{subsubsection:CDM-Intro-NN}. The text encoder $\Et: \cXt \to \R^\state$ is given by the pre-trained CLIP representations, as defined in Eq.~(\ref{eq:simscore_architecture}) and (\ref{eq:clip-empirical-risk-minimization}).

The transformer $\TF_{\cdm}: \R^{D \times \dimimage} \to \R^{D \times \dimimage}$ is a trainable $(2\layer + 1)$-layer model, defined in \Cref{def:transformer_CLIP}, with parameter $\bW_\cdm$ adapted to the $(2L + 1)$-layer structure, as detailed in Eq.~(\ref{eqn:W_matrix_CLIP_encoder}). The adapter network, $\Adapter:\R^{\state}\mapsto\R^{\state}$ is implemented as a simple network:
\begin{align}
\Adapter(\bv) \defn\adaweighta\softmax(\log(\readsimscore(\adaweightb\,\softmax(\bv) )\,)\,), ~~~~\forall \bv\in\R^\state, \label{eq:adapter_def_cdm}
\end{align} 
where $\adaweighta \in \R^{\state \times \diminadapt}$ and $\adaweightb \in \R^{\diminadapt \times \state}$ are trainable weights. This adapter network structure leverages the canonical representation of the GHM framework, as described in Example~\ref{example:separator}. We note that using an adapter network on top of CLIP representations is consistent with practice in prior work \cite{radford2021learning, esser2024scaling}. Following the pre-training fine-tuning paradigm, we consider the fine-tuning phase where the parameters $\Par = (\bW_\cdm, \adaweighta, \adaweightb)$ are optimized, while $\read_{\cdm}$, $\encode_{\cdm}$, and the CLIP encoder $\Et$ remain fixed.

\paragraph{The ERM estimator.}
Given a pre-trained text encoder $\Et:\cXt\mapsto\R^{\state}$, the goal is to obtain the conditional denoising function. To achieve this, we solve the empirical risk minimization problem defined by the following objective: 
\begin{align}\label{eq:cdm-empirical-risk-minimization}
\textstyle \what \Par = \underset{\Par \in \Parspacecdm }{\argmin} \Big\{ \what\risk_{\cdm, t}(\Mt^\Par, \Et) \defn 
\frac{1}{n}\sum_{i=1}^n
\big\| \xi^{(i)} - \Mt^\Par(\bz_t^{(i)}, \Et(\xt^{(i)})) \big\|_2^2 \Big\}, 
\end{align}
where the parameter space is defined as
{
\begin{align}\label{eqn:CDM_param_space}
&~\Parspacecdm \defn \Big\{ \bW_{\cdm} \text{ as defined in Eq.~(\ref{eqn:W_matrix_CLIP_encoder})}, \adaweighta,\adaweightb \text{ as defined in Eq.~(\ref{eq:adapter_def_cdm})};
\\
& ~~\nrmps{\btheta} \defn \lops{\adaweighta}\vee\lops{\adaweightb} \vee \max_{i \in [\depth+1], \ell \in [2\layer+1]}\{ \| \ffnweight_{i, \cdm}^{(\ell)} \|_{\op}, \| {\querymatrix}_{,\cdm}^{(\ell)}\|_{\op}, \| {\keymatrix}_{,\cdm}^{(\ell)} \|_{\op}, \| {\valuematrix}_{,\cdm}^{(\ell)} \|_{\op} \}  \le B \Big\}.
\end{align}
}

The following theorem provides an estimation error bound on the conditional denoiser:
\begin{theorem}[Estimation error of conditional denoising function]\label{theorem:CDM-Generalization-two-step}
Suppose that \Cref{assumption:factorization} and \Cref{ass:bounded_transition} hold, and assume \Cref{ass:well_posed_representation} (a) holds for the image representation $\trueEi(\xi) = [ \P(\bs | \xi) /\P(\bs)]_{\bs \in \cS} \in \R^{\state}$ where $\cS$ is the set of root nodes. Let $\Et$ and $\scorefun$ be obtained from the CLIP minimization. For simplicity, assume $t=1$. Let $\Parspacecdm$ be the set defined in Eq.~(\ref{eqn:CDM_param_space}), where $J = \bigOtil(L)$, $D = \bigO(\state \layer)$, $D' =\bigOtil(\mmaintextmax \state \layer^3)$, and $\bound = \bigOtil( \Binvbd+(\state\layer+\mmaintextmax^2)\sqrt{\diminadapt})$. Let  $\EstPar$ be the empirical risk minimizer defined in Eq.~(\ref{eq:cdm-empirical-risk-minimization}). Then, with probability at least $1 - 1/n$, we have
{
\begin{align}\label{eqn:excess_risk_bound_CDM}
\E_{(\xi, \xt, \bz_t)}
\Big[\frac{1}{\dimimage} \big\| \truebbmcdm(\bz_t, \xt) - \Mt^{\EstPar}(\bz_t, \Et(\xt)) \big\|_2^2 \Big]
\le \bigOtil \Bigg( \sqrt{\frac{ (\state \layer^{8}\mmaintextmax^2+\diminadapt)\state^5\layer^3}{n}} 
+
\state^7\Binvbd^2\Big(\suffmeasure(\scorefun)+\frac{1}{\diminadapt}\Big)
\Bigg),
\end{align}
}
where $\bigOtil$ hides polynomial factors in $(\log(\mmaintextmax \state \layer \Binvbd n), (\transbd)^{\mmaintextone})$.
\end{theorem}
See the proof of \Cref{theorem:CDM-Generalization-two-step} in~\Cref{sec:pf_theorem:CDM-Generalization-two-step}. The main step in the proof involves constructing transformers to approximate the conditional denoiser, similar to \Cref{theorem:CLIP-Generalization}. 

The estimation error bound has two terms. The first term, which scales as $n^{-1/2}$, comes from the approximation and generalization errors during the training of the conditional denoising function with the CLIP text representation fixed. The second term, which scales with $(\suffmeasure(\scorefun) + \diminadapt^{-1})$, is caused by the near-sufficiency of the CLIP representation. The term $\suffmeasure(\scorefun)$ can be controlled by the excess risk of CLIP training, as shown in \Cref{theorem:CLIP-Generalization}, while the term $\diminadapt^{-1}$ decreases as we increase the width of the adapter network. If the conditional denoising function $\TF_{\cdm}$ and the CLIP text representation $\Et$ are jointly trained (eliminating the need for $\Adapter$), the second term vanishes, as shown in Appendix~\ref{sec:joint_train_cdm}. 

By integrating over $t$ and using Girsanov's theorem, this estimation error bound can be converted into a sampling error bound for diffusion sampling, as illustrated in \Cref{cor:diffusion_sampling_error}. 
Additionally, we perform similar analyses for the sampling error of vision-language models in \Cref{sec:VLM_GHM}.

\section{Experiments}\label{sec:simulations}

We conduct experiments using transformer architectures to train CLIP encoders and downstream tasks for image-text distribution under JGHMs. 

\if\showfigure1
\begin{figure}[!ht]
\centering
\begin{minipage}{0.25\textwidth}
\includegraphics[width=\linewidth]{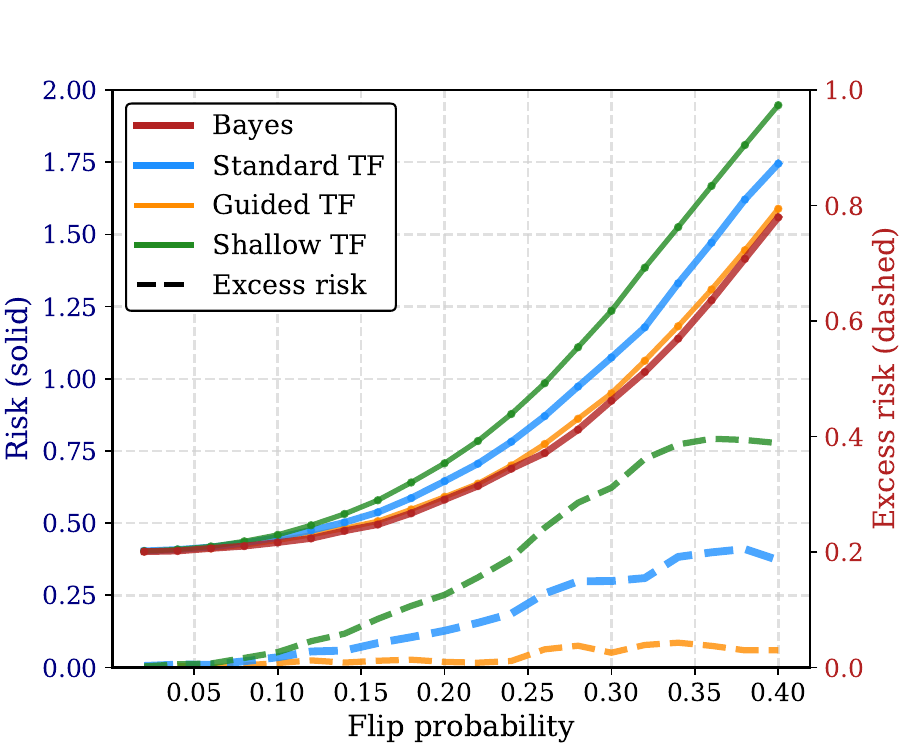}
\subcaption{ CLIP risk}
\label{fig:clip-train}
\end{minipage}
\hspace{-0.015\textwidth}
\begin{minipage}{0.25\textwidth}
\includegraphics[width=\linewidth]{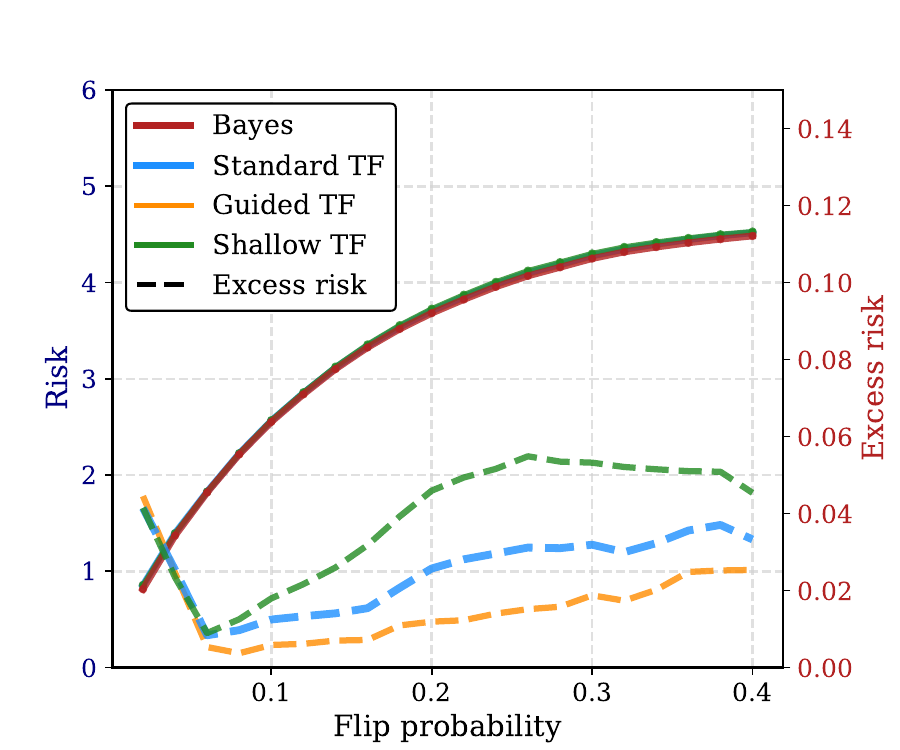}
\subcaption{ ZSC risk}
\label{fig:zsc-train}
\end{minipage}
\hspace{-0.015\textwidth}
\begin{minipage}{0.25\textwidth}
\includegraphics[width=\linewidth]{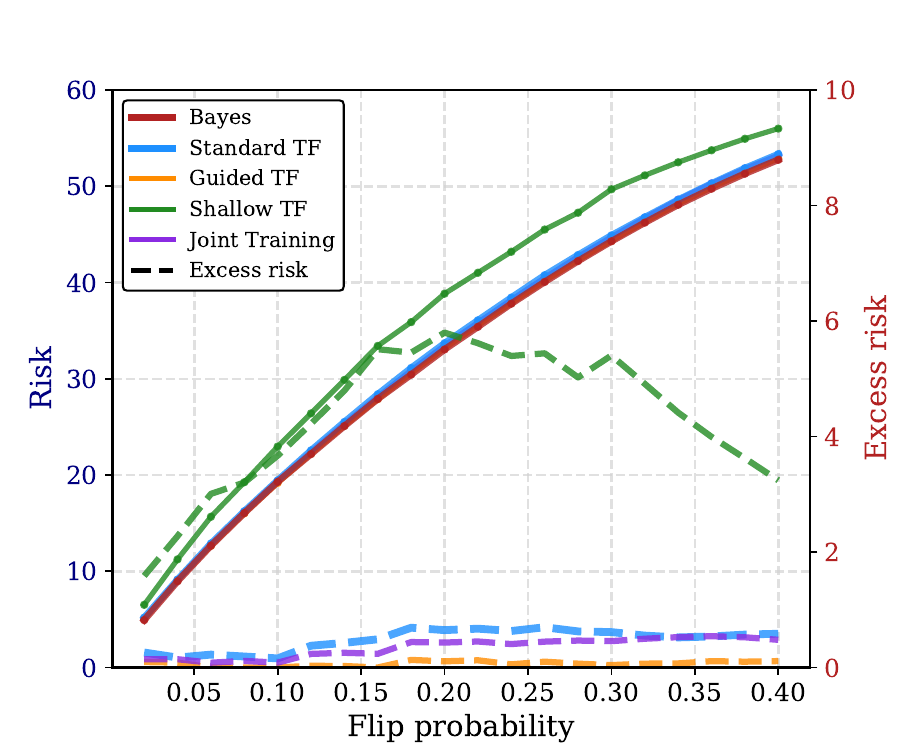}
\subcaption{ CDM risk}
\label{fig:cdm-train}
\end{minipage}
\hspace{-0.015\textwidth}
\begin{minipage}{0.25\textwidth}
\includegraphics[width=\linewidth]{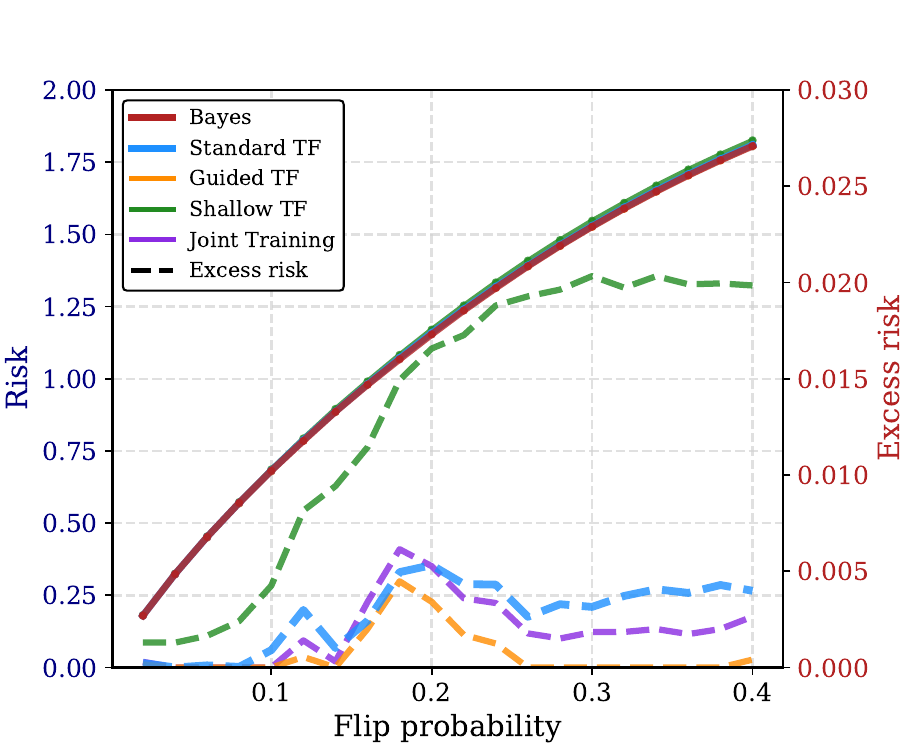}
\subcaption{ VLM risk}
\label{fig:vlm-train}
\end{minipage}
\caption{{ Risks (solid curves) and excess risks (dashed curves) as a function of the parameter $\pflip$ for CLIP training, ZSC, CDM, and VLM. The training setups for the different curves are described in \Cref{sec:simulations}. Across all setups, the excess risks of \GTF{} approach zero. The  risks of  \ATF{}  are close to the Bayes risk in ZSC, CDM, and VLM, demonstrating that CLIP representations can effectively adapt to downstream tasks.}}\label{fig:in-distribution}
\end{figure}
\fi

\paragraph{Training data distribution.} We sample the image and text data from the JGHM described in \Cref{sec:GHMs} with parameters $L=4$, $\cS=[10]$, and $\nchild_\square^{(\ell)} = 3$ for $\square \in \{ \image, \txt\}$ and all $\ell$, following the factorization assumption (\Cref{assumption:factorization}). The transition probabilities $(\psi_\root^{(0)}, \{ \psi^{(\ell)}_{\image,\iota},  \psi^{(\ell)}_{\txt,\iota}\}_{\iota \in [\state], \ell \in [L]})$ are randomly generated from a specific distribution using a fixed random seed (details provided in \Cref{sec:simu-detail-jghm}). These probabilities are governed by the parameter $\pflip \in [0, 1]$, which controls the conditional entropy of the leaf nodes $(\bx_{\image}, \bx_{\txt})$ given the root node $x_{\root}$. When $\pflip = 0$, $(\bx_{\image}, \bx_{\txt})$ are deterministic functions of $x_{\root}$, while $\pflip = 1$ results in high conditional entropy for $(\bx_{\image}, \bx_{\txt})$ given $x_{\root}$. As $\pflip$ increases, predicting $x_{\root}$ from $(\bx_{\image}, \bx_{\txt})$ becomes progressively more challenging. In our experiments, the values of $\pflip$ are chosen from the range $0.02$ to $0.4$ in increments of $0.02$. 

\paragraph{Training setup.} The CLIP encoders and conditional denoising functions are implemented using encoder transformers, while conditional next-token prediction functions (VLMs) are parameterized by decoder transformers. Detailed architectural specifications are provided in \Cref{sec:simu-detail-network}. For training the CLIP encoders, we consider three setups: (1) \ATF{}: A $5$-layer transformer trained using the standard CLIP loss. (2) \GTF{}: A $5$-layer transformer trained with the CLIP loss, supplemented by a guided loss encouraging the model to emulate the belief propagation algorithm (details in \Cref{sec:simu-detail-bp}). (3) \STF{}: A $1$-layer transformer trained using the standard CLIP loss. 

For CDMs and VLMs, we consider the following setups: (1) \ATF{}: The CLIP encoder trained under the \ATF{} setup is fixed, and a $9$-layer transformer is trained on top of it using a standard supervised loss. (2) \STF{}: The CLIP encoder trained under the \ATF{} setup is fixed, and a $1$-layer transformer is trained on top of it with a standard supervised loss. (3) \JT{}: Jointly train the CLIP encoder and the conditional denoiser/next-token predictor with a standard supervised loss. (4) \GTF{}: Jointly train the CLIP encoder and the conditional denoiser/next-token predictor with a supervised loss augmented by a guided loss. (5) \Bayes{}: The Bayes-optimal predictor. 

All models are trained using AdamW for $30,000$ steps, with each step using a fresh batch of size $128$. Details on network architectures (which could be different from architectures used in theorems), learning rates, and other hyperparameters are provided in \Cref{sec:simu-detail-adam}.

\paragraph{Experimental results.} \Cref{fig:in-distribution} shows the risk (solid curve) and excess risk (dashed curve) as functions of the parameter $\pflip$ across different setups: CLIP training (\Cref{fig:clip-train}), ZSC (\Cref{fig:zsc-train}), CDM (\Cref{fig:cdm-train}), and VLM (\Cref{fig:vlm-train}). 

Standard training of CLIP (\Cref{fig:clip-train}) exhibits a non-vanishing excess  risk, likely due to the training dynamics failing to converge to a global minimizer of the CLIP loss. Despite this excess risk in CLIP training, standard training (\ATF{}) results in small excess risks in ZSC, CDM, and VLM tasks (\Cref{fig:zsc-train,fig:cdm-train,fig:vlm-train}). This suggests that CLIP representations can effectively adapt to these downstream tasks, supporting our theoretical results, even when the conditions of our theory are not fully satisfied. 

Guided training (\GTF{}) for CLIP (\Cref{fig:clip-train}) significantly reduces excess risk to nearly zero, in line with our approximation theory. In the ZSC, CDM, and VLM setups, \GTF{} outperforms \ATF{} by a considerable margin, indicating that guided training promotes better convergence to the global minimizer of the CLIP loss. Across all settings, \ATF{} consistently outperforms \STF{} by a wide margin, as expected. This suggests that shallow networks are insufficient for approximating the Bayes predictor, which relies on the belief propagation algorithm. In the CDM and VLM setups (\Cref{fig:cdm-train,fig:vlm-train}), both sequential training (\ATF{}) and joint training (\JT{}) yield small excess risks, indicating that CLIP pre-training may not always be necessary in this simulated environment.
Further ablation studies are presented in \Cref{sec:ablations}.

\section{Conclusion}\label{sec:conclusion}

This paper presents a theoretical framework explaining the success of contrastive pre-training in multi-modal generative AI. It shows that near-minimizers of contrastive loss serve as approximate sufficient statistics, adaptable to diverse tasks like zero-shot classification and conditional diffusion models. The Joint Generative Hierarchical Model (JGHM) illustrates how transformers efficiently approximate functions via belief propagation, breaking the curse of dimensionality. These findings provide guarantees on the sample efficiency and generalization of contrastive pre-training, validated by numerical simulations.

Approximate sufficient statistics are central to this framework, providing a foundation for understanding contrastive pre-training. Future research could examine how this concept extends to other learning paradigms. Another promising avenue is exploring single-modal contrastive learning frameworks, where data augmentations serve as positive samples. Additionally, extending the JGHM to model more realistic generative processes for image and text distributions holds significant potential for further advancements. 

\if\showacknowledgement1
\section*{Acknowledgement}
We thank Xiangxiang Xu and Daniel Hsu for their insightful discussions. This project was supported by NSF grants DMS-2210827, CCF-2315725, CAREER DMS-2339904, ONR grant N00014-24-S-B001, DARPA HR001124S0029-AIQ-FP-003, JST ACT-X JP-MJAX23C4, an Amazon Research Award, a Google Research Scholar Award, an Okawa Foundation Research Grant, and an Nvidia Academic Grant Program Award. 
\fi

\bibliographystyle{alpha}
\bibliography{reference}

\newcommand{\etalchar}[1]{$^{#1}$}
\begin{thebibliography}{POVDO{\etalchar{+}}19}

\bibitem[ADL{\etalchar{+}}22]{alayrac2022flamingo}
Jean-Baptiste Alayrac, Jeff Donahue, Pauline Luc, Antoine Miech, Iain Barr,
  Yana Hasson, Karel Lenc, Arthur Mensch, Katherine Millican, Malcolm Reynolds,
  et~al.
\newblock Flamingo: a visual language model for few-shot learning.
\newblock {\em Advances in neural information processing systems},
  35:23716--23736, 2022.

\bibitem[AGKM21]{ash2021investigating}
Jordan~T Ash, Surbhi Goel, Akshay Krishnamurthy, and Dipendra Misra.
\newblock Investigating the role of negatives in contrastive representation
  learning.
\newblock {\em arXiv preprint arXiv:2106.09943}, 2021.

\bibitem[AZL23]{allen2023physics}
Zeyuan Allen-Zhu and Yuanzhi Li.
\newblock Physics of language models: Part 1, context-free grammar.
\newblock {\em arXiv preprint arXiv:2305.13673}, 2023.

\bibitem[Bac17]{bach2017breaking}
Francis Bach.
\newblock Breaking the curse of dimensionality with convex neural networks.
\newblock {\em The Journal of Machine Learning Research}, 18(1):629--681, 2017.

\bibitem[Bar93]{barron1993universal}
Andrew~R Barron.
\newblock Universal approximation bounds for superpositions of a sigmoidal
  function.
\newblock {\em IEEE Transactions on Information theory}, 39(3):930--945, 1993.

\bibitem[BBC{\etalchar{+}}23]{brohan2023rt}
Anthony Brohan, Noah Brown, Justice Carbajal, Yevgen Chebotar, Xi~Chen,
  Krzysztof Choromanski, Tianli Ding, Danny Driess, Avinava Dubey, Chelsea
  Finn, et~al.
\newblock Rt-2: Vision-language-action models transfer web knowledge to robotic
  control.
\newblock {\em arXiv preprint arXiv:2307.15818}, 2023.

\bibitem[BCW{\etalchar{+}}24]{bai2024transformers}
Yu~Bai, Fan Chen, Huan Wang, Caiming Xiong, and Song Mei.
\newblock Transformers as statisticians: Provable in-context learning with
  in-context algorithm selection.
\newblock {\em Advances in neural information processing systems}, 36, 2024.

\bibitem[BHB19]{bachman2019learning}
Philip Bachman, R~Devon Hjelm, and William Buchwalter.
\newblock Learning representations by maximizing mutual information across
  views.
\newblock {\em Advances in neural information processing systems}, 32, 2019.

\bibitem[BS16]{bun2016concentrated}
Mark Bun and Thomas Steinke.
\newblock Concentrated differential privacy: Simplifications, extensions, and
  lower bounds.
\newblock In {\em Theory of cryptography conference}, pages 635--658. Springer,
  2016.

\bibitem[CDLG23]{chen2023understanding}
Zixiang Chen, Yihe Deng, Yuanzhi Li, and Quanquan Gu.
\newblock Understanding transferable representation learning and zero-shot
  transfer in clip.
\newblock {\em arXiv preprint arXiv:2310.00927}, 2023.

\bibitem[CKNH20]{chen2020simple}
Ting Chen, Simon Kornblith, Mohammad Norouzi, and Geoffrey Hinton.
\newblock A simple framework for contrastive learning of visual
  representations.
\newblock In {\em International conference on machine learning}, pages
  1597--1607. PMLR, 2020.

\bibitem[CSDS21]{changpinyo2021cc12m}
Soravit Changpinyo, Piyush Sharma, Nan Ding, and Radu Soricut.
\newblock {Conceptual 12M}: Pushing web-scale image-text pre-training to
  recognize long-tail visual concepts.
\newblock In {\em Proceedings of the IEEE/CVF Conference on Computer Vision and
  Pattern Recognition (CVPR)}, 2021.

\bibitem[CW24]{cagnetta2024towards}
Francesco Cagnetta and Matthieu Wyart.
\newblock Towards a theory of how the structure of language is acquired by deep
  neural networks.
\newblock {\em arXiv preprint arXiv:2406.00048}, 2024.

\bibitem[CZG{\etalchar{+}}20]{chen2020neural}
Yanzhi Chen, Dinghuai Zhang, Michael Gutmann, Aaron Courville, and Zhanxing
  Zhu.
\newblock Neural approximate sufficient statistics for implicit models.
\newblock {\em arXiv preprint arXiv:2010.10079}, 2020.

\bibitem[DDS{\etalchar{+}}09]{deng2009imagenet}
Jia Deng, Wei Dong, Richard Socher, Li-Jia Li, Kai Li, and Li~Fei-Fei.
\newblock Imagenet: A large-scale hierarchical image database.
\newblock {\em Proceedings of the IEEE Conference on Computer Vision and
  Pattern Recognition (CVPR)}, pages 248--255, 2009.

\bibitem[EAMS22]{el2022sampling}
Ahmed El~Alaoui, Andrea Montanari, and Mark Sellke.
\newblock Sampling from the sherrington-kirkpatrick gibbs measure via
  algorithmic stochastic localization.
\newblock In {\em 2022 IEEE 63rd Annual Symposium on Foundations of Computer
  Science (FOCS)}, pages 323--334. IEEE, 2022.

\bibitem[EKB{\etalchar{+}}24]{esser2024scaling}
Patrick Esser, Sumith Kulal, Andreas Blattmann, Rahim Entezari, Jonas
  M{\"u}ller, Harry Saini, Yam Levi, Dominik Lorenz, Axel Sauer, Frederic
  Boesel, et~al.
\newblock Scaling rectified flow transformers for high-resolution image
  synthesis.
\newblock In {\em Forty-first International Conference on Machine Learning},
  2024.

\bibitem[Eld13]{eldan2013thin}
Ronen Eldan.
\newblock Thin shell implies spectral gap up to polylog via a stochastic
  localization scheme.
\newblock {\em Geometric and Functional Analysis}, 23(2):532--569, 2013.

\bibitem[Fis22]{fisher1922mathematical}
Ronald~A Fisher.
\newblock On the mathematical foundations of theoretical statistics.
\newblock {\em Philosophical transactions of the Royal Society of London.
  Series A, containing papers of a mathematical or physical character},
  222(594-604):309--368, 1922.

\bibitem[GBMMS24]{garnier2024transformers}
Jerome Garnier-Brun, Marc M{\'e}zard, Emanuele Moscato, and Luca Saglietti.
\newblock How transformers learn structured data: insights from hierarchical
  filtering.
\newblock {\em arXiv preprint arXiv:2408.15138}, 2024.

\bibitem[GH10]{gutmann2010noise}
Michael Gutmann and Aapo Hyv{\"a}rinen.
\newblock Noise-contrastive estimation: A new estimation principle for
  unnormalized statistical models.
\newblock In {\em Proceedings of the thirteenth international conference on
  artificial intelligence and statistics}, pages 297--304. JMLR Workshop and
  Conference Proceedings, 2010.

\bibitem[GRS{\etalchar{+}}23]{giannou2023looped}
Angeliki Giannou, Shashank Rajput, Jy-yong Sohn, Kangwook Lee, Jason~D Lee, and
  Dimitris Papailiopoulos.
\newblock Looped transformers as programmable computers.
\newblock {\em arXiv preprint arXiv:2301.13196}, 2023.

\bibitem[GSA{\etalchar{+}}20]{grill2020bootstrap}
Jean-Bastien Grill, Florian Strub, Florent Altch{\'e}, Corentin Tallec, Pierre
  Richemond, Elena Buchatskaya, Carl Doersch, Bernardo Avila~Pires, Zhaohan
  Guo, Mohammad Gheshlaghi~Azar, et~al.
\newblock Bootstrap your own latent-a new approach to self-supervised learning.
\newblock {\em Advances in neural information processing systems},
  33:21271--21284, 2020.

\bibitem[HFLM{\etalchar{+}}18]{hjelm2018learning}
R~Devon Hjelm, Alex Fedorov, Samuel Lavoie-Marchildon, Karan Grewal, Phil
  Bachman, Adam Trischler, and Yoshua Bengio.
\newblock Learning deep representations by mutual information estimation and
  maximization.
\newblock {\em arXiv preprint arXiv:1808.06670}, 2018.

\bibitem[HFW{\etalchar{+}}20]{he2020momentum}
Kaiming He, Haoqi Fan, Yuxin Wu, Saining Xie, and Ross Girshick.
\newblock Momentum contrast for unsupervised visual representation learning.
\newblock In {\em Proceedings of the IEEE/CVF conference on computer vision and
  pattern recognition}, pages 9729--9738, 2020.

\bibitem[HHG{\etalchar{+}}20]{hewitt2020rnns}
John Hewitt, Michael Hahn, Surya Ganguli, Percy Liang, and Christopher~D
  Manning.
\newblock Rnns can generate bounded hierarchical languages with optimal memory.
\newblock {\em arXiv preprint arXiv:2010.07515}, 2020.

\bibitem[HJA20]{ho2020denoising}
Jonathan Ho, Ajay Jain, and Pieter Abbeel.
\newblock Denoising diffusion probabilistic models.
\newblock {\em Advances in Neural Information Processing Systems},
  33:6840--6851, 2020.

\bibitem[HWGM21]{haochen2021provable}
Jeff~Z HaoChen, Colin Wei, Adrien Gaidon, and Tengyu Ma.
\newblock Provable guarantees for self-supervised deep learning with spectral
  contrastive loss.
\newblock {\em Advances in Neural Information Processing Systems},
  34:5000--5011, 2021.

\bibitem[HYZJ21]{huang2021towards}
Weiran Huang, Mingyang Yi, Xuyang Zhao, and Zihao Jiang.
\newblock Towards the generalization of contrastive self-supervised learning.
\newblock {\em arXiv preprint arXiv:2111.00743}, 2021.

\bibitem[JYX{\etalchar{+}}21]{jia2021scaling}
Chao Jia, Yinfei Yang, Ye~Xia, Yi-Ting Chen, Zarana Parekh, Hieu Pham, Quoc Le,
  Yun-Hsuan Sung, Zhen Li, and Tom Duerig.
\newblock Scaling up visual and vision-language representation learning with
  noisy text supervision.
\newblock In {\em International conference on machine learning}, pages
  4904--4916. PMLR, 2021.

\bibitem[Kee10]{keener2010theoretical}
Robert~W Keener.
\newblock {\em Theoretical statistics: Topics for a core course}.
\newblock Springer Science \& Business Media, 2010.

\bibitem[KGMS23]{kadkhodaie2023learning}
Zahra Kadkhodaie, Florentin Guth, St{\'e}phane Mallat, and Eero~P Simoncelli.
\newblock Learning multi-scale local conditional probability models of images.
\newblock {\em arXiv preprint arXiv:2303.02984}, 2023.

\bibitem[KGSM23]{kadkhodaie2023generalization}
Zahra Kadkhodaie, Florentin Guth, Eero~P Simoncelli, and St{\'e}phane Mallat.
\newblock Generalization in diffusion models arises from geometry-adaptive
  harmonic representation.
\newblock {\em arXiv preprint arXiv:2310.02557}, 2023.

\bibitem[Kin14]{kingma2014adam}
Diederik~P Kingma.
\newblock Adam: A method for stochastic optimization.
\newblock {\em arXiv preprint arXiv:1412.6980}, 2014.

\bibitem[KSCE24]{karan2024unrolled}
Aayush Karan, Kulin Shah, Sitan Chen, and Yonina~C Eldar.
\newblock Unrolled denoising networks provably learn optimal bayesian
  inference.
\newblock {\em arXiv preprint arXiv:2409.12947}, 2024.

\bibitem[LAG{\etalchar{+}}22]{liu2022transformers}
Bingbin Liu, Jordan~T Ash, Surbhi Goel, Akshay Krishnamurthy, and Cyril Zhang.
\newblock Transformers learn shortcuts to automata.
\newblock {\em arXiv preprint arXiv:2210.10749}, 2022.

\bibitem[LBM23]{lin2023transformers}
Licong Lin, Yu~Bai, and Song Mei.
\newblock Transformers as decision makers: Provable in-context reinforcement
  learning via supervised pretraining.
\newblock {\em arXiv preprint arXiv:2310.08566}, 2023.

\bibitem[Lin88]{linsker1988self}
Ralph Linsker.
\newblock Self-organization in a perceptual network.
\newblock {\em Computer}, 21(3):105--117, 1988.

\bibitem[LLLL24]{liu2024improved}
Haotian Liu, Chunyuan Li, Yuheng Li, and Yong~Jae Lee.
\newblock Improved baselines with visual instruction tuning.
\newblock In {\em Proceedings of the IEEE/CVF Conference on Computer Vision and
  Pattern Recognition}, pages 26296--26306, 2024.

\bibitem[LLSZ21]{lee2021predicting}
Jason~D Lee, Qi~Lei, Nikunj Saunshi, and Jiacheng Zhuo.
\newblock Predicting what you already know helps: Provable self-supervised
  learning.
\newblock {\em Advances in Neural Information Processing Systems}, 34:309--323,
  2021.

\bibitem[LLWL24]{liu2024visual}
Haotian Liu, Chunyuan Li, Qingyang Wu, and Yong~Jae Lee.
\newblock Visual instruction tuning.
\newblock {\em Advances in neural information processing systems}, 36, 2024.

\bibitem[LLXH22]{li2022blip}
Junnan Li, Dongxu Li, Caiming Xiong, and Steven Hoi.
\newblock Blip: Bootstrapping language-image pre-training for unified
  vision-language understanding and generation.
\newblock In {\em International conference on machine learning}, pages
  12888--12900. PMLR, 2022.

\bibitem[Los17]{loshchilov2017decoupled}
I~Loshchilov.
\newblock Decoupled weight decay regularization.
\newblock {\em arXiv preprint arXiv:1711.05101}, 2017.

\bibitem[LZS{\etalchar{+}}24]{lu2024f}
Yiwei Lu, Guojun Zhang, Sun Sun, Hongyu Guo, and Yaoliang Yu.
\newblock $ f $-micl: Understanding and generalizing infonce-based contrastive
  learning.
\newblock {\em arXiv preprint arXiv:2402.10150}, 2024.

\bibitem[Mei24]{mei2024u}
Song Mei.
\newblock U-nets as belief propagation: Efficient classification, denoising,
  and diffusion in generative hierarchical models.
\newblock {\em arXiv preprint arXiv:2404.18444}, 2024.

\bibitem[MLLR23]{marwah2023neural}
Tanya Marwah, Zachary~Chase Lipton, Jianfeng Lu, and Andrej Risteski.
\newblock Neural network approximations of pdes beyond linearity: A
  representational perspective.
\newblock In {\em International Conference on Machine Learning}, pages
  24139--24172. PMLR, 2023.

\bibitem[MLR21]{marwah2021parametric}
Tanya Marwah, Zachary Lipton, and Andrej Risteski.
\newblock Parametric complexity bounds for approximating pdes with neural
  networks.
\newblock {\em Advances in Neural Information Processing Systems},
  34:15044--15055, 2021.

\bibitem[MM09]{mezard2009information}
Marc Mezard and Andrea Montanari.
\newblock {\em Information, physics, and computation}.
\newblock Oxford University Press, 2009.

\bibitem[Mon23]{montanari2023sampling}
Andrea Montanari.
\newblock Sampling, diffusions, and stochastic localization.
\newblock {\em arXiv preprint arXiv:2305.10690}, 2023.

\bibitem[Mos16]{mossel2016deep}
Elchanan Mossel.
\newblock Deep learning and hierarchal generative models.
\newblock {\em arXiv preprint arXiv:1612.09057}, 2016.

\bibitem[MW23]{mei2023deep}
Song Mei and Yuchen Wu.
\newblock Deep networks as denoising algorithms: Sample-efficient learning of
  diffusion models in high-dimensional graphical models.
\newblock {\em arXiv preprint arXiv:2309.11420}, 2023.

\bibitem[Ney36]{neyman1936teorema}
Jerzy Neyman.
\newblock {\em Su un teorema concernente le cosiddette statistiche
  sufficienti}.
\newblock Istituto Italiano degli Attuari, 1936.

\bibitem[NGD{\etalchar{+}}23]{nakada2023understanding}
Ryumei Nakada, Halil~Ibrahim Gulluk, Zhun Deng, Wenlong Ji, James Zou, and
  Linjun Zhang.
\newblock Understanding multimodal contrastive learning and incorporating
  unpaired data.
\newblock In {\em International Conference on Artificial Intelligence and
  Statistics}, pages 4348--4380. PMLR, 2023.

\bibitem[NI20]{nakada2020adaptive}
Ryumei Nakada and Masaaki Imaizumi.
\newblock Adaptive approximation and generalization of deep neural network with
  intrinsic dimensionality.
\newblock {\em Journal of Machine Learning Research}, 21(174):1--38, 2020.

\bibitem[OLV18]{oord2018representation}
Aaron van~den Oord, Yazhe Li, and Oriol Vinyals.
\newblock Representation learning with contrastive predictive coding.
\newblock {\em arXiv preprint arXiv:1807.03748}, 2018.

\bibitem[Ope22]{dalle2}
OpenAI.
\newblock Openai announces dall-e 2.
\newblock {\em \url{https://openai.com/index/dall-e-2/}}, 2022.

\bibitem[Ope23]{gpt4v}
OpenAI.
\newblock Openai announces gpt 4v.
\newblock {\em \url{https://openai.com/index/gpt-4v-system-card/}}, 2023.

\bibitem[PCT{\etalchar{+}}23]{petrini2023deep}
Leonardo Petrini, Francesco Cagnetta, Umberto~M Tomasini, Alessandro Favero,
  and Matthieu Wyart.
\newblock How deep neural networks learn compositional data: The random
  hierarchy model.
\newblock {\em arXiv preprint arXiv:2307.02129}, 2023.

\bibitem[Pea82]{pearl2022reverend}
Judea Pearl.
\newblock Reverend bayes on inference engines: A distributed hierarchical
  approach.
\newblock In {\em Probabilistic and Causal Inference: The Works of Judea
  Pearl}, pages 129--138. 1982.

\bibitem[POVDO{\etalchar{+}}19]{poole2019variational}
Ben Poole, Sherjil Ozair, Aaron Van Den~Oord, Alex Alemi, and George Tucker.
\newblock On variational bounds of mutual information.
\newblock In {\em International Conference on Machine Learning}, pages
  5171--5180. PMLR, 2019.

\bibitem[RBL{\etalchar{+}}22]{rombach2022high}
Robin Rombach, Andreas Blattmann, Dominik Lorenz, Patrick Esser, and Bj{\"o}rn
  Ommer.
\newblock High-resolution image synthesis with latent diffusion models.
\newblock In {\em Proceedings of the IEEE/CVF conference on computer vision and
  pattern recognition}, pages 10684--10695, 2022.

\bibitem[RDN{\etalchar{+}}22]{ramesh2022hierarchical}
Aditya Ramesh, Prafulla Dhariwal, Alex Nichol, Casey Chu, and Mark Chen.
\newblock Hierarchical text-conditional image generation with clip latents.
\newblock {\em arXiv preprint arXiv:2204.06125}, 1(2):3, 2022.

\bibitem[RKH{\etalchar{+}}21]{radford2021learning}
Alec Radford, Jong~Wook Kim, Chris Hallacy, Aditya Ramesh, Gabriel Goh,
  Sandhini Agarwal, Girish Sastry, Amanda Askell, Pamela Mishkin, Jack Clark,
  et~al.
\newblock Learning transferable visual models from natural language
  supervision.
\newblock In {\em International conference on machine learning}, pages
  8748--8763. PMLR, 2021.

\bibitem[SCL{\etalchar{+}}23]{shi2023trade}
Zhenmei Shi, Jiefeng Chen, Kunyang Li, Jayaram Raghuram, Xi~Wu, Yingyu Liang,
  and Somesh Jha.
\newblock The trade-off between universality and label efficiency of
  representations from contrastive learning.
\newblock {\em arXiv preprint arXiv:2303.00106}, 2023.

\bibitem[SCS{\etalchar{+}}22]{saharia2022photorealistic}
Chitwan Saharia, William Chan, Saurabh Saxena, Lala Li, Jay Whang, Emily~L
  Denton, Kamyar Ghasemipour, Raphael Gontijo~Lopes, Burcu Karagol~Ayan, Tim
  Salimans, et~al.
\newblock Photorealistic text-to-image diffusion models with deep language
  understanding.
\newblock {\em Advances in Neural Information Processing Systems},
  35:36479--36494, 2022.

\bibitem[SDS18]{sharma2018conceptual_captions}
Piyush Sharma, Nan Ding, and Radu Soricut.
\newblock Conceptual captions: A cleaned, hypernym-predicted dataset of
  image–text pairs.
\newblock In {\em Proceedings of the 56th Annual Meeting of the Association for
  Computational Linguistics (ACL)}, 2018.

\bibitem[SDWMG15]{sohl2015deep}
Jascha Sohl-Dickstein, Eric Weiss, Niru Maheswaranathan, and Surya Ganguli.
\newblock Deep unsupervised learning using nonequilibrium thermodynamics.
\newblock In {\em International Conference on Machine Learning}, pages
  2256--2265. PMLR, 2015.

\bibitem[SE19]{song2019generative}
Yang Song and Stefano Ermon.
\newblock Generative modeling by estimating gradients of the data distribution.
\newblock {\em Advances in neural information processing systems}, 32, 2019.

\bibitem[SFLW24]{sclocchi2024probing}
Antonio Sclocchi, Alessandro Favero, Noam~Itzhak Levi, and Matthieu Wyart.
\newblock Probing the latent hierarchical structure of data via diffusion
  models.
\newblock {\em arXiv preprint arXiv:2410.13770}, 2024.

\bibitem[SFW24]{sclocchi2024phase}
Antonio Sclocchi, Alessandro Favero, and Matthieu Wyart.
\newblock A phase transition in diffusion models reveals the hierarchical
  nature of data.
\newblock {\em arXiv preprint arXiv:2402.16991}, 2024.

\bibitem[SH20]{schmidt2020nonparametric}
Johannes Schmidt-Hieber.
\newblock Nonparametric regression using deep neural networks with relu
  activation function.
\newblock 2020.

\bibitem[Soh16]{sohn2016improved}
Kihyuk Sohn.
\newblock Improved deep metric learning with multi-class n-pair loss objective.
\newblock {\em Advances in neural information processing systems}, 29, 2016.

\bibitem[SPA{\etalchar{+}}19]{saunshi2019theoretical}
Nikunj Saunshi, Orestis Plevrakis, Sanjeev Arora, Mikhail Khodak, and
  Hrishikesh Khandeparkar.
\newblock A theoretical analysis of contrastive unsupervised representation
  learning.
\newblock In {\em International Conference on Machine Learning}, pages
  5628--5637. PMLR, 2019.

\bibitem[SSDK{\etalchar{+}}20]{song2020score}
Yang Song, Jascha Sohl-Dickstein, Diederik~P Kingma, Abhishek Kumar, Stefano
  Ermon, and Ben Poole.
\newblock Score-based generative modeling through stochastic differential
  equations.
\newblock {\em arXiv preprint arXiv:2011.13456}, 2020.

\bibitem[Suz18]{suzuki2018adaptivity}
Taiji Suzuki.
\newblock Adaptivity of deep relu network for learning in besov and mixed
  smooth besov spaces: optimal rate and curse of dimensionality.
\newblock {\em arXiv preprint arXiv:1810.08033}, 2018.

\bibitem[Tel16]{telgarsky2016benefits}
Matus Telgarsky.
\newblock Benefits of depth in neural networks.
\newblock In {\em Conference on learning theory}, pages 1517--1539. PMLR, 2016.

\bibitem[TKH21a]{tosh2021contrastive1}
Christopher Tosh, Akshay Krishnamurthy, and Daniel Hsu.
\newblock Contrastive estimation reveals topic posterior information to linear
  models.
\newblock {\em Journal of Machine Learning Research}, 22(281):1--31, 2021.

\bibitem[TKH21b]{tosh2021contrastive2}
Christopher Tosh, Akshay Krishnamurthy, and Daniel Hsu.
\newblock Contrastive learning, multi-view redundancy, and linear models.
\newblock In {\em Algorithmic Learning Theory}, pages 1179--1206. PMLR, 2021.

\bibitem[TKI20]{tian2020contrastive}
Yonglong Tian, Dilip Krishnan, and Phillip Isola.
\newblock Contrastive multiview coding.
\newblock In {\em Computer Vision--ECCV 2020: 16th European Conference,
  Glasgow, UK, August 23--28, 2020, Proceedings, Part XI 16}, pages 776--794.
  Springer, 2020.

\bibitem[TW24]{tomasini2024deep}
Umberto Tomasini and Matthieu Wyart.
\newblock How deep networks learn sparse and hierarchical data: the sparse
  random hierarchy model.
\newblock {\em arXiv preprint arXiv:2404.10727}, 2024.

\bibitem[TYCG20]{tian2020understanding}
Yuandong Tian, Lantao Yu, Xinlei Chen, and Surya Ganguli.
\newblock Understanding self-supervised learning with dual deep networks.
\newblock {\em arXiv preprint arXiv:2010.00578}, 2020.

\bibitem[UST{\etalchar{+}}24]{uesaka2024understanding}
Toshimitsu Uesaka, Taiji Suzuki, Yuhta Takida, Chieh-Hsin Lai, Naoki Murata,
  and Yuki Mitsufuji.
\newblock Understanding multimodal contrastive learning through pointwise
  mutual information.
\newblock {\em arXiv preprint arXiv:2404.19228}, 2024.

\bibitem[Vas17]{vaswani2017attention}
A~Vaswani.
\newblock Attention is all you need.
\newblock {\em Advances in Neural Information Processing Systems}, 2017.

\bibitem[Ver18]{vershynin2018high}
Roman Vershynin.
\newblock {\em High-dimensional probability: An introduction with applications
  in data science}, volume~47.
\newblock Cambridge university press, 2018.

\bibitem[Wai19]{wainwright_2019}
Martin~J. Wainwright.
\newblock {\em High-Dimensional Statistics: A Non-Asymptotic Viewpoint}.
\newblock Cambridge Series in Statistical and Probabilistic Mathematics.
  Cambridge University Press, 2019.

\bibitem[WCM22]{wei2022statistically}
Colin Wei, Yining Chen, and Tengyu Ma.
\newblock Statistically meaningful approximation: a case study on approximating
  turing machines with transformers.
\newblock {\em Advances in Neural Information Processing Systems},
  35:12071--12083, 2022.

\bibitem[WI20]{wang2020understanding}
Tongzhou Wang and Phillip Isola.
\newblock Understanding contrastive representation learning through alignment
  and uniformity on the hypersphere.
\newblock In {\em International conference on machine learning}, pages
  9929--9939. PMLR, 2020.

\bibitem[WJ{\etalchar{+}}08]{wainwright2008graphical}
Martin~J Wainwright, Michael~I Jordan, et~al.
\newblock Graphical models, exponential families, and variational inference.
\newblock {\em Foundations and Trends{\textregistered} in Machine Learning},
  1(1--2):1--305, 2008.

\bibitem[WL21]{wen2021toward}
Zixin Wen and Yuanzhi Li.
\newblock Toward understanding the feature learning process of self-supervised
  contrastive learning.
\newblock In {\em International Conference on Machine Learning}, pages
  11112--11122. PMLR, 2021.

\bibitem[WZW{\etalchar{+}}22]{wang2022chaos}
Yifei Wang, Qi~Zhang, Yisen Wang, Jiansheng Yang, and Zhouchen Lin.
\newblock Chaos is a ladder: A new theoretical understanding of contrastive
  learning via augmentation overlap.
\newblock {\em arXiv preprint arXiv:2203.13457}, 2022.

\bibitem[XZ24]{xu2024dependence}
Xiangxiang Xu and Lizhong Zheng.
\newblock Dependence induced representations.
\newblock In {\em 2024 60th Annual Allerton Conference on Communication,
  Control, and Computing}, pages 1--8. IEEE, 2024.

\bibitem[YPPN21]{yao2021self}
Shunyu Yao, Binghui Peng, Christos Papadimitriou, and Karthik Narasimhan.
\newblock Self-attention networks can process bounded hierarchical languages.
\newblock {\em arXiv preprint arXiv:2105.11115}, 2021.

\bibitem[YZAS21]{yan2021videogpt}
Wilson Yan, Yunzhi Zhang, Pieter Abbeel, and Aravind Srinivas.
\newblock Videogpt: Video generation using vq-vae and transformers.
\newblock {\em arXiv preprint arXiv:2104.10157}, 2021.

\bibitem[ZLZ{\etalchar{+}}23a]{zhang2023speechgpt}
Dong Zhang, Shimin Li, Xin Zhang, Jun Zhan, Pengyu Wang, Yaqian Zhou, and
  Xipeng Qiu.
\newblock Speechgpt: Empowering large language models with intrinsic
  cross-modal conversational abilities.
\newblock {\em arXiv preprint arXiv:2305.11000}, 2023.

\bibitem[ZLZ{\etalchar{+}}23b]{zhao2023bubogpt}
Yang Zhao, Zhijie Lin, Daquan Zhou, Zilong Huang, Jiashi Feng, and Bingyi Kang.
\newblock Bubogpt: Enabling visual grounding in multi-modal llms.
\newblock {\em arXiv preprint arXiv:2307.08581}, 2023.

\bibitem[ZPGA23]{zhao2023transformers}
Haoyu Zhao, Abhishek Panigrahi, Rong Ge, and Sanjeev Arora.
\newblock Do transformers parse while predicting the masked word?
\newblock {\em arXiv preprint arXiv:2303.08117}, 2023.

\bibitem[ZSS{\etalchar{+}}21]{zimmermann2021contrastive}
Roland~S Zimmermann, Yash Sharma, Steffen Schneider, Matthias Bethge, and
  Wieland Brendel.
\newblock Contrastive learning inverts the data generating process.
\newblock In {\em International Conference on Machine Learning}, pages
  12979--12990. PMLR, 2021.

\end{thebibliography}

\clearpage

\if\showappendix1
\appendix


\begin{center}
    {\Large \bf Appendix }
    \end{center}

\tableofcontents

\addtocontents{toc}{\protect\setcounter{tocdepth}{2}}

\if\showprelim1
\section{Background on CLIP, ZSC, CDM, and VLM}\label{sec:prelim}

\paragraph{Contrastive Language-Image Pre-Training (CLIP) and Zero-Shot Classification (ZSC).} CLIP \cite{radford2021learning} trains two transformer-based neural network encoders—one for images and one for text—using an extensive dataset of 400 million image-caption pairs sourced from the internet. The training objective is based on the principle that representations of paired images and captions should be similar, while representations of non-paired images and captions should be dissimilar. Let $\Ei: \cX_i \to \R^\dimclip$ denote the image encoder and $\Et: \cXt \to \R^\dimclip$ the text encoder, both parameterized by neural networks. Given a user-defined similarity score function, $\scorelink: \R^\dimclip \times \R^\dimclip \to \R$, and available image-caption pairs $(\xi^{(i)}, \xt^{(i)})_{i \in [n]} \subseteq \cXi \times \cXt$, CLIP trains the encoders $(\Ei, \Et)$ by maximizing $\scorelink(\Ei(\xi^{(i)}), \Et(\xt^{(i)}))$ for paired images and captions, while minimizing $\scorelink(\Ei(\xi^{(i)}), \Et(\xt^{(j)}))$ for non-paired instances, as illustrated in \Cref{fig:CLIP_loss_illustration}. This alignment is achieved by minimizing the InfoNCE loss \cite{oord2018representation}, defined in Eq.~(\ref{eqn:CLIP_risk}), a cross-entropy loss that distinguishes paired image-caption from non-paired ones. 

\cite{radford2021learning} showed that CLIP’s learned representations achieve strong performance on downstream image classification tasks, such as ImageNet, in a zero-shot manner. In a zero-shot classification (ZSC) task with images and labels $(\xi, y) \in \cX_i \times \cY$, each label $y \in \cY$ is converted into a text prompt $\xt(y)$ through a mapping $\xt: \cY \to \cXt$. For instance, if $y$ is ``dog'', then $\xt(y)$ becomes ``A photo of a dog''. Given any new image $\xi$ from the ImageNet dataset, the ZSC prediction selects the label that maximizes similarity with the image representation, $\hat y = \arg\max_{y \in \cY} \scorelink(\Ei(\xi), \Et(\xt(y)))$, where $(\Ei, \Et)$ are the trained CLIP encoders. This approach is illustrated in \Cref{fig:CLIP_zero_shot_illustration}. Remarkably, \cite{radford2021learning} demonstrated that ZSC with CLIP encoders matches the accuracy of the original ResNet-50 on ImageNet, without using any of its 1.28 million training examples, achieving surprisingly high performance. In this paper, we aim to provide a theoretical explanation for why CLIP encoders perform so well on the ZSC task.

\paragraph{Vision-Language Models (VLMs).} Vision-language models are generative models that process both image and text inputs to generate text outputs. Notable VLMs include BLIP \cite{li2022blip}, Flamingo \cite{alayrac2022flamingo}, and Llava \cite{liu2024visual, liu2024improved}, with applications spanning image captioning, visual question answering, and cross-modal retrieval. VLMs are typically based on transformer architectures that incorporate the CLIP image representations, denoted as $\Ei(\xi)$, as input tokens. This is formalized as $\{ \mu(x_{\txt, i}| \Ei(\xi), x_{\txt, 1:i-1}) \}_{i \in [\dim]}$, a sequence of distributions over text tokens $x_{\txt, i}$ conditioned on the image embedding $\Ei(\xi)$ and previous text tokens $x_{\txt, 1:i-1}$. VLMs are trained on large datasets of image-text pairs $(\xi^{(j)}, \xt^{(j)})_{j \in [n]}$ with a next-token prediction loss, defined as 
\begin{equation}
\textstyle \hat \mu = \arg\min_{\mu} \Big\{\what \risk_{\vlm}(\mu) = -\frac{1}{n} \sum_{j \in [n]} \sum_{i \in [\dim]} \log \mu(x_{\txt, i}^{(j)}| \Ei(\xi^{(j)}), x_{\txt, 1:i-1}^{(j)}) \Big\}.
\end{equation}
After training, given a new image $\xi$ and a text prompt $x_{\txt, 1:i}$, the VLM generates subsequent tokens by sequentially sampling $x_{\txt, i+1} \sim \hat \mu(\cdot | \Ei(\xi), x_{\txt, 1:i})$ for each $i \in [\dim]$. An illustration of the VLM framework is shown in \Cref{fig:VLM_illustration}. 

Assuming infinite samples and unlimited representational power of the neural network, theoretical results suggest that the generated text $\xt$ produced by VLMs follows the conditional distribution $\P( \xt | \Ei(\xi))$. In this paper, we investigate: (1) the conditions under which VLMs can be effectively learned with finite network capacity and finite samples, and (2) how closely the conditional distribution of the generated text approximates the true conditional distribution $\P(\xt | \xi)$.

\if\showfigure1
\begin{figure}
\begin{subfigure}[t]{0.48\textwidth}
\centering 
\caption{ Contrastive Language-Image Pre-training.}\label{fig:CLIP_loss_illustration}
\includegraphics[width=\linewidth]{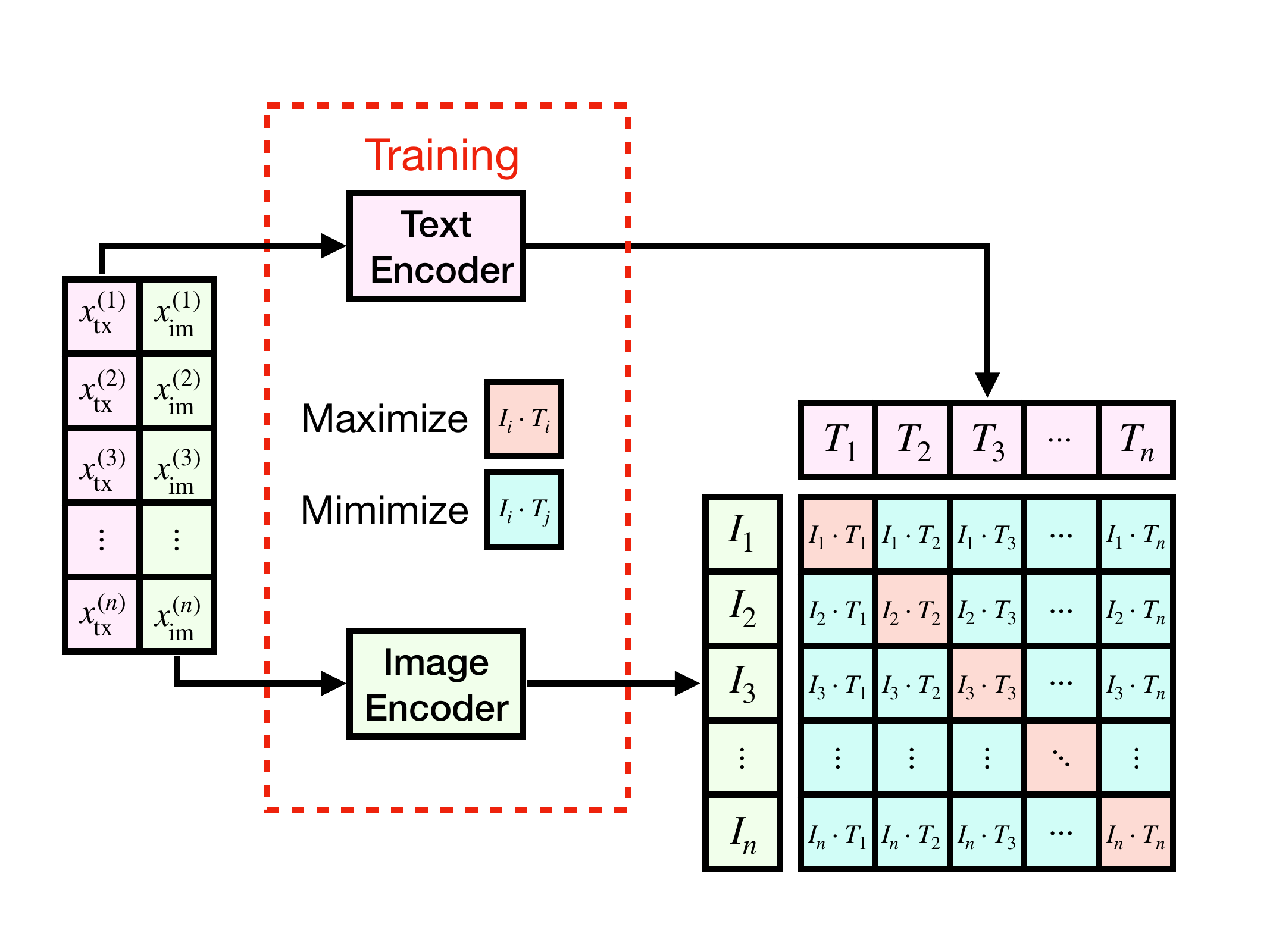}
\end{subfigure}\hfill
\begin{subfigure}[t]{0.48\textwidth}
\centering 
\caption{ Zero-shot classification.}\label{fig:CLIP_zero_shot_illustration}
\includegraphics[width=\linewidth]{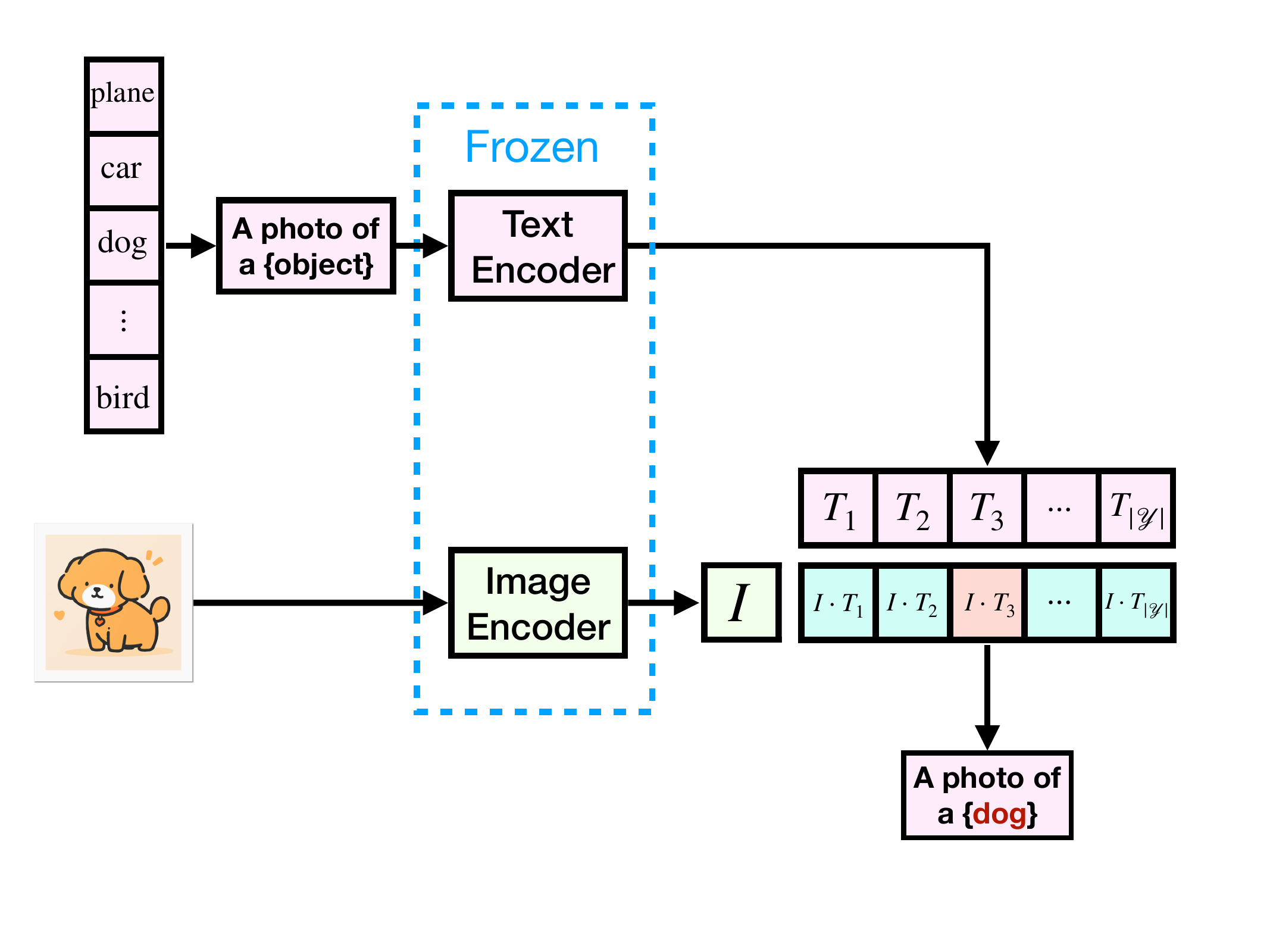}
\end{subfigure}
\caption{ Illustration of CLIP and zero-shot classification. CLIP trains a text encoder and an image encoder by maximizing the similarity of paired image-caption representations and minimizing the similarity of non-paired representations. After pre-training, zero-shot classification predicts the label whose representation has the highest similarity with the image representation. This figure reproduces Figure 1 from \cite{radford2021learning}.}\label{fig:CLIP_illustration}
\end{figure}
\fi

\paragraph{Conditional Diffusion Models (CDMs).} Conditional diffusion models are generative models that, when applied to image-text tasks, use diffusion processes to generate image samples conditioned on text inputs. These models have gained attention for their impressive performance in tasks such as image generation, super-resolution, and inpainting. Notable CDMs include DALL-E \cite{ramesh2022hierarchical}, StableDiffusion \cite{rombach2022high}, and Imagen \cite{saharia2022photorealistic}. CDMs typically operate by iteratively refining noise into a clear image using a series of conditional denoising functions, which incorporate the CLIP text embedding $\Et(\xt)$ as input. To illustrate CDMs, consider a specific diffusion model, stochastic localization \cite{eldan2013thin, el2022sampling}. The conditional denoising function, represented as $\{ \Mt(\bz, \Et(\xt)) \}_{t \ge 0}$, is typically parameterized by a U-Net or a transformer that approximates the conditional expectation of the clean image $\xi$ given noisy observations $\bz \sim \cN(t \cdot \xi, t \cdot  \id_{\dim})$ and the text embedding $\Et(\xt)$. These models are trained on large datasets of image-text pairs $(\xi^{(j)}, \xt^{(j)})_{j \in [n]}$ using a regression loss: 
\begin{equation}
\textstyle  \Mthat = \arg\min_{\Mt} \Big\{\what \risk_{\cdm, t}(\Mt) = \frac{1}{n} \sum_{j \in [n]} \big\| \xi^{(j)} - \Mt(t \cdot \xi^{(j)} + \sqrt{t} \cdot \bg^{(j)}, \Et(\xt^{(j)})) \big\|_2^2 \Big\},
\end{equation}
where $\{ \bg^{(j)}\}_{j \in [n]}$ are independent Gaussian noises. After training, given a new prompt $\xt$, the CDM generates an image as $\bz_T / T$ for large $T$, where $\bz_t$ is a solution to the stochastic differential equation
\[
\de \bz_t =  \Mthat(\bz_t, \Et(\xt)) \de t + \de \bW_t,~~~~ \bz_0 = \bzero,~~~~ \bW_t \text{ is Brownion motion. }
\]
An illustration of the CDM framework is shown in \Cref{fig:CDM_illustration}. 

Similar to VLMs, assuming infinite samples and unlimited neural network capacity, theoretical results suggest that as $T \to \infty$, the generated image $\xi$ produced by CDMs follows the conditional distribution $\P(\xi | \Et(\xt))$ \cite{sohl2015deep}. In this paper, we investigate: (1) the conditions under which CDMs can be effectively learned with finite network capacity and finite samples, and (2) how closely the conditional distribution of the generated image approximates the true conditional distribution $\P(\xi | \xt)$.

\if\showfigure1
\begin{figure}
\begin{subfigure}[t]{0.48\textwidth}
\centering 
\caption{ Vision-Language Models.}\label{fig:VLM_illustration}
\includegraphics[width=\linewidth]{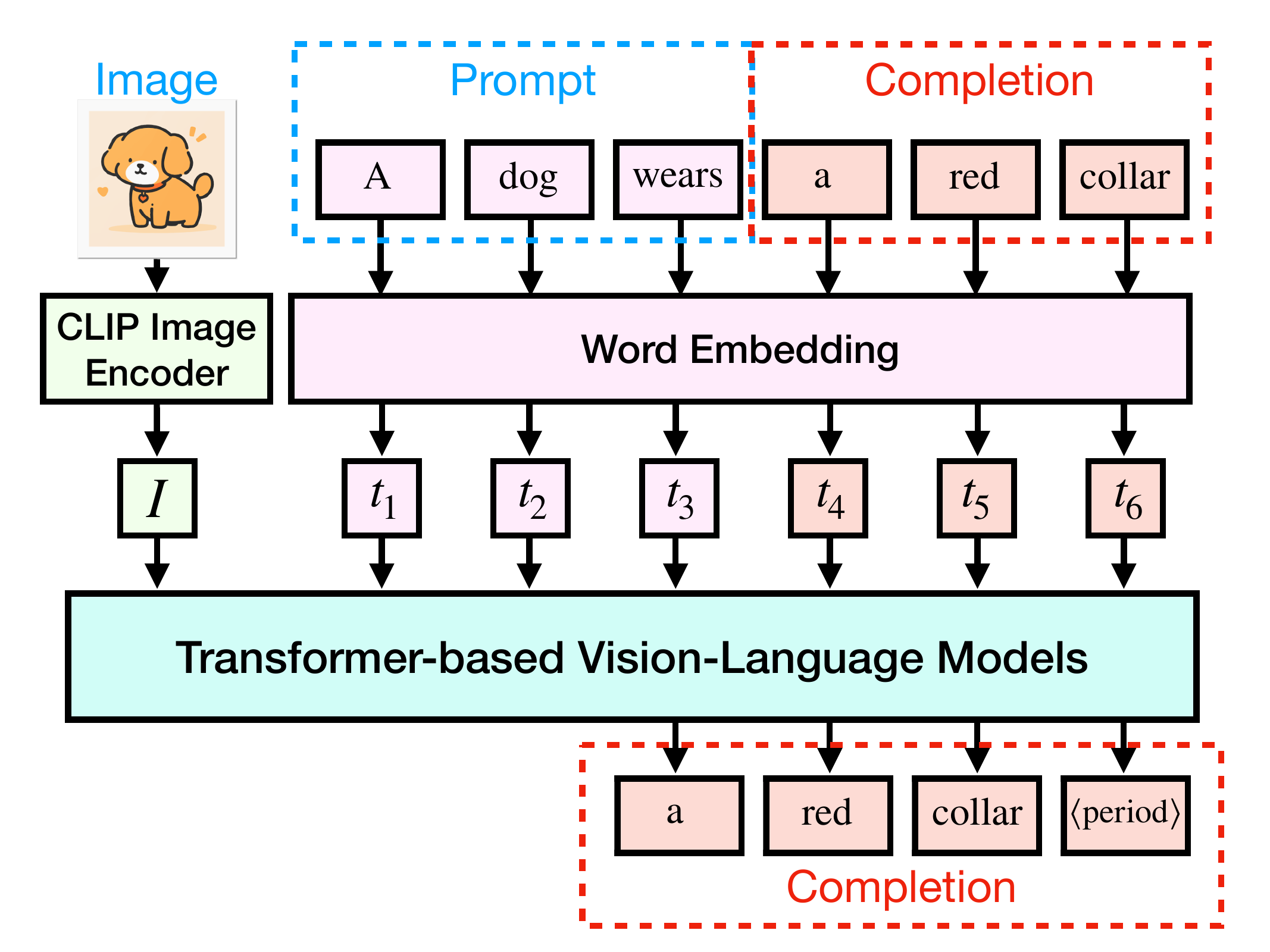}
\end{subfigure}\hfill
\begin{subfigure}[t]{0.48\textwidth}
\centering 
\caption{ Conditional Diffusion Models.}\label{fig:CDM_illustration}
\includegraphics[width=\linewidth]{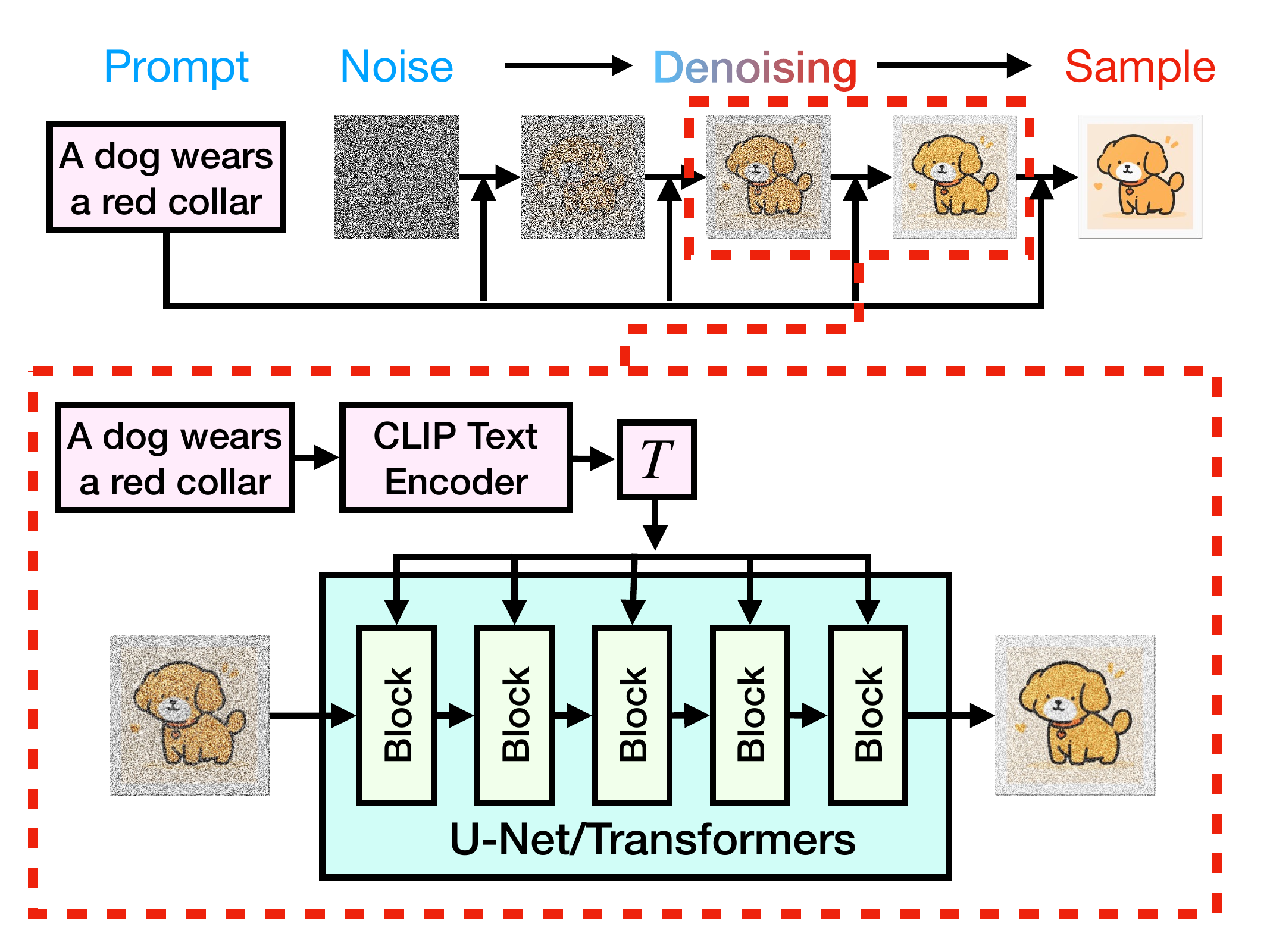}
\end{subfigure}
\caption{ Illustration of VLMs and CDMs. VLMs use neural networks to approximate the conditional distribution of each next token, given prior tokens and image embeddings. CDMs employ neural networks to approximate the conditional expectation of a clear image, given a noisy input image and text embeddings. }\label{fig:VLM_CDM_illustration}
\end{figure}
\fi

\fi

\section{Further related literature}\label{sec:related-literature}

\paragraph{CLIP and contrastive learning.} CLIP \cite{radford2021learning} and ALIGN \cite{jia2021scaling} are representation learning methods that extract visual and textual embeddings through large-scale contrastive pretraining. Central to these approaches are loss functions such as NCE \cite{gutmann2010noise}, InfoNCE \cite{oord2018representation}, and Multi-class N-pair loss \cite{sohn2016improved}, which use cross-entropy loss to distinguish between paired and non-paired samples. In single-modal contexts, similar contrastive learning methods like SimCLR \cite{chen2020simple}, MoCo (Momentum Contrast) \cite{he2020momentum}, and BYOL (Bootstrap Your Own Latent) \cite{grill2020bootstrap} employ data augmentations, momentum encoders, and self-distillation techniques to learn robust visual representations in a self-supervised manner. 

\paragraph{Multimodal learning.} Conditional Diffusion Models generate realistic images from text prompts \cite{sohl2015deep, ho2020denoising, song2019generative, song2020score}, with notable large-scale implementations such as DALL-E \cite{dalle2} and Stable Diffusion \cite{esser2024scaling}. Vision-Language Models produce natural language descriptions based on text prompts and image inputs, with examples like Flamingo \cite{alayrac2022flamingo}, BLIP \cite{li2022blip}, and Llava \cite{liu2024visual, liu2024improved}. Beyond traditional image and text modalities, multimodal learning also incorporates additional modalities such as speech \cite{zhao2023bubogpt, zhang2023speechgpt}, video \cite{yan2021videogpt}, and action \cite{brohan2023rt}. Contrastive pre-training plays a crucial role in extracting useful representations within these multimodal learning frameworks.


\paragraph{Theories of Contrastive Learning and CLIP.} Numerous studies have shown that InfoNCE loss (derived from the InfoMax principle \cite{linsker1988self}) maximizes a lower bound on mutual information between positive sample pairs \cite{oord2018representation, poole2019variational, hjelm2018learning, bachman2019learning, tian2020contrastive, zimmermann2021contrastive, lu2024f}, which aligns with \Cref{lem:global_min_contrastive_loss} and \Cref{prop:approximate_global_min_contrastive_loss}. \cite{wang2020understanding} interpret contrastive loss through the concepts of alignment and uniformity, where alignment ensures that positive pairs have similar representations, and uniformity encourages a broader spread of representations across the feature space. \cite{saunshi2019theoretical, wang2022chaos, ash2021investigating} provide generalization bounds for InfoNCE minimizers in downstream classification tasks that are comprised of a subset of the same set of latent classes. \cite{tosh2021contrastive1} adopt a topic modeling perspective, demonstrating that contrastive loss minimizers reveal underlying topic posterior information to linear models, while \cite{tosh2021contrastive2} shows that linear functions of learned representations perform nearly optimally on downstream tasks when the two views contain redundant label information. \cite{haochen2021provable} utilize a spectral clustering perspective to offer a generalization bound for spectral (square-style) contrastive loss. \cite{huang2021towards} introduce a measure to quantify data augmentation and provide an error bound for downstream tasks. \cite{shi2023trade} discover a trade-off between label efficiency and universality in contrastive learning with linear probing. Regarding training dynamics, \cite{tian2020understanding} prove the emergence of hierarchical features, while \cite{wen2021toward} show that proper augmentations enable ReLU networks to learn desired sparse features. \cite{lee2021predicting} quantify how the approximate independence of pretext task components facilitates learning representations adaptable to downstream tasks. \cite{nakada2023understanding} examined CLIP within specific linear representation settings and emphasized its connection to singular value decomposition. 

Our work diverges from these existing theories of contrastive learning in three key ways: (1) While many studies provide “absolute risk bounds” for downstream tasks under structural conditions, our work offers “excess risk bounds,” which require more refined statistical analysis; (2) We analyze the multimodal learning, including zero-shot prediction task, conditional diffusion models, and vision-language models, which have not been addressed in these work; and (3) We proposed a data distribution for image and text pairs and provided end-to-end statistical efficiency guarantees for multimodal learning through neural networks. 

The works \cite{uesaka2024understanding, chen2023understanding} are the most closely related to our work. \cite{uesaka2024understanding} adopt a similar point-wise mutual information perspective to establish an upper bound on the excess risk for downstream classification tasks. \cite{chen2023understanding} examine the properties of the CLIP minimizer under the completeness condition and demonstrate the strong zero-shot classification capabilities of CLIP loss. In contrast to these studies, our work (1) adopts a sufficient statistics perspective to interpret the CLIP approach, (2) reveals additional properties of the learned CLIP representations, and (3) presents a unified approach with an end-to-end theory for multimodal learning, including vision-language Models and conditional diffusion models.

\paragraph{Approximate sufficient statistics.} The concept of approximate sufficient statistics was mentioned in \cite{chen2020neural}, which proposed an approach to find them. However, this work did not provide a formal definition of approximate sufficient statistics or explore its theoretical properties. The relationship between contrastive loss minimizers and sufficient statistics was examined in \cite{xu2024dependence}, but the notion of approximate sufficient statistics was not considered. After an extensive review of the literature, we conclude that the definition of approximate sufficient statistics and its connection to the approximate minimizer of CLIP loss, to the best of the authors’ knowledge, is novel.  

\paragraph{Neural networks as algorithms.} A recent line of work has investigated the expressiveness of neural networks from the perspective of algorithm approximation \cite{wei2022statistically, bai2024transformers, giannou2023looped, liu2022transformers, marwah2021parametric, marwah2023neural, lin2023transformers, mei2023deep, karan2024unrolled}. In particular, \cite{wei2022statistically, bai2024transformers, giannou2023looped, liu2022transformers, lin2023transformers} demonstrate that transformers can efficiently approximate various classes of algorithms, including gradient descent, reinforcement learning algorithms, and even Turing machines. In the context of diffusion models, \cite{mei2023deep, mei2024u} show that ResNets and U-Nets can efficiently approximate the score function of high-dimensional graphical models by approximating the variational inference algorithm. 

\paragraph{Generative hierarchical models (GHMs).} Generative hierarchical modeling of data distributions has been explored in a series of studies \cite{mossel2016deep, petrini2023deep, sclocchi2024phase, tomasini2024deep, cagnetta2024towards, garnier2024transformers, kadkhodaie2023learning, kadkhodaie2023generalization}. Notably, \cite{mossel2016deep} established the distinction between deep and shallow algorithms in GHMs, indicating that a deep network is essential for efficiently approximating belief propagation algorithms. GHMs are closely related to Dyck languages and context-free grammars in the context of language modeling \cite{hewitt2020rnns, yao2021self, zhao2023transformers, allen2023physics}. The diffusion model for multi-scale image distribution representations has been investigated in \cite{kadkhodaie2023learning, kadkhodaie2023generalization}, showing that U-Nets are effective for modeling denoising algorithms. Furthermore, the theoretical and empirical findings presented in \cite{petrini2023deep, sclocchi2024phase, tomasini2024deep, cagnetta2024towards, sclocchi2024probing, garnier2024transformers, mei2024u} highlight the ability of GHMs to capture the combinatorial properties of image and text datasets, demonstrating that neural networks can effectively represent and learn belief propagation algorithms within GHMs. 



\section{Results for vision-language models}

We include results upon vision-language models in this section.

\subsection{Error bound for vision-language models}\label{sec:error_bound_VLM}

VLMs take both image and text inputs and generate text outputs by sequentially sampling from a transformer-based model trained to approximate the conditional next-token probability, denoted as 
\begin{equation}\label{eqn:VLM_true_mu}
\trueprobnext(\, \cdot \,| \, \square\,,\, \star \, ) = \P_{\image, \txt}( \xtsub{i} = \, \cdot \, |  \xtsub{1:i-1} = \, \square\,,\xi = \, \star \, ).
\end{equation}
The transformer model achieves this by minimizing the risk over $\probnextspace \subseteq \cup_{i \in [\dimtxt]} \{ \probnext:  \cXtsub{1:i-1}\times \R^\dimclip  \to \cP(\cXtsub{i}) \}$, a function class with inputs consisting of the CLIP image representation and the text prompt. The population risk minimization is formulated as 
\begin{align}\label{eqn:VLM_with_CLIP}
\estprobnext = \arg\min_{\probnext \in \probnextspace} \Big\{ \risk_{\vlm}(\probnext, \Ei) \defn \E_{ (\xi, \xt) \sim \P_{\image,\txt}}
\Big[ \sum_{i \in [\dimtxt]} - \log \probnext(\xtsub{i}| \xtsub{1:i-1},\Ei(\xi)) \Big] \Big\}.
\end{align}

Notice that the global minimizer of this formulation, when $\probnextspace$ includes all measurable conditional probability functions, is given by $\estprobnext(\, \cdot \,| \square\,,\sE \,) = \P_{\image, \txt}( \xtsub{i} = \, \cdot \, |  \xtsub{1:i-1} = \, \square\,,\Ei(\xi) = \sE )$. This differs from the true conditional next-token probability $\trueprobnext(\, \cdot \,| \, \square\,,\, \star \, )$ as defined in Eq.~(\ref{eqn:VLM_true_mu}). Nevertheless, \Cref{prop:VLM_CLIP_error} below shows that the error of $\estprobnext$, measured by 
\begin{equation}\label{eqn:error_rate_VLM_mu}
\distance(\trueprobnext, \probnext)
\defn \E_{ (\xi, \xt) \sim \P_{\image,\txt}}
\Big[ \sum_{i \in [\dimtxt]} \KL\Big( \trueprobnext(\xtsub{i}| \xtsub{1:i-1} , \xi) \Big|\Big| \probnext(\xtsub{i}| \xtsub{1:i-1},\Ei(\xi)) \Big) \Big],
\end{equation}
is bounded by the sufficiency of the image encoder $\Ei$, and hence bounded by the CLIP excess risk. 

\begin{proposition}[Error bound for VLMs]\label{prop:VLM_CLIP_error}
Let $\probnextspace$ include all measurable conditional probability functions. Then, the error rate of $\estprobnext$, as defined in (\ref{eqn:VLM_with_CLIP}) and (\ref{eqn:error_rate_VLM_mu}), is bounded by the sufficiency of the encoder $\Ei$: 
\begin{equation}\label{eqn:NTP_generalization_CLIP}
\begin{aligned}
\distance(\trueprobnext, \estprobnext) \le \suffmeasure(\Ei). 
\end{aligned}
\end{equation} 
\end{proposition}
The proof of \Cref{prop:VLM_CLIP_error} is provided in \Cref{app:proof_VLM_CLIP_error}. Briefly, the error rate $\distance(\trueprobnext, \estprobnext)$ can be directly bounded as follows 
\[
\distance(\trueprobnext, \estprobnext) = \E_{\xt}[\KL(\P(\xt | \xi), \P(\xt | \Ei(\xi)))] \le \suffmeasure(\Ei),
\]
where the first inequality follows from the tensorization property of KL divergence, and the second inequality follows by the definition of $\suffmeasure(\Ei)$.

\subsection{Sample-efficient learning of VLMs}\label{sec:VLM_GHM}

In this section, we investigate the vision-language models (VLMs) within the JGHM framework. Suppose we are given $n$ i.i.d. samples $(\xi^{(i)},\xt^{(i)})_{i \in [n]}$ drawn from the joint distribution of image and text, denoted as $\trueprobnext\defn\P_{\image,\txt}$. Given an image representation $\Et(\xt)\in\R^\dimclip$ (e.g., a CLIP-based embedding), the goal is to learn next-token predictors $\{ \probnext(\xtsub{j}| \xtsub{1:j-1},\Ei(\xi)) \}_{j\in[\dim_\txt]}$. Under an appropriate loss function, the optimal predictors are the conditional next-token probabilities $\{\trueprobnext(\xtsub{j}|\xi,\xtsub{1:j-1}) \}_{j\in[\dim_\txt]}$. This section focuses on analyzing the sample complexity of learning these conditional next-token predictors using empirical risk minimization over a class of transformers.

\paragraph{The neural network architecture.} The conditional next-token predictors are modeled as
\[
\probnext^\Par(\,\cdot \,| \xtsub{1:j-1},\Ei(\xi)) = \read_{\vlm} \circ \TF_{\vlm} \circ \encode_{\vlm}(\xtsub{1:j-1}, \Adapter(\Ei(\xi))), 
\] 
where each component is defined as follows. The function $\read_{\vlm}: \R^{D \times \star} \to \R^{\state}$ maps the transformer output to the predicted probabilities for the next token. The embedding function $\encode_{\vlm}: \R^{\star} \times \R^{\state} \to \R^{D \times \cdot}$ maps the input features into a transformer-compatible embedding, with specific details provided in Appendix~\ref{subsubsection:NWP-Intro-NN}. The image encoder $\Ei: \cXi \to \R^\state$ is given by the pre-trained CLIP representations, as defined in Eq.~(\ref{eq:simscore_architecture})~and~(\ref{eq:clip-empirical-risk-minimization}). 

The transformer $\TF_{\vlm}: \R^{D \times \star} \to \R^{D \times \star}$ is a trainable $2\layer+2$-layer model parameterized by $\bW_\vlm$, 
where each layer consists of the first feed forward layer, self-attention, second feed forward layer, and normalization
(see Appendix~\ref{subsubsection:NWP-Intro-NN}). 
There are two feed forward networks in a single layer, and the parameters of the two feed forward networks in the $\ell$-th layer are denoted by $\{W_{1,i}^{(\ell)}\}_{i=1}^{J+1}$ and $\{W_{2,i}^{(\ell)}\}_{i=1}^{J+1}$.
The adapter network $\Adapter:\R^{\state}\mapsto\R^{\state}$, also trainable, is parameterized by $\adaweighta \in \R^{\state \times \diminadapt}$ and $\adaweightb \in \R^{\diminadapt \times \state}$, and is defined identically to that in CDMs, as described in Eq.~(\ref{eq:adapter_def_cdm}). Following the pre-training fine-tuning paradigm, we consider the fine-tuning phase where the parameters $\Par = (\bW_\vlm, \adaweighta, \adaweightb)$ are optimized, while $\read_{\vlm}$, $\encode_{\vlm}$, and the CLIP encoder $\Ei$ remain fixed.

\paragraph{The ERM estimator.}
Given a pre-trained image encoder $\Ei:\cXi\mapsto\R^{\state}$, the goal is to obtain the conditional next-token predictors. To achieve this, we solve the empirical risk minimization problem defined by the following objective:
\begin{align}\label{eqn:VLM_empirical_risk_minimization}
\EstPar = \arg\min_{\Par \in \Parspacevlm} \Big\{ \what\risk_{\vlm}(\probnext^\Par, \Ei) \defn \frac{1}{n}\sum_{i=1}^n
\Big[ \sum_{j \in [\dimtxt]} - \log \probnext^\Par(\xtsub{j}| \xtsub{1:j-1},\Ei(\xi)) \Big] \Big\},
\end{align}
where the parameter space is defined as (see also \Cref{definition:VLM_param_space})
\begin{align}\label{eqn:VLM_param_space}
&~\Parspacecdm \defn \Big\{ \bW_{\vlm}, \adaweighta,\adaweightb \text{ as defined in Eq.~(\ref{eq:adapter_def_cdm})};
\\
&\nrmps{\btheta} \defn \lops{\adaweighta}\vee\lops{\adaweightb} \vee \max_{j\in [2],i \in [\numfflayer+1], \ell \in [2\layer+2]}\!\{ \| \ffnweight_{j,i, \vlm}^{(\ell)} \|_{\op}, \| {\querymatrix}_{,\vlm}^{(\ell)}\|_{\op}, \| {\keymatrix}_{,\vlm}^{(\ell)} \|_{\op}, \| {\valuematrix}_{,\vlm}^{(\ell)} \|_{\op} \}  \le \bound \Big\}.
\end{align}


The following theorem establishes a bound on the sampling error of the conditional next-token predictors in terms of the conditional KL divergence. 
\begin{theorem}[Sampling error of the conditional next-token predictors]~\label{thm:vlm_two_steps}
Suppose that \Cref{assumption:factorization} and \Cref{ass:bounded_transition} hold, and assume \Cref{ass:well_posed_representation} (a) holds for the text representation $\trueEt(\xi) = [ \P(\bs | \xt) /\P(\bs)]_{\bs \in \cS} \in \R^{\state}$ where $\cS$ is the set of root nodes. Let $\Ei$ and $\scorefun$ be obtained from the CLIP minimization. Let $\Parspacevlm$ be the set defined in Eq.~(\ref{eqn:VLM_param_space}), where $\numfflayer = \bigOtil(\layer)$, $D = \bigO(\state \layer)$, $D' =\bigOtil(\mmaintextmax \state \layer^3)$, and $\bound = \bigOtil( \Binvbd+(\state\layer+\mmaintextmax^2)\sqrt{\diminadapt})$. Let  $\EstPar$ be the empirical risk minimizer defined in Eq.~(\ref{eqn:VLM_empirical_risk_minimization}). Then, with probability at least $1 - 1/n$, we have
\begin{align}\label{eqn:excess_risk_bound_VLM}
\distance(\trueprobnext, \probnext^\EstPar)
&\defn
\E_{ (\xi, \xt) \sim \P_{\image,\txt}}
\Big[ \sum_{i \in [\dimtxt]} \KL\Big( \trueprobnext(\xtsub{i}| \xtsub{1:i-1}, \xi ) \Big|\Big| \probnext^\EstPar(\xtsub{i}| \xtsub{1:i-1},\Ei(\xi)) \Big) \Big]\\
&
\le
\dim_\txt\cdot
\bigOtil \Bigg( \sqrt{\frac{ (\state \layer^{8}\mmaxb^2+\diminadapt)\state\layer^3}{n}} 
+
\sqrt{\state^{5} \cdot \Binvbd^2 \cdot \Big( \suffmeasure(\scorefun) + \frac{1}{\diminadapt}\Big )}
\Bigg),
\end{align}
where $\bigOtil$ hides polynomial factors in $(\log(\mmaxb \state \layer\Binvbd n), (\transbd)^{\mmaxone})$.
\end{theorem}
The proof of \Cref{thm:vlm_two_steps} is provided in \Cref{sec:proof-vlm_two_steps}. Again, the key step involves constructing transformers that approximate the conditional next-token probabilities by emulating the belief propagation algorithm. The interpretation of the two terms in the upper bound aligns with the explanation provided following \Cref{theorem:CDM-Generalization-two-step}.

\section{Proofs in Section~\ref{sec:CLIP_minimizer}}

We start with introducing an alternative data distribution on the random variables $(\xib,(\xtb_{,j})_{j\in[K]},\kb)$, viz.,
\begin{align}
(\xtb_{,j})_{j\in[K]}\simiid\P_\txt\indep\kb\sim \mathrm{Unif}\{1,\ldots,K\},~~ \xib\sim \P_{\image|\txt}(\cdot|\xtb_{,\kb}).
\end{align}
Note that conditioned on $\kb$, $(\xib,(\xtb_{,j})_{j\in[K]})$ and $(\xi_{,1},(\xt_{,j})_{j\in[K]})$ have the same distribution up to some permutation of the samples. Therefore,
\begin{align}
\E_{\xi_{,1},(\xt_{,j})_{j\in[K]}} \Big[ - \log \frac{\exp(\simscore(\xi_{,1}, \xt_{,1})  ) }{\sum_{j \in [K]} \exp(\simscore(\xi_{,1}, \xt_{, j}))} \Big] 
=
\E_{\xib,(\xtb_{,j})_{j\in[K]},\kb} \Big[ - \log \frac{\exp(\simscore(\xib, \xtb_{,\kb})  ) }{\sum_{j \in [K]} \exp(\simscore(\xib, \xtb_{, \kb}))} \Big]. 
\end{align}
Similarly, we introduce the distribution on  $(\xtu,(\xiu_{,j})_{j\in[K]},\ku)$ as  
\begin{align}
(\xiu_{,j})_{j\in[K]}\simiid\P_\image\indep\ku\sim \mathrm{Unif}\{1,\ldots,K\},~~ \xtu\sim \P_{\txt|\image}(\cdot|\xiu_{,\ku}).
\end{align}
Then the CLIP risk function
\begin{align}
    \orisk_{\clip,K}(\scorefun) 
    &= 
    \E_{\xib,(\xtb_{,j})_{j\in[K]},\kb} \Big[ - \log \frac{\exp(\simscore(\xib, \xtb_{,\kb})  ) }{\sum_{j \in [K]} \exp(\simscore(\xib, \xtb_{, \kb}))} \Big]\\
    &\qquad
    +
    \E_{\xtu,(\xiu_{,j})_{j\in[K]},\ku} \Big[ - \log \frac{\exp(\simscore(\xtu, \xiu_{,\ku})  ) }{\sum_{j \in [K]} \exp(\simscore(\xtu, \xiu_{, \ku}))} \Big].\label{eq:clip_risk_alternative_expr}
\end{align}
To simplify the proofs,  in this section, we will use the alternative expression in Eq.~\eqref{eq:clip_risk_alternative_expr} for the CLIP risk function. 

Moreover, throughout this section, we use $\polyshort>0$ to denote constants that depend polynomially in $c_1$ in Assumption~\ref{ass:bounded_score}. We allow the value of $\polyshort$ to vary from place to place.

\subsection{Proof of \Cref{lem:global_min_contrastive_loss}}\label{app:proof_global_min_contrastive_loss}

Define 
\begin{align}
    \orisk_{\clip,\image,K}(S) \defn  \E_{\xib, (\xtb_{, j})_{j \in [K]}, \kb} \Big[ - \log \frac{\exp(S(\xib, \xtb_{, \kb})  ) }{\sum_{j \in [K]} \exp(S(\xib, \xtb_{, j}))} \Big] . \label{eq:claim_minimizer_pf_1}
\end{align}
We will show that $\scorefun$ is a minimizer of $ \orisk_{\clip,\image,K}(S)$ if and only if \begin{align}\scorefun(\xi,\xt)=\log \Big[ \P_{\image, \txt}( \xi, \xt) / [ \P_{\image}(\xi) \cdot \P_{\txt}(\xt)] \Big]+h(\xi)
\end{align}
for some arbitrary function $h:\cXt\mapsto\R$.
Similarly, we can  define $ \orisk_{\clip,\txt,K}(S) $ and conclude that 
 $\scorefun$ is a minimizer of $ \orisk_{\clip,\txt,K}(S)$ if and only if \begin{align}\scorefun(\xi,\xt)=\log \Big[ \P_{\image, \txt}( \xi, \xt) / [ \P_{\image}(\xi) \cdot \P_{\txt}(\xt)] \Big]+h(\xt).\label{eq:claim_minimizer_pf_2}
\end{align}
Noting that 
$\orisk_{\clip,K} =\orisk_{\clip,\image,K}+\orisk_{\clip,\txt,K}$ and taking the intersection of the two sets of minimizers yields Eq.~(\ref{eqn:global_min_clip_risk}) in Lemma~\ref{lem:global_min_contrastive_loss}.

To establish the second part of Lemma~\ref{lem:global_min_contrastive_loss}, note that
\begin{align}
    \lim_{K \to \infty} \big[- 
 \inf_{\simscore} \orisk_{\clip, K,\image}(\simscore) + \log K\big]
 &=
  \lim_{K \to \infty}\E_{\xib, (\xtb_{, j})_{j \in [K]}, \kb} \Big[  \log \frac{\exp(S(\xib, \xtb_{, \kb})  ) }{\sum_{j \in [K]} \exp(S(\xib, \xtb_{, j}))/K} \Big]
 \\
 &= \mutinfor(\xi,\xt)-\lim_{K \to \infty}\E_{\xib, (\xtb_{, j})_{j \in [K]}, \kb} \Big[  \log\Big( {\frac{1}{K}\sum_{j \in [K]} \frac{\P(\xib, \xtb_{, j})}{\P_\image(\xib)\P_\txt(\xtb_{,j})}}\Big) \Big]
 \\
 &= \mutinfor(\xi,\xt)-
 \lim_{K \to \infty}\E_{\xi_{,1}, (\xt_{, j})_{j \in [K]}} \Big[  \log\Big( {\frac{1}{K}\sum_{j \in [K]} \frac{\P(\xi_{,1}, \xt_{, j})}{\P_\image(\xi_{,1})\P_\txt(\xt_{,j})}}\Big) \Big]\\
 &=\mutinfor(\xi,\xt),
\end{align}
where the second equality follows from plugging in the optimal score function and the last inequality uses the boundedness assumption of the density ratio and the bounded convergence theorem.
Combined this with a similar calculation for $\orisk_{\clip,K,\txt}$ yields the second part of Lemma~\ref{lem:global_min_contrastive_loss}.\\
\quad

It remains to establish Eq.~\eqref{eq:claim_minimizer_pf_1}~and~\eqref{eq:claim_minimizer_pf_2}. We only present the proof of Eq.~\eqref{eq:claim_minimizer_pf_1} here since Eq.~\eqref{eq:claim_minimizer_pf_2} follows from a similar argument.
Let $\mathcal{F}$ be the class of functions $f:[K]\times\cXi\times\cXt^{\otimes K}\mapsto \R$ such that $f(\kb,\xib,  (\xtb_{,j} )_{j \in [K]}) \ge 0$ and 
\begin{align}\sum_{\kb=1}^K f(\kb,\xib,  (\xtb_{,j})_{j \in [K]} ) = 1.
\end{align}
 For any $f\in\mathcal{F}$, consider the objective
\begin{align}
R_{\image}(f)
&\defn
\E_{\xi_{,1}, (\xt_{, j})_{j \in [K]}, \kb} \Big[ - \log f(\kb, \xib_, ( \xtb_{,j} )_{j \in [K]})
\Big]\\
&=\E_{\xi_{,1}, (\xt_{, j})_{j \in [K]}}[\KL(\P(\cdot|\xi_{,1}, (\xt_{, j})_{j \in [K]})|| f(\cdot, \xib_, ( \xtb_{,j} )_{j \in [K]}))]\\
&\quad
-
\E_{\xi_{,1}, (\xt_{, j})_{j \in [K]},\kb}[\log\P(\kb|\xi_{,1}, (\xt_{, j})_{j \in [K]})]
\end{align}
Therefore, the unique minimizer of $R_\image(f)$ on $\cF$ is
\[
f_\star(\kb, \xib, ( \xtb_{,j} )_{j \in [K]}) = \P( \kb | \xib, ( \xtb_{,j} )_{j \in [K]}). 
\]
For any score function $\scorefun$, define $f_\scorefun( \xib, ( \xtb_{,j} )_{j \in [K]})={\exp(\simscore(\xib, \xtb_{,\kb})  ) }/{\sum_{j \in [K]} \exp(\simscore(\xib, \xtb_{, \kb}))} $. Then $f_\scorefun\in\cF$ and $R_\image(f_\scorefun)=\orisk_{\clip,K,\image}(\scorefun)$. Thus, if the set $\mathcal{M}_\image\defn\{\scorefun:f_\scorefun=f_\star\}$ is non-empty, then $\scorefun$ is a minimizer of 
$\orisk_{\clip,K,\image}(\scorefun)$ if and only if $\scorefun\in\mathcal{M}_\image.$

To find $\mathcal{M}_\image$, we first calculate $f_\star( \kb , \xib, ( \xtb_{,j} )_{j \in [K]})$. Note that
\begin{align}
\P(\kb | \xib, ( \xtb_{,j} )_{j \in [K]}) =&~ \frac{\P(\kb, \xib, ( \xtb_{,j} )_{j \in [K]})}{\sum_{j = 1}^K \P(j, \xib, ( \xtb_{,j} )_{j \in [K]})} = \frac{\P(\kb) \P(\xib | \xtb_{,\kb} ) \prod_{j} \P(\xtb_{,j})}{\sum_{j=1}^K \P(j) \P(\xib | \xtb_{,j}) \prod_{j} \P(\xtb_{,j})} \\
=&~ \frac{\P(\xib, \xtb_{,\kb} ) / [ \P(\xib)\cdot \P(\xtb_{,\kb} )] }{\sum_{j=1}^K \P(\xib, \xtb_{,j}) / [ \P(\xib)\cdot \P(\xtb_{,j})]}. \label{eq:posterior_expr}
\end{align}
As a consequence, $\scorefun(\xi, \xt) = \log [ \P( \xi, \xt) / [ \P(\xt) \cdot \P(\xi)] ]+h(\xi)\in\mathcal{M}_\image$ for any function $h.$

Lastly, we conclude that $\mathcal{M}_\image$ only include such score functions. If $\widetilde\scorefun,\scorefun\in\mathcal{M}_\image$, then by properties of the softmax function, we have $\widetilde\scorefun(\xi,\xt_{1})-\widetilde\scorefun(\xi,\xt_{2})=\scorefun(\xi,\xt_{,1})-\scorefun(\xi,\xt_{,2})$ for any $(\xi,\xt_{,1},\xt_{,2})$. Thus, there must exist some function $\widetilde h$ such that $\widetilde\scorefun(\xi,\xt)=\scorefun(\xi,\xt)+h(\xi)$.  

Putting pieces together, we conclude that the set of minimizers of $\orisk_{\clip,\image,K}(\scorefun)$ is 
\begin{align}\mathcal{M}_\image = \{\scorefun: \scorefun =  \log [ \P( \xi, \xt) / [ \P(\xt) \cdot \P(\xi)] ]+h(\xi),\text{ for some }h\}.
\end{align}

\subsection{Proof of~\Cref{prop:approximate_global_min_contrastive_loss}}\label{app:approximate_global_min_contrastive_loss}

\begin{proof}[Proof of \Cref{prop:approximate_global_min_contrastive_loss}]

Similar to the proof of Lemma~\ref{lem:global_min_contrastive_loss}, we introduce 
\begin{align}
    \orisk_{\clip,\image,K}(\simscore) \defn  \E_{\xib, (\xtb_{, j})_{j \in [K]}, \kb} \Big[ - \log \frac{\exp(\simscore(\xib, \xtb_{, \kb})  ) }{\sum_{j \in [K]} \exp(\simscore(\xib, \xtb_{, j}))} \Big] .
\end{align}
We will show 
 \begin{align}
  \Big|
  \Big[ \orisk_{\clip, \image, K}(\simscore) - \orisk_{\clip,  \image, K}(\truesimscore) \Big] - \E_{\xi\sim\P_\image}\Big[ \KL\Big( \P(\cdot | \xi) \Big|\Big|  \softmaxscoreinf{\scorefun}(\cdot | \xi) \Big) \Big]
  \Big|
  \leq \frac{\polyshort}{{K}}
  \cdot 
  (\orisk_{\clip, \image, K}(\simscore) - \orisk_{\clip,  \image, K}(\truesimscore))
  \label{eq:proof_prop1_11}
\end{align} under \Cref{ass:bounded_score}. Likewise, we can define $\orisk_{\clip,\txt,K}(\simscore)$ and derive a bound similar to equation~\eqref{eq:proof_prop1_11} by the symmetry between $\xi$ and $\xt$. 
Proposition~\ref{lem:global_min_contrastive_loss} then follows from combining two bounds with a  triangle inequality.

Therefore, it remains to prove equation~\eqref{eq:proof_prop1_11}.
At a high level, the proof  consists of two steps: (a) we first simplify the expressions for $ \orisk_{\clip, \image, K}(\simscore) - \orisk_{\clip,  \image, K}(\truesimscore) $ and $ \E_{\xi\sim\P_\image}[ \KL( \P(\cdot | \xi) ||  \softmaxscoreinf{\scorefun}(\cdot | \xi) )]$ by some basic algebra; (b) we then establish an upper bound on the difference of the simplified expressions.\\

\myunder{Simplifying the expressions.}
By definition, we have
\begin{align}
    &~~~~~
    \orisk_{\clip, \image, K}(\simscore) - \orisk_{\clip, \image, K}(\truesimscore) \\
    &=
    \E_{\xib, (\xtb_{, j})_{j \in [K]}, \kb} \Big[ - \log \frac{\exp(\simscore(\xib, \xtb_{, \kb})  ) }{\sum_{j \in [K]} \exp(\simscore(\xib, \xtb_{, j}))}+
    \log \frac{\exp(\truesimscore(\xib, \xtb_{, \kb})  ) }{\sum_{j \in [K]} \exp(\truesimscore(\xib, \xtb_{, j}))} 
    \Big]
    \\
     &=
     \E_{\xib, (\xtb_{, j})_{j \in [K]}, \kb} \Big[   \log \frac{\exp(\truesimscore(\xib, \xtb_{, \kb})  ) }{\exp(\simscore(\xib, \xtb_{, \kb})  )}   
      - \log \frac{\sum_{j \in [K]} \exp(\truesimscore(\xib, \xtb_{, j})) }{\sum_{j \in [K]} \exp(\simscore(\xib, \xtb_{, j}))} 
    \Big]\\
    &=
    \Term_{a1}-\Term_{a2},
\end{align}  where
\begin{align}
    \Term_{a1} 
    &\defn\E_{(\xib,\xtb)\sim\P_{\image,\txt}}[\truesimscore(\xib, \xtb)  -\simscore(\xib, \xtb)  ],\\
    \Term_{a2}
    &\defn
   \E_{\xib, (\xtb_{, j})_{j \in [K]}, \kb} \Big[ \log \frac{\sum_{j \in [K]} \exp(\truesimscore(\xib, \xtb_{, j})) }{\sum_{j \in [K]} \exp(\simscore(\xib, \xtb_{, j}))} 
    \Big] .
\end{align}
Similarly,
\begin{align}
    E_{\xi}\Big[ \KL\Big( \P(\xt | \xi) \Big|\Big|  \softmaxscoreinf{\scorefun}(\xt | \xi) \Big) \Big]
    &=
    \E_{(\xi,\xt)\sim\P_{\image,\txt}}\Big[\log\frac{ \P(\xt | \xi) }{ \softmaxscoreinf{\scorefun}(\xt | \xi)}\Big]\\
    &=
  \E_{(\xi,\xt)
  \sim\P_{\image,\txt}}\Biggr[\log\frac{  \frac{\exp(\truesimscore(\xi, \xt)) \P(\xt)}{\sum_{\xt'} \exp( \truesimscore(\xi, \xt')) \P(\xt')}}{  \frac{\exp( \simscore(\xi, \xt)) \P(\xt)}{\sum_{\xt'} \exp( \simscore(\xi, \xt')) \P(\xt')}
  }\Biggr]
  \\
  &
  =
   \Term_{b1}-\Term_{b2},
\end{align}
where
\begin{align}
  \Term_{b1}&=   \E_{(\xi,\xt)\sim\P_{\image,\txt}}[\truesimscore(\xi, \xt)  -\simscore(\xi, \xt))  ],\\
  \Term_{b2}&= \E_{\xi\sim\P_{\image}}\Big[\log \frac{\E_{\xt\sim\P_{\txt}} \exp(\truesimscore(\xi, \xt)) }{\E_{\xt\sim\P_{\txt}} \exp(\simscore(\xi, \xt))} \Big].
\end{align}
Since $\Term_{a1}=\Term_{b1}$, it suffices to bound the difference $|\Term_{a2}-\Term_{b2}|$.\\

\myunder{Bounding $|\Term_{a2}-\Term_{b2}|$.}
Without loss of generality, we assume $\scorefun,\scorefun_\star$ are chosen such that the conditional mean
\begin{align}
\E_{\xt|\xi\sim \P_{\txt|\image}}[\scorefun(\xi,\xt)] = \E_{\xt|\xi\sim \P_{\txt|\image}}[\scorefun_\star(\xi,\xt)]=0
\end{align}
for all $\xi\in\cX_\image$. Note that this can be done by 
substracting the conditional mean (which is a function of $\xi$) from $\scorefun$ (or $\scorefun_\star$).  In this case, \Cref{ass:bounded_score} still holds  but with a different constant $c_1'>1$ that is polynomially dependent on $\boundscoreconst$.

For any function $\widetilde\perturb:\cX_\image\times\cX_\txt\mapsto \R$, we define its norm
\begin{align}
    \nrmp{\widetilde\perturb} \defn  \sqrt{\E_{(\xi,\xt)\sim \P_{\image,\txt}}[\widetilde\perturb(\xi,\xt)]^2}.
\end{align}
In addition, for any score function $\widetilde\scorefun$, we introduce
the distributions 
\begin{align}
   \softmaxscore_{\widetilde\scorefun}(k|\xib, (\xtb_{, j})_{j \in [K]} ) 
   &\defn 
   \frac{\exp(\widetilde\scorefun(\xib,\xtb_{, k}))}{\sum_{j \in [K]} \exp(\widetilde\scorefun(\xib, \xtb_{, j}))}~~~\text{for all}~ k\in[K],  \text{~~~and recall that}
   \\
   \softmaxscoreinf{\widetilde\scorefun}(\xt|\xi) & \defn
   \frac{\exp(\widetilde\scorefun(\xi,\xt))\P(\xt)}{\E_{\xt'\sim\P_\txt}{\exp(\widetilde\scorefun(\xi,\xt'))}}~~~\text{for all}~ \xt\in\cXt.
\end{align}
Note that $\softmaxscore_{\scorefun_\star}$ is the posterior distribution of $\kb$ conditioned on $\xib, (\xtb_{, j})_{j \in [K]}$ as shown in equation~\eqref{eq:posterior_expr} in the proof of Lemma~\ref{lem:global_min_contrastive_loss}; moreover, we have $  \softmaxscoreinf{\scorefun_\star}=\P_{\txt|\image}$.

We begin by claiming that
\begin{align}
  \nrmp{\scorefun-\scorefun_\star}
  \leq \polyshort\cdot\sqrt{  \orisk_{\clip, \image, K}(\scorefun)-\orisk_{\clip, \image, K}(\scorefun_\star)}\label{eq:alter_clip_general_claim1}
\end{align} 
for some constant $\polyshort>0.$ The proof of this claim can be found in Lemma~\ref{lm:bound_score_diff}. Moreover, we argue that
\begin{align}
    |\Term_{a2}-\Term_{b2}|\leq
    \frac{\polyshort}{K}\cdot\nrmp{\scorefun-\scorefun_\star}^2.
    \label{eq:alter_clip_general_claim2}
\end{align}
Combining equation~\eqref{eq:alter_clip_general_claim1}~and~\eqref{eq:alter_clip_general_claim2} yields the desired result.

Therefore, it remains to establish equation~\eqref{eq:alter_clip_general_claim2}.
 Write $\scorefun=\scorefun_{\star}+r \perturb$ with $r=\nrmp{\scorefun-\scorefun_\star}$ and $\perturb= (\scorefun-\scorefun_\star)/\nrmp{\scorefun-\scorefun_\star}$, and define
 \begin{align}
     \Term(r)\defn 
      \E_{{\xib, (\xtb_{, j})_{j \in [K]}, \kb}}\Big[\log \frac{\sum_{j \in [K]}  \exp(\simscore_r(\xib, \xtb_{, j})) /K}{\E_{\xt\sim\P_{\txt}} \exp(\simscore_r(\xib, \xt))} \Big],
 \end{align}
 where $\scorefun_r=\scorefun_\star+r\perturb$. Then we have $|\Term_{a2}-\Term_{b2}|=|\Term(\nrmp{\scorefun-\scorefun_\star})-\Term(0)|$.
Performing a second-order Taylor expansion on $\Term(r)$ w.r.t. $r$ at $r=0$, and noting that $r=0$ is a stationary point, we obtain
\begin{align}
   |\Term_{a2}-\Term_{b2}|
    &=
     \Big|\Big\{ \E_{\xib, (\xtb_{, j})_{j \in [K]}}\Big[\Var_{k\sim \softmaxscore_{\scorefun_\star+\widetilde r\perturb}}\Big[\frac{\scorefun(\xib,\xtb_{,k})-\scorefun_\star(\xib,\xtb_{,k})}{\nrmp{\scorefun-\scorefun_\star}}\Big]\Big]
    \\
    &\quad
    -
    \E_{\xi\sim\P_\image} \Var_{\xt\sim\softmaxscoreinf{\scorefun_\star+\widetilde r\perturb}}
     \Big[
\frac{\scorefun(\xi,\xt)-\scorefun_\star(\xi,\xt)}{\nrmp{\scorefun-\scorefun_\star}}\Big]
     \Big\}\cdot
     \nrmp{\scorefun-\scorefun_\star}^2
     \Big|\\
     &=
|\Term_{d2}(\widetilde r)| 
\end{align}
for some $\widetilde r\in[0,\nrmp{\scorefun-\scorefun_\star}]$,
where
\begin{align}
    \Term_{d2}( r)
    &\defn
              \E_{\xib, (\xtb_{, j})_{j \in [K]}} \Big[\Var_{k\sim \softmaxscore_{\scorefun_\star+ r\perturb}}\Big[\scorefun(\xib,\xtb_{,k})
              -\scorefun_\star(\xib,\xtb_{,k})\Big]\Big]
         -\\
              &~\quad
   \E_{\xi\sim\P_\image}\Big[ \Var_{\xt\sim\softmaxscoreinf{\scorefun_\star+ r\perturb }}
         \Big[
    {\scorefun(\xi,\xt)-\scorefun_\star(\xi,\xt)}\Big]\Big].\label{eq:def_term_d2}
\end{align}
Equation~\eqref{eq:alter_clip_general_claim2} then follows immediately from Lemma~\ref{lm:bound_diff_second}, which states that
\begin{align}
    |\Term_{d2}( r)|\leq \polyshort\cdot \nrmp{\scorefun-\scorefun_\star}^2/K  \label{eq:alter_clip_claim2_pf2}
\end{align} for some constant $\polyshort>0$   for all $r\in[0,\nrmp{\scorefun-\scorefun_\star}]$.

\end{proof}

{

\subsection{An alternative to Proposition~\ref{prop:approximate_global_min_contrastive_loss}}\label{sec:alternative_bounded_score}
An alternative additive error bound on the sufficiency $\suffmeasure(\simscore)$ can be established without the boundedness conditions in Assumption~\ref{ass:bounded_score}. To this end, we introduce the following weaker assumption.
\begin{assumption}[Bounded expected score]\label{ass:bounded_score_alternative}
    There exists some constant $\boundscoreconst>0$ such that 
    \begin{align}
        \E_{(\xi,\xt)\sim\mathcal{P}}[\exp(4A(\xi,\xt))] \leq \boundscoreconst, \text{ for any } A\in\{\pm\simscore,\pm\truesimscore\}~\text{and}~\mathcal{P}\in\{\P_{\image,\txt},\P_{\image}\times\P_{\txt}\},
    \end{align}
where $\truesimscore(\xi,\xt)\defn \log \frac{\P_{\image, \txt}(\xi, \xt) }{ \P_{\image}(\xi) \P_{\txt}(\xt) } $.
\end{assumption}
Note that Assumption~\ref{ass:bounded_score_alternative} is implied by Assumption~\ref{ass:bounded_score}. Under this condition, we have the following result.
\begin{proposition}\label{prop:sufficiency_excess_risk_alternative}
    Under Assumption~\ref{ass:bounded_score_alternative} and the notations in Proposition~\ref{prop:approximate_global_min_contrastive_loss}, for any $K\geq 2$, we have 
    \begin{align}\label{eqn:sufficiency_excess_risk_alternative}
        \lim_{K' \to \infty}\Big[ \orisk_{\clip,  K'}(\simscore) - \orisk_{\clip,  K'}(\truesimscore)\Big] = \suffmeasure(\simscore) \leq \underbrace{\Big[ \orisk_{\clip,  K}(\simscore) - \orisk_{\clip,  K}(\truesimscore) \Big]}_{\textup{CLIP excess risk}}+\frac{\polyshort}{K}
        \end{align} 
        for some constant $\polyshort>0$ depending polynomially on $\boundscoreconst$.
\end{proposition}
As a consequence, the similarity score function $\simscore$ is near-sufficient when the CLIP excess risk is small and the batch size $K$ is sufficiently large. 
\footnote{{In practice, a large batch size $K$ (e.g., $K=32768$) is often used to improve the performance of CLIP~\cite{radford2021learning}.}}

\begin{proof}[Proof of Proposition~\ref{prop:sufficiency_excess_risk_alternative}]
    Throughout the proof, we use  $\polyshort>0$ to denote constants that depends polynomially on $\boundscoreconst$ in Assumption~\ref{ass:bounded_score_alternative}. We allow the value of $\polyshort$ to vary from place to place.
   Following the  proof of~\Cref{prop:approximate_global_min_contrastive_loss} in Section~\ref{app:approximate_global_min_contrastive_loss} and the notations therein, we have
   \begin{align}
    \Big|
    \Big[ \orisk_{\clip, \image, K}(\simscore) - \orisk_{\clip,  \image, K}(\truesimscore) \Big] - \suffmeasure(\simscore)
    \Big|
    \leq |\Term_{a2}-\Term_{b2}|,
    \label{eq:proof_prop1_11_alternative}
  \end{align}
  where 
  \begin{align}
    \Term_{a2}&=    \E_{\xi, (\xt_{, j})_{j \in [K]}} \Big[ \log \frac{\sum_{j \in [K]} \exp(\truesimscore(\xi, \xt_{, j})) }{\sum_{j \in [K]} \exp(\simscore(\xi, \xt_{, j}))} 
    \Big],\\
    \Term_{b2}&= \E_{\xi\sim\P_{\image}}\Big[\log \frac{\E_{\xt\sim\P_{\txt}} \exp(\truesimscore(\xi, \xt)) }{\E_{\xt\sim\P_{\txt}} \exp(\simscore(\xi, \xt))} \Big].
  \end{align}
  It suffices to show that 
  \begin{align}
    |\Term_{a2}-\Term_{b2}|
    \leq 
   \polyshort/K\label{eq:proof_suff_minimizer_alternative_1}
  \end{align}
  for some constant $\polyshort>0$ depending polynomially on $\boundscoreconst$.
  By some basic algebra, we have
  \begin{align}
    |\Term_{a2}-\Term_{b2}|
    &
    \leq 
    |\Term_{3}|+|\Term_{4}|,
      \end{align}
      where 
      \begin{align}
        \Term_{3}&=
        \E_{(\xi,\xt)\sim\P_{\image,\txt},\{\xt_{,j}\}_{j=1}^{K-1}\sim\P_\txt}\Big[\log \frac{\exp(S_\star(\xi, \xt))/K+\sum_{j \in [K-1]}  \exp(S_\star(\xi, \xt_{, j}))/K }{\E_{\xt\sim\P_{\txt}} \exp(S_\star(\xi, \xt)) } \Big],\\
        \Term_{4}&=
        \E_{(\xi,\xt)\sim\P_{\image,\txt},\{\xt_{,j}\}_{j=1}^{K-1}\sim\P_\txt}\Big[\log \frac{\exp(S(\xi, \xt))/K+\sum_{j \in [K-1]}  \exp(S(\xi, \xt_{, j}))/K }{\E_{\xt\sim\P_{\txt}} \exp(S(\xi, \xt)) } \Big].
      \end{align}
    We will show below that $|\Term_{3}|\leq \polyshort/K$  for some constant $\polyshort>0$. The same bound holds for $|\Term_{4}|$ following the same argument. Thus, combining the bounds yields Eq.~\eqref{eq:proof_suff_minimizer_alternative_1}.\\

    \myunder{An upper bound on $\Term_{3}.$} Note that
      \begin{align}
       &\quad \Term_{3}\\
        &\overset{(i)}{\leq}  
        \E_{(\xi,\xt)\sim\P_{\image,\txt},\{\xt_{,j}\}_{j=1}^{K-1}\sim\P_\txt}\Big[ \frac{\frac{1}{K}\sum_{j \in [K-1]}  \exp(S_\star(\xi, \xt_{, j}))+\frac{\exp(S_\star(\xi, \xt))}{K}-\E_{\xt\sim\P_{\txt}} \exp(S_\star(\xi, \xt))}{\E_{\xt\sim\P_{\txt}} \exp(S_\star(\xi, \xt)) } \Big]\\
        &=
        \frac{1}{K}
        \E_{(\xi,\xt)\sim\P_{\image,\txt}}\Big[ \frac{\exp(S_\star(\xi, \xt))}{\E_{\xt\sim\P_{\txt}} \exp(S_\star(\xi, \xt)) } \Big]-  \frac{1}{K},
      \end{align}
      where step~(i) uses $\log(1+a)\leq a $ for any $a>-1.$ By Cauchy-Schwarz inequality and Jensen's inequality, we further have 
   \begin{align}   
   &\quad \E_{(\xi,\xt)\sim\P_{\image,\txt}}\Big[ \frac{\exp(S_\star(\xi, \xt))}{\E_{\xt\sim\P_{\txt}} \exp(S_\star(\xi, \xt)) } \Big]\\
    &\leq
   \sqrt{\E_{\P_{\image,\txt}} {\exp(2S_\star(\xi, \xt))}}
   \cdot \sqrt{\E_{\P_{\image}} \frac{1}{(\E_{\xt\sim\P_{\txt}} {\exp(S_\star(\xi, \xt))})^2}}\\
   &\leq
   \sqrt{\E_{\P_{\image,\txt}} {\exp(2S_\star(\xi, \xt))}}
   \cdot \sqrt{\E_{\P_{\image}\times\P_{\txt}} \frac{1}{{\exp(2S_\star(\xi, \xt))}}}\\
   &\leq \polyshort,
   \end{align}  
   where the last inequality follows from Assumption~\ref{ass:bounded_score_alternative}. Putting pieces together yields $\Term_{3}\leq \polyshort/K$.
   \\

   \myunder{An lower bound on $\Term_{3}.$}
      On the other hand, we have the lower bound
      \begin{align}
        &\quad \Term_{3}
        \\
        &\overset{(ii)}{\geq}  
      \E_{(\xi,\xt)\sim\P_{\image,\txt},\{\xt_{,j}\}_{j=1}^{K-1}\sim\P_\txt}\Big[
      \frac{\frac{1}{K}\sum_{j \in [K-1]}  \exp(S_\star(\xi, \xt_{, j}))+\frac{\exp(S_\star(\xi, \xt))}{K}-\E_{\xt\sim\P_{\txt}} \exp(S_\star(\xi, \xt))}{\sum_{j \in [K-1]}  \exp(S_\star(\xi, \xt_{, j}))/K +\exp(S_\star(\xi, \xt))/K }  \Big]\\
      &=: \Term_{5}+\Term_{6},
      \end{align}
      where step~(ii) uses $\log(1+a)\geq a/(a+1)$ for any $a> -1$, and
      \begin{align}
  \Term_{5}
  &\defn
        \frac{1}{K}\Big[
        \E_{(\xi,\xt)\sim\P_{\image,\txt}}\Big[ \frac{\exp(S_\star(\xi, \xt))}{\E_{\xt\sim\P_{\txt}} \exp(S_\star(\xi, \xt)) } \Big]-1\Big],\\
        \Term_{6}
        &\defn
        -
       \E_{(\xi,\xt)\sim\P_{\image,\txt},\{\xt_{,j}\}_{j=1}^{K-1}\sim\P_\txt}\Big[
      \frac{[\frac{\sum_{j \in [K-1]}  \exp(S_\star(\xi, \xt_{, j}))}{K}+\frac{\exp(S_\star(\xi, \xt))}{K}-\E_{\xt\sim\P_{\txt}} \exp(S_\star(\xi, \xt))]^2}{[\frac{\sum_{j \in [K-1]}  \exp(S_\star(\xi, \xt_{, j}))}{K}+\frac{\exp(S_\star(\xi, \xt))}{K}]\cdot \E_{\xt\sim\P_{\txt}} \exp(S_\star(\xi, \xt))}  \Big].
      \end{align}
      Note that $\Term_{5}\geq -1/T$. To prove that $\Term_3\geq -\polyshort/K$, it suffices to show that $|\Term_{6}|\leq \polyshort/K$ for some $c_1$-dependent constant $\polyshort>0$. By Cauchy-Schwarz inequality, we have 
  \begin{align}
      |\Term_{6}|
      &\leq
     \Term_{6x}\cdot \Term_{6y}\cdot \Term_{6z},
  \end{align}
  where 
  \begin{align}
  \Term_{6x}&\defn
  \sqrt{\E\Bigg[
    {\Big[\frac{\sum_{j \in [K-1]}  \exp(S_\star(\xi, \xt_{, j}))}{K}+\frac{\exp(S_\star(\xi, \xt))}{K}-\E_{\xt\sim\P_{\txt}} \exp(S_\star(\xi, \xt))\Big]^4} \Bigg]},\\
    \Term_{6y}
    &\defn
   \Bigg({\E\Bigg[
    {\Big[\frac{\sum_{j \in [K-1]}  \exp(S_\star(\xi, \xt_{, j}))}{K}+\frac{\exp(S_\star(\xi, \xt))}{K}\Big]^{-4}} \Bigg]}\Bigg)^{1/4},\\
    \Term_{6x}&\defn
   \Bigg({\E\Bigg[
      {\Big[\E_{\xt\sim\P_{\txt}} \exp(S_\star(\xi, \xt))\Big]^{-4}} \Bigg]}\Bigg)^{1/4}.
  \end{align}
  Applying Jensen's inequality on $\Term_{6y},\Term_{6z}$ and using Assumption~\ref{ass:bounded_score_alternative}, it is readily verified that $\Term_{6y},\Term_{6z}\leq \polyshort$ for some constant $\polyshort>0$. 
For $\Term_{6x}$, 
  introduce the shorthand  $\Delta(\xi,\xt)= \exp(\truesimscore(\xi,\xt))-\E_{\xt\sim\P_{\txt}} \exp(\truesimscore(\xi,\xt))$. Then we have $\E_{\xt\sim\P_{\txt}} [\Delta(\xi, \xt)]=0$ and
  \begin{align}
    \Term_{6x}
    &\leq
    c\Bigg(
       \sqrt{\E\Big[\frac{\sum_{j \in [K-1]} \Delta(\xi, \xt_{, j})}{K}\Big]^4}
      +
       \frac{1}{K^2}    \sqrt{\E_{(\xi,\xt)\sim\P_{\image,\txt}}[\Delta(\xi, \xt)^4]} 
    \Bigg)\\
    &\leq
    c\Bigg(
      \frac{1}{K^2} \sqrt{\E\Big[{\sum_{i,j \in [K-1]}  \Delta(\xi, \xt_{, i})^2\Delta(\xi, \xt_{, j})^2}\Big]}+
       \frac{\polyshort}{K^2} 
    \Bigg)\\
    &\leq
    c\Bigg(
      \frac{1}{K} \sqrt{\E\Big[{\frac{1}{K}\sum_{i \in [K-1]}  \Delta(\xi, \xt_{, i})^4}\Big]}+
       \frac{\polyshort}{K^2} 
    \Bigg)\leq \polyshort/K
   \end{align} for some universal constant $c>0$, where the first line uses the fact that $\sqrt{\E|X+Y|^4}\leq c(\sqrt{\E|X|^4}+\sqrt{\E|Y|^4})$ for some universal constant $c>0$, the second line follows Assumption~\ref{ass:bounded_score_alternative}, the conditional independence of $\{\xt_{,j}\}_{j=1}^{K-1}$  given $\xi$, and the property that $\E_{\xt\sim\P_{\txt}} [\Delta(\xi, \xt)]=0$. 
   The last line uses Jensen's inequality and Assumption~\ref{ass:bounded_score_alternative}. Combining the bounds on $\Term_{6x},\Term_{6y},\Term_{6z}$ yields $|\Term_{6}|\leq \polyshort/K$ and therefore $\Term_3\geq -\polyshort/K$.

\end{proof}

}

\subsection{Proof of \Cref{prop:validity_zero_shot_classification}}\label{app:proof_validity_zero_shot_classification}

\begin{proof}[Proof of \Cref{prop:validity_zero_shot_classification}]
Recall the conditional probabilities
\begin{align}
\softmaxscoreinf{\scorefun}( \xt | \xi) 
=
\frac{\exp( S(\xi, \xt)) \P(\xt)}{\sum_{\xt'} \exp( S(\xi, \xt')) \P(\xt')},~~~
\clipfntprob{M}{\scorefun}( y | \xi) \defn \frac{\sum_{j=1}^M \exp( S(\xi, \xt^\jth(y))) \P( y)}{\sum_{y\in\cY}\sum_{j=1}^M \exp( S(\xi, \xt^\jth(y))) \P( y)}\notag.
\end{align}
and define the infinite-sample probability
\begin{align}
\softmaxscoreinf{\scorefun}( y | \xi) 
\defn 
\frac{\sum_{\xt'} \exp( S(\xi, \xt')) \P(\xt', y)}{\sum_{\xt'} \exp( S(\xi, \xt')) \P(\xt')}\notag.
\end{align}

We will prove that\footnote{We abuse the notation $\P(\cdot)$ for $\P_{\cls|\image}(\cdot),\P_{\txt|\image}(\cdot)$ when it is clear from the context.}
\begin{subequations}
\begin{align}
    \E_{\xi\sim\P_\image}\Big[ \KL\Big( \P_{\cls|\image}(y | \xi) \Big|\Big|  \softmaxscoreinf{\scorefun}(y | \xi) \Big) \Big] \
    &\le 
    \E_{\xi\sim\P_\image}\Big[ \KL\Big( \P(\xt | \xi) \Big|\Big|  \softmaxscoreinf{\scorefun}(\xt | \xi) \Big) \Big] \label{eq:scratch_prop_concentration_1},\text{~~and~~}
    \\
     \E_{\xi\sim\P_\image}\Big[
    \Renyitwo\Big( \softmaxscoreinf{\scorefun}(y | \xi) \Big|\Big|  \clipfntprob{M}{\simscore}(y | \xi) \Big)
    \Big]
    &\leq
    \polyshort   \frac{\log(2/\delta)}{{M}}~ \text{with probability at least }1-\delta,
\label{eq:scratch_prop_concentration_2}
\end{align}
where we define the
$\alpha$-Rényi divergence 
\begin{align}
\Renyi\Big( \softmaxscoreinf{\scorefun}(y | \xi) \Big|\Big|  \clipfntprob{M}{\simscore}(y | \xi) \Big)
\defn 
\frac{1}{\alpha-1}\log\Big(\E_{y\sim\softmaxscoreinf{\scorefun}(\cdot|\xi)}\Big[\Big(\frac{\softmaxscoreinf{\scorefun}(y | \xi)}{\clipfntprob{M}{\simscore}(y | \xi)}\Big)^{\alpha-1}
\Big]\Big)
\end{align} for any $\alpha>1.$

Given these results,
\Cref{prop:validity_zero_shot_classification} follows from~\Cref{prop:approximate_global_min_contrastive_loss}, 
combined with a triangle-like inequality for KL divergence (see e.g., Lemma~26 of Bun and Steinke~\cite{bun2016concentrated}), which states that for any tuple of  distributions $(\P,\mathbb{Q},\R)$,
\begin{align}
   \KL\Big( \P \Big|\Big| \R \Big) \leq
     \KL\Big( \P \Big|\Big| \mathbb{Q} \Big) 
     +
       \Renyitwo\Big(\mathbb{Q} \Big|\Big| \R \Big) .
\end{align}
\end{subequations}\\

\myunder{Proof of bound~\eqref{eq:scratch_prop_concentration_1}.}
Observe that 
\begin{align}
    \P( y | \xi)
 &= 
 \sum_{\xt'} \P( y | \xt',\xi)\cdot \P(\xt'|\xi)
\overset{(i)}{=}
\sum_{\xt'} \P( y | \xt')\cdot \P(\xt'|\xi),\notag
\\
   \softmaxscoreinf{\scorefun}( y | \xi)
 &
\overset{(ii)}{=}
\sum_{\xt'} \P( y | \xt')\cdot \softmaxscoreinf{\scorefun}(\xt'|\xi)\notag,
\end{align}
where step~(i) follows from \Cref{ass:conditional_independence} and step~(ii) uses the definitions of  $\softmaxscoreinf{\scorefun}( y | \xi)$ and $\softmaxscoreinf{\scorefun}(\xt'|\xi)$.  Therefore,  it follows from the data-processing inequality that
\begin{align}
 \KL\Big( \P(y | \xi) \Big|\Big|  \softmaxscoreinf{\scorefun}(y | \xi) \Big) 
 \le  \KL\Big( \P(\xt | \xi) \Big|\Big|  \softmaxscoreinf{\scorefun}(\xt | \xi) \Big),~~~\text{ for all } \xi\in\cXi.
\end{align}
Taking expectation over $\xi$ yields
bound~\eqref{eq:scratch_prop_concentration_1}. \\

\myunder{Proof of bound~\eqref{eq:scratch_prop_concentration_2}.}
 To prove bound~\eqref{eq:scratch_prop_concentration_2}, 
a key component is to establish
\begin{align}
   \E_{\xi}\Big[\sum_{y\in\cY}\Big[ \frac{ |\softmaxscoreinf{\scorefun}(y | \xi)-\clipfntprob{M}{\scorefun}(y | \xi) |^2 }{\clipfntprob{M}{\scorefun}(y | \xi) }\Big]\Big]
    &\leq 
   \polyshort\cdot \frac{\log(2/\delta)}{{M}} 
    \label{eq:alpha_divergence_2}
\end{align} with probability at least $1-\delta$ for some constant $\polyshort>0$ polynomially depending on $\boundscoreconst$ in Assumption~\ref{ass:bounded_score}. We will prove this at the end of the  section. Using claim~\eqref{eq:alpha_divergence_2}, we have
\begin{align}
 \E_{\xi}\Big[ \Renyitwo\Big( \softmaxscoreinf{\scorefun}(y | \xi) \Big|\Big|  \clipfntprob{M}{\simscore}(y | \xi) \Big)\Big]
   &=
 \E_{\xi}\Big[\log\E_{y\sim\softmaxscoreinf{\scorefun}(y|\xi)}\Big[ \frac{ \softmaxscoreinf{\scorefun}(y | \xi) }{\clipfntprob{M}{\scorefun}(y | \xi) }\Big]\Big]\\
 &\leq 
\E_{\xi}\E_{y\sim\softmaxscoreinf{\scorefun}(y|\xi)}\Big[ \frac{ \softmaxscoreinf{\scorefun}(y | \xi)-\clipfntprob{M}{\scorefun}(y | \xi)  }{\clipfntprob{M}{\scorefun}(y | \xi) }\Big]\\
 &= 
\E_{\xi}\Big[\sum_{y\in\cY}\Big[ \frac{ |\softmaxscoreinf{\scorefun}(y | \xi)-\clipfntprob{M}{\scorefun}(y | \xi) |^2 }{\clipfntprob{M}{\scorefun}(y | \xi) }\Big]\Big]\\
   &\leq
   \polyshort\cdot
   \frac{\log(2/\delta)}{{M}} \label{eq:alpha_divergence_3} 
\end{align}
with probability at least $1-\delta.$ This concludes the proof of bound~\eqref{eq:scratch_prop_concentration_2}.\\

Now, it remains to establish bound~\eqref{eq:alpha_divergence_2}. By properties of sub-exponential variables, it suffices to show the Orlicz norm (see e.g.,~\cite{vershynin2018high,wainwright_2019})
\begin{align}
    \orliczbig{ \E_{\xi}\Big[\sum_{y\in\cY}\Big[ \frac{ |\softmaxscoreinf{\scorefun}(y | \xi)-\clipfntprob{M}{\scorefun}(y | \xi) |^2 }{\clipfntprob{M}{\scorefun}(y | \xi) }\Big]\Big]
    }{1} \leq 
   \frac{\polyshort}{{M}}\label{eq:pf_orlicz_bound_1}
\end{align}
for some constant $\polyshort>0.$

Define the quantities
\begin{align}
    \shortempprob_1(y)
    &\defn 
   \sum_{\xt'} \exp( S(\xi, \xt')) \P(\xt', y),~~~~ \shortempprob_2
   \defn {\sum_{\xt'} \exp( S(\xi, \xt'))
   \P(\xt')},
   \\
    \shortempprob_3(y)
    &\defn 
  \frac{1}{M}\sum_{j=1}^M \exp( S(\xi, \xt^\jth(y))) \P( y),~~~~ \shortempprob_4
   \defn
   {\sum_{y\in\cY}\frac{1}{M}\sum_{j=1}^M \exp( S(\xi, \xt^\jth(y))) \P( y)}.
\end{align} 
Then $\softmaxscoreinf{\scorefun}(y | \xi)=\shortempprob_1(y)/\shortempprob_2,\clipfntprob{M}{\scorefun}(y | \xi)=\shortempprob_3(y)/\shortempprob_4$ and 
\begin{align}
  \frac{ |\softmaxscoreinf{\scorefun}(y | \xi)-\clipfntprob{M}{\scorefun}(y | \xi) |^2 }{\clipfntprob{M}{\scorefun}(y | \xi) }
  =
  \frac{(\shortempprob_1(y)\shortempprob_4-\shortempprob_2\shortempprob_3(y))^2}{\shortempprob_2^2\shortempprob_3(y)\shortempprob_4}
  \leq
  \frac{2[(\shortempprob_1(y)-\shortempprob_3(y))\shortempprob_4]^2
  +
  2[(\shortempprob_4-\shortempprob_2)\shortempprob_3(y)]^2}{\shortempprob_2^2\shortempprob_3(y)\shortempprob_4}. 
  \label{eq:chi_square_diff_expr1} 
\end{align}

\begin{subequations}
By \Cref{ass:bounded_score} and concentration properties of bounded random variables, there exists  constant $\polyshort>0$ such that   the Orlicz norm
\begin{align}
    \orlicz{ \shortempprob_3(y)-\shortempprob_1(y)}{2}\leq 
    \polyshort\cdot\frac{\P(y)}{\sqrt{M}} \label{eq:alpha_divergence_pf_1}
\end{align}
for all $y\in\cY,\xi\in\cXi$. Summing over  $y\in\cY$ and using the triangle inequality, we obtain\begin{align}
  \orlicz{\shortempprob_4-\shortempprob_2}{2}
  \leq
\frac{\polyshort}{\sqrt{{M}}}
   \label{eq:alpha_divergence_pf_2}.
\end{align}
Moreover, \Cref{ass:bounded_score} implies that
\begin{align}
  \shortempprob_2, {\shortempprob_4}
   &\in
   [1/\polyshort,\polyshort],~~~\text{and ~~~}
  \shortempprob_1(y), {\shortempprob_3(y)}
 \in
 \Big[\frac{\P(y)}{\polyshort},\polyshort\cdot\P(y)\Big]
   \label{eq:alpha_divergence_pf_3}
\end{align}
for all $\xi\in\cX_\image,y\in\cY$ for some constant $\polyshort>0$. 
\end{subequations}

Substituting equation~\eqref{eq:alpha_divergence_pf_1},~\eqref{eq:alpha_divergence_pf_2}~and~\eqref{eq:alpha_divergence_pf_3} into equation~\eqref{eq:chi_square_diff_expr1} and using properties of the Orlicz norm, we find
\begin{align}
    \orliczbig{ \frac{ |\softmaxscoreinf{\scorefun}(y | \xi)-\clipfntprob{M}{\scorefun}(y | \xi) |^2 }{\clipfntprob{M}{\scorefun}(y | \xi) }}{1}\leq \polyshort\cdot\frac{\P(y)}{{M}}\label{eq:pf_orlicz_bound_2}
\end{align} for all $\xi\in\cX_\image,y\in\cY$ for some constant $\polyshort>0$. Finally, summing equation~\eqref{eq:pf_orlicz_bound_2} over $y\in\cY$ and invoking Jensen's inequality yields equation~\eqref{eq:pf_orlicz_bound_1}.

\end{proof}

\subsection{Proof of \Cref{prop:CDM_CLIP_error}}\label{app:proof_CDM_CLIP_error}

\begin{proof}[Proof of \Cref{prop:CDM_CLIP_error}]
Recall that  $\bz_t=t \cdot \xi + \sqrt{t} \cdot \bg$.
By definition,
\begin{align}
&~\quad\E_{( \xt, \bz_t)}
\Big[ \big\| \bbm_t(\bz_t, \xt) - \Mthat(\bz_t, \Et(\xt)) \big\|_2^2 \Big]\\
&=
\E_{( \xt, \bz_t)}
\Big[ \big\| \E[\xi| \bz_t, \xt] -  \E[\xi| \bz_t, \Et(\xt)]\big\|_2^2 \Big]\\
&\overset{(i)}{\leq}
4\dim_\image\boundimage^2\cdot 
\E_{(\xt,\bz_t)}[\TV^2(\P(\xi | \xt, \bz_t), \P(\xi | \Et(\xt), \bz_t))] \\
&\overset{(ii)}{\leq}
2\dim_\image\boundimage^2\cdot 
\E_{(\xt,\bz_t)}[\KL(\P(\xi | \xt, \bz_t)|| \P(\xi | \Et(\xt), \bz_t))] 
\\
&\overset{(iii)}{\leq}
2\dim_\image\boundimage^2\cdot 
\E_{\xt}[\KL(\P(\xi | \xt)||\P(\xi | \Et(\xt)))] 
=
2\dim_\image\boundimage^2\cdot \suffmeasure(\Et),
\end{align}
where step~(i) follows since
\begin{align}
&\qquad\big\| \E[\xi| \bz_t, \xt] -  \E[\xi| \bz_t, \Et(\xt)]\big\|_2^2\\
&=
\Big\|\int \xi [\P(\xi|  \bz_t, \xt)
-\P(\xi| \bz_t, \Et(\xt))]d\xi
\Big\|_2^2 \\
&\leq 
\Big(\int \|\xi\|_2 \cdot|\P(\xi|  \bz_t, \xt)
-\P(\xi| \bz_t, \Et(\xt))|d\xi\Big)^2\\
&\leq 4\dim_\image\boundimage^2\cdot\TV^2(\P(\xi|  \bz_t, \xt)
,\P(\xi| \bz_t, \Et(\xt))),
\end{align}
step~(ii) uses Pinsker's inequality, step~(iii) uses the data processing inequality.
\end{proof}

\subsubsection{Proof of \Cref{cor:diffusion_sampling_error}}\label{sec:proof_diffusion_sampling_error}

Consider the process
\begin{align}
    \bwimage_t=t\cdot\xi+\de\bnoise_t
\end{align} where $\xi\sim\P_{\image|\txt}(\cdot|\xt)$ and $(\bnoise_t)_{t\geq0}$ is the Brownian motion on $\R^{\dimimage}$.
Since $\|\xi\|_2\leq \sqrt{\dim_\image}\times\boundimage$, it follows from Proposition 1 in~\cite{montanari2023sampling} that $(\bwimage_t)_{t\geq0}$ is the unique solution to the  stochastic differential equation (assuming $\bwimage_0=\mathbf{0}$)
\begin{align}
   \de\bwimage_t = \truepostmean_t(\bwimage_t,\xt)\de t+\de\bnoise_t,~~t\geq 0,  
\end{align}where $\truepostmean_t(\bz,\xt)=\E[\xi|\bz = t\xi+\sqrt{t}\gnoise,\xt]$. Let $\bwimagedist{T}=\bwimagedist{T}(\cdot|\xt)$ denote the distribution of $\bwimage_T/T$.  It follows immediately that $\probngauss{\frac{1}{\sqrt{T}}}=\bwimagedist{T}$. 
Therefore, 
\begin{align}
   \E_{\xt}\KL(\probngauss{\frac{1}{\sqrt{T}}}(\cdot | \xt)||\abwimagedist{T}(\cdot | \xt))  
    &\overset{(i)}{=}
   \E_{\xt} \KL(\P_{\bwimage_T}||\P_{\abwimage_T} )
   \overset{(ii)}{\leq}
    \frac{1}{2}\E_{\xt}\int_{0}^T\E_{\bwimage_t}\|\truepostmean_t(\bwimage_t,\xt)-\Mthat(\bwimage_t,\Et(\xt))\|_2^2\de t,
    \\
    &\overset{(iii)}{\leq}
    \dim_\image\boundimage^2T\cdot\suffmeasure(\Et),
\end{align}
where step~(i) uses the scale invariance of KL divergence, step~(ii) uses Girsanov theorem (Lemma~\ref{lm:girsanov}), and step~(iii) follows from Proposition~\ref{prop:CDM_CLIP_error} and the fact that $(\xt,\bwimage_t)\overset{d}{=}(\xt,\bz_t)$, where $\bz_t=t\cdot\xi+\sqrt{t}\cdot\gnoise.$

\subsection{Proof of \Cref{prop:VLM_CLIP_error}}\label{app:proof_VLM_CLIP_error}
We claim that the minimizer $\estprobnext$ in~\eqref{eqn:VLM_with_CLIP} is
\begin{align}
    \estprobnext(\, \cdot \,|\sE , \, \square\,) = \P_{\image, \txt}( \xtsub{i} = \, \cdot \, | \Ei(\xi) = \sE , \xtsub{1:i-1} = \, \square\,).\label{eq:vlm_pf_clipprob_claim}
\end{align}
By the tensorization property of KL divergence and the expression of $\trueprobnext,\estprobnext$ (in Eq.~\ref{eqn:VLM_true_mu},~\ref{eq:vlm_pf_clipprob_claim}), we have
\begin{align}
    \distance(\trueprobnext, \estprobnext)
&=
\E_{ (\xi, \xt) \sim \P_{\image,\txt}}
\Big[ \sum_{i \in [\dimtxt]} \KL\Big( \trueprobnext(\xtsub{i}| \xi, \xtsub{1:i-1} ) \Big|\Big| \estprobnext(\xtsub{i}| \Ei(\xi),\xtsub{1:i-1}) \Big) \Big]
\\
&=
\E_{ \xi \sim \P_{\image}}
\Big[\KL\Big( \prod_{i=1}^{\dimtxt}\trueprobnext(\xtsub{i}| \xi, \xtsub{1:i-1} ) \Big|\Big| \prod_{i=1}^{\dimtxt}\estprobnext(\xtsub{i}| \Ei(\xi),\xtsub{1:i-1}) \Big) \Big]
\\
&=
\E_{ \xi \sim \P_{\image}}
[\KL(\P_{\image|\txt}(\cdot|\xi)||\P_{\image|\txt}(\cdot|\Ei(\xi))\,)\,]
=
\suffmeasure(\Ei).
\end{align}

It remains to establish Eq.~\eqref{eq:vlm_pf_clipprob_claim}. Note that this follows immediately from the fact that
\begin{align}
   & \quad\arg\min_{\probnext \in \probnextspace} \Big\{ \risk_{\vlm}(\probnext, \Ei) \defn \E_{ (\xi, \xt) \sim \P_{\image,\txt}}
\Big[ \sum_{i \in [\dimtxt]} - \log \probnext(\xtsub{i}| \Ei(\xi),\xtsub{1:i-1}) \Big] \Big\}\\
&=
 \arg\min_{\probnext \in \probnextspace} \Big\{ \risk_{\vlm}(\probnext, \Ei) \defn \E_{ (\xi, \xt) \sim \P_{\image,\txt}}
\Big[ \sum_{i \in [\dimtxt]}\KL(\P(\xtsub{i}| \xi,\xtsub{1:i-1})||\probnext(\xtsub{i}| \Ei(\xi),\xtsub{1:i-1}) \Big] \Big\},\label{eq:vlm_pf_clipprob_claim_eq1}
\end{align}
and for each $i\in[\dimtxt]$, the KL divergence in~\eqref{eq:vlm_pf_clipprob_claim_eq1} is minimized when \begin{align}\probnext(\xtsub{i}| \Ei(\xi),\xtsub{1:i-1})=\P_{\image,\txt}(\xtsub{i}| \Ei(\xi),\xtsub{1:i-1})~~\text{for all~~}\xtsub{i}\in\cXtsub{i}.
\end{align}

\subsection{Proof of \Cref{prop:representation_equivalence}}\label{app:proof_representation_equivalence}

Define $\trueEmatrix\defn\big(\P(\xi)\trueEi^\top(\xi)\big)_{\xi\in\cXi}\in\R^{|\cXi|\times\truedimclip}$and  introduce the pseudoinverse 
\begin{align}\trueEmatrix^\dagger\defn \big[\big(\E_{\xi \sim \P_{\image}} [\trueEi(\xi) \trueEi(\xi)^\sT]\big)^{-1}\trueEi(\xi)~\big]_{\xi\in\cXi}\in\R^{\truedimclip\times|\cXi|}.
\end{align}
It can be verified that $\trueEmatrix^\dagger\trueEmatrix=\id_{\truedimclip}$ and 
\begin{align}
    \E_{\xi\sim\P_\image}\|\trueEmatrix^\dagger(\xi)\|_2^2
   =
  \tr\big( \big(\E_{\xi \sim \P_{\image}} [\trueEi(\xi) \trueEi(\xi)^\sT]\big)^{-1}\,\big)
\leq 
\Binvbd^2\truedimclip.
\end{align}

Define the embedding
\begin{align}
    \widetilde \Et(\xt) \defn \trueEmatrix^\dagger \diag(\P(\xi))  \big(\truescorelink^{-1}(\scorefun(\xi,\xt)-\log\E_{\xi\sim\P_\image}[\exp(\scorefun(\xi,\xt))])\,\big)_{\xi,\xt}.
\end{align}
In the proof we  bound the differences $\E_{\xt}[\|\widetilde \Et(\xt)-\trueEt(\xt)\|_2^2]$ and $\E_{\xt}[\|\widehat \Et(\xt)-\widetilde\Et(\xt)\|_2^2]$, respectively. Namely, we will show that
\begin{align}
\E_{\xt}[\|\widetilde \Et(\xt)-\trueEt(\xt)\|^2_2]\label{eq:repr_diff_claim_1}
&\leq \polyshort\Binvbd^2\truedimclip\tslinvLip^2\cdot{\E_{\xt\sim\P_\txt}\Big[ \KL\Big( \P_{\image|\txt}(\cdot | \xt) \Big|\Big|  \softmaxscoreinf{\scorefun}(\cdot | \xt) \Big) \Big]},
\end{align}
and there exists some parameters $(\adaweighta,\adaweightb)$ such that
\begin{align}
\E_{\xt}[\|\widehat \Et(\xt)-\widetilde \Et(\xt)\|^2_2]
&\leq 
\frac{\polyshort\Binvbd^2\truedimclip\tslinvLip^2}{{\diminadapt}}
,\label{eq:repr_diff_claim_2}
\end{align} 
and $\lops{\adaweighta}\leq\polyshort\boundadapt,\lops{\adaweightb}\leq \polyshort\Binvbd/\sqrt{\diminadapt}$.
Combining two bounds, we obtain 
\begin{align}
    \E_{\xt}[\| \what \Et(\xt) - \trueEt(\xt) \|_2^2] &\le
     \polyshort \cdot \Binvbd^2 \cdot \tslinvLip^2 \cdot \truedimclip\cdot \Big( {\E_{\xt\sim\P_\txt}\Big[ \KL\Big( \P_{\image|\txt}(\cdot | \xt) \Big|\Big|  \softmaxscoreinf{\scorefun}(\cdot | \xt) \Big) \Big]} + \frac{1}{\diminadapt}\Big)\label{eq:strong_adapt_bound}\\
     &\le \polyshort \cdot \Binvbd^2 \cdot \tslinvLip^2 \cdot \truedimclip\cdot ( \suffmeasure(\scorefun) + \diminadapt^{-1} ),
\end{align} where the second line uses~\Cref{def:sufficiency}. \\

\myunder{Proof of Eq.~\eqref{eq:repr_diff_claim_1}.}
By definition, we have
\begin{align}
   (\truescorelink^{-1}(\scorefun_\star(\xi,\xt)))_{\xi,\xt}=(\<\trueEi(\xi),\trueEt(\xt)\>)_{\xi,\xt}\in\R^{|\cXi|\times|\cXt|}.
\end{align}
Multiplying both sides by $\trueEmatrix^\dagger \diag(\P(\xi))$, we find
\begin{align}
    \trueEt(\xt) &=
    \trueEmatrix^\dagger \diag(\P(\xi))  (\truescorelink^{-1}(\scorefun_\star(\xi,\xt)))_{\xi,\xt}\\
    &=
    \trueEmatrix^\dagger \diag(\P(\xi))  \big(\truescorelink^{-1}(\scorefun_\star(\xi,\xt)-\log\E_{\xi\sim\P_\image}[\exp(\scorefun_\star(\xi,\xt))])\,\big)_{\xi},
\end{align}
where the last line follows since $\E_{\xi}[\exp(\scorefun_\star(\xi,\xt))] = \sum_{\xi}\P(\xi|\xt)=1.$
Introduce the shorthand
\begin{align}
    \Term(\scorefun(\xi,\xt))
    &\defn  \scorefun(\xi,\xt)-\log\E_{\xi\sim\P_\image}[\exp(\scorefun(\xi,\xt))]
\end{align} and let $\Delta^\Gamma(\xi,\xt)\defn\truescorelink^{-1}(    \Term(\scorefun(\xi,\xt)))-\truescorelink^{-1}(    \Term(\scorefun_\star(\xi,\xt))).$
Therefore,
\begin{align}
  \E_{\xt}[\|\widetilde \Et(\xt)-\trueEt(\xt)\|^2_2]
  &=  \E_{\xt}[\|\E_{\xi\sim\P_\image}[\trueEmatrix^\dagger(\xi) \Delta^\Gamma(\xi,\xt)]
  \|^2_2]\\
  &\leq
   \E_{\xt}\Big[\E_{\xi\sim\P_\image}\|\trueEmatrix^\dagger(\xi) 
\|_2^2 \cdot \E_{\xi\sim\P_\image}|\Delta^\Gamma(\xi,\xt)|^2\Big]\\
&
\leq
{\E_{\xi\sim\P_{\image}}[\|\trueEmatrix^\dagger(\xi) 
\|_2^2]\cdot\E_{(\xt,\xi)\sim\P_{\txt}\times\P_{\image}}[\Delta^\Gamma(\xi,\xt)^2]}\\
&
=\Binvbd^2\truedimclip\cdot{\E_{(\xt,\xi)\sim\P_{\txt}\times\P_{\image}}[\Delta^\Gamma(\xi,\xt)^2]},\label{eq:diff_encoder_pf_05}
\end{align}
where the second line uses Cauchy-Schwartz inequality, and the last line follows from the assumption on $\trueEmatrix^\dagger.$ Moreover,
\begin{align}
  \E_{(\xt,\xi)\sim\P_{\txt}\times\P_{\image}}[\Delta^\Gamma(\xi,\xt)^2]
&\leq
 \tslinvLip^2 \cdot\E_{\xt}\E_{\xi\sim\P_\image}\Big[ |   \Term(\scorefun(\xi,\xt))-    \Term(\scorefun_\star(\xi,\xt))|^2
\Big]
\\
&\leq \polyshort
 \tslinvLip^2 \cdot\E_{\xt}\E_{\xi\sim\P_{\image|\txt}(\cdot|\xt)}\Big[ |   \Term(\scorefun(\xi,\xt))-    \Term(\scorefun_\star(\xi,\xt))|^2
\Big]
.
\label{eq:diff_encoder_pf_1}
\end{align}
where the second line uses the fact that $\P(\xi)/\P(\xi|\xt)\leq \polyshort$ for some $\polyshort>0$  implied by Assumption~\ref{ass:bounded_score}.  It remains to bound $\E_{(\xi,\xt)\sim\P_{\image\times\txt}}\Big[|   \Term(\scorefun(\xi,\xt))-   \Term(\scorefun_\star(\xi,\xt))|^2\Big].$

Since adding any function of $\xi$ does not change the value of $\Term(\scorefun)$, w.l.o.g., we assume \begin{align}\E_{\xi\sim\P(\cdot|\xt))}[\scorefun(\xi,\xt)]=\E_{\xi\sim\P(\cdot|\xt))}[\scorefun_\star(\xi,\xt)]=0
\end{align}
and write  $\scorefun=\scorefun_{\star}+r \perturb$ with $r=\nrmp{\scorefun-\scorefun_\star}$ and $\perturb= (\scorefun-\scorefun_\star)/\nrmp{\scorefun-\scorefun_\star}$,
where 
\begin{align}
\nrmp{f}\defn\sqrt{\E_{(\xi,\xt)\sim\P_{\xi,\xt}}[f(\xi,\xt)^2]}
\end{align}

Similar to the proof of~\Cref{prop:approximate_global_min_contrastive_loss}, by a Taylor expansion w.r.t. $r$ at $0$, we find
\begin{align}
    &\qquad\E_{(\xi,\xt)}\Big[|   \Term(\scorefun(\xi,\xt))-   \Term(\scorefun_\star(\xi,\xt))|^2\Big]
    \\
    &\leq 
     \E_{(\xi,\xt)}\Big[\big|\scorefun_\star(\xi,\xt)-\scorefun(\xi,\xt)-(\log\E_{\xi\sim\P_\image}[\exp(\scorefun_\star(\xi,\xt))]
     -
     \log\E_{\xi\sim\P_\image}[\exp(\scorefun(\xi,\xt))])\big|^2
     \Big]\\
     &\leq  2\nrmp{\scorefun-\scorefun_\star}^2
     +
     2\E_{\xt}\big[\big|\E_{\xi\sim\softmaxscoreinf{\scorefun_\star+\widetilde r \perturb}}(\scorefun(\xi,\xt)-\scorefun_\star(\xi,\xt))\big|^2\big]\label{eq:diff_encoder_pf_2}
\end{align}
for some $\widetilde r=\widetilde r(\xt) \in[0,\nrmp{\scorefun-\scorefun_\star}]$, where for any given $\xt\in\cXt$ and score $\widetilde\scorefun$
\begin{align}
   \softmaxscoreinf{\widetilde\scorefun}(\xi|\xt)\defn \frac{\P(\xi)\cdot\exp(\widetilde\scorefun(\xi,\xt))}{\E_{\xi'\sim\P_\image}[\exp(\widetilde\scorefun(\xi',\xt))]}.
\end{align}
Since $\sup_{\xi\in\cXi,\xt\in\cXt}\softmaxscoreinf{\scorefun_\star+\widetilde\perturb}(\xi|\xt)/\P(\xi|\xt)\leq\polyshort$ for some constant $\polyshort>0$ by Assumption~\ref{ass:bounded_score}, it follows that
\begin{align}
&\quad\E_{\xt}\big[\big|\E_{\xi\sim\softmaxscoreinf{\scorefun_\star+\widetilde r \perturb}}(\scorefun(\xi,\xt)-\scorefun_\star(\xi,\xt))\big|^2\big]
\leq 
\polyshort\cdot\E_{\xt}\big[\E_{\xi\sim\P_{\image|\txt}}|\scorefun(\xi,\xt)-\scorefun_\star(\xi,\xt)|^2\big]\\
&\leq
\polyshort\cdot
\nrmp{\scorefun-\scorefun_\star}^2
\overset{(i)}{\leq} 
\polyshort\cdot\lim_{K\to\infty}\Big({\orisk_{\clip,\txt,K}(\scorefun)-\orisk_{\clip,\txt,K}(\scorefun_\star)}\Big)
\overset{(ii)}{=} 
\polyshort\cdot
\E_{\xt\sim\P_\txt}\Big[ \KL\Big( \P(\cdot | \xt) \Big|\Big|  \softmaxscoreinf{\scorefun}(\cdot | \xt) \Big) \Big],
\label{eq:diff_encoder_pf_3}
\end{align}
where step~(i) follows from Lemma~\ref{lm:bound_score_diff} for any $K\geq3$,  step~(ii) follows from the proof of   Proposition~\ref{prop:approximate_global_min_contrastive_loss}.
Combining Eq.~\eqref{eq:diff_encoder_pf_05},~\eqref{eq:diff_encoder_pf_1},~\eqref{eq:diff_encoder_pf_2}~and~\eqref{eq:diff_encoder_pf_3}  yields Eq.~\eqref{eq:repr_diff_claim_1}.\\

\myunder{Proof of Eq.~\eqref{eq:repr_diff_claim_2}.}
Let $\xi_{,1},\ldots,\xi_{,\diminadapt}$ be i.i.d. samples from $\P(\xi)$. 
We choose $\adaweighta=\trueEmatrix^\dagger_{\diminadapt}/\diminadapt\in\R^{\truedimclip\times \diminadapt}$ be the matrix consisting of the columns of $\trueEmatrix^\dagger$ that correspond to the samples  $\{\xi_{,j}\}_{j=1}^{\diminadapt}$. We choose $\adaweightb=(\Ei(\xi_{,j})\in\R^{\dimclip})_{j\in[\diminadapt]}\in\R^{\diminadapt\times\dimclip}$.
For any $\xt\in\cXt$, define
\begin{align}
    \Term(\xt) 
    &\defn (\scorelink(\Ei(\xi),\Et(\xt))-\log\E_{\xi\sim\P_\image}[\exp(\scorelink(\Ei(\xi),\Et(\xt)))]\,)_{\xi\in\cXi}\in\R^{|\cXi|},
    \\
    \Term^{(\diminadapt)}(\xt) 
    &\defn (\scorelink(\adaweightb_{,j:},\Et(\xt))-\log[\frac{1}{\diminadapt}\sum_{k=1}^{\diminadapt}\exp(\scorelink(\adaweightb_{,k:},\Et(\xt)))]\,)_{j\in[\diminadapt]}\in\R^{\diminadapt},
    \\
     \widetilde\Term^{(\diminadapt)}(\xt) 
    &\defn (\scorelink(\adaweightb_{,j:},\Et(\xt))-\log\E_{\xi\sim\P_\image}[\exp(\scorelink(\Ei(\xi),\Et(\xt)))]\,)_{j\in[\diminadapt]}\in\R^{\diminadapt}.
\end{align}
Then we have 
\begin{align}
  \E_{\xt}[\|\widetilde \Et(\xt)-\widehat\Et(\xt)\|^2_2]
  &\leq
  2\residual_{E1}+ 2\residual_{E1},
  \end{align}
  where
    \begin{align}
     \residual_{E1}
     &\defn \E_{\xt}[\|\trueEmatrix^\dagger \diag(\P(\xi))\truescorelink^{-1}(    \Term(\xt))-\frac{1}{\diminadapt}\trueEmatrix_{\diminadapt}^\dagger \truescorelink^{-1}(    \widetilde\Term^{(\diminadapt)}(\xt))
  \|^2_2],\\
    \residual_{E2}
     &\defn \frac{1}{\diminadapt^2}\E_{\xt}[\|\trueEmatrix_{\diminadapt}^\dagger \truescorelink^{-1}(    \widetilde\Term^{(\diminadapt)}(\xt))-\trueEmatrix_{\diminadapt}^\dagger \truescorelink^{-1}(    \Term^{(\diminadapt)}(\xt))
  \|^2_2].
  \end{align}
For $\residual_{E1}$, since
\begin{align}
   \E[ \trueEmatrix_{\diminadapt}^\dagger \truescorelink^{-1}(    \widetilde\Term^{(\diminadapt)}(\xt))] = \trueEmatrix^\dagger \diag(\P(\xi))\truescorelink^{-1}(    \Term(\xt)),
\end{align} it follows that
\begin{align}
   \E[\residual_{E1}] 
   &= \E_{\xt}\E[\|\trueEmatrix^\dagger \diag(\P(\xi))\truescorelink^{-1}(    \Term(\xt))-\frac{1}{\diminadapt}\trueEmatrix_{\diminadapt}^\dagger \truescorelink^{-1}(    \widetilde\Term^{(\diminadapt)}(\xt))
  \|^2_2]\\
  &=
  {  \E_{\xt}
  \Big[\sum_{i\in[\truedimclip]} \Var\big[\frac{1}{\diminadapt}\trueEmatrix_{i,\diminadapt}^\dagger \truescorelink^{-1}(    \widetilde\Term^{(\diminadapt)}(\xt))\big]\Big]}
.
\end{align} Since the variance in the above equation is invariant under any translation of 
$\truescorelink^{-1}$, we can w.l.o.g. assume there exists a point $v_\star\in\R$ in the feasible range of $\truescorelink^{-1}$ such that $\truescorelink^{-1}(v_\star)\leq\tslinvLip.$ It follows immediately that
$\Var[\trueEmatrix_{i,\diminadapt}^\dagger \truescorelink^{-1}(    \widetilde\Term^{(\diminadapt)}(\xt))
 ]\leq\E[|\trueEmatrix_{i,1}^\dagger \truescorelink^{-1}(    \widetilde\Term^{(1)}(\xt))|^2]/\diminadapt
 \leq\polyshort\tslinvLip^2\E[\trueEmatrix_{i,1}^{\dagger2}]/\diminadapt$ for all $i\in[\truedimclip]$. Therefore, we further have
 \begin{align}
    \E[\residual_{E1}] 
   &\leq 
   \frac{\polyshort\tslinvLip^2}{{\diminadapt}}\cdot { \E_{\xt} \Big[\E\sum_{i\in[\truedimclip]}[\|\trueEmatrix_{i,1}^\dagger \|_2^2
 ]\Big] }
 \leq  
 \frac{\polyshort\tslinvLip^2\Binvbd^2\truedimclip}{{\diminadapt}}
 .
 \end{align}

Let $\widetilde\Delta^{\Gamma}(\xt)\defn \truescorelink^{-1}(    \widetilde\Term^{(\diminadapt)}(\xt))- \truescorelink^{-1}(    \Term^{(\diminadapt)}(\xt))\in\R^{\diminadapt}.$ Note that all entries of $\widetilde\Delta^{\Gamma}(\xt)$ are equal. 
For $\residual_{E2}$, we have
\begin{align}
    \E[\residual_{E2}]
    &=
     \frac{1}{\diminadapt^2}\E_{\xt}\E[\|\trueEmatrix_{\diminadapt}^\dagger \widetilde\Delta^{\Gamma}(\xt)
  \|^2_2]
 \overset{(i)}{\leq}
\E_{\xt}\E[\|\trueEmatrix_1^\dagger\widetilde\Delta_1^{\Gamma}(\xt)
  \|^2_2]\\
  & \overset{(ii)}{=}
  {\E_{\xt}\E_{\xi_{,1}}[\|\trueEmatrix^\dagger(\xi_{,1})\|_2^2
\cdot
\E_{(\xi_{,i})_{i=2}^{\diminadapt}}|\widetilde\Delta_1^{\Gamma}(\xt)
  |^2]}
  ,
\end{align}where step~(i) follows from the symmetry of on $\xi_{,1},\ldots,\xi_{,\diminadapt}$ and step~(ii) follows from properties of conditional expectation

Since by Assumption~\ref{ass:bounded_score}, concentration of bounded random variables and Lipschitz continuity of $f(x)=\log(x)$ on $[1/\boundscoreconst,\boundscoreconst]$, we have
\begin{align}
    \orlicz{|\Term^{(\diminadapt)}_i(\xt)-\widetilde\Term^{(\diminadapt)}_i(\xt)|}{2}
    \leq
    \frac{\polyshort}{\sqrt{\diminadapt}}
\end{align} for all fixed $\xt\in\cXt,\xi_{,1}\in\cXi$  and $i\in[\diminadapt]$.
It follows from properties of sub-Gaussian random variables and Assumption~\ref{ass:well_posed_representation} that
\begin{align}
    \E_{(\xi_{,i})_{i=2}^{\diminadapt}}[\E|\widetilde\Delta_1^{\Gamma}(\xt)
  |^2]
  &=\E_{(\xi_{,i})_{i=2}^{\diminadapt}}\big|\truescorelink^{-1}(    \widetilde\Term_1^{(\diminadapt)}(\xt))-\truescorelink^{-1}(    \Term_1^{(\diminadapt)}(\xt))
  \big|^2\\
  &\leq 
  \tslinvLip^2\cdot \E_{(\xi_{,i})_{i=2}^{\diminadapt}}|    \widetilde\Term_1^{(\diminadapt)}(\xt)-    \Term_1^{(\diminadapt)}(\xt)
  |^2
  \leq
  \frac{\polyshort\tslinvLip^2}{\diminadapt}.
\end{align}
Putting pieces together and using the Assumption on $\trueEmatrix^\dagger$ yields $\E[\residual_{E2}]\leq\polyshort\Binvbd^2\truedimclip\tslinvLip^2/{\diminadapt}.$

Lastly, under our choice of $\adaweighta,\adaweightb,$ we have
\begin{align}
   \E\lops{\adaweighta}^2
   \leq
    \E\|\adaweighta\|_F^2= {\diminadapt}\E \|\Ematrix^\dagger_{1}\|_2^2\leq \frac{\Binvbd^2}{\diminadapt},
    \\
     \E\lops{\adaweightb}^2
   \leq
    \E\|\adaweightb\|_F^2={\diminadapt}\E \|\Ei(\xi_{,1})\|_2^2=\boundadapt^2.
\end{align}
Combining these with
  the bounds on $\E[\residual_{E1}],\E[\residual_{E2}]$, we may find samples $(\xi_{,j})_{j\in[\diminadapt]}$ such that 
 \begin{align}
\tslinvLip^2\truedimclip\lops{\adaweighta}^2+\frac{\Binvbd^2\truedimclip\tslinvLip^2}{\boundadapt{\diminadapt}}\lops{\adaweightb}^2+ \residual_{E1}+\residual_{E2}\leq \frac{\polyshort\Binvbd^2\truedimclip\tslinvLip^2}{\diminadapt}.
\end{align}
Choosing $(\adaweighta,\adaweightb)$ based on these samples gives an encoder $\widehat\Et$ such that Eq.~\eqref{eq:repr_diff_claim_2} holds.

{
\begin{remark}[An improved bound]
The error bound in Eq.~\eqref{eq:bound_adaptation} can be improved to 
\begin{align}
\E_{\xt}[\| \what \Et(\xt) - \trueEt(\xt) \|_2^2] \le \polyshort \cdot \Binvbd^2 \cdot \tslinvLip^2 \cdot \truedimclip\cdot ( \suffmeasure(\Et) + \diminadapt^{-1} )\label{eq:bound_adaptation_improved}
\end{align}
if the link function $\scorelink$ and the image embedding $\Ei(\xi)$ are chosen such that \begin{align}
    \simscore(\xi,\xt)
    =
\scorelink(\Ei(\xi),\Et(\xt)) \defn \log \frac{\P_{\image|\txt}(\xi|\Et(\xt))}{\P_{\image}(\xi)}.\label{eq:score_choice_improved}
\end{align} 
Note that such a pair of $\scorelink$ and $\Ei(\xi)$ always exists as one can choose $\Ei(\xi)=\xi$ and $\scorelink(\xi,\Et(\xt))\\=\log \frac{\P_{\image|\txt}(\xi|\Et(\xt))}{\P_{\image}(\xi)}$.

To establish Eq.~\eqref{eq:bound_adaptation_improved}, echoing the notations in~\Cref{def:sufficiency},
it suffices to note that
\begin{align}
    \suffmeasure(\Et) 
    &= \E_{\xt\sim\P_\txt}\Big[\KL\Big( \P_{\image | \txt}(\cdot | \xt) \Big|\Big|  \what \P_\simscore(\cdot  | \xt) \Big)\Big]
\end{align}
under the choices in Eq.~\eqref{eq:score_choice_improved}, and recall that we have proved a stronger bound than Eq.~\eqref{eq:bound_adaptation} in Eq.~\eqref{eq:strong_adapt_bound}.
\end{remark}
}

\subsection{Details in the proof of Corollary~\ref{cor:sufficient_stats_FNF}}\label{sec:proof_corollary_1}
In the section, we explain how the Neyman-Fisher factorization theorem can be used to prove Corollary~\ref{cor:sufficient_stats_FNF}.
Recall the Neyman-Fisher factorization theorem (see e.g., Theorem~3.6~in~\cite{keener2010theoretical}):
\begin{quote}
{\bf Theorem 3.6~\cite{keener2010theoretical}.}  
Let $\mathcal{P} = \{P_\theta : \theta \in \Theta\}$ be a family of distributions dominated by a measure $\mu$, with densities $p_\theta$.  
A statistic $T(X)$ is sufficient for $\theta$ if and only if there exist measurable functions $g_\theta\geq0$ and $h\geq0$ such that
\[
p_\theta(x) = g_\theta(T(x)) \, h(x),  \,~~~ \text{for a.e. } x \text{ under } \mu.
\]
\end{quote}
Also, recall that we have the following decomposition in the proof of Corollary~\ref{cor:sufficient_stats_FNF}:
\[
\P_{\image| \txt}(\xi | \xt) = \exp\{- \const\} \cdot  \P_{\image}(\xi) \cdot \exp\{ \scorelink(\trueEi(\xi), \trueEt(\xt)) \}.
\]
In our setting, we can choose the parameter  $\theta = \xt$, the sample $X=\xi$, and the dominating measure $\mu$ be the counting measure. Moreover, we let
\[
T(\xi) = \trueEi(\xi),~
g_\theta(T(\xi)) = \exp\{\scorelink(\trueEi(\xi), \trueEt(\xt))\},~  
h(\xi) = \P_{\image}(\xi)\exp\{-\const\},
\]
so by Theorem~3.6~in~\cite{keener2010theoretical}, $\trueEi(\xi)$ is sufficient for $\xt$. The argument for $\trueEt(\xt)$ is symmetric.

\subsection{Properties of approximate sufficiency}\label{sec:properties_approx_suff}
\begin{lemma}\label{lm:equiv_diff_mi_suff}
    Under Definition~\ref{def:sufficiency}, we have
    \begin{subequations}
    \begin{align}
        \suffmeasure(\Ei) &= \mutinfor(\xi, \xt) - \mutinfor(\Ei(\xi), \xt),
         \label{eq:suff_equiv_def_c1}\\
          \suffmeasure(\Et)
          &=
          \mutinfor(\xi, \xt) - \mutinfor(\xi, \Et(\xt)). \label{eq:suff_equiv_def_c2}
    \end{align} Note that $\Et(\xt)$ (resp. $\Ei(\xi)$) is a sufficient statistics if and only if $\suffmeasure(\Et)$ (resp. $\suffmeasure(\Ei)$) is zero.
    Moreover,
    \begin{align}
        \suffmeasure(\Ei) &=\inf_{\Q: \R^p \to \Delta(\cXt)} \E_{\xi\sim\P_\image}\Big[\KL\Big( \P_{\txt | \image}(\cdot | \xi) \Big|\Big|  \Q(\cdot  | \Ei(\xi)) \Big) \Big],\label{eq:suff_equiv_def_c3} \\
  \suffmeasure(\Et) 
  &=\inf_{\Q: \R^p \to \Delta(\cXi)} \E_{\xt\sim\P_\txt}\Big[\KL\Big( \P_{\image | \txt}(\cdot | \xt) \Big|\Big|  \Q(\cdot  | \Et(\xt)) \Big) \Big]. \label{eq:suff_equiv_def_c4}
    \end{align}
    \end{subequations}
\end{lemma}
\begin{proof}[Proof of Lemma~\ref{lm:equiv_diff_mi_suff}]
Note that
\begin{align}
  &\mutinfor(\xi, \xt) - \mutinfor(\Ei(\xi), \xt)\\ 
  &=
\E_{(\xi,\xt)\sim\P_{\image,\txt}}\Big[\log\frac{\P(\xi,\xt)}{\P(\xi)\P(\xt)}\Big]
  -
  \E_{\Ei(\xi),\xt}\Big[\log\frac{\P(\Ei(\xi),\xt)}{\P(\Ei(\xi))\P(\xt)}\Big]
  \\
  &=
  \E_{\xi,\xt,\Ei(\xi)}\Big[\log\frac{\P(\xi,\xt)}{\P(\xi)\P(\xt)}
  -
  \log\frac{\P(\Ei(\xi),\xt)}{\P(\Ei(\xi))\P(\xt)}
  \Big]
 \\
 &=
 \E_{\xi,\xt,\Ei(\xi)}\Big[\log{\P(\xt|\xi)}
  -
  \log{\P(\xt|\Ei(\xi))}
  \Big]\\
   &=
\E_{\xi,\xt}\Big[\log{\P(\xt|\xi)}
  -
  \log{\P(\xt|\Ei(\xi))}
  \Big]=\suffmeasure(\Ei).
\end{align}
This gives the Eq.~\eqref{eq:suff_equiv_def_c1}.
Eq.~\eqref{eq:suff_equiv_def_c2} follows from the symmetry between image and text.

To establish Eq.~\eqref{eq:suff_equiv_def_c3}~and~\eqref{eq:suff_equiv_def_c4}, we note that
\begin{align}
    &\qquad\E_{\xi\sim\P_\image}\Big[\KL\Big( \P_{\txt| \image}(\cdot | \xi) \Big|\Big|  \Q(\cdot  | \Ei(\xi)) \Big) \Big]
    \\
    &=
     \E_{\xi\sim\P_\image}\Big[\KL\Big( \P_{ \txt|\image}(\cdot | \xi) \Big|\Big|  \P(\cdot  | \Ei(\xi)) \Big) \Big]
     +
    \E_{\xi\sim\P_\image   }\Big[\E_{\xt\sim\P_{\txt|\image } (\cdot|\xi)}\Big( \frac{\log \P(\xt  | \Ei(\xi)) }{\log  \Q(\xt  | \Ei(\xi)) } \Big) \Big]
    \\
    &\overset{(i)}{=}
     \E_{\xi\sim\P_\image}\Big[\KL\Big( \P_{ \txt|\image}(\cdot | \xi) \Big|\Big|  \P(\cdot  | \Ei(\xi)) \Big) \Big]
     +
     \E_{\xi\sim\P_\image}\Big[\E_{\xt\sim\P_{\txt|\image} (\cdot|\Ei(\xi))}\Big( \frac{\log \P(\xt  | \Ei(\xi)) }{\log  \Q(\xt  | \Ei(\xi)) } \Big) \Big]\\
     &=
     \suffmeasure(\Ei)
     +\E_{\xi\sim\P_\image}\Big[\KL\Big( \P_{\txt | \image}(\cdot | \Ei(\xi) )\Big|\Big|  \Q(\cdot  | \Ei(\xi)) \Big) \Big]\geq   \suffmeasure(\Ei),
\end{align}
where step~(i) follows since for any function $f(\xt,\Ei(\xi))$, we have
\begin{align}
    \E_{\xi}[\E_{\xt\sim\P_{\txt|\image}(\cdot|\xi)}f(\xt,\Ei(\xi))]
    &=
    \E_{\xi}[\E_{\xt\sim\P_{\txt|\image}(\cdot|\xi)}f(\xt,\Ei(\xi))|\Ei(\xi)]\\
    &=
  \E_{\xi}[  \sum_{\xt\in\cXt} \E[ 
     \P_{\txt|\image}(\xt|\xi) 
    |\Ei(\xi) ]f(\xt,\Ei(\xi))]\\
      &=
  \E_{\xi}[  \sum_{\xt\in\cXt}  
     \P_{\txt|\image}(\xt 
    |\Ei(\xi) )f(\xt,\Ei(\xi))]\\
    &=  \E_{\xi}[\E_{\xt\sim\P_{\txt|\image}(\cdot|\Ei(\xi))}f(\xt,\Ei(\xi))].
\end{align}
\end{proof}

\subsection{Auxiliary lemmas}

\begin{lemma}[Bounds on $\nrmp{\scorefun-\scorefun_\star}$]\label{lm:bound_score_diff}
 Under the assumptions and notations in \Cref{prop:approximate_global_min_contrastive_loss} and  its proof, 
we have
    \begin{align}
  \nrmp{\scorefun-\scorefun_\star}
  &\leq \polyshort\cdot \sqrt{  \orisk_{\clip, \star, K}(\scorefun)-\orisk_{\clip, \star, K}(\scorefun_\star)} ~~~\text{for} ~~~\star\in\{\image,\txt\},\\
 & \leq
  \polyshort\cdot \sqrt{  \orisk_{\clip,  K}(\scorefun)-\orisk_{\clip, K}(\scorefun_\star)} 
\end{align}
for some constant $\polyshort>0$ depending polynomially in $c_1$.
\end{lemma}
\begin{proof}[Proof of Lemma~\ref{lm:bound_score_diff}]
We only prove the lemma for $\star=\image$. The other case follows by symmetry between image and text. 
Note that  
\begin{align}
\orisk_{\clip, \image, K}(\scorefun)
&= \E_{\xib, (\xtb_{, j})_{j \in [K]}, \kb} \Big[ - \log \frac{\exp(\scorefun(\xib, \xtb_{, \kb})  ) }{\sum_{j \in [K]} \exp(\scorefun(\xib, \xtb_{, j}))} \Big]\\
&=
\E \Big[  \log {\sum_{j \in [K]} \exp(\scorefun(\xib, \xtb_{, j}))} \Big],
\end{align}
where the last line follows since $(\xib,\xtb_{,\kb})\sim\P_{\image,\txt}$ and we assume  
\begin{align}
 \E_{\xt|\xi\sim \P_{\txt|\image}}[\scorefun(\xi,\xt)]=\E_{\xt|\xi\sim \P_{\txt|\image}} [\scorefun_\star(\xi,\xt)]=0
 \end{align}
 for all $\xi\in\cX_\image$ in the proof of \Cref{prop:approximate_global_min_contrastive_loss}.

 Write $\scorefun=\scorefun_{\star}+\scorediffnorm \perturb$ with $\scorediffnorm=\nrmp{\scorefun-\scorefun_\star}$ and $\perturb= (\scorefun-\scorefun_\star)/\nrmp{\scorefun-\scorefun_\star}$.
 For any function $\perturb:\cXi\times\cXt\mapsto\R$ such that $\E_{\xt|\xi\sim \P_{\txt|\image}}[\perturb(\xi,\xt)]=0$ for all $\xi\in\cX_\image$  and $\E_{(\xi,\xt)\sim \P_{\image,\txt}}[\perturb^2(\xi,\xt)]=1$, 
it can be verified that for any $r\in\R$
\begin{subequations}
\begin{align}
    \partial_r \orisk_{\clip, \image, K}(\scorefun_\star+r\perturb) 
    &=
    \E_{\xib, (\xtb_{, j})_{j \in [K]}}\Big[\E_{k\sim \softmaxscore_{\scorefun_\star+r\perturb}}[\perturb(\xib,\xtb_{,k})]\Big],\label{eq:r_grad}
    \\
     \partial^2_r \orisk_{\clip, \image, K}(\scorefun_\star+r\perturb) 
    &=
    \E_{\xib, (\xtb_{, j})_{j \in [K]}}\Big[\Var_{k\sim \softmaxscore_{\scorefun_\star+r\perturb}}[\perturb(\xib,\xtb_{,k})]\Big],\label{eq:r_hessian}
    \\
    \partial^3_r \orisk_{\clip, \image, K}(\scorefun_\star+r\perturb) 
    &=
    \E_{\xib, (\xtb_{, j})_{j \in [K]}}\Big[\E_{k\sim \softmaxscore_{\scorefun_\star+r\perturb}}\big[\perturb(\xib,\xtb_{,k})
    -
    \E_{k\sim \softmaxscore_{\scorefun_\star+r\perturb}}[\perturb(\xib,\xtb_{,k})]
    \big]^3\Big].\label{eq:r_third}
\end{align}
\end{subequations}
We claim that
\begin{subequations}
    \begin{itemize}
        \item[(a).] $\orisk_{\clip, \image, K}(\scorefun_\star+r\perturb)$ is globally convex in $r\in\R$ and is strongly  convex at the minimizer $r=0$, namely, there exists some constant $\polyshort>0$ such that
        $ \partial^2_r \orisk_{\clip, \image, K}(\scorefun_\star+r\perturb) |_{r=0}  \geq 1/\polyshort$.
        \item[(b).] There exists some constant $\polyshort>0$ such that $|\partial^3_r \orisk_{\clip, \image, K}(\scorefun_\star+r\perturb) |\leq \polyshort/r_0$ for any $|r|\leq \scorediffnorm. $
    \end{itemize}
\end{subequations}
We will prove these claims later in this section.
With these two claims at hand, it follows from properties of convex functions that
\begin{align}
   \orisk_{\clip, \image, K}(\scorefun_\star+r\perturb)-\orisk_{\clip, \image, K}(\scorefun_\star) \geq
\begin{cases} 
   r^2/\polyshort  &\text{if}~~ |r|< r_0/\polyshortprime,\\
    r_0|r|/\polyshort  &\text{if}~~ |r|\geq r_0/\polyshortprime,
\end{cases} \label{eq:one_dim_cvx_bounds}
\end{align}
for some constants $\polyshort,\polyshortprime>1$ polynomially dependent on $\boundscoreconst$ in Assumption~\ref{ass:bounded_score}. The proof of equation~\eqref{eq:one_dim_cvx_bounds} is deferred to the end of this section. Finally, choosing $r=r_0$ in equation~\eqref{eq:one_dim_cvx_bounds} yields Lemma~\ref{lm:bound_score_diff}.\\

\myunder{Proof of claim~(a).}
The global convexity of $\orisk_{\clip, \image, K}(\scorefun_\star+r\perturb)$ follows immediately from equation~\eqref{eq:r_hessian} and the fact that the variance is non-negative. $r=0$ is a global minimizer of  $\orisk_{\clip, \image, K}(\scorefun_\star+r\perturb)$ because
\begin{align}
       \partial_r \orisk_{\clip, \image, K}(\scorefun_\star+r\perturb)|_{r=0} 
    &=
    \E_{\xib, (\xtb_{, j})_{j \in [K]}}\Big[\E_{k\sim \softmaxscore_{\scorefun_\star}}[\perturb(\xib,\xtb_{,k})]\Big]
    \overset{(i)}{=}\E_{\xib, (\xtb_{, j})_{j \in [K]},\kb}[\perturb(\xib,\xtb_{,\kb})]=0,
\end{align}
where step~(i) uses the fact that $\softmaxscore_{\scorefun_\star}$ is the posterior distribution of $\kb$ conditioned on $\xib, (\xtb_{, j})_{j \in [K]}.$

It remains to establish a lower bound on $ \partial_r^2 \orisk_{\clip, \image, K}(\scorefun_\star+r\perturb)|_{r=0}$. Note that
\begin{align}
    \Var_{k\sim\P_{\scorefun_\star}}(\perturb(\xib,\xtb_{,k})) 
    &= \sum_{i\neq j}\softmaxscore_{\scorefun_\star}(i)\softmaxscore_{\scorefun_\star}(j) \cdot (\perturb(\xib,\xtb_{,i})-\perturb(\xib,\xtb_{,j}))^2\\
    &\geq 
    \frac{1}{\polyshort K^2}\sum_{i,j\in[K]}(\perturb(\xib,\xtb_{,i})-\perturb(\xib,\xtb_{,j}))^2
\end{align}for some constant $\polyshort>0$ that depends on $\boundscoreconst$ polynomially,
where the second line uses Assumption~\ref{ass:bounded_score}, which implies that $\softmaxscore_{\scorefun_\star}(i)\in[1/\polyshortprime/K,\polyshortprime/K]$ for all $i\in[K]$ and some constant $\polyshortprime>0$. Therefore,
\begin{align}
    \partial_r^2 \orisk_{\clip, \image, K}(\scorefun_\star+r\perturb)|_{r=0}
    &=
      \E_{\xib, (\xtb_{, j})_{j \in [K]}}\Big[\Var_{k\sim \softmaxscore_{\scorefun_\star}}[\perturb(\xib,\xtb_{,k})]\Big]\\
      &\geq 
     \frac{1}{\polyshort K^2}\sum_{i,j\in[K]}  \E_{\xib, (\xtb_{, j})_{j \in [K]}} (\perturb(\xib,\xtb_{,i})-\perturb(\xib,\xtb_{,j}))^2\\
     &\overset{(ii)}{\geq}  
    \frac{1}{\polyshort} \cdot \E_{\xi}\big[\Var_{\xt\sim\P_{\txt|\image}}(\perturb(\xi,\xt))\big]
     =\frac{1}{\polyshort},
\end{align}
where step~(ii) uses the fact that for any $i\neq j$,
\begin{align}
&\quad\E_{\xib, (\xtb_{, j})_{j \in [K]}} (\perturb(\xib,\xtb_{,i})-\perturb(\xib,\xtb_{,j}))^2\\
&\geq 
\min\{\E_{\xi}\Var_{\xt\sim\P_{\txt|\image}}(\perturb(\xi,\xt)),\E_{\xi}\Var_{\xt\sim\P_{\txt}}(\perturb(\xi,\xt))\},
\end{align} 
and 
\begin{align}
   &\quad 
   \E_{\xi}\big[\Var_{\xt\sim\P_{\txt}}(\perturb(\xi,\xt))\big]\\
   &=
   \frac{1}{2}\E_{\xi}\big[\E_{\xt_{,1},\xt_{,1}\sim\P_{\txt}}[(\perturb(\xi,\xt_{,1})-\perturb(\xi,\xt_{,2}))^2]\big]\\
   &\geq
 \frac{1}{2} \E_{\xi}\Big[ \inf_{(\xt_{,1},\xt_{,2})\in\cXt^2} \frac{\P_{\txt}(\xt_{,1})\times\P_{\txt}(\xt_{,2})}{\P_{\txt|\image}(\xt_{,1})\times\P_{\txt|\image}(\xt_{,2})} \cdot 
 \E_{\xt_{,1},\xt_{,2}\sim\P_{\txt|\image}}[(\perturb(\xi,\xt_{,1})-\perturb(\xi,\xt_{,2}))^2]\Big]\\
 &\geq
 \frac{1}{\polyshort} \cdot \E_{\xi}\big[\Var_{\xt\sim\P_{\txt|\image}}(\perturb(\xi,\xt))\big].
\end{align}
Here the second inequality follows from the boundedness assumption on $\scorefun_\star$ in Assumption~\ref{ass:bounded_score}. This completes the proof of claim~(a).\\

\myunder{Proof of claim~(b).}
By definition,
\begin{align}
    &\quad \partial^3_r \orisk_{\clip, \image, K}(\scorefun_\star+r\perturb) \\
    &=
    \E_{\xib, (\xtb_{, j})_{j \in [K]}}\Big[\E_{k\sim \softmaxscore_{\scorefun_\star+r\perturb}}\big[\perturb(\xib,\xtb_{,k})
    -
    \E_{k\sim \softmaxscore_{\scorefun_\star+r\perturb}}[\perturb(\xib,\xtb_{,k})]
    \big]^3\Big]\\
    &\leq 
2\sup_{(\xi,\xt)\sim\cXi\times\cXt}\frac{|\scorefun(\xi,\xt)-\scorefun_\star(\xi,\xt)|}{\nrmp{\scorefun-\scorefun_\star}}
    \cdot 
      \E_{\xib, (\xtb_{, j})_{j \in [K]}}\Big[\Var_{k\sim \softmaxscore_{\scorefun_\star+r\perturb}}[\perturb(\xib,\xtb_{,k})]\Big]\\
      &\leq
      \frac{\polyshort}{\nrmp{\scorefun-\scorefun_\star}}\cdot 
       \E_{\xib, (\xtb_{, j})_{j \in [K]}}\Big[\Var_{k\sim \softmaxscore_{\scorefun_\star+r\perturb}}[\perturb(\xib,\xtb_{,k})]\Big]\leq \frac{\polyshort}{\nrmp{\scorefun-\scorefun_\star}},
\end{align}
where the second inequality uses Assumption~\ref{ass:bounded_score} and noting that 
\begin{align}
    &\quad\E_{\xib, (\xtb_{, j})_{j \in [K]}}\Big[\Var_{k\sim \softmaxscore_{\scorefun_\star+r\perturb}}[\perturb(\xib,\xtb_{,k})]\Big]
   \leq
     \E_{\xib, (\xtb_{, j})_{j \in [K]}}\Big[\E_{k\sim \softmaxscore_{\scorefun_\star+r\perturb}}[\perturb(\xib,\xtb_{,k})^2]\Big]\\
     &\overset{(i)}{\leq}
     \polyshort
      \E_{\xib, (\xtb_{, j})_{j \in [K]}}\Big[\E_{k\sim \softmaxscore_{\scorefun_\star}}[\perturb(\xib,\xtb_{,k})^2]\Big]
     =
   \polyshort  \E_{\xib, (\xtb_{, j})_{j \in [K]},\kb}[\perturb(\xib,\xtb_{,\kb})^2]= \polyshort,
\end{align}
where step~(i) uses Assumption~\ref{ass:bounded_score}, which implies that $\softmaxscore_{\scorefun_\star+r\perturb}/\softmaxscore_{\scorefun_\star}\in[1/\polyshort,\polyshort]$ for some $\polyshort>0$ depending polynomially on $\boundscoreconst$.

\myunder{Proof of claim~\eqref{eq:one_dim_cvx_bounds}.}
Using claim~(a),~(b)  and the properties of convex functions, we have
\begin{align}
    \orisk_{\clip, \image, K}(\scorefun_\star+r\perturb)-\orisk_{\clip, \image, K}(\scorefun_\star)
    &\geq \frac{1}{2}r^2-\frac{\polyshort}{\nrmp{\scorefun-\scorefun_\star}}|r|^3
    \end{align}
    for some constant $\polyshort>0$. It follows immediately that
    \begin{align}
     \orisk_{\clip, \image, K}(\scorefun_\star+r\perturb)-\orisk_{\clip, \image, K}(\scorefun_\star)
    &\geq \frac{1}{4}|r|^2
    \end{align}
    for $|r|\leq \nrmp{\scorefun-\scorefun_\star}/\polyshort$ for some constant $\polyshort>0.$
    Moreover, by claim~(a),~(b) and Newton-Leibniz formula
    \begin{align}
   \partial_r^2
   \orisk_{\clip, \image, K}(\scorefun_\star+r\perturb)|_{r=r_1}
  \geq
   -\frac{\polyshort}{\nrmp{\scorefun-\scorefun_\star}}\cdot|r_1|
   +
 \partial_r^2
   \orisk_{\clip, \image, K}(\scorefun_\star+r\perturb)|_{r=0}\geq \frac{1}{\polyshort}
\end{align}
when  $|r_1|\leq \nrmp{\scorefun-\scorefun_\star}/\polyshortprime$  for some constant $\polyshortprime>0.$ It then follows immediately that at $r=\pm\nrmp{\scorefun-\scorefun_\star}/\polyshortprime$
\begin{align}
   |  \partial_r
   \orisk_{\clip, \image, K}(\scorefun_\star+r\perturb)|\geq \frac{\nrmp{\scorefun-\scorefun_\star}}{\polyshort}.
\end{align}
for some constant $\polyshort>0.$

Now, let $\proj_{c}(x)=\argmin_{y\in[-c,c]}|x-y|$ be the  projection of $x$ to the interval $[-c,c]$. 
Putting pieces together, we can find some constant $\polyshortprime>0$ such that for any $|r|\leq\nrmp{\scorefun-\scorefun_\star}/\polyshortprime$,
\begin{align}
     \orisk_{\clip, \image, K}(\scorefun_\star+r\perturb)-\orisk_{\clip, \image, K}(\scorefun_\star)
    &\geq \frac{1}{4}|r|^2,
\end{align}
and for any $|r|>\nrmp{\scorefun-\scorefun_\star}/\polyshortprime$,
    \begin{align}
    \orisk_{\clip, \image, K}(\scorefun_\star+r\perturb)-\orisk_{\clip, \image, K}(\scorefun_\star)
    &\geq
    \orisk_{\clip, \image, K}(\scorefun_\star+r\perturb)- \orisk_{\clip, \image, K}(\scorefun_\star+\proj_{\nrmp{\scorefun-\scorefun_\star}/\polyshortprime}(r)\perturb)
    +
   \frac{\nrmp{\scorefun-\scorefun_\star}^2}{\polyshort}\\
   &\geq
   |  \partial_r
   \orisk_{\clip, \image, K}(\scorefun_\star+r\perturb)|\cdot
   |r-\proj_{\nrmp{\scorefun-\scorefun_\star}/\polyshortprime}(r)|+
    \frac{\nrmp{\scorefun-\scorefun_\star}^2}{\polyshort}\\
    &\geq 
     \frac{\nrmp{\scorefun-\scorefun_\star}\cdot |r|}{\polyshort},
    \end{align}
    where the second line uses properties of convex functions.
\end{proof}

\begin{lemma}[Bound on $\Term_{d2}$]\label{lm:bound_diff_second}
Recall the definition of $\Term_{d2}$ in equation~\eqref{eq:def_term_d2}. 
Under the assumptions and notations in Proposition~\ref{prop:approximate_global_min_contrastive_loss} and  its proof,   for some constant $\polyshort>0$
\begin{align}
    |\Term_{d2}(r)|\leq\frac{\polyshort\cdot\nrmp{\scorefun-\scorefun_\star}^2}{K}
\end{align} for all $r\in[0,\nrmp{\scorefun-\scorefun_\star}]$.
\end{lemma}
\begin{proof}[Proof of Lemma~\ref{lm:bound_diff_second}]
 Write $\scorefun=\scorefun_{\star}+r \perturb$ with $r=\nrmp{\scorefun-\scorefun_\star}$ and $\perturb= (\scorefun-\scorefun_\star)/\nrmp{\scorefun-\scorefun_\star}$. 
By the scaling property of variance and noting that $\Var_P(X)=\E_{X,Y\simiid P}(X-Y)^2/2$, it suffices to show 
 \begin{align}
 |\Varterm_1-\Varterm_2|\leq \frac{\polyshort}{K},\label{eq:var_diff_claim}
 \end{align}
 where
 \begin{subequations}
 \begin{align}
  \Varterm_1&\defn  \E_{\xib, (\xtb_{, j})_{j \in [K]},\kb;k_1,k_2\sim \softmaxscore_{\scorefun_\star+ r\perturb}}[\perturb(\xib,\xtb_{,k_1})-\perturb(\xib,\xtb_{,k_2})]^2
             ,\label{eq:varterm_1_def}\\
  \Varterm_2&\defn
\E_{\xib\sim\P_\image;\xt_{,1},\xt_{,2}\sim\softmaxscoreinf{\scorefun_\star+ r\perturb }}
         [\perturb(\xib,\xt_{,1})-\perturb(\xib,\xt_{,2})]^2.\label{eq:varterm_2_def}
 \end{align}
 \end{subequations}
 Let $\scorejoint(\cdot,\cdot|\xib)$ denote the joint distribution of $(\xtb_{,k_1},\xtb_{,k_2})$ conditioned on $\xib$ in the definition of $\Varterm_1$, and let $\scorejointinf(\cdot,\cdot|\xi)$ denote the joint distribution of $(\xt_{,1},\xt_{,2})$ conditioned on $\xib$ in the definition of  $\Varterm_2$.
 
 We claim that there exists some constant $\polyshort,\polyshortprime>0$ such that when $K\geq \polyshortprime$
\begin{align}
   \frac{\big|\scorejoint(\xt_{,a},\xt_{,b}|\xib)- 
    \scorejointinf(\xt_{,a},\xt_{,b}|\xib)\big|}
    { \scorejointinf(\xt_{,a},\xt_{,b}|\xib)}\leq \frac{\polyshort}{K}\label{eq:var_dist_diff_claim_1}
\end{align} 
for all $\xt_{,a},\xt_{,b}\in\cXt$ such that $\xt_{,a}\neq\xt_{,b}$.
 Given claim~\eqref{eq:var_dist_diff_claim_1} and adopting the shorthand notation $\diffperturb(\xib,\xt_{,1},\xt_{,2})=\perturb(\xib,\xt_{,1})-\perturb(\xib,\xt_{,2})$, we immediately obtain
 \begin{align}
  & |\Varterm_1-\Varterm_2|
  \\ &=
   \big|\E_{\xib\sim\P_\image;(\xt_{,1},\xt_{,2})\sim\scorejoint} [\diffperturb(\xib,\xt_{,1},\xt_{,1})^2]
   -
   \E_{\xib\sim\P_\image;(\xt_{,1},\xt_{,2})\sim\scorejointinf} [\diffperturb(\xib,\xt_{,1},\xt_{,1})^2]
   \big|\\
   &\leq
\E_{\xib\sim\P_\image}\big|\E_{(\xt_{,1},\xt_{,2})\sim\scorejoint} [\diffperturb(\xib,\xt_{,1},\xt_{,1})^2]
   -
   \E_{(\xt_{,1},\xt_{,2})\sim\scorejointinf} [\diffperturb(\xib,\xt_{,1},\xt_{,1})^2]
   \big|\\
   &\overset{(i)}{\leq}
   \frac{\polyshort}{K}\cdot
   \E_{\xib\sim\P_\image}\big[ \E_{(\xt_{,1},\xt_{,2})\sim\scorejointinf} [\diffperturb(\xib,\xt_{,1},\xt_{,1})^2]\big]
   =
    \frac{\polyshort}{K}\cdot \Varterm_2
   \overset{(ii)}{\leq}  \frac{\polyshort}{K},
 \end{align}
 where step~(i) uses equation~\eqref{eq:var_dist_diff_claim_1} and the fact that $\diffperturb(\xtb,\xt_{,1},\xt_{,2})=0$ when $\xt_{,1}=\xt_{,2}$; step~(ii) follows from Assumption~\ref{ass:bounded_score}, which implies $\sup_{\xt\in\cXt}\big[\softmaxscoreinf{\scorefun_\star+r\perturb}(\xt)/\P_{\txt|\image}(\xt|\xib)\big]\leq\polyshort$  for some constant $\polyshort>0$, and  the fact that $\E_{(\xt,\xi)\sim\P_{\txt,\image}}[\perturb(\xi,\xt)^2]=1$. 
 
 When $K< \polyshortprime$, it can be readily verified by Assumption~\ref{ass:bounded_score} and noting $\E_{(\xt,\xi)\sim\P_{\txt,\image}}[\perturb(\xi,\xt)^2]=1$ that equation~\eqref{eq:var_diff_claim} holds.
 This completes the proof of equation~\eqref{eq:var_diff_claim} and hence Lemma~\ref{lm:bound_diff_second}.\\

 \myunder{Proof of claim~\eqref{eq:var_dist_diff_claim_1}.}
In the expression of $\Varterm_1$ in equation~\eqref{eq:varterm_1_def}, we note that the  distribution of $(k_1,k_2)$ conditioned on $(\xib, (\xtb_{, j})_{j \in [K]},\kb)$ remains unchanged under any permutation of $(\xtb_{, j})_{j \in [K]}$. Therefore, without loss of generality, we can drop the implicit dependence on $\kb$  and assume
\begin{align}
    (\xtb_{, j})_{2\leq j \leq K}\overset{i.i.d.}{\sim} \P_{\txt},~~~\text{and}~~ \xtb_{,1}\sim\P_{\txt|\image}(\cdot|\xib).
\end{align}
To provide an overview, the proof consists of three  steps. First, we  
rewrite the expressions for $\scorejoint,\scorejointinf$ in terms of the  expectation of certain quantities conditioned on $\xib$. Second, we 
 introduce an additional distribution on $\cXt\times\cXt$, denoted by $\scorejointloo$, which connects two distributions  $\scorejoint,\scorejointinf$, and we bound the differences $|\scorejoint-\scorejointloo|,|\scorejointloo-\scorejointinf|$ separately. Finally, we combine the bounds to obtain claim~\eqref{eq:var_dist_diff_claim_1}.

 \paragraph{Rewriting the expressions for $\scorejoint$ and $\scorejointinf$. }
Adopt the shorthand notation $\scorefun_r=\scorefun_\star+r\perturb.$
By the definition of $\scorejoint$, for any $(\xt_{,a},\xt_{,b})\in\cXt\times\cXt$
\begin{align}
    \scorejoint(\xt_{,a},\xt_{,b}|\xib)
&=
\sum_{i,j=1}^K\E[\mathrm{1}_{\{k_1=i,k_2=j,\xtb_{,i}=\xt_{,a},\xtb_{,j}=\xt_{,b}\}}|\xib]\\
&=
\sum_{i,j=1}^K\E\big[\E[\mathrm{1}_{\{k_1=i,k_2=j,\xtb_{,i}=\xt_{,a},\xtb_{,j}=\xt_{,b}\}}|(\xtb_{,k})_{k\in[K]},\xib]|\xib\big]\\
&\revdef
\tscorejoint(\xt_{,a},\xt_{,b},\xib),
\end{align}
where
\begin{align}
\tscorejoint(\xt_{,a},\xt_{,b},\xib) 
= 
\sum_{i,j=1}^K\E\Bigg[\frac{\exp(\scorefun_r(\xib,\xt_{,a}))\cdot \exp(\scorefun_r(\xib,\xt_{,b}))}{[\sum_{k\in[K]}
\exp(\scorefun_r(\xib,\xt_{,k}))]^2}\mathrm{1}_{\{\xtb_{,i}=\xt_{,a},\xtb_{,j}=\xt_{,b}\}}
\Bigg|\xib\Bigg].
\end{align}
On the other hand, we have
\begin{align}
    \scorejointinf(\xt_{,a},\xt_{,b}|\xib)
    &=
   \frac{\exp(\scorefun_r(\xib,\xt_{,a}))\P(\xt_{,a})\cdot \exp(\scorefun_r(\xib,\xt_{,b}))\P(\xt_{,b})}{[\E_{\xt\sim\P_\txt}{\exp(\scorefun_r(\xib,\xt))}]^2}\\
   &\propto  
   \exp(\scorefun_r(\xib,\xt_{,a}))\P(\xt_{,a})\cdot \exp(\scorefun_r(\xib,\xt_{,b}))\P(\xt_{,b})\\
   &\overset{(i)}{\propto}~
   \E\Bigg[\frac{\exp(\scorefun_r(\xib,\xt_{,a}))\cdot \exp(\scorefun_r(\xib,\xt_{,b}))}{[\sum_{k\in[K-2]}
\exp(\scorefun_r(\xib,\xt_{,k}))]^2}\mathrm{1}_{\{\xtb_{,K-1}=\xt_{,a},\xtb_{,K}=\xt_{,b}\}}
\Bigg|\xib\Bigg]\\
 &\overset{(ii)}{\propto}~
 \tscorejointinf(\xt_{,a},\xt_{,b},\xib),
 \end{align}
 where
 \begin{align}
 &\tscorejointinf(\xt_{,a},\xt_{,b},\xib)\\ 
 &\defn
  \sum_{i\neq j;2\leq i,j\leq K} \E\Bigg[\frac{\exp(\scorefun_r(\xib,\xt_{,a}))\cdot \exp(\scorefun_r(\xib,\xt_{,b}))}{[\sum_{k\in[K]\backslash\{i,j\}}
\exp(\scorefun_r(\xib,\xt_{,k}))]^2}\mathrm{1}_{\{\xtb_{,i}=\xt_{,a},\xtb_{,j}=\xt_{,b}\}}
\Bigg|\xib\Bigg].
\end{align}
Above, step~(i) follows from the conditional independence between $(\xtb_{,K-1},\xtb_{,K})$  and $(\xtb_{,k})_{k\leq K-2}$, and the distributional assumption on $\xtb_{,K-1},\xtb_{,K}$; step~(ii) follows from the symmetry across the $K-1$ indices. 

To control the different between $\tscorejoint$ and $\tscorejointinf$, we introduce the function
\begin{align}
    &\tscorejointloo(\xt_{,a},\xt_{,b},\xib) \\ 
&\defn 
\sum_{i\neq j ;2\leq i,j\leq K}\E\Bigg[\frac{\exp(\scorefun_r(\xib,\xt_{,a}))\cdot \exp(\scorefun_r(\xib,\xt_{,b}))}{[\sum_{k\in[K]}
\exp(\scorefun_r(\xib,\xt_{,k}))]^2}\mathrm{1}_{\{\xtb_{,i}=\xt_{,a},\xtb_{,j}=\xt_{,b}\}}
\Bigg|\xib\Bigg],
\end{align}and define the conditional distribution $\scorejointloo$ on $\cXt\times\cXt$ to be the distribution proportional to $\tscorejointloo$, namely, 
\begin{align}
\scorejointloo(\xt_{,a},\xt_{,b}|\xib)~\propto~\tscorejointloo(\xt_{,a},\xt_{,b},\xib).
\end{align} We will bound the differences between $\scorejoint$ and $\scorejointloo$, $\scorejointloo$ and $\scorejointinf$ in the following.

\paragraph{Bounding the differences.}
We first control the difference between $\scorejoint$ and $\scorejointloo.$
By Assumption~\ref{ass:bounded_score}, we have
\begin{align}
 0&\leq  \tscorejoint(\xt_{,a},\xt_{,b},\xib)- \tscorejointloo(\xt_{,a},\xt_{,b},\xib)
  \\
  &=
  \sum_{i=j\text{ or }i=1\text{ or }j=1}
  \E\Bigg[\frac{\exp(\scorefun_r(\xib,\xt_{,a}))\cdot \exp(\scorefun_r(\xib,\xt_{,b}))}{[\sum_{k\in[K]}
\exp(\scorefun_r(\xib,\xt_{,k}))]^2}\mathrm{1}_{\{\xtb_{,i}=\xt_{,a},\xtb_{,j}=\xt_{,b}\}}
\Bigg|\xib\Bigg]\\
&\leq 
\frac{\polyshort}{K^2}\cdot
\sum_{i=j\text{ or }i=1\text{ or }j=1}
\P(\xtb_{,i}=\xt_{,a},\xtb_{,j}=\xt_{,b}|\xib)
\\
&\leq
\frac{\polyshort}{K}\cdot\Big[
\mathrm{1}_{\{\xt_{,a}=\xt_{,b}\}}(\P_{\txt|\image}(\xt_{,a}|\xib)+\P_{\txt}(\xt_{,a}))
+
\P_{\txt|\image}(\xt_{,a}|\xib)\cdot \P_\txt(\xt_{,b})
\\ &\quad\quad +
\P_{\txt|\image}(\xt_{,b}|\xib)\cdot \P_\txt(\xt_{,a})
\Big]\\
&\leq
\frac{\polyshort}{K}\cdot\Big[
\mathrm{1}_{\{\xt_{,a}=\xt_{,b}\}}\P_{\txt|\image}(\xt_{,a}|\xib)
+
\scorejointinf(\xt_{,a},\xt_{,b}|\xib)
\Big],\label{eq:tscore_bound_1}
\end{align} where the first inequality follows from the boundedness assumption of $\exp(\scorefun_r)$ implied by Assumption~\ref{ass:bounded_score}, and the second and third inequalities use the boundedness of $\exp(\scorefun_\star).$ Summing equation~\eqref{eq:tscore_bound_1} over $\xt_{,a},\xt_{,b}$ and recalling that $\scorejoint=\tscorejoint$, we find
\begin{align}
    1&\geq
    \sum_{\xt_{,a},\xt_{,b}}\tscorejointloo(\xt_{,a},\xt_{,b},\xib)\\
    &
    \geq 
    1-\sum_{\xt_{,a},\xt_{,b}}\frac{\polyshort}{K}\cdot\Big[
\mathrm{1}_{\{\xt_{,a}=\xt_{,b}\}}\P_{\txt|\image}(\xt_{,a}|\xib)
+
\scorejointinf(\xt_{,a},\xt_{,b}|\xib)
\Big]\\
&= 1-\frac{2\polyshort}{K}.
\end{align} Thus, when $K\geq 4\polyshort$ in the equation above, it follows from the triangle inequality that
\begin{align}
  &\quad |\scorejoint(\xt_{,a},\xt_{,b}|\xib)- \scorejointloo(\xt_{,a},\xt_{,b}|\xib)  |
  \\&
   \leq
   \frac{\polyshortprime}{K}\cdot\Big[
\mathrm{1}_{\{\xt_{,a}=\xt_{,b}\}}\P_{\txt|\image}(\xt_{,a}|\xib)
+
\scorejointinf(\xt_{,a},\xt_{,b}|\xib)+
\scorejointloo(\xt_{,a},\xt_{,b}|\xib) 
\Big]\label{eq:tscore_bound_125}
\end{align} for some constant $\polyshortprime>0.$

Next, we  bound the difference between $\scorejointloo$ and $\scorejointinf.$ 
Introduce the shorthand notations 
\begin{align}
    \shortscoreratio\defn  \frac{\exp(\scorefun_r(\xib,\xt_{,a}))\cdot \exp(\scorefun_r(\xib,\xt_{,b}))}{[\sum_{k\in[K]}
\exp(\scorefun_r(\xib,\xt_{,k}))]^2},~~~
\shortscoreratio_{i,j}\defn  \frac{\exp(\scorefun_r(\xib,\xt_{,a}))\cdot \exp(\scorefun_r(\xib,\xt_{,b}))}{[\sum_{k\in[K]\backslash\{i,j\}}
\exp(\scorefun_r(\xib,\xt_{,k}))]^2},~\text{ for } i\neq j.
\end{align}
Then $\shortscoreratio_{i,j}\geq\shortscoreratio$ and  \begin{align}
{\shortscoreratio_{i,j}-\shortscoreratio}
    &=\frac{[\sum_{k\in[K]}
\exp(\scorefun_r(\xib,\xt_{,k}))]^2-[\sum_{k\in[K]\backslash\{i,j\}}
\exp(\scorefun_r(\xib,\xt_{,k}))]^2}{[\sum_{k\in[K]}
\exp(\scorefun_r(\xib,\xt_{,k}))]^2\cdot[\sum_{k\in[K]\backslash\{i,j\}}
\exp(\scorefun_r(\xib,\xt_{,k}))]^2}
    \leq \frac{\polyshort\cdot K}{K^4}\\
    &\leq 
    \frac{\polyshort}{K}\cdot \frac{\exp(\scorefun_r(\xib,\xt_{,a}))\cdot \exp(\scorefun_r(\xib,\xt_{,b}))}{[\sum_{k\in[K]\backslash\{i,j\}}
\exp(\scorefun_r(\xib,\xt_{,k}))]^2}
=
  \frac{\polyshort}{K}\cdot \shortscoreratio_{i,j}
\end{align} for all $i\neq j$, where the inequalities follow from Assumption~\ref{ass:bounded_score}.
Therefore, 
we obtain
\begin{align}
  0&\leq
  \tscorejointinf(\xt_{,a},\xt_{,b},\xib)
 -\tscorejointloo(\xt_{,a},\xt_{,b},\xib)
  \\
  &=
  \sum_{i\neq j;2\leq i,j\leq K}
  \E\Bigg[(\shortscoreratio_{i,j}-\shortscoreratio)\cdot\mathrm{1}_{\{\xtb_{,i}=\xt_{,a},\xtb_{,j}=\xt_{,b}\}}
\Bigg|\xib\Bigg]\\
&\leq 
\frac{\polyshort}{K}\cdot
 \sum_{i\neq j;2\leq i,j\leq K}
  \E\Bigg[\shortscoreratio_{i,j}\cdot\mathrm{1}_{\{\xtb_{,i}=\xt_{,a},\xtb_{,j}=\xt_{,b}\}}
\Bigg|\xib\Bigg]
=
\frac{\polyshort}{K}\cdot
\tscorejointinf(\xt_{,a},\xt_{,b},\xib)\label{eq:tscore_bound_15}
\end{align}for all $\xt_{,a},\xt_{,b}\in\cXt$.

Since  $ \tscorejointinf, \tscorejointloo$ are proportional to the conditional distributions $ \scorejointinf, \scorejointloo$, when $K\geq 4\polyshort$ in equation~\eqref{eq:tscore_bound_15}, we have
\begin{align}
  |\scorejointinf(\xt_{,a},\xt_{,b}|\xib)
 -\scorejointloo(\xt_{,a},\xt_{,b}|\xib) | \leq \frac{\polyshortprime}{K}\cdot
\scorejointinf(\xt_{,a},\xt_{,b}|\xib).\label{eq:tscore_bound_2}
\end{align}for some constant $\polyshortprime>0.$

\paragraph{Combining the bounds.}
Combining bounds~\eqref{eq:tscore_bound_125}~and~\eqref{eq:tscore_bound_2} with the triangle inequality, when $K\geq \polyshort$ for some constant $\polyshort>0$ depending polynomially on $\boundscoreconst$ in Assumption~\ref{ass:bounded_score}, we obtain
\begin{align}
 &\quad|\scorejoint(\xt_{,a},\xt_{,b}|\xib)
 -\scorejointinf(\xt_{,a},\xt_{,b}|\xib) | \\
 &\leq 
 \frac{\polyshortprime}{K}\cdot
\Big[
\mathrm{1}_{\{\xt_{,a}=\xt_{,b}\}}\P_{\txt|\image}(\xt_{,a}|\xib)
+
\scorejointinf(\xt_{,a},\xt_{,b}|\xib)+
\scorejointloo(\xt_{,a},\xt_{,b}|\xib) 
\Big]\\
 &\leq 
 \frac{\polyshortprime}{K}\cdot
\Big[
\mathrm{1}_{\{\xt_{,a}=\xt_{,b}\}}\P_{\txt|\image}(\xt_{,a}|\xib)
+
\scorejointinf(\xt_{,a},\xt_{,b}|\xib)
\Big]
\end{align}
for some constant $\polyshortprime>0$ for all $\xt_{,a},\xt_{,b},$ where the last inequality uses Eq.~\eqref{eq:tscore_bound_2}.  This yields claim~\eqref{eq:var_dist_diff_claim_1}.

\end{proof}

\begin{lemma}[Girsanov theorem]\label{lm:girsanov}
Let $\{ \bmu_t, \bgamma_t \}_{t \ge 0} \subseteq \R^d \to \R^d$. Consider two stochastic differential equation
\[
\begin{aligned}
\de \bx_t =&~ \bmu_t(\bx_t) \de t + \de \bW_t,~~~~ \bx_0 = \bzero, \\
\de \by_t =&~ \bgamma_t(\by_t) \de t + \de \bW_t,~~~~ \by_0 = \bzero. \\
\end{aligned}
\]
Let $\sP_T$ be the distribution of $\bx_T $, and $\sQ_T$ be the distribution of $\by_T $. Then we have 
\[
\KL(\sP_T || \sQ_T) \le \frac{1}{2}\int_0^T \E_{\bx_t}\| \bmu_t(\bx_t) - \bgamma_t(\bx_t) \|_2^2 \de t. 
\]
\end{lemma}
\begin{proof}[Proof of Lemma~\ref{lm:girsanov}]
  We provide a proof here for completeness. Let $\sP,\sQ$  denote the distributions of $\bx_{0:T},\by_{0:T}$, respectively. Girsanov theorem implies that for any $\bz_{0:T}$
  \begin{align}
      \log \frac{\sP(\bz_{0:T})}{\sQ(\bz_{0:T})}
      =\int_{0}^T 
  (\bmu_t(\bz_t) - \bgamma_t(\bz_t)) \de \bnoise_t
      +
      \frac{1}{2}\int_0^T \| \bmu_t(\bz_t) - \bgamma_t(\bz_t) \|_2^2 \de t.
  \end{align}
 Therefore,  
 \begin{align}
   \KL(\sP_T||\sQ_T) \leq \KL(\sP||\sQ)
   &=\E_{\bx_{0:T}}\Big[\int_{0}^T 
  (\bmu_t(\bx_t) - \bgamma_t(\bx_t)) \de \bnoise_t
      +
      \frac{1}{2}\int_0^T \| \bmu_t(\bx_t) - \bgamma_t(\bx_t) \|_2^2 \de t\Big]\\
      &= 
      \frac{1}{2}  \E_{\bx_{0:T}}\int_0^T \| \bmu_t(\bx_t) - \bgamma_t(\bx_t) \|_2^2 \de t
      =
      \frac{1}{2} \int_0^T \E_{\bx_{t}} \| \bmu_t(\bx_t) - \bgamma_t(\bx_t) \|_2^2 \de t,
 \end{align}
 where the first inequality uses data-processing inequality.
\end{proof}

\section{Proof of \Cref{theorem:CLIP-Generalization}}\label{app:proof_CLIP-Generalization}

\subsection{Overview}\label{sec:clip_pf_overview}
The generalization error analysis of neural networks is typically conducted by first constructing a neural network that approximates a certain function (or algorithm) and then evaluating the complexity of the class containing that network.
This subsection explains the whole architecture and the proof strategy that constructs a pipeline approximating the target algorithm. 


The transfomer-based architectures in the following will process vectors related to each node by concatenating them into a matrix.
Thus associating nodes with integers will make the discussion easier.
For this, with a slight abuse of notation, we identify the nodes with the positive integer defined as follows.
\begin{definition}[Numbering of nodes]\label{definition:Numbering}
For $\square\in\{\image,\txt\}$, 
a node $v \in \cV_\square^{(L)}$ is identified as a integer defined as
\begin{align}
        \iota(v) + m^{(L)}_\square(\iota(\pa(v))-1) + m^{(L-1)}_\square m^{(L)}_\square(\iota(\pa^{(2)}(v))-1) + \cdots + (m^{(2)}_\square \cdots m^{(L)}_\square)(\iota(\pa^{(L-1)}(v))-1).
        \label{eq:Int(v)}
\end{align}
Here $\pa^{(\ell)}(v)$ means the $\ell$-th grand parent of $v$. 
We also identify intermediate nodes $v\in \cV^{(\ell)}_\square (\ell= L-1,\dots,0)$ as a positive integer
\begin{align}
    (m^{(L)}_\square \cdots m^{(\ell+1)}_\square) \big [\iota(v) + m^{(\ell)}_\square(\iota(\pa(v))-1)  + \cdots + (m^{(2)}_\square \cdots m^{(\ell)}_\square)(\iota(\pa^{(\ell-1)}(v))-1) \big ].
    \label{eq:Int(v)-2}
\end{align}
\end{definition}
This allows us to compare two nodes $u,v$ in different levels (say, $u\in \cV^{(\ell)}_\image, v\in \cV^{(\ell')}_\image$) like $u>v$ or $u=v$. 
However, treating a node $ v \in \cV^{(\ell)}_\image $ as a node in another level $\ell'$ sometimes leads to confusion, as $ \cN $, $ \cC $, and ``$\pa$'' no longer points a unique node. 
Therefore, when there is a risk of confusion, we explicitly indicate the level of the node by referring to the node as $v^{(\ell)}$.

\subsubsection{Belief propagation and message passing algorithms} \label{subsubsection:CLIP-Intro-BP}
Now we outline our approach to approximating the algorithm. Let us recall the message passing algorithm we aim to approximate.
For the text part, it starts with $h^{(L)}_{\txt,v}=x_{\txt,v}\ (v\in \cV_\txt^{(L)})$,  and computes $(q^{(\ell)}_{\txt,v})_{v\in \cV_\txt^{(\ell)}}$ and $(h^{(\ell)}_{\txt,v})_{v\in \cV_\txt^{(\ell)}}$ in the decreasing order of $\ell$ to obtain $h^{(0)}_{\txt,\root} \in \R^S$: 
   \begin{align}
    \begin{aligned}
    \textstyle q_{\txt,v}^{(\ell)} &\textstyle= f_{\txt,\iota(v)}^{(\ell)}(h^{(\ell)}_{\txt,v}) \in \R^S,
    && v\in \cV^{(\ell)}_\txt,\ \ell = L,\dots,1,
   \\ 
    \textstyle h^{(\ell-1)}_{\txt,v} &\textstyle = \normalize \big(\sum_{u\in \cC(v)}q_{\txt,u}^{(\ell)}\big) \in \R^S,
     &&v\in \cV^{(\ell-1)}_\txt,\  \ell = L,\dots,1.
       \end{aligned}\label{eq:Appendix-CLIP-ErrorPropagation-15}
    \end{align}
    Computation of $q_{\txt,v}^{(\ell)}$ and $h^{(\ell-1)}_{\txt,v}$ from $h^{(\ell)}_{\txt,v}$ is called the $\ell$-th step. 
   Here, $f_{\txt,\iota}^{(\ell)}$ are defined as
    \begin{align}
    \begin{aligned}
      &  (f_{\txt, \iota}^{(L)}(x))_s = \log \psi_{\txt, \iota}^{(L)}(s,x),
        &&  x\in [S],\ s\in [S],
        \\
    & \textstyle   (f_{\txt, \iota}^{(\ell)}(h))_s = \log \sum_{a\in [S]}\psi_{\txt, \iota}^{(\ell)}(s,a)e^{h_a}, 
        && h\in \R^S,\ s\in [S],\ \ell = L-1,\dots,2,
   \\  & \textstyle   (f_{\txt, \iota}^{(1)}(h))_s = \log \sum_{a\in [S]}(\P[s])^{\frac{1}{\nchild_{\txt}^{(1)}}}\psi_{\txt, \iota}^{(1)}(s,a)e^{h_a}, 
        && h\in \R^S,\ s\in [S],
    \end{aligned}\label{eq:Appendix-CLIP-ErrorPropagation-14}
    \end{align} 
    and $\normalize(h)_s=x_s - \max_{s'}h_{s'}$ for $h\in \R^S$. 
    In the same way, the image part yields $h^{(0)}_{\image,\root} \in \R^S$. 
    Combining them finally yields
    \begin{align}
     \scorefun_{\MP} &\textstyle = f^{(0)}(\softmax(h_{\image,\root}^{(0)}),\softmax(h_{\txt,\root}^{(0)})) 
    \end{align}
    where 
    \begin{align}
        f^{(0)}(h,h') = \log \sum_{s} h_s h'_s (\P[s])^{-1},  \quad\quad h,h'\in [0,1]^S.
        \label{sq:scorefunction}
    \end{align}
    The correctness of this algorithm is formally stated as follows.
    \begin{lemma}[MP yields the optimal similarity 
 score]\label{lem:MP_score_exact}
Applying the message passing algorithm above, it holds that $\softmax(h_{\image,\root}^{(0)})_s = \P[s| \xi]$, $\softmax(h_{\txt,\root}^{(0)})_s =\P[s| \xt]$, and $\scorefun_{\MP} = \scorefun_\star(\xi,\xt)+(\text{const.})$.
    \end{lemma}
    \begin{proof}
    According to \Cref{lem:global_min_contrastive_loss}, 
    the optimal similarity score function is defined as
    \begin{align}
       \scorefun_{\star}(\xi,\xt) =  \log \Bigg[ \frac{\P[ \xi, \xt] }{  \P[\xt] \cdot \P[\xi] }\Bigg].
    \end{align}
    From Proposition 2 of \cite{mei2024u}, it holds that $\softmax(h_{\image,\root}^{(0)})_s = \P[s| \xi]$ and $\softmax(h_{\txt,\root}^{(0)})_s =\P[s| \xt]$. (Note that their definition of $\psi^{(0)}_{\iota}$ includes $\P[s]$ while our $\psi^{(0)}_{\image,\iota}$ and $\psi^{(0)}_{\txt,\iota}$ do not, which results in the $\P[s]^{\frac{1}{m^{(1)}}}$ term in the definition of $f^{(1)}_{\txt,\iota}$.) 
    Because of the Bayes rule,
    \begin{align}
      \frac{\P[ \xi, \xt] }{  \P[\xt] \cdot \P[\xi] }
      &=
      \frac{\sum_{s\in [S]}\P[ \xi|s]\P[ \xt|s]\P[s]}{\P[\xt] \cdot \P[\xi] }
    =
      \sum_{s\in [S]}\frac{\P[s| \xi]\P[ s|\xt]}{\P[s]}
   \\ & = \sum_{s\in [S]}\softmax(h_{\image,\root}^{(0)})_s\softmax(h_{\image,\root}^{(0)})_s(\P[s])^{-1}.
    \end{align}
    By taking the logarithm of this yields $\scorefun_{\star}(\xi,\xt) =  \log \Big[ \frac{\P[ \xi, \xt] }{  \P[\xt] \cdot \P[\xi] }\Big]$.
    \end{proof}

\subsubsection{Approximation with transformer networks}\label{subsubsection:CLIP-Overview-Transformer}
    We construct a transformer-based pipeline to replicate the message passing algorithm.
    It consists of three components: a transformer encoder for images $\NN_\image^{\bW_{\image}}$; a transformer encoder for text $\NN_\txt^{\bW_{\txt}}$; and a parameterized link function $\NNs^{\linkweight}(h,h')$. 
    The two transformers $\NN_\image^{\bW_{\image}}$ and $\NN_\txt^{\bW_{\txt}}$ approximately compute $h_{\image,\root}^{(0)}$ and $h_{\txt,\root}^{(0)}$, respectively, by following the message passing algorithm \eqref{eq:Appendix-CLIP-ErrorPropagation-15}.
     Because $\NN_\txt^{\bW_{\txt}}$ and $\NN_\image^{\bW_{\image}}$ follows the same construction, we will sometimes omit the subscripts ``$\txt$'' and ``$\image$'' in the following to discuss these networks.
    Finally we put them into the link function $\NNs^{\linkweight}(h,h')$.

    After the embedding  (positional encoding)  layer $\encodeclip$, the network obtains the initial matrix $\hmatrix^{(L)}$ of size $(d_\feature+d_\positional)\times d$, with $d_\feature=2SL+1$ and $d_\positional=2L$.
    Here $d_\positional=2L$ is the dimension corresponding to the positional encoding, and the rest $d_\feature=2SL+1$ corresponds to the ``features''.
    Specifically, this matrix is written as
    \begin{align}
        \hmatrix^{(L)} = \encodeclip(\bx) = 
        \begin{bmatrix}
            &&\boldzero&
            \\
            x_1 & x_2 & \cdots & x_d
            \\
            &&\pomatrix&
        \end{bmatrix},
    \end{align}
    so that it consists of the positional encoding $\pomatrix\in \R^{d_\positional \times d}$, the text variable $\xt$, and the zeros reserved for later calculation $\boldzero \in \R^{(d_\feature-1) \times d}$. 
 
    Starting from $\hmatrix^{(L)}$, 
    the transformer network applies $L$ transformer blocks.
    These blocks are indexed by $\ell=L,\dots,1$ in in the decreasing order, so that the $\ell$-th layer  corresponds with the $\ell$-th step of the message passing algorithm.
    They sequentially calculate $\qmatrix^{(\ell)}\ (\ell=L,\dots,1)$ and $\hmatrix^{(\ell)}\ (\ell=L,\dots,0)$ defined as
    \begin{align}
        \hmatrix^{(\ell)} = 
        \begin{bmatrix}
        &  & \boldzero & 
        \\
            \hnetwork^{(\ell)}_1 & \hnetwork^{(\ell)}_2 & 
             \cdots
            &
            \hnetwork^{(\ell)}_d
            \\
            \vdots & \vdots & 
             \ddots
            &
            \vdots
            \\
            \qnetwork^{(L)}_1 & \qnetwork^{(L)}_2 & 
             \cdots
            &
            \qnetwork^{(L)}_d
            \\
            \hnetwork^{(L)}_1 & \hnetwork^{(L)}_2 & 
             \cdots
            &
            \hnetwork^{(L)}_d
            \\
            &&\pomatrix&
        \end{bmatrix},
        \quad
        \qmatrix^{(\ell)} = 
        \begin{bmatrix}
         & & \boldzero & 
        \\
            \qnetwork^{(\ell)}_1 & \qnetwork^{(\ell)}_2 & 
             \cdots
            &
            \qnetwork^{(\ell)}_d
            \\
            \vdots & \vdots & 
             \ddots
            &
            \vdots
            \\
            \qnetwork^{(L)}_1 & \qnetwork^{(L)}_2 & 
             \cdots
            &
            \qnetwork^{(L)}_d
            \\
            \hnetwork^{(L)}_1 & \hnetwork^{(L)}_2 & 
             \cdots
            &
            \hnetwork^{(L)}_d
            \\
            &&\pomatrix&
        \end{bmatrix},
    \end{align}
    where $\hnetwork^{(\ell)}_v\in \R^S$ and $\qnetwork^{(\ell)}_v\in \R^S$ except for $\hnetwork^{(L)}_v\in [S]$. 

    The $\ell$-th block approximates the $\ell$-th step of the message passing algorithm. 
    It consists of a position-wise feed forward layer $\FF^{(\ell)}$ with skip connection and self-attention layer $\Attn^{(\ell)}$ with skip connection and normalization.
    The feed forward layer $\FF^{(\ell)}$, a fully-connected ReLU network, receives $\hmatrix^{(\ell)}$ and outputs $\qmatrix^{(\ell)}$ by computing $\qnetwork^{(\ell)}_v$ from $\hnetwork^{(\ell)}_v$:
    \begin{align}
        \qmatrix^{(\ell)} = 
        \underbrace{\hmatrix^{(\ell)}}_{\text{skip connection}}
        + 
        \feedforward^{(\ell)}(\hmatrix^{(\ell)})
        =
        \hmatrix^{(\ell)}
        +
        \begin{bmatrix}
        \multicolumn{4}{l}{\boldzero\ (\in \R^{((2\ell-1)S) \times d})} \\ 
       \qnetwork^{(\ell)}_1 & \qnetwork^{(\ell)}_2 &  \cdots & \qnetwork^{(\ell)}_d \\
        \multicolumn{4}{l}{ \boldzero\ (\in \R^{(d_\positional+1+2(L-\ell)S) \times d})} 
        \end{bmatrix}
        .
    \end{align}
    The self-attention layer $\Attn^{(\ell)}$, uses this $\qmatrix^{(\ell)}$ and outputs $\hmatrix^{(\ell-1)}$ by computing $\hnetwork^{(\ell-1)}_v$:
    \begin{align}
        \hmatrix^{(\ell-1)} = 
        \normalize\Big(\underbrace{\qmatrix^{(\ell)}}_{\text{skip connection}}
        +\Attn^{(\ell)}(\qmatrix^{(\ell)})
        \Big)
        =
        \normalize\Bigg(
        \qmatrix^{(\ell)}
        + 
        \begin{bmatrix}
            \boldzero & (\in \R^{((2\ell-2)S) \times d})\quad \quad\quad \ \ \ 
            \\
            {\star} &(\in \R^{S \times d})\quad \quad \quad \quad \quad\quad \  \ \ 
            \\
            \boldzero &(\in \R^{(d_\positional+1+(2L-2\ell+1)S) \times d})
        \end{bmatrix}
        \Bigg)
        .
    \end{align} 
    Here $\star$ means $[\hnetwork_1^{(\ell-1)}\ \hnetwork_2^{(\ell-1)}\ \cdots\  \hnetwork_d^{(\ell-1)}]$ before normalization. 
    In this way, we iteratively compute $\qnetwork_v^{(\ell)}$ and $\hnetwork_v^{(\ell)}$ to fill zeros of the previous matrices.
    These $\hnetwork^{(\ell)}_v\in \R^S$ and $\qnetwork^{(\ell)}_v\in \R^S$ approximate $h^{(\ell)}_u$ and $h^{(\ell)}_u$ as
    \begin{align}
    \begin{aligned}
    &  \qnetwork^{(\ell)}_{v} \approx q^{(\ell)}_{\pa^{(L-\ell)}(v)}, && v\in \cV^{(L)}, 
    \ \ell = L,\dots,1,
      \\
    & \hnetwork^{(\ell)}_{v} \approx h^{(\ell)}_{\pa^{(L-\ell)}(v)}, &&   v\in \cV^{(L)}, 
    \ \ell = L,\dots,0.
    \end{aligned}
    \label{eq:CLIP-Intro-approx-target}
    \end{align}
    
    After we obtain $\hmatrix^{(0)}$, we extract $\hnetwork^{(0)}_{d}$ (this is an approximation of $h_\root^{(0)}$) to output $\readclip(\hmatrix^{(0)})=\hnetwork^{(0)}_{d}$. 
  
    We remark that our transformer block applies the feed forward layer first to emphasize the correspondence with the message passing. 
    If adhering to a typical structure where self-attention comes first, we can implement the pipeline with $(L+1)$-blocks.

We now formally define each component of the network and explain key lemmas to confirm \eqref{eq:CLIP-Intro-approx-target} iteratively. 

\paragraph{Embedding $\encodeclip$.}
When the network receives the input $\bx\in [S]^{d}$, it first passes it through the embedding layer $\encode$, where it concatenate the input $\bx$ with the positional encoding $\pomatrix$ and the zeros $\boldzero$.
The $v$-th column of $\pomatrix$ is denoted by $\pnetwork_v$. 
This $\pnetwork_v\in \R^{d_\positional}\ (d_\positional=2L)$ is defined as
\begin{align}
  & \!\!\!\! \pnetwork_v = 
  \\ & \!\!\!\!\textstyle\Big[\sin \big( \frac{2\pi\iota(v)}{m^{(L)}}\big)\  \cos \big(\frac{2\pi\iota(v)}{m^{(L)}}\big)\ \sin \big( \frac{2\pi\iota(\pa(v))}{m^{(L-1)}}\big)\  \cos \big( \frac{2\pi\iota(\pa(v))}{m^{(L-1)}}\big)\  \cdots  \  \sin \big(\frac{2\pi\iota(\pa^{(L-1)}(v))}{m^{(1)}}\big)\  \cos \big( \frac{2\pi\iota(\pa^{(L-1)}(v))}{m^{(1)}}\big)\Big]^\top\!\!.
    \label{eq:positional_encoding}
\end{align}

\paragraph{Position-wise feed forward layer.}

Consider the feed forward network $\FF^{(\ell)}$ of the $\ell$-th block $(\ell=L,\dots,1)$, which computes $\qnetwork^{(\ell)}_v$ from $\hnetwork^{(\ell)}_v$.
We will show that, for each $\hnetwork^{(\ell)}_{v} (v\in \cV^{(L)})$, the network can identify its (ancestor's) rank $\iota=\iota(\pa^{(L-\ell)}(v))$ and apply $\fnetwork^{(\ell)}_{\iota}$, which is a neural network approximation of $f^{(\ell)}_{\iota}$. 
The identification of $\iota(\pa^{(L-\ell)}(v))$ can be implemented with no errors. 
Therefore, the feed forward layer at the $\ell$-th layer yields
\begin{align}
   \qnetwork^{(\ell)}_{v} 
   =\hnetwork^{(\ell)}_{v} + \fnetwork_{\iota(\pa^{(L-\ell)}(v))}^{(\ell)}(\hnetwork^{(\ell)}_{v}), \quad v\in \cV^{(L)}.
\end{align}
When $\hnetwork^{(\ell)}_{v}\approx h_{\pa^{(L-\ell)}(v)}^{(\ell)}$ for $v\in \cV^{(L)}$ and $\fnetwork_{\iota}^{(\ell)}\approx f_\iota^{(\ell)}$, we have $\qnetwork^{(\ell)}_{v}\approx q_{\pa^{(L-\ell)}(v)}^{(\ell)}$. 

We define a class of full connected networks with the ReLU activation as follows.
For an $l$-dimensional vector $x\in \R^l$, we write $[x;1]=(x_1,\dots,x_l,1)^\top$. 
\begin{definition}[A class of fully connected networks]\label{definition:FullyConnected}
    For $\depth \in \N$, $\widthvector=(\width_1,\dots,\width_{L+2})\in \N^{\depth+2}$, and $B>0$, we define a class of full connected networks with the ReLU activation as
    \begin{align}
        \mathcal{F}(\depth, \widthvector, B) &= \Big\{
        \bW^{(\depth+1)}[\cdot; 1]\circ \ReLU (\bW^{(\depth)} [\cdot; 1]) \circ \ReLU (\bW^{(\depth-1)} [\cdot; 1]) \circ \cdots \circ \ReLU (\bW^{(1)} [\cdot; 1])\ \Big|\ 
       \\ & \quad \quad \bW^{(1)} \in \R^{\width_2 \times (\width_1+1)},
       \bW^{(2)} \in \R^{\width_3 \times (\width_2+1)},
       \cdots,
       \bW^{(\depth+1)}\in \R^{\width_{\depth+2}\times (\width_{\depth+1}+1)}
         , \max_{j\in [J+1]}\max_{k,l}|\bW^{(j)}_{k,l}|\leq B
       \Big\}.
    \end{align}
    Each element implements a function from $\R^{\width_1}$ to $\R^{\width_{\depth+2}}$.  
\end{definition}
We will show the following approximation error guarantee of $f^{(\ell)}_\iota$. 
Since it is easy to concatenate zeros to the first and last layer matrices and adjust the input and output dimensions to be $d_\feature+d_\positional$, the network $\NN$ in the following is presented as a function from $\R^{S}\times \R^{d_\positional}$ (or $[S]\times \R^{d_\positional}$ for $\ell=L$) to $\R^S$, focusing only on relevant dimensions.
The proof will be found in \Cref{subsection:FF}. 
\begin{lemma}[Approximation error of feed forward layer]\label{lemma:Approximation-FF}
Fix $\ell\in [L]$ and $\delta>0$.
Assume that $\transbd^{-1} \leq \psi^{(\ell)}_{\iota}(s,a) \leq \transbd$ for all $s,a\in [\staten]$.
When $\ell=1$, also assume that $\transbd^{-1}\leq \P[s]\leq \transbd$ for all $s$. 
Then, there exists an $\NN\in \mathcal{F}(\depth,\widthvector,B)$ such that
\begin{align}
    \|\NN([h;\pnetwork_v]) - f^{(\ell)}_{\iota(\pa^{(L-\ell)}(v))}(h)\|_\infty \leq \delta, \quad v\in \cV^{(L)},
\end{align}
for all $h\in \R^S$ with $\max_s h_s = 0$ ($\ell\leq L-1$) or $h\in [S]$ ($\ell=L$).
The network parameters $\depth,\widthvector$ and $B$ are bounded as follows:
\begin{align}
    \depth &\lesssim (\log\log(S\transbd/\delta))\log(S\transbd/\delta),
    \\
    \|\widthvector\|_\infty  
    &\lesssim m^{(\ell)}S(\log(S\transbd/\delta))^3+\layer+d_\positional,
    \\
    B  &\lesssim 2S(\transbd^2+\log(S\transbd/\delta)) + (m^{(\ell)})^2.
\end{align}

\end{lemma}

The bound uses polylogarithmic depth with respect to the approximation error $\delta$.
It is known that deep neural networks can achieve significantly finer approximations \cite{schmidt2020nonparametric,suzuki2018adaptivity} than ones with constant depth 
 \cite{telgarsky2016benefits}.
Although this differs from real-world transformers, which use feed forward layers of constant depth, we can achieve the same result while keeping the feed forward layers of each block constant depth by using multiple blocks to approximate a single $f_\iota^{(\ell)}$ instead of increasing $J$ (ignoring intermediate self-attention layers). 
We chose not to adopt such a way of presentation because we prioritized keeping the correspondence between the $\ell$-th block of the transformer and the index $\ell$ in the message passing algorithm.
Also, please refer to ``Approximation with constant depth'' paragraph \Cref{subsection:FF} for details on using feed forward layers of constant depth with $L$ blocks.


\paragraph{Self-attention block.}
Consider the self-attention layer $\AT^{(\ell)}$ of the $\ell$-th block $(\ell=L,\dots,1)$.
Ignoring irrelevant dimensions, it takes $\qnetwork_v^{(\ell)}$ and $\pnetwork_v$ as inputs, and computes $\sum_{u\in \cC(\pa^{(L-\ell+1)}(v))}\qnetwork^{(\ell)}_{u}$ for each $v\in \cV^{(L)}$. 
For each $v\in \cV^{(L)}$, we denote the error by $\bdelta^{(\ell-1)}_v\in \R^S$. 
As a result, the self-attention block 
yields
\begin{align}
\textstyle
   \hnetwork_v^{(\ell-1)} 
   = \normalize\Bigg(
   \underset{\iota(\pa^{(L-\ell')}(u))=\iota(\pa^{(L-\ell')}(v))\ (\ell'\ne \ell)}{\sum}\qnetwork^{(\ell)}_{u} +\bdelta^{(\ell-1)}_v\Bigg).
\end{align}
Here $\normalize(x)_s=x_s - \max x_{s'}$.

We explain interpretation of $\sum_{\iota(\pa^{(L-\ell')}(u))=\iota(\pa^{(L-\ell')}(v))\ (\ell'\ne \ell)}$.
For each $v\in \cV^{(L)}$, the nodes $u$ that satisfy $\iota(\pa^{(L-\ell')}(u))=\iota(\pa^{(L-\ell')}(v))\ (\ell'\ne \ell)$ are descendants of $\pa^{(L-\ell+1)}(u)=\pa^{(L-\ell+1)}(v) $ whose ancestors' ranks are the same as $v$ except for the $\ell$-th level.
Thus, for each $v'\in \cC(\pa^{(L-\ell+1)}(v))$, the summation selects exactly one of the descendants of $v'$. 
This implies that, when $\qnetwork_v^{(\ell)}\approx q_{\pa^{(L-\ell)}(v)}^{(\ell)}$ for all $v\in \cV^{(L)}$, we have 
\begin{align}
    \textstyle
    \hnetwork_v^{(\ell-1)} 
    =
    \normalize\big(\sum_{u\in \cC(\pa^{(L-\ell+1)}(v))}\qnetwork^{(\ell)}_{u^{(L)}}+\bdelta_v^{(\ell-1)}\big)
    \approx \normalize\big(\sum_{u\in \cC(\pa^{(L-\ell+1)}(v))}q^{(\ell)}_{u}\big) = 
    h_{\pa^{(L-\ell+1)}(v)}^{(\ell-1)}. 
\end{align}
To further clarify the correspondence with the message passing, 
for $v\in \cV^{(\ell-1)}$, this is simplified as
\begin{align}
    \textstyle
    \hnetwork_{v^{(L)}}^{(\ell-1)} = \normalize\big(\sum_{u\in \cC(v)}\qnetwork^{(\ell)}_{u^{(L)}}+\bdelta^{(\ell-1)}_{v^{(L)}}\big).
\end{align}


We define a class of self-attention block as follows.
\begin{definition}[A class of self-attention blocks]\label{definition:Self-Attn}
    We define a class of self-attention blocks as
    \begin{align}
        \cA(D,B) = &\Big\{
        (\bW_V\ \cdot)\ \softmax((\bW_K\ \cdot)^\top(\bW_Q \ \cdot))\ \Big|\\ &\quad \
        \bW_K, \bW_Q, \bW_V\in \R^{D\times D},\
       \max_{i,j}|(\bW_K)_{i,j}|, \max_{i,j}|(\bW_Q)_{i,j}|, \max_{i,j}|(\bW_V)_{i,j}|
       \leq B\Big\}.
    \end{align}
    Each element implements a function that takes a matrix of size $D\times d'$ ($d'$: arbitrary) and maps it to a matrix of size $D\times d'$.
\end{definition}
Then, we obtain the following approximation error guarantee.
See \Cref{subsection:Selt-attention} for the proof. . 
\begin{lemma}[Approximation error of self-attention layer]\label{lemma:Self-attention-CLIP}
    For $\ell\in [L]$, there exists $\Attn\in \cA(D,B)$ with $D=d_\feature+d_\positional$ and $B\lesssim \log(d\delta^{-1})+m^{(\ell)}$ such that
    \begin{align}
    \Attn(\qmatrix^{(\ell)})=
    \begin{bmatrix}
            \boldzero&  (\in \R^{(2\ell-2)S})\quad \quad\quad \ \ \ 
            \\
            \underset{\iota(\pa^{(L-\ell')}(u))=\iota(\pa^{(L-\ell')}(v))\ (\ell'\ne \ell)}{\sum}\qnetwork^{(\ell)}_{u} + \bdelta^{(\ell-1)}_v & (\in \R^{S})\ \quad \quad \quad \quad\quad \  \ \ 
            \\
            \boldzero &(\in \R^{d_\positional+1+(2L-2\ell+1)S})
        \end{bmatrix}_{v\in \cV^{(L)}}, 
    \end{align} 
    where $\bdelta_{v}^{(\ell-1)}\in \R^S$ satisfies $\|\bdelta_{v}^{(\ell-1)}\|_\infty \leq \delta   \max_{v'}\|\qnetwork^{(\ell)}_{v'}\|_\infty$.
\end{lemma}

\paragraph{Normalization.}
In the attention network, since column vectors of $\hmatrix^{(\ell)}$ and $ \qmatrix^{(\ell)} $ are collections of multiple $ \hnetwork^{(\ell)}_v $ and $ \qnetwork^{(\ell)}_v $, we adopt a slightly different definition of ``$\normalize$'' for these column vectors, from the one for $S$-dimensional vectors.
Specifically, for $\xnetwork =[\hnetwork^{(0)}\ \qnetwork^{(1)}\ \hnetwork^{(1)}\ \dots\ \qnetwork^{(L)}\ \hnetwork^{(L)}\ \pnetwork]\in \R^{d_\feature+d_\positional}$ with $\hnetwork^{(L)}\in [S], \hnetwork^{(\ell)}\in \R^S\ (\ell=L-1,\dots,0),$ and $\qnetwork^{(\ell)}\in \R^S$, we define
\begin{align}
    \normalize(\xnetwork)
    = 
    \begin{bmatrix}
        \hnetwork^{(0)} \\ \qnetwork^{(1)}- \ones_S\max_{s\in S}\qnetwork^{(1)}_s \\ \hnetwork^{(1)} \\ \qnetwork^{(2)} - \ones_S\max_{s\in S} \qnetwork^{(2)}_s \\ \vdots \\ \qnetwork^{(L)} - \ones_S\max_{s\in S} \qnetwork^{(L)}\\ \hnetwork^{(L)} \\ \pnetwork
    \end{bmatrix}\in \R^{d_\feature+d_\positional},\quad \ones_S = \begin{bmatrix}
        1 \\ 1 \\ \vdots \\ 1
    \end{bmatrix}\in \R^S.\label{eq:clip_normalize_def}
\end{align}
For a matrix with its column dimension $d_\feature+d_\positional$, it is applied in a column-wise manner.

\paragraph{Readout layer $\read$.}
In the readout layer, we extract $\hnetwork_{d}^{(0)}$ as $\hnetwork_{d}^{(0)}=\read(\hmatrix^{(0)})=\read(\TF^{\bW}(\encode(\bx)))$.

\paragraph{Similarity score.}
    The link function $\NNs^{\linkweight}$ is defined as
    \begin{align}\textstyle
       \NNs^{\linkweight}(h,h') = \log(\readsimscore(\sum_{s\in [S]}h_sh_s'w_s)),
    \end{align}
    where
    \begin{align}\label{eq:readout_def}
      \readsimscore(z)\defn \proj_{[-\exp(-\readbd),\exp(\readbd)]}(z)
    \end{align} is the function that projects $z$ onto the interval $[\exp(-\readbd),\exp(\readbd)]$ for any 
     $z\in\R\cup\{-\infty\}.$ 
    We choose the threshold $\readbd\defn4\mmaxone\log\transbd$. As shown in Lemma~\ref{lemma:BoundScore}, the threshold $\readbd$ is chosen sufficiently large to ensure the truncation does not occur in our construction (when the approximation error is sufficiently small).
    Thus, setting $w_s = \mathbb{P}[s]^{-1}$ yields the exact $f^{(0)}$, i.e., $\NNs^{\linkweight}=f^{(0)}$ (see \eqref{sq:scorefunction} to remember the definition of $f^{(0)}$). 
    Under \Cref{ass:bounded_transition}, this $\linkweight_s$ satisfies $\|\linkweight\|_\infty \leq \transbd$.

\paragraph{The whole pipeline.}
Putting it all together, the whole pipeline, starting from $\hnetwork^{(L)}_{\txt,v}=x_{\txt,v}\ (v\in \cV_\txt)$, is written as
    \begin{align}
    \begin{aligned}
    \qnetwork_{\txt,v}^{(\ell)} &
           \textstyle = \fnetwork_{\txt,\iota(\pa^{(L-\ell)}(v))}^{(\ell)}(\hnetwork^{(\ell)}_{\txt,v}) \in \R^S,&& v\in \cV_\txt^{(L)}, \ \ell = L,\dots,1,
           \\
    \textstyle
   \hnetwork_{\txt,v}^{(\ell-1)} 
   & \textstyle= \normalize\Bigg(
   \underset{\iota(\pa^{(L-\ell')}(u))=\iota(\pa^{(L-\ell')}(v))\ (\ell'\ne \ell)}{\sum}\qnetwork^{(\ell)}_{u} +\bdelta^{(\ell-1)}_v\Bigg)\in \R^S,&& v\in \cV_\txt^{(L)},\ \ell = L,\dots,1.
    \end{aligned}
    \label{eq:thewholepipeline}
    \end{align}
    We can write this alternatively to emphasize the connection to the message passing algorithm \eqref{eq:Appendix-CLIP-ErrorPropagation-15} and \eqref{eq:Appendix-CLIP-ErrorPropagation-14} (see ``Self-attention block'' paragraph).
    \begin{align}
    \begin{aligned}
    \qnetwork_{\txt,v^{(L)}}^{(\ell)} &
           \textstyle = \fnetwork_{\txt,\iota(v)}^{(\ell)}(\hnetwork^{(\ell)}_{\txt,v^{(L)}}) \in \R^S,&& v\in \cV_\txt^{(\ell)}, \ \ell = 1,2,\dots,L,
    \\ 
           \hnetwork^{(\ell-1)}_{\txt,v^{(L)}} &
    \textstyle = \normalize \big(\sum_{u\in \cC(v)}\qnetwork_{\txt,{u}^{(L)}}^{(\ell)}+\bdelta_{\txt,v^{(L)}}^{(\ell-1)}\big) \in \R^S,&& v\in \cV_\txt^{(\ell-1)},\ \ell = 1,2,\dots,L.
    \end{aligned}
    \end{align}
    
    The image part is defined in the same way.
    Finally, with the link function $\NNs^{\linkweight}$ that exactly represents $f^{(0)}$, we get
    \begin{align}
    \scorefun_{\NN} &\textstyle = \NNs^{\linkweight}(\softmax(\hnetwork_{\image,d}^{(0)}),\softmax(\hnetwork_{\txt,d}^{(0)})).
    \label{eq:thewholepipeline-2}
    \end{align}

    Under \Cref{ass:bounded_transition} (because below we use $\transbd$), the hypothesis class to which a tuple $(\NN_\image^{\bW_{\image}},\NN_\txt^{\bW_{\txt}},\NNs^\linkweight)$ belongs is defined as follows, formally restating (\ref{eqn:CLIP_param_space}).
    \begin{definition}[Eq.~(\ref{eqn:CLIP_param_space}), restated]
    We say the collection of the parameters of  $(\NN_\image^{\bW_{\image}},\NN_\txt^{\bW_{\txt}},\tau^\bw)$ belongs to $\Parspaceclip$ if the following holds:
    For transformer networks $\NN_\image^{\bW_{\image}}$ and $\NN_\txt^{\bW_{\txt}}$, they have $L$ blocks of feed forward (\Cref{definition:FullyConnected}), self-attention (\Cref{definition:Self-Attn}), and normalization.
    In each block, its feed forward $\FF$ and self-attention $\Attn$ satisfy
        \begin{align}
            &\FF\in \cF (\depth,\widthvector=(D,*,\cdots,*,D),B), \text{ with } \|\widthvector\|_\infty \leq D',
            \quad \Attn\in \cA (D,B).
        \end{align}
    For the link function $\NNs^\linkweight$, its weight satisfies $\|\linkweight\|_\infty \leq B$. 
    \end{definition}

    The subsequent sections consist as follows.
    \Cref{subsection:Proof-of-CLIP} combines \Cref{lemma:Approximation-FF} and \Cref{lemma:Self-attention-CLIP}, as well as the bound on the propagation of the intermediate errors (\Cref{lemma:Appendix-CLIP-errorpropagation}) to prove \Cref{theorem:CLIP-Generalization}. 
    \Cref{subsection:FF} and \Cref{subsection:Selt-attention} will provide the proof of \Cref{lemma:Approximation-FF} and \Cref{lemma:Self-attention-CLIP}, respectively.
    The proof of the error propagation lemma (\Cref{lemma:Appendix-CLIP-errorpropagation}) will be found in \Cref{subsection:ErrorPropagation}.  \Cref{subsection:Appendix-CLIP-Auxiliary} provides some useful properties about the message passing algorithm.

\subsection{Proof of \Cref{theorem:CLIP-Generalization}}\label{subsection:Proof-of-CLIP}
We prove \Cref{theorem:CLIP-Generalization} by combining \Cref{lemma:Approximation-FF},  \Cref{lemma:Self-attention-CLIP}, and \Cref{lemma:Appendix-CLIP-errorpropagation}, as well as several auxiliary lemmas.
By definition, the excess risk $\excess_K(\scorefun^{\EstPar}_\NN,\scorefun_\star)$ has the following decomposition:
\begin{align}
\excess_K(\scorefun^{\EstPar}_{\NN},\scorefun_\star)
&
= \orisk(\scorefun^\EstPar_{\NN})-\orisk(\scorefun_\star)
\\
&=\underbrace{\inf_{\Par\in\Parspaceclip}\orisk(\scorefun^{\Par}_{\NN})-\orisk(\scorefun_\star)}_{\text{approximation error}}
+
\underbrace{\orisk(\scorefun^\EstPar_{\NN})
-
\inf_{\Par\in\Parspaceclip}\orisk(\scorefun^{\Par}_{\NN})}_{\text{generalization error}}.
\end{align}
We claim that
\begin{itemize}
    \item [(a).] If we choose $J = \bigOtil(L)$, $D = \bigO(\state \layer)$, $D' =\bigOtil( \mmaxb\state \layer^{3})$, and $\bound = \bigOtil( \state \layer+\mmaxb^2)$, then the approximation error satisfies
    \begin{align}
\inf_{\Par\in\Parspaceclip}\orisk_{\clip,K}(\scorefun^{\Par}_{\NN})-\orisk_{\clip,K}(\scorefun_\star) 
\leq
\bigOtil\Bigg(\sqrt{\frac{\state^2\layer^{11}\mmaxb^2}{n}}\Bigg)
    \end{align}
    \item[(b).] Under the same choice of model class $\Parspaceclip$, the generalization error satisfies
    \begin{align}
    \orisk(\scorefun^\EstPar_{\NN})
-
\inf_{\Par\in\Parspaceclip}\orisk(\scorefun^{\Par}_{\NN})
\leq 
\bigOtil\Bigg(\sqrt{\frac{\state^2\layer^{11}\mmaxb^2}{n}}\Bigg)
    \end{align}with probability at least $1-1/n.$
\end{itemize}
Putting pieces together yields Theorem~\ref{theorem:CLIP-Generalization}. The remainder of this section is devoted to proving these claims.
\\
\paragraph{(a) Approximation error.}
 Note that
\begin{align}
    &\orisk_{\clip,K}(\scorefun_\NN^\Par)-\orisk_{\clip,K}(\scorefun_\star)
  \\  &\leq
     \E \Big|
     \log \frac{\exp(\scorefun_\NN^\Par(\xi_{,1}, \xt_{,1})  ) }{\sum_{j \in [K]} \exp(\scorefun_\NN^\Par(\xi_{,1}, \xt_{, j}))} 
     -
     \log \frac{\exp(\simscore_\star(\xi_{,1}, \xt_{,1})  ) }{\sum_{j \in [K]} \exp(\simscore_\star(\xi_{,1}, \xt_{, j}))} \Big| \\
 &    
~~~+ 
\E \Big| 
 \log \frac{\exp(\scorefun_\NN^\Par(\xi_{,1}, \xt_{,1})  ) }{\sum_{j \in [K]} \exp(\scorefun_\NN^\Par(\xi_{, j}, \xt_{,1}))} 
- 
\log \frac{\exp(\simscore_\star(\xi_{,1}, \xt_{,1})  ) }{\sum_{j \in [K]} \exp(\simscore_\star(\xi_{, j}, \xt_{,1}))} 
\Big|\\
&\leq
2 \E\max_{j\in[K]}\Big|
    \scorefun_\NN^\Par(\xi_{,1}, \xt_{, j}) 
     -
    \simscore_\star(\xi_{,1}, \xt_{,j}) \Big|
    +
    2 \E\max_{j\in[K]}\Big|
    \scorefun_\NN^\Par(\xi_{,j}, \xt_{, 1}) 
     -
    \simscore_\star(\xi_{,j}, \xt_{,1}) \Big|
    \\
    &\leq
    4\max_{(\xi,\xt)\in\cXi\times\cXt}
    | \scorefun_\NN^\Par(\xi, \xt) 
     -
    \simscore_\star(\xi, \xt)|,
\end{align}
where the second inequality follows from Lemma~\ref{lm:log_softmax}. Therefore, it remains to find some parameter $\Par\in\Parspaceclip$ such that $\max_{(\xi,\xt)\in\cXi\times\cXt}
    | \scorefun_\NN^\Par(\xi, \xt) 
     -
    \simscore_\star(\xi, \xt)|\leq \tO(\sqrt{\state^2\layer^{11}\mmaxb^2/n})$.

    Take some $\delta'>0$ which will be defined later.
    For the feed forward layers, we use  \Cref{lemma:Approximation-FF} with $\delta=\delta'\ll 1$.
    For the self-attention layers, we use \Cref{lemma:Self-attention-CLIP} with $\delta=\frac{\delta'}{\max_{v}\|\qnetwork^{(\ell)}_{v}\|_\infty}$.
    According to \Cref{lemma:Boundedness-1}, $f^{(\ell)}_{\iota(\pa^{(L-\ell)}(v))}(\hnetwork^{(\ell)}_{v})$ are all bounded by $2\log S\transbd$ with the $\|\cdot\|_\infty$-norm, and the approximation error from \Cref{lemma:Self-attention-CLIP} is $\delta$.
    Thus $\qnetwork^{(\ell)}_{v}=f^{(\ell)}_{\iota(\pa^{(L-\ell)}(v))}(\hnetwork^{(\ell)}_{v})$ is bounded by $2\log S\transbd+\delta\leq 3(1\lor\log S\transbd)$, and $\delta$ in \Cref{lemma:Self-attention-CLIP} is bounded by $\frac{\delta'}{3(1\lor\log S\transbd)}$. 

    The error from each operation is then bounded by $\delta'$ in the $\|\cdot\|_\infty$-norm.
    Now we can apply \Cref{lemma:Appendix-CLIP-errorpropagation} to obtain that
    %
    \begin{align}
&\max_{(\xi,\xt)\in\cXi\times\cXt}|\scorefun^\Par_\NN(\xi,\xt)-\scorefun_\star(\xi,\xt)| 
    \\ &\textstyle\leq 
    \delta' \times 
     \big[\prod_{1\leq \ell\leq L}(2m^{(\ell)}+3) + \prod_{1\leq \ell\leq L}(2m^{(\ell)}+3)\big]
     \leq 
     2 \cdot 5^L \delta'd,
     \label{eq:ApproxError-1}
    \end{align}
    where we set $d = \max\{d_\image,d_\txt\}$. 
        
    Choose 
    \begin{align}
        \delta'=\frac{\sqrt{\state^2\layer^{11}\mmaxb^2}}{5^Ld\sqrt{n}}
    \end{align}
    with $\mmax=\max\{\max_{k}m^{(k)}_\txt, \max_{k}m^{(k)}_\image\}$, so that the approximation error \eqref{eq:ApproxError-1} is bounded by $\sqrt{\frac{ \state^2 \layer^{11}\mmaxb^2}{n}}$.
    According to \Cref{lemma:Approximation-FF} and \Cref{lemma:Self-attention-CLIP}, 
    we now know that there exists some parameter $\Par\in\Parspaceclip$ such that 
    Eq.~\eqref{eq:ApproxError-1} is satisfied, where
    \begin{align}
        D&\leq d_\feature+d_\positional = 2(S+1)L+1=\bigO(\state\layer),
        \\
        J &\lesssim (\log\log(S\transbd/\delta'))\log(S\transbd/\delta')=\bigOtil(\layer),
        \\
        D'=\|\widthvector\|_\infty &\lesssim \mmax S(\log(S\transbd/\delta'))^3+d_\feature+d_\positional
        =\bigOtil(\mmaxb\state\layer^3),
        \\
        B &\lesssim S(\transbd^2+\log(S\transbd/\delta'))+\mmax^2+\log\frac{d\log(S\transbd)}{\delta'}=\bigOtil(\state\layer+\mmaxb^2).
    \end{align}
  
\paragraph{(b) Generalization error analysis.}
Since $\EstPar$ is the minimizer of $\what\risk_{\clip,K}(\scorefun_\NN^\Par)$ defined in Eq.~\eqref{eq:clip-empirical-risk-minimization}, we have 
\begin{align}
    \orisk_{\clip,K}(\scorefun_\NN^{\EstPar})-\inf_{\Par\in\Parspaceclip}\orisk_{\clip,K}(\scorefun_\NN^\Par)
    &\leq
    2\sup_{\Par\in\Parspaceclip}|\what\risk_{\clip,K}(\scorefun_\NN^\Par)
    -
    \orisk_{\clip,K}(\scorefun_\NN^\Par)|.\label{eq:erm_bound_clip}
\end{align}

 Next, we verify the conditions required for Lemma~\ref{lm:generalization_lemma} and then  apply the lemma to obtain an upper bound for the R.H.S. of Eq.~\eqref{eq:erm_bound_clip}.

{
 In Lemma~\ref{lm:generalization_lemma}, take $\Theta=\Parspaceclip,\rho(\Par,\Par')=\nrmp{\Par-\Par'}$, $z_i=(\xi_{,j}^{(i)},\xt_{,j}^{(i)})_{j\in[K]}$, and
   \begin{align}
   f(z_i;\Par)=- \frac{1}{K}\sum_{k=1}^{K}\log \frac{\exp(\simscore_\NN^\Par(\xi^{(i)}_{,k}, \xt^{(i)}_{,k})  ) }{\sum_{j \in [K]} \exp(\simscore_\NN^\Par(\xi^{(i)}_{,k}, \xt^{(i)}_{, j}))} 
 - \frac{1}{K}\sum_{k=1}^{K}\log \frac{\exp(\simscore_\NN^\Par(\xi^{(i)}_{,k}, \xt^{(i)}_{,k})  ) }{\sum_{j \in [K]} \exp(\simscore_\NN^\Par(\xi^{(i)}_{, j}, \xt^{(i)}_{,k}))} .
 \end{align}
 for $i=1,\dots,\numbatch$.
   \\

\myunder{Verification of condition~(a) in Lemma~\ref{lm:generalization_lemma}.}
We note that the set $\Parspaceclip$ with metric $\rho(\Par,\Par')=\nrmp{\Par-\Par'}$ has a diameter $\bound_\rho\defn2\bound$. Moreover, $\Parspaceclip$ has a dimension bounded by $d_\rho\defn (\numfflayer+3)\layer(D+D'+1)^2+\state=\bigOtil(\state^2\layer^8\mmaxb^2)$. Thus, by Example 5.8 in~\cite{wainwright_2019}, we have
$\log\cN(\Delta;\Parspaceclip,\nrmp{\cdot})\leq d_\rho\log(1+2 r/\Delta)\leq d_\rho\log(2 A_\rho r/\Delta)$ for $\Delta\in(0,2r]$ with $A_\rho=2$.\\

\myunder{Verification of condition~(b) in Lemma~\ref{lm:generalization_lemma}.}
Since $\simscore_\NN^\Par$ is $\readbd$-bounded with $\readbd =4\mmaxone\log\transbd $ by Lemma~\ref{lm:simscore_lip}, it follows that $f(z_i;\Par)-\E[f(z_i;\Par)]$ is $\sigma=c\readbd$-sub-Gaussian for all $\Par\in\Parspaceclip$  for some numerical constant $c>0$ from Lemma~\ref{lm:empirical_infonce_loss_bounded_diff_concen}.\\

\myunder{Verification of condition~(c) in Lemma~\ref{lm:generalization_lemma}.}
By Lemmas~\ref{lm:simscore_lip},~\ref{lm:log_softmax}~and~\ref{lm:empirical_infonce_loss_bounded_diff_concen}, we have
\begin{align}
    |f(z_i;\Par) - f(z_i;\Par')|
    &\leq 4
    \Big|
    \scorefun_\NN^\Par(\xi_{,1}, \xt_{, j}) 
     -
    \simscore_\NN^{\Par'}(\xi_{,1}, \xt_{,j}) \Big|
    +
    4 \Big|
    \scorefun_\NN^{\Par}(\xi_{,j}, \xt_{, 1}) 
     -
    \simscore_\NN^{\Par'}(\xi_{,j}, \xt_{,1}) \Big|\\
    & \leq \bound_f \nrmp{\Par-\Par'},~~~~~~~~\text{where}~~ \bound_f\defn 4((c\bound)^{18\numfflayer\layer}\state^4)^{\layer+1}.
\end{align}Therefore, we may choose $\sigma'=\bound_f$ and condition (c) is hence satisfied.

Now, invoking Lemma~\ref{lm:generalization_lemma} and plugging in the values of $d_\rho,\sigma,\sigma',A_\rho,\bound_\rho$, we find
\begin{align}
 \sup_{\Par\in\Parspaceclip}|\what\risk_{\clip,K}(\scorefun_\NN^\Par)
    -
    \orisk_{\clip,K}(\scorefun_\NN^\Par)|
    &\leq c \sigma \sqrt{\frac{d_\rho \log \left(2A_\rho \left(1 + B_\rho \sigma' / \sigma\right)\right) + \log(1 / \eta)}{\numbatch}}\\
    &\leq \bigOtil \Bigg( \sqrt{\frac{ \state^2 \layer^{11}\mmaxb^2+\log(1/\eta)}{\numbatch}} \Bigg)
\end{align} with probability at least $1-\eta.$ Setting $\eta=1/n$ completes the proof.
}

\subsubsection{Proof of Corollary~\ref{cor:ZSC_GHM}}\label{sec:clip_cor_pf}
By Lemma~\ref{lemma:BoundScore} and the definition of the readout function $\readsimscore(\cdot)$ in $\NNs^\linkweight(\cdot)$, we have Assumption~\ref{ass:bounded_score} is satisfied with $\boundscoreconst=(\transbd)^{4\mmaxone}$. Thus, Corollary~\ref{cor:ZSC_GHM} follows immediately from combining Theorem~\ref{theorem:CDM-Generalization-two-step} and Proposition~\ref{prop:validity_zero_shot_classification}.

\subsection{Position-wise feed forward layer (proof of \Cref{lemma:Approximation-FF})}\label{subsection:FF}
%
Now we construct ReLU networks that approximate $f^{(\ell)}_{\iota(\pa^{(L-\ell)}(v))}$ to prove \Cref{lemma:Approximation-FF}. 
We first approximate each $f^{(\ell)}_\iota$ as follows. 
\Cref{lemma:Appendix-CLIP-Existence-log-sum-exp} covers $\ell\leq L-1$ and \Cref{lem:existence_ReLU_log_Psi} covers $\ell=L$.
\begin{lemma}[ReLU network approximation of log-sum-exponential]\label{lemma:Appendix-CLIP-Existence-log-sum-exp}
Fix $\ell\in [L-1]$ and $\delta>0$.
Assume that $\transbd^{-1} \leq \psi^{(\ell)}_{\iota}(s,a) \leq \transbd$ for all $s,a\in [S]$.
When $\ell=1$, also assume that $\transbd^{-1}\leq \P[s]\leq \transbd$ for all $s$. 
Then, there exists an $\NN\in \mathcal{F}(\depth,\widthvector=(S,\dots,S),B)$ such that
\begin{align}
    \|\NN(h) - f^{(\ell)}_{\iota}(h)\|_\infty \leq \delta, \text{ for all $h\in \R^S$ with $\max_s h_s = 0$}.
\end{align}
Here, the network parameters $\depth,\widthvector$ and $B$ are bounded by
\begin{align}
    \depth \lesssim (\log\log(\staten\transbd/\delta))\log(\staten\transbd/\delta),\quad
    \|\widthvector\|_\infty  \lesssim \staten(\log(\staten\transbd/\delta))^3,
    \quad
    B  \lesssim 2\staten(\transbd^2+\log(\staten\transbd/\delta)).
\end{align}
\end{lemma}

\begin{lemma}[ReLU network approximation of log-Psi]\label{lem:existence_ReLU_log_Psi}
Let $\delta>0$.
Assume that $\transbd^{-1} \leq \psi^{(L)}_{\iota}(s,a) \leq \transbd$ for all $s,a\in [S]$. 
Then, there exists an $\NN\in \mathcal{F}(\depth,\widthvector=(1,\dots,S),B)$ such that
\begin{align}
    \|\NN(s) - f^{(L)}_{\iota}(s)\|_\infty \leq \delta, \text{ for all $s\in [S]$}.
\end{align}
Here, the network parameters $\depth,\widthvector$ and $B$ are bounded by 
\begin{align}
    \depth \lesssim (\log \log (\log(\transbd)/\delta)) \log (\log(\transbd)/\delta), \quad \|\widthvector\|_\infty \leq S (\log (\log (\transbd)/\delta))^2 \log \transbd,\quad B \lesssim \transbd+S.
\end{align}
\end{lemma}
Their proofs are deferred to \Cref{subsubsection:CLIP-FF-1}. 
In addition to approximating a single $f^{(\ell)}_\iota$, the ReLU network should identify $\iota(\pa^{(\ell)}(v))$ using the positional encoding and apply the correct $\fnetwork^{(\ell)}_{\iota}$ to $\hnetwork^{(\ell)}_v$.
The following lemma states that this can be implemented with no errors.
\begin{lemma}[Identify the rank from positional encoding]\label{lemma:IdeitifyTheRank}
    Fix $\ell\in [L]$. 
    Suppose that we have $m^{(\ell)}$ different networks $\NN_1\in \mathcal{F}(\depth_1,\widthvector_1,B_1),\dots,\NN_{m^{(\ell)}}\in \mathcal{F}(\depth_{m^{(\ell)}},\widthvector_{m^{(\ell)}},B_{m^{(\ell)}})$ with the shared input and the same output dimension $k$, and the outputs of these networks are bounded by $C$ with the $\|\cdot\|_\infty$-norm. 
    Then, for $v\in \cV^{(L)}$, there exists a ReLU network $\NN$ that selects $\NN_{\iota(\pa^{(L-\ell)}(v))}$ given $\pnetwork_{v}$, i.e., 
    \begin{align}
        \NN([h; \pnetwork_{v}]) = \NN_{\iota(\pa^{(L-\ell)}(v))}(h).
    \end{align}
    This network satisfies that 
    \begin{align}
    \textstyle
    \depth=\max_i\depth_i, 
    \ \|\widthvector\|_\infty \leq m^{(\ell)}+2\sum_i^{m^{(\ell)}}\|\widthvector_i\|, 
    \
    B = \max_iB_i + (m^{(\ell)})^2+C.
    \end{align}
 \end{lemma}
Using \Cref{lemma:IdeitifyTheRank} together with \Cref{lemma:Appendix-CLIP-Existence-log-sum-exp,lem:existence_ReLU_log_Psi}, for any $\ell\in [S]$, there exists a ReLU network such that, for all $v \in \cV^{(L)}$, it takes $\hnetwork^{(\ell)}_{v}$ and $\pnetwork^{(\ell)}_{v}$ and outputs $f^{(\ell)}_{\iota(\pa^{(L-\ell)}(v))}(\hnetwork^{(\ell)}_{v})$ with the $\|\cdot\|_\infty$-error at most $\delta$, where
\begin{align}
    \depth \lesssim (\log\log(S\transbd/\delta))\log(S\transbd/\delta),\
    \|\widthvector\|_\infty  \lesssim m^{(\ell)}S(\log(S\transbd/\delta))^3,
    \
    B  \lesssim S(\transbd^2+\log(S\transbd/\delta)) + (m^{(\ell)})^2.
\end{align}
Note that $f^{(\ell)}_{\iota(\pa^{(L-\ell)}(v))}$ (and thus its neural network approximation) is bounded by $O(\log (S\transbd))$ according to \Cref{lemma:Boundedness-1}, which gives $C \lesssim \log (SK)$ in the application of \Cref{lemma:IdeitifyTheRank}. 
Now, we have obtained \Cref{lemma:Approximation-FF}. 

 \paragraph{Approximation with constant depth.}
While we used the networks with polylogarithmic depth, we add a remark on how the analysis changes when we use networks with constant depth.

The networks that approximate basic functions are changed as follows:
\begin{itemize}
    \item[](Approximation of logarithm function.)\  
    There exists a network that achieves the same bound as \Cref{lem:ReLU_logarithm}, where  
    \begin{align}
        \depth = 1, \quad
        \|\widthvector\|_\infty \leq \lceil 2 A/\delta \rceil + 1
        ,\quad
        B \leq e^A.
    \end{align}
    This two-layer approximation is obtained as a modification of Lemma 9 of \cite{mei2024u}, where we choose $e_j = 2Aj/(M-1)-A$, $b_j = - \exp(e_j)$ for $j \in [M-1]$, $a_1 = (e_{2} - e_{1})/(b_{2} - b_1)$ and $a_j = (e_{j+1} - e_{j})/(b_{j+1} - b_j) - (e_{j} - e_{j-1})/(b_{j} - b_{j-1})$ for $2 \le j \le M - 2$ to obtain $\mathrm{NN}_{\log}(x) = \sum_{j = 1}^{M-2} a_j \ReLU(x +  b_j) + \ReLU(- x + e_1) - \ReLU(- x)$.
    \item[](Approximation of exponential function)\  
    According to Lemma 8 of \cite{mei2024u}, there exists a network that achieves the same bound as \Cref{lem:ReLU_exponential}, where  
    \begin{align}
        J=1,\quad \|\widthvector\|_\infty = \lceil\delta^{-1} \rceil + 1, \quad B \leq \log (\lceil\delta^{-1} \rceil + 1),
    \end{align}
\end{itemize}

By using these networks, 
we can obtain \Cref{lemma:Approximation-FF} with the following bound:
\begin{align}
    \depth =3,\
    \|\widthvector\|_\infty  \lesssim m^{(\ell)}S^2\transbd^5 \delta^{-1}\log (S\transbd),
    \
    B  \lesssim 
    (m^{(\ell)})^3S^5\transbd^8 \delta^{-1}(\log (S\transbd/\delta))^2.
\end{align}

The problem here is that the dependency on $\delta$ is $\delta^{-1}$, while in the original bound it was polylogarithmic. 
In the proof of \Cref{theorem:CLIP-Generalization}, we need to take $\delta \lesssim d^{-1}$. Then the parameter $\|\widthvector\|_\infty$ linearly depends on $d$, which incurs linear dependency on $d$ in the generalization error bound.
This problem is avoided when we are allowed to have polylogarithmic depth as in the original \Cref{lemma:Approximation-FF} (or polylogarithmic number of blocks with each feed forward layer being constant depth). 

\subsubsection{Proofs of Lemmas~\ref{lemma:Appendix-CLIP-Existence-log-sum-exp},~\ref{lem:existence_ReLU_log_Psi},~and~\ref{lemma:IdeitifyTheRank}}\label{subsubsection:CLIP-FF-1}
Lemmas~\ref{lemma:Appendix-CLIP-Existence-log-sum-exp},~\ref{lem:existence_ReLU_log_Psi},~and~\ref{lemma:IdeitifyTheRank} compose modules that approximate basic functions such as logarithm and exponential. 
The proofs of these basic modules will be deferred to \Cref{subsubsection:basicfunctionapproximation}.
First we review basic operations which we will use without a proof (borrowed from Appendix B.1.1. of \cite{nakada2020adaptive}).
\paragraph{Composition of two networks.}
When we want to construct a composition of two networks $\mathrm{NN}_1\in \mathcal{F}(\depth_1,\widthvector_1=(\dots,j),B_1)$ and $\mathrm{NN}_2\in \mathcal{F}(\depth_2,\widthvector_2=(j,\dots),B_2)$, a naive way is to multiply the last layer's matrix of one network with the first layer's matrix with the other, where the parameter bound $B_{1+2}$ of the new network is  $jB_1B_2$. 
However, the following construction can bound $B_{1+2}$ more tightly with one additional layer. 
Let $\bW_i^{(k)}$ be the parameters of the $k$th layer of $\NN_i\ (i=1,2)$.
We define 
\begin{align}
    \label{eq:composition-of-two-networks}
    \mathrm{NN}_{1+2} &= \bW_2^{(\depth+1)}\ReLU (\bW_2^{(\depth)}[\cdot ;1 ]) \circ \cdots \circ \ReLU ([\bW_2^{(1)}\ \widetilde{\bW}_2^{(1)}][\cdot ;1 ]) 
    \\ & \quad \circ \ReLU \bigg(\begin{bmatrix}
         \bW_1^{(\depth+1)} \\ - \bW_1^{(\depth+1)}
    \end{bmatrix}
    [\cdot ;1 ]
    \bigg) \circ \cdots \circ \ReLU (\bW_1^{(1)}[\cdot ;1 ]).
\end{align}
Here $\widetilde{\bW}_2^{(1)}$ is a matrix such that $(\widetilde{\bW}_2^{(1)})_{k,l}=-(\bW_2^{(1)})_{k,l}$ for all $k,l$, except that $(\widetilde{\bW}_2^{(1)})_{k,l}=(\bW_2^{(1)})_{k,l}$ in the column corresponding to the bias term.
It is easy to check that $\mathrm{NN}_{1+2}$ implements the composition of $\NN_1$ and $\NN_2$, considering that  either the first half or the latter half columns of $\ReLU \big(\begin{bmatrix}
         \bW_1^{(\depth+1)} \\ - \bW_1^{(\depth+1)}
    \end{bmatrix}
    [\cdot ;1 ]
    \big)$ is zero. 
Moreover, we have $\NN_{1+2}\in \mathcal{F}(\depth_{1+2},\widthvector_{1+2},B_{1+2})$ with
$\depth_{1+2}=\depth_1+\depth_2+1$, $\|\widthvector_{1+2}\|_\infty \leq 2 \max\{\|\widthvector_1\|_\infty, \|\widthvector_2\|_\infty\}$, and $B_{1+2} \leq \max\{B_1,B_2\}$. 

\paragraph{Identify function.}
The identity function for $d$-dimensional inputs is implemented as a ReLU network with arbitrary depth:
    \begin{align}
         \begin{bmatrix}
        \bI_d 
        \\ -\bI_d
        \end{bmatrix}\ReLU\bigg(
        \begin{bmatrix}
        \bI_d & 0
        \\ 0 & \bI_d
        \end{bmatrix}\cdot
        \bigg)\circ \cdots \circ\ReLU\bigg(
        \begin{bmatrix}
        \bI_d & 0
        \\ 0 & \bI_d
        \end{bmatrix} \cdot\bigg) \ReLU\Big(
        \begin{bmatrix}
        \bI_d & - \bI_d
        \end{bmatrix}\cdot 
        \Big).
    \end{align}
\paragraph{Parallelization.}
When there are multiple networks $\mathrm{NN}_i\in \mathcal{F}(\depth_i,\widthvector_i,B_i)\ (i=1,\dots,I)$ that share the input $x$, we can construct a larger network $\NN\in \mathcal{F}(\depth,\widthvector,B)$ that outputs $[\mathrm{NN}_1;\cdots;\mathrm{NN}_I]$, where $\depth=\max_i \depth_i$, $\|\widthvector\|_\infty \leq 2\sum_{i=1}^I \|\widthvector_i\|_\infty$, and $B \leq \max_i B_i$. 
Specifically, we first unify the depths of these networks by composing an identity function with each $\NN_i$.
We then concatenate these networks, by making a block diagonal matrix where the block diagonal parts are matrices of the original networks.

Now we provide the proofs of Lemmas~\ref{lemma:Appendix-CLIP-Existence-log-sum-exp},~\ref{lem:existence_ReLU_log_Psi},~and~\ref{lemma:IdeitifyTheRank} in order.
\begin{proof}[Proof of Lemma~\ref{lemma:Appendix-CLIP-Existence-log-sum-exp}]
We focus on the case of $\ell=1$, as the proof for $\ell\geq 2$ follows similarly (just delete all the $\P[s]^{\frac{1}{m^{(1)}}}$ terms).
We utilize two ReLU networks $\mathrm{NN}_{\log}(x)$ and $\mathrm{NN}_{\exp}(x)$, defined in \Cref{lem:ReLU_logarithm} and \Cref{lem:ReLU_exponential}.
We will determine the values of $\delta'$ and $A$ later.
\begin{itemize}
    \item $\mathrm{NN}_{\log}(x)$, which approximates $\log (x)$ within the error of $\delta'$ for $e^{-A}\leq x \leq e^A$, with \\ 
    $\depth \lesssim (\log\log(A/\delta'))\log (A/\delta')$, $\|\widthvector\|_\infty \lesssim A (\log (A/\delta'))^2$, and $B\leq e^A$ (see \Cref{lem:ReLU_logarithm} for construction). 
    \item $\mathrm{NN}_{\exp}(x)$, which approximates $\log (x)$ within the error of $\delta'$ for $x \leq 0$, 
    $\depth \lesssim (\log\log(1/\delta'))\log (1/\delta')$, $\|\widthvector\|_\infty \lesssim (\log (1/\delta'))^3$, and $B\lesssim \log (1/\delta')$ (see \Cref{lem:ReLU_exponential} for construction).
\end{itemize}
We define $\NN_1$ and $\NN_2$ by parallelizing $S$ instances of $\mathrm{NN}_{\exp}$ and $\mathrm{NN}_{\log}$, respectively.

The function we want to implement is $f^{(1)}_\iota$, which is
\begin{align}
\textstyle
    f^{(1)}_\iota(h)_s = \log \sum_{a\in [S]}\P[s]^{\frac{1}{m^{(1)}}}\psi_{\iota}^{(1)}(s,a)e^{h_a},\quad h\in \R^S.
\end{align} 
Therefore, we combine $\NN_1$, $\Psi=(\P[s]^{\frac{1}{m^{(1)}}}\psi_{\iota}^{(1)}(s,a))_{s,a}\in \R^{S\times S}$, and $\NN_2$ to yield the desired network. 
The network parameters are bounded by
\begin{align}
    \depth \lesssim (\log \log (A/\delta')) \log (A/\delta'), \quad \|\widthvector\|_\infty \leq S A(\log (A/\delta'))^2 + S(\log (1/\delta'))^3,\quad B \lesssim e^A + \log (1/\delta')+\transbd^2.
\end{align}

Let us determine the value of $\delta'$ and $A$.
Note that, for $h\in \R^S$ with $\max_a h_a=0$, we have
\begin{align}
    \textstyle
   \transbd^{-2}(1-\delta') \leq \sum_{a\in [S]}\P[s]^{\frac{1}{m^{(1)}}}\psi_{\iota}^{(1)}(s,a) \NN_{\exp}(h_a) \leq S\transbd^2(1+\delta').
   \label{eq:Boundedness-of-output-1}
\end{align}
Because we will take $\delta' \ll 1$, we can assume that (\ref{eq:Boundedness-of-output-1}) 
is bounded by $S\transbd^2(1+\delta')\leq 2S\transbd^2$ and $\transbd^{-2}(1-\delta')\geq (2\transbd^2)^{-1}$.
Thus we let $A = \log (2SK^2)$ in the definition of $\mathrm{NN}_{\log}(x)$.
Also, the overall error $\|\NN(h) - f^{(1)}_{\iota}(h)\|_\infty$ is bounded by
\begin{align}
\textstyle
   \delta' + \frac{\transbd^2}{1-\delta'} (2S\transbd^2)\delta'. 
\label{eq:CLIP-target-error}
\end{align}
Here, the first $\delta'$ is the approximation error of $\log$, and $\frac{\transbd^2}{1-\delta'}$ is the smoothness of $\log(t)$ in $\transbd^{-2}(1-\delta')\leq t$, $2S\transbd^2$ is the amplification rate of the approximation error of $\exp$ (by $\sum_{a\in [S]}$ and multiplying $\P[s]^{\frac{1}{m^{(1)}}}\psi_{\iota}^{(1)}(s,a)$), and the final $\delta'$ corresponds to the approximation error of $\exp$. 
It suffices to take $\delta' \leq \frac{\delta}{8S\transbd^4}$ to achieve $\eqref{eq:CLIP-target-error} \leq \delta$.

Now, evaluating the parameters of the desired network with these $\delta'$ and $A$, we obtain the desired bound. 
\end{proof} 

\begin{proof}[Proor of \Cref{lem:existence_ReLU_log_Psi}]
We use the following basic networks.
\begin{itemize}
    \item $\mathrm{NN}_{\log}(x)$ from \Cref{lem:ReLU_logarithm}, which approximates $\log (x)$ within the error of $\delta$ for $e^{-A}\leq x \leq e^A$, with 
$\depth \lesssim (\log\log(A/\delta))\log (A/\delta)$, $\|\widthvector\|_\infty \lesssim A (\log (A/\delta))^2$, and $B\leq e^A$. 
    \item $\NN_{\Ind[s]}(x)$ from \Cref{lem:ReLU_indicator}, which implements $\indicator[x=s]$ exactly, whose parameters are bounded by $\depth=1, \|\widthvector\|_\infty \lesssim S$, and $B=S+1$.
\end{itemize}
We define $\NN_1$ by parallelizing $S$ instances of $\mathrm{NN}_{\log}$, while the input $x$ is shared.
Also, we define $\NN_2$ by parallelizing $\NN_{\Ind[s]}\ (s=1,\dots,S)$).

The function we want to implement is $f^{(L)}_\iota$, which is
\begin{align}
    \textstyle
    f^{(L)}_\iota(x) = \log \sum_{s\in [S]}\psi_{\iota}^{(L)}(s,a)\indicator[x=s],\ \in \R^S, \quad x\in [S].
\end{align}

By combining $\NN_1$, a matrix $\Psi=(\psi_{\iota}^{(L)}(s,a))_{s,a}\in \R^{S\times S}$, and $\NN_2$, 
The network parameters are bounded by
\begin{align}
    \depth \lesssim (\log \log (A/\delta)) \log (A/\delta), \quad \|\widthvector\|_\infty \lesssim S A(\log (A/\delta))^2 ,\quad B \lesssim e^A + \transbd+S.
\end{align}

Finally, let us determine the value of $A$. 
Because it holds that
\begin{align}
    \textstyle
   \transbd^{-1}\leq \sum_{a\in [S]}\psi_{\iota}^{(\ell)}(s,a) \NN_{\Ind}(s) \leq \transbd.
    \label{eq:Boundedness-of-output-3}
\end{align}
it suffices to take $A = \log (\transbd)$.
Also, because the computation of $\sum_{a\in [S]}\psi_{\iota}^{(\ell)}(s,a) \NN_{\Ind}(s)$ is exact, the approximation error only comes from $\NN_{\log}$. Thus, the approximation error is bounded by $\delta$. 

Now, evaluating the parameters of the desired network with $A = \log (\transbd)$, we have the desired bound. 
\end{proof}
\begin{proof}[Proof of \Cref{lemma:IdeitifyTheRank}] 
    Suppose that we have a network $\overline{\NN}\in \cF(\bar{J},\bar{\widthvector},\bar{B})$ that takes $[h;\pnetwork_{v^{(L)}}]$ and outputs
    \begin{align}
    \begin{bmatrix}
        \NN_1(h) & \cdots &\NN_{m^{(\ell)}}(h) & \indicator[\iota(\pa^{(L-\ell)}(v))=1]
        & \cdots
        & \indicator[\iota(\pa^{(L-\ell)}(v))=m^{(\ell)}]
    \end{bmatrix}^\top .
          \label{eq:Identify-rank-of-nodes-1}
    \end{align}
    Then we compose this with a one layer ReLU network, with the first layer matrix
    \begin{align}
    \left[ \begin{array}{c|c}
        \bI_{km^{(\ell)}} & 
        \begin{matrix} 
        \bzero_k & -C \boldsymbol{1}_k& \cdots & -C \boldsymbol{1}_k
        \\ -C \boldsymbol{1}_k & \bzero_k & \cdots & -C \boldsymbol{1}_k
        \\ \vdots & \vdots & \ddots & \vdots
        \\ -C \boldsymbol{1}_k & -C \boldsymbol{1}_k & \cdots & \bzero_k
        \end{matrix}  \\
        \hline
        -\bI_{km^{(\ell)}}  & \begin{matrix} 
        \bzero_k & -C \boldsymbol{1}_k& \cdots & -C \boldsymbol{1}_k
        \\ -C \boldsymbol{1}_k & \bzero_k & \cdots &-C \boldsymbol{1}_k
        \\ \vdots & \vdots & \ddots & \vdots
        \\ -C \boldsymbol{1}_k & -C \boldsymbol{1}_k & \cdots & \bzero_k
        \end{matrix} 
    \end{array}\right]\in \R^{2km^{(\ell)} \times (k+1)m^{(\ell)}},  \quad (\bzero_k,  \boldsymbol{1}_k \in \R^k),
            \label{eq:Identify-rank-of-nodes-2}
    \end{align}
    and the first layer bias $\bzero$, and the second layer matrix
    \begin{align}
        \begin{bmatrix}
            \bI_{km^{(\ell)}} & -\bI_{km^{(\ell)}}
        \end{bmatrix}.
    \end{align}
    In \eqref{eq:Identify-rank-of-nodes-2}, applying the left columns to \eqref{eq:Identify-rank-of-nodes-1} yields $[\NN_1(\bx) \ \cdots \ \NN_{m^{(\ell)}}(\bx)\ -\NN_1(\bx) \ \cdots - \NN_{m^{(\ell)}}]^\top$. 
    From the boundedness assumption, each element of the obtained vector is in $[-C,C]$. 
    On the other hand, the right columns yields $-C$ in all coordinates but those corresponding to $\NN_{\iota(\pa^{(L-\ell)}(v))}(h)$ and $-\NN_{\iota(\pa^{(L-\ell)}(v))}(h)$.
    After applying the ReLU and the second layer matrix, we can single out only one $\NN_{\iota(\pa^{(L-\ell)}(v))}(\bx)$.

    Therefore, when we have a network $\overline{\NN}\in \cF(\bar{J},\bar{\widthvector},\bar{B})$ that computes 
    \eqref{eq:Identify-rank-of-nodes-1}, we get the desired network with size
    \begin{align}
         \depth=\bar{\depth}+2, \quad \|\widthvector\|_\infty \lesssim \|\bar{\widthvector}\|, 
        \quad B = \bar{B} +C.
        \label{eq:Identify-rank-of-nodes-3}
    \end{align}

    Finally, let us construct $\overline{\NN}$ to bound its network parameters $\bar{J},\bar{\widthvector},\bar{B}$.
    We use the $2(L-\ell+1)$th dimension of $\pnetwork_{v}$, which is $\cos(\frac{2\pi \iota(v)}{m^{\ell}})$, to identify the correct rank $\iota$. 
    According to \Cref{lem:ReLU_indicator}, there exists a ReLU network that implements $\indicator[x=\cos(\frac{2\pi \iota}{m^{\ell}})]$ for each $\iota = 1,2,\dots,m^{(\ell)}$, where $\depth = 1$, $\|\widthvector\|_\infty = 3$, and $B = m^{(\ell)}+2\delta^{-1}$. 
    We need to take $\delta = \min_i |\cos(\frac{2\pi i}{m^{\ell}}) - \cos(\frac{2\pi (i+1)}{m^{\ell}})| = 2 \sin^2(\frac{2\pi }{m^{(\ell)}})$. 
    By parallelizing this, there exists a ReLU network with $\depth = 1$, $\|\widthvector\|_\infty = 3m^{(\ell)}$, and $B = m^{(\ell)}+4\sin^{-2}(\frac{2\pi }{m^{\ell}})$, that takes $\pnetwork_{v}$ and outputs an $m^{(\ell)}$-dimensional vector $(\indicator[\iota(\pa^{(L-\ell)}(v))=1],\dots,\indicator[\iota(\pa^{(L-\ell)}(v))=m^{(\ell)}])$. 
    Concatenating this indicator network and $\NN_1,\dots,\NN_{m^{(\ell)}}$, 
    we have $\overline{\NN}$ with $\bar{\depth} = \max_i\depth_i$, $\|\bar{\widthvector}\|_\infty = m^{(\ell)}+2\sum_{i=1}^{m^{(\ell)}}\|\widthvector_i\|+3$, and $\bar{B} =\max_i B_i + 2m^{(\ell)}+4\sin^{-2}(\frac{2\pi }{m^{\ell}})$. 
    Putting these bounds into \eqref{eq:Identify-rank-of-nodes-3} yields the desired bound. 
\end{proof}




\subsubsection{ReLU network approximation of basic functions}\label{subsubsection:basicfunctionapproximation}
Here we construct ReLU newtorks that approximate basic functions for \Cref{subsubsection:CLIP-FF-1}.
\begin{lemma}[ReLU network approximation of logarithm function]\label{lem:ReLU_logarithm}
For any $A \in \N$, $\delta > 0$, there exists a network $\NN_{\log}(x)\colon \R\to\R$ that approximates $\log (x)$ within the error of $\delta$ for all $x\in [e^{-A},e^A]$, and that belongs to $\mathcal{F}(\depth,\widthvector=(1,j_2,j_3,\dots,1),B)$, where
\begin{align}
    \depth = 4+(\lceil \log_2 (12 A \lceil \log_2 (6A/\delta)\rceil^2/\delta)\rceil+5)\lceil \log_2 (\lceil \log_2 (6A/\delta)\rceil )\rceil
    ,\quad
    \|\widthvector\|_\infty &\leq 36 A \lceil \log_2 (6A/\delta)\rceil^2
    ,\quad 
    B \leq e^A.
\end{align}
Moreover, the network satisfies $-A-\delta \leq \NN_{\log}(x) \leq A+\delta$ for all $x\in \R$.
\end{lemma}
\begin{proof}

{\bf (1) Piece-wise polynomial approximation.}
Let us define $p_0 = e^{-A}, p_1 = e^{-A+\frac13}, p_2 = e^{-A+\frac23},\dots, p_{6A} = e^A$.
By defining $q_0 = -A$ and 
\begin{align}
    q_i (x) &= \log(\min\{\max\{x,p_{i-1}\},p_i\})-\log p_{i-1}
    \\
    & = \ReLU(\log (-\ReLU(-\ReLU(x-p_{i-1})-p_{i-1}+p_{i})+p_{i})-\log p_{i-1} )
    ,
\end{align}
we have
\begin{align}
    \textstyle
    \log (x) = \sum_{i=0}^{6A} q_i(x), \quad e^{-A}\leq x \leq e^A,
\end{align}
$(\mathrm{RHS}) = -A \ (x\leq e^{-A})$, and $(\mathrm{RHS}) = A \ (e^{A}\leq x)$.

For $1\leq i\leq 6A$, consider the Taylor expansion of $\log (x)-\log p_{i-1}$ at $p_{i-1} = e^{-A+\frac{i-1}{3}}$ as
\begin{align}
\textstyle
    \log (x)-\log p_{i-1} = \sum_{k=1}^{K} \frac{(-1)^{k+1}}{k} \left( \frac{x - p_{i-1}}{p_{i-1}} \right)^k + \frac{(-1)^{K+2}}{K+1} \left( \frac{y(K,y) - p_{i-1}}{p_{i-1}} \right)^{K+1},
\end{align}
where $p_{i-1} \leq y(x,K)\leq x$.
When $p_{i-1} \leq x\leq p_i$, the approximation error by the first $K$ terms is bounded by 
\begin{align}
\textstyle
   \Big| \frac{(-1)^{K+2}}{K+1} \left( \frac{y(K,y) - p_{i-1}}{p_{i-1}} \right)^{K+1} \Big| \leq \frac1{K+1} (e^{1/3}-1)^{K+1} \leq 2^{-(K+1)}. 
    \label{eq:Appendix-BasicFunctions-log-1}
\end{align}

Let $r_{i,k}(x) = \frac{(-1)^{k+1}}{k} \left( \frac{x - p_{i-1}}{p_{i-1}} \right)^k \ (1\leq k \leq K)$. 
If $r_{i,k}(x)$ is approximated by a function $\tilde{r}_{i,k}$ in $p_{i-1}\leq x\leq p_i$ within the error of $\delta'$, then
\begin{align}
\textstyle
    \ReLU\Big(q_0 + \sum_{i=1}^{6A} \ReLU(\sum_{k=1}^{K} \tilde{r}_{i,k}(-\ReLU(-\ReLU(x-p_{i-1})-p_{i-1}+p_{i})+p_{i})-\log p_{i-1} )\Big)
    \label{eq:Appendix-BasicFunctions-log-3}
\end{align}
approximates $\log (x)\ (e^{-A}\leq x \leq e^A)$ within the error of
\begin{align}
    6AK\delta' + 6A 2^{-(K+1)}.
    \label{eq:Appendix-BasicFunctions-log-2}
\end{align}
Here the first term comes from approximation of $r_{i,k}$ and the second term from Taylor expansion \eqref{eq:Appendix-BasicFunctions-log-1}.
Also, $-A-\eqref{eq:Appendix-BasicFunctions-log-2}\leq \eqref{eq:Appendix-BasicFunctions-log-3} \leq A+\eqref{eq:Appendix-BasicFunctions-log-2}$ holds for all $x$.
Thus, from now, our goal is to construct ReLU networks that approximate $r_{i,k}(x)$ within the error of $\delta'$.

If this goal is achieved, we take
\begin{align}
    K = \lceil \log_2 (6A/\delta)\rceil ,
\end{align}
and 
\begin{align}
    \delta' = \frac{\delta}{12 AK} = \frac{\delta}{12 A \lceil \log_2 (6A/\delta)\rceil}
\end{align}
so that the approximation error \eqref{eq:Appendix-BasicFunctions-log-2} of $\log x$ by \eqref{eq:Appendix-BasicFunctions-log-3} is bounded by $\delta$.

\noindent
{\bf (2) ReLU network approximation of monomials.}
According to Lemma A.4 of \cite{schmidt2020nonparametric} (focusing on only one $\boldsymbol{\alpha}$), there exists a neural network $\mathrm{Mult}_m^k(x)$ belonging to $\mathcal{F}\big(1+(m+5)\lceil \log_2 k\rceil, (1,6k,6k,\dots,6k,1), 1\big)$ such that
\begin{align}
   \sum_{0\leq x \leq 1} |\mathrm{Mult}_m^k(x) - x^k|\leq k^2 2^{-m}.
\end{align}
Then, because $p_{i-1}\leq x \leq p_i$ implies $0\leq x/p_{i-1}-1 \leq 1$, we have
\begin{align}
    \sum_{p_{i-1}\leq x \leq p_i} \bigg|\frac{(-1)^{k+1}}{k}\mathrm{Mult}_m^k (x/p_{i-1}-1) - r_{i,k}(x)\bigg| \leq k 2^{-m}.
    \label{eq:Appendix-BasicFunctions-log-4}
\end{align}
We take $m = \lceil \log_2 (K/\delta')\rceil = \lceil \log_2 (12 A \lceil \log_2 (6A/\delta)\rceil^2/\delta)\rceil$ so that \eqref{eq:Appendix-BasicFunctions-log-4} is bounded by $\delta'$ for all $i=1,\dots,6A$ and $k=1,2,\dots,K$.

We now know that there exists a network belonging to $\mathcal{F}\big(1+(m+5)\lceil \log_2 k\rceil, (1,6k,6k,\dots,6k,1), e^A\big)$ that approxmates $r_{i,k}$ within the error of $\delta'$.
As a result,  \eqref{eq:Appendix-BasicFunctions-log-3} using these networks yields the desired network belonging to $\mathcal{F}\big(4+(m+5)\lceil \log_2 K\rceil, \widthvector = (1,\dots,1), e^A\big)$, where $\|\widthvector\|_\infty \leq 36AK^2$.
\end{proof}

\begin{lemma}[ReLU network approximation of exponential function]\label{lem:ReLU_exponential}
For any $\delta > 0$, there exists a network $\NN_{\exp}(x)\colon \R\to\R$ that approximates $\exp (x)$ within the error of $\delta$ for all $x\leq 0$, and that belongs to $\mathcal{F}(\depth,\widthvector=(1,j_2,j_3,\dots,1),B)$, where
\begin{align}
    \depth &= 4+(\lceil \log_2 (8\lceil\log_2 (4\lceil \log 2\delta^{-1} \rceil/\delta)\rceil^2\lceil \log 2\delta^{-1} \rceil/\delta)\rceil+5)\lceil \log_2 (\lceil\log_2 (4\lceil \log 2\delta^{-1} \rceil/\delta)\rceil)\rceil,
    \\
    \|\widthvector\|_\infty &\leq 12 \lceil \log 2\delta^{-1} \rceil \lceil \log_2 (4\lceil \log 2\delta^{-1} \rceil/\delta)\rceil^2,
    \\ 
    B &\leq \lceil \log2\delta^{-1}\rceil \lor 2.
\end{align}
Moreover, the network satisfies $-\delta\leq \NN_{\exp}(x) \leq 1+\delta$ for all $x\in \R$.
\end{lemma}
\begin{proof}
    The proof 
    basically follows that of  \Cref{lem:ReLU_logarithm}. 
    We will show how to obtain the counterpart of \eqref{eq:Appendix-BasicFunctions-log-3}, and omit the rest.
    
    Let $p_0 =-\lceil \log 2\delta^{-1} \rceil,p_1 = p_0 +\frac12, p_2 = p_1+1,\dots,p_A = 0$ with $A = 2 \lceil \log 2\delta^{-1} \rceil$.
    By defining $q_0 = e^{-p_0}$ and 
    \begin{align}
        q_i(x) &= \exp (\min\{\max\{x,p_{i-1}\}p_{i}\}) - \exp(p_{i-1})
        \\ &=\ReLU(\exp (-\ReLU(-\ReLU(x-p_{i-1})-p_{i-1}+p_{i})+p_{i})-\log p_{i-1} ),
    \end{align}
    we have
    \begin{align}
    \textstyle
        \exp(x) = \sum_{i=0}^A q_i(x), p_0 \leq x \leq 0,
    \end{align}
    $(\mathrm{RHS}) = e^{p_0} \leq \frac{\delta}{2} \ (x\leq p_0)$ and $(\mathrm{RHS}) = 0 \ (0\leq x)$.

    For $1\leq i \leq A$, we consider Taylor expansion of $\exp(x)-\exp(p_{i-1})$ as
    \begin{align}
    \textstyle
        \exp(x)-\exp(p_{i-1}) = \exp(p_{i-1})\Big[\sum_{k=1}^K \frac{(x-p_{i-1})^k}{k!} + \frac{(y(K,x)-p_{i-1})^{K+1}}{(K+1)!}\Big],
    \end{align}
    where $p_{i-1}\leq y(K,x)\leq p_{i}$. Thus, the approximation error by the first $K$ terms is bounded by $2^{-(K+1)}$. 

    Let $r_{i,k}(x) =\frac{\exp(p_{i-1})}{k!}(x-p_{i-1})^k \ (1\leq k \leq K)$. 
    Then, if $r_{i,k}(x)$ is approximated by a function $\tilde{r}_{i,k}$ in $p_{i-1}\leq x\leq p_i$ within the error of $\delta'$, 
    \begin{align}
    \textstyle
        \ReLU\Big(q_0 + \sum_{i=1}^{6A} \sum_{k=1}^{K} \tilde{r}_{i,k}(-\ReLU(-\ReLU(x-p_{i-1})-p_{i-1}+p_{i})+p_{i})-\log p_{i-1} )\Big)
        \label{eq:Appendix-BasicFunctions-exp-1}
    \end{align}
    approximates $\exp (x)\ (x \leq 0)$ within the error of
\begin{align}
   \frac\delta2+ 2AK\delta' + 2A 2^{-(K+1)}.
\end{align}
Also, $\eqref{eq:Appendix-BasicFunctions-exp-1} \leq 1+\delta$ for all $x$.

    The rest of the argument follows that of \Cref{lem:ReLU_logarithm}.
    Specifically, we take $K=\lceil\log_2 (2A/\delta)\rceil$ and $\delta' =\frac{\delta}{4AK}=\frac{\delta}{8\lceil\log_2 (2A/\delta)\rceil\lceil \log 2\delta^{-1} \rceil}$ in the part (2) of \Cref{lem:ReLU_logarithm}, and all the others are identical.
\end{proof}

\begin{lemma}[ReLU approximation of indicator function]\label{lem:ReLU_indicator}
Let $a\in \R$, and $\delta>0$. 
A one-layer neural network $\NN_{\indicator[a]}$ defined by
\[
\textstyle \NN_{\indicator[a]}(x) = \frac{1}{\delta}\ReLU(x - (a-\delta)) + \frac{1}{\delta} \ReLU(x -(a+\delta)) - \frac{2}{\delta} \ReLU(x-\delta),
\]
satisfies
\[
\NN_{\indicator[a]}(x) = \indicator[x=a],~~~~~ \text{for all $x$ such that $x \leq a-\delta, x = a,$ or $x \geq a + \delta$.}
\]
\end{lemma}

\begin{proof}[Proof of Lemma~\ref{lem:ReLU_indicator}]
The lemma holds by direct calculation. 
\end{proof}

\subsection{Self-attention layer (proof of \Cref{lemma:Self-attention-CLIP})}
\label{subsection:Selt-attention}
We use the following lemma to prove \Cref{lemma:Self-attention-CLIP}.
\begin{lemma}\label{lemma:QK-matrix}
    Fix $\ell\in [L]$. 
    There exist matrices $\bWauxiliary^{(\ell)}_{K},\ \bWauxiliary^{(\ell)}_{Q} \in \R^{d_\positional \times d_\positional}$ 
    with 
    $\max_{i,j}|(\bWauxiliary^{(\ell)}_{Q})_{i,j}|,  \max_{i,j}|(\bWauxiliary^{(\ell)}_{K})_{i,j}|\leq \log \frac{d}{\delta(1-\cos(2\pi d^{-1})}$ such that
    \begin{align}
    \textstyle|(\softmax((\bWauxiliary^{(\ell)}_{K}\bP)^\top
        (\bWauxiliary^{(\ell)}_{ Q}\bP)) - \frac{1}{m^{(\ell)}}\bI^{(\ell)} )_{u,v}| \leq \delta,\quad u,v\in \cV^{(L)},
        \label{eq:Softmax-Identity}
    \end{align}
    where $\bI^{(\ell)}\in \R^{d\times d}$ is a matrix such that $\bI^{(\ell)}_{u,v}=1$ if 
    $\iota(\pa^{(\ell')}(u))=\iota(\pa^{(\ell')}(v))\ (\ell'\ne L-\ell)$, and $0$ otherwise.
\end{lemma}
    By using this lemma, \Cref{lemma:Self-attention-CLIP} is shown as follows.
    Use $\bWauxiliary^{(\ell)}_{K}$  and $\bWauxiliary^{(\ell)}_{Q}$ from \Cref{lemma:QK-matrix} to construct
    \begin{align}
        \bW^{(\ell)}_K 
        = 
        \begin{bmatrix}
        \bzero &   
            \bWauxiliary^{(\ell)}_{K} 
        \end{bmatrix}, 
        \quad 
        \bW^{(\ell)}_Q 
        = 
        \begin{bmatrix}
         \bzero &   \bWauxiliary^{(\ell)}_{Q} 
        \end{bmatrix}, 
    \end{align}
    where $\bzero\in \R^{(d_\feature+d_\positional)\times d_\feature}$. 
    
    Then, let $\bW^{(\ell)}_{V}$ be
    \begin{align}
        (\bW^{(\ell)}_{V})_{i,j} = 
        \begin{cases}
            m^{(\ell)} & i=j,\ (2\ell-1) S+2\leq i\leq 2\ell S+1
            \\
            0 & \text{otherwise},
        \end{cases}
    \end{align}
    so that $\bW^{(\ell)}_{V}$ extracts $\qnetwork_v^{(\ell)}$. 

    Then, the $v$-th column of $(\bW_V^{(\ell)}\qmatrix^{(\ell)})\softmax\big((\bW_K^{(\ell)}\qmatrix^{(\ell)})^\top(\bW_Q^{(\ell)}\qmatrix^{(\ell)})\big)$ implements the average of $\qnetwork_u^{(\ell)}$ over $u$ satisfying that $\bI^{(\ell)}_{u,v}=1$ defined in \Cref{lemma:QK-matrix} multiplied by $m^{(\ell)}$, within the error of $m^{(\ell)}\delta$. 
    The number of such $u$ is exactly $m^{(\ell)}$, thus the average multiplied by $m^{(\ell)}$ is the summation. 
    Now the error is $m^{(\ell)}\delta$, so letting $\delta \leftarrow (m^{(\ell)})^{-1} \delta$ yields the assertion.

\begin{proof}[Proof of \Cref{lemma:QK-matrix}]
    Define the key and the query matrix $\bWauxiliary^{(\ell)}_{K}=\bI_{d_\positional}\in \R^{d_\positional \times d_\positional}$ and $\bWauxiliary^{(\ell)}_{Q}\in  \R^{d_\positional \times d_\positional}$ as
    \begin{align}
       (\bWauxiliary^{(L-1)}_{Q})_{i,j} = 
        \begin{cases}
            \alpha & \text{if $i=j$ and $i\ne 2(L-\ell)+1, 2(L-\ell)+2$},
            \\
            0 & \text{otherwise},
        \end{cases}
    \end{align}
    for some $\alpha>0$ which will be defined later.
    (From now, we will focus on the case when $L\geq 2$. When $L=1$, it is obvious to see that the assertion still holds because $((\bWauxiliary^{(L)}_{K}\bP)^\top
        (\bWauxiliary^{(L)}_{ Q}\bP))_{u,v}=0$ for all $u,v$.)
    
    Then, we have
    \begin{align}
      & ((\bWauxiliary^{(\ell)}_{K}\bP)^\top
        (\bWauxiliary^{(\ell)}_{ Q}\bP))_{u,v}
    \\   & \textstyle=\alpha 
       \sum_{\ell'\ne \ell} \Big[\sin\big(\frac{2\pi \iota(\pa^{(L-\ell')}(u))}{m^{(\ell')}}\big)\sin\big(\frac{2\pi \iota(\pa^{(L-\ell')}(v))}{m^{(\ell')}}\big)+\cos\big(\frac{2\pi \iota(\pa^{(L-\ell')}(u))}{m^{(\ell')}}\big)\cos\big(\frac{2\pi \iota(\pa^{(L-\ell')}(v))}{m^{(\ell')}}\big)\Big]
       \\ & \textstyle =\alpha 
       \sum_{\ell'\ne \ell} \cos\big(\frac{2\pi \iota(\pa^{(L-\ell')}(v))-\iota(\pa^{(L-\ell')}(u))}{m^{(\ell')}}\big)
\\ & \textstyle =
       \begin{cases}
           (L-1)\alpha \quad\quad (\text{if 
           $\iota(\pa^{(L-\ell')}(u))=\iota(\pa^{(L-\ell')}(v))\ (\ell'\ne \ell)$})
           \\
           (L-1)\alpha - \alpha\min_{\ell'\ne \ell}\big( 1- \cos\big(\frac{2\pi}{m^{(\ell')}}\big)\big)
           \quad\quad  (\text{otherwise}).
       \end{cases}
    \end{align}
    
    Let us recall the property of $\softmax$.
    For $a\in \R^d$ with $a_1=\cdots=a_m>a_{m+1}\geq \cdots \geq a_d$ with $a_m-a_{m+1} = A>0$, it holds that $\softmax(a)_1\geq \frac{1}{m^{(\ell)}}\cdot \frac{1}{1+de^{-A}}\geq 1-de^{-A}$ and $\softmax(a)_i\leq e^{-A}\ (i=2,\dots,d)$. 
    Therefore, for $\delta<1$, by taking $\alpha = \log \frac{d}{\delta (1-\cos(2\pi d^{-1}))}$, we have
    \begin{align}
    \textstyle|(\softmax((\qnetwork^{(\ell)}\bWauxiliary^{(\ell)}_{K})^\top
        (\qnetwork^{(\ell)}\bWauxiliary^{(\ell)}_{Q})) - \frac{1}{m^{(\ell)}}\bI^{(\ell)})_{u,v} | \leq \delta.
        \label{eq:Softmax-Identity_b}
    \end{align}
    %
\end{proof}

\subsection{Evaluation of error propagation}\label{subsection:ErrorPropagation}
    To control the approximation error on the optimal similarity score function, we need to convert an approximation error of each component $f_{\image, \iota}^{(\ell)}$, $f_{\txt, \iota}^{(\ell)}$ by evaluating how component-wise approximation error propagates in the pipeline.
    The proof of this lemma requires Lipschitzness of the basic operations (\Cref{lemma:Appendix-CLIP-Lipschitz-normalization,lemma:Appendix-CLIP-Lipschitz-logsumexp,lemma:Appendix-CLIP-Lipschitz-logsumsoftmax} in \Cref{subsection:Lip-Basic-Functions}), to ensure that the propagated errors do not explode.
    We use $\NNs^{\linkweight}=f^{(0)}$ so there is no error when considering the link function.

\begin{lemma}[Evaluation of error propagation]\label{lemma:Appendix-CLIP-errorpropagation}
    Assume we have functions 
     $\fnetwork_{\txt, \iota}^{(\ell)}\ (1\leq \ell \leq L,\iota\in [m_\txt^{(\ell)}])$ such that
    \begin{align}
    \begin{aligned}
        \|f_{\txt,\iota}^{(L)}(x)-\fnetwork_{\txt,\iota}^{(L)}(x)\|_\infty
        &\leq \delta,\quad \forall x\in [S],
        \\ 
        \|f_{\txt,\iota}^{(\ell)}(h)-\fnetwork_{\txt,\iota}^{(\ell)}(h)\|_\infty
        &\leq \delta,\quad \forall h\in \R^S\text{ such that $\max_{s\in S}h_s= 0$, }\ell\in [L-1]   ,    
    \end{aligned}
     \label{eq:Appendix-CLIP-ErrorPropagation-3}
    \end{align}
    and $f^{(\ell)}_{\image,\iota}$ in the same way.
    Also, assume that $\|\bdelta^{(\ell)}_{v}\|_\infty\leq \delta$ holds for all $\ell=L-1,\dots,0$ and $v\in \cV^{(\ell)}$. 
    Let us take $\NNs^\linkweight=f^{(0)}$. 

    Consider the update in \eqref{eq:thewholepipeline} and \eqref{eq:thewholepipeline-2}.
    Then, we have the following bound on the error propagation:
    \begin{align}
      \textstyle 
    \max_{v\in \cV_\txt^{(L)}}\|\hnetwork_{\txt,v}^{(\ell)}-h_{\txt,\pa^{(L-\ell)}(v)}^{(\ell)}\|_\infty
    &  \textstyle \leq \delta \times (2m^{(\ell+1)}_\txt+2)\prod_{\ell+2\leq k\leq L}(2m^{(k)}_\txt+3),&& \ell = L-1,\dots,0,
     \label{eq:Appendix-CLIP-ErrorPropagation-1}
    \\
     \textstyle 
      \max_{v\in \cV_\txt^{(L)}}\|\qnetwork_{\txt,v}^{(\ell)}-q_{\txt,\pa^{(L-\ell)}(v)}^{(\ell)}\|_\infty
    &  \textstyle \leq \delta \times \prod_{\ell+1\leq k\leq L}(2m^{(k)}_\txt+3),&&
    \ell = L,\dots,1,
    \label{eq:Appendix-CLIP-ErrorPropagation-12}
    \end{align}
    and the bounds on the image part follows in the same way.
    Furthermore, we have
    \begin{align}
      \textstyle  |\scorefun_{\NN}-\scorefun_{\MP}| &  \textstyle \leq 
      \delta 
      \times 
     \big[\prod_{1\leq \ell\leq L}(2m_{\image}^{(\ell)}+3) + \prod_{1\leq \ell\leq L}(2m_{\txt}^{(\ell)}+3)\big].
     \label{eq:Appendix-CLIP-ErrorPropagation-9}
    \end{align}
\end{lemma}
\begin{proof}
    First, we prove \eqref{eq:Appendix-CLIP-ErrorPropagation-1} and \eqref{eq:Appendix-CLIP-ErrorPropagation-12}.
    We focus on the language model and the bounds on the vision model follows in the same way.

    We use the induction. 
    Let us check \eqref{eq:Appendix-CLIP-ErrorPropagation-1} for $\ell=L-1$ and \eqref{eq:Appendix-CLIP-ErrorPropagation-12} for $\ell=L$. Because $\hnetwork_{\txt,v}^{(L)}=h_{\txt,v}^{(L)}$, \eqref{eq:Appendix-CLIP-ErrorPropagation-3} implies that
    \begin{align}
         \| q^{(L)}_{\txt,v}-\qnetwork^{(L)}_{\txt,v}\|_\infty \leq\delta.
    \end{align}
    By \Cref{lemma:Appendix-CLIP-Lipschitz-normalization} and $\|\bdelta_{\txt,v}^{(L-1)}\|_\infty \leq \delta$, we have
    \begin{align}
        \| h^{(L-1)}_{\txt,\pa(v)}-\hnetwork^{(L-1)}_{\txt,v}\|_\infty \leq 2m_\txt^{(L)} \max_{u\in \cV_\txt^{(L)}}\| f^{(L)}_{\txt,\iota(u)}(x_{\txt,u})-\fnetwork^{(L)}_{\txt,\iota(u)}(x_{\txt,u})\|_\infty + 2\delta \leq 2(m^{(L)}_\txt+1)\delta,
    \end{align}
    for all $v\in \cV_\txt^{(L)}$, which confirms \eqref{eq:Appendix-CLIP-ErrorPropagation-1} for $\ell=L-1$ and \eqref{eq:Appendix-CLIP-ErrorPropagation-12} for $\ell=L$.

    Assume \eqref{eq:Appendix-CLIP-ErrorPropagation-1} for $\ell=L,\dots,\ell$ and \eqref{eq:Appendix-CLIP-ErrorPropagation-12} for $\ell=L,\dots,\ell+1$ and prove \eqref{eq:Appendix-CLIP-ErrorPropagation-1} for $\ell$ and \eqref{eq:Appendix-CLIP-ErrorPropagation-12} for $\ell-1$.
    \begin{align}
       & \| q^{(\ell)}_{\txt,\pa^{(L-\ell)}(v)}-\qnetwork^{(\ell)}_{\txt,v}\|_\infty
       \\ & =\max_{u\in \cV_\txt^{(L)}}\| f^{(\ell)}_{\txt,\pa^{(L-\ell)}(u)}(h_{\txt,\pa^{(L-\ell)}(u)}^{(\ell)})-\fnetwork^{(\ell)}_{\txt,\pa^{(L-\ell)}(u)}(\hnetwork_{\txt,u}^{(\ell)})\|_\infty 
        \\ &\leq 
        \max_{u\in \cV_\txt^{(L)}}\| f^{(\ell)}_{\txt,\pa^{(L-\ell)}(u)}(\hnetwork_{\txt,u}^{(\ell)})-\fnetwork^{(\ell)}_{\txt,\pa^{(L-\ell)}(u)}(\hnetwork_{\txt,u}^{(\ell)})\|_\infty \\ 
        &\quad\quad 
        +
        \| f^{(\ell)}_{\txt,\pa^{(L-\ell)}(u)}(h_{\txt,\pa^{(L-\ell)}(u)}^{(\ell)})-f^{(\ell)}_{\txt,\pa^{(L-\ell)}(u)}(\hnetwork_{\txt,u}^{(\ell)})\|_\infty
\\ & \leq \delta
        +
        \max_{u\in \cV_\txt^{(L)}}\|h_{\txt,\pa^{(L-\ell)}(u)}^{(\ell)}-\hnetwork_{\txt,u}^{(\ell)}\|_\infty
 \\ &\textstyle\leq \delta + \delta \times (2m^{(\ell+1)}_\txt+2)\prod_{k=\ell+2}^L(2m^{(k)}_\txt+3)
 \\ & \textstyle \leq \delta \times \prod_{k=\ell+1}^L(2m_\txt^{(k)}+3),
 \label{eq:Appendix-CLIP-ErrorPropagation-13}
    \end{align}
where we used \Cref{lemma:Appendix-CLIP-Lipschitz-logsumexp} for the second inequality.
Also, 
    \begin{align}
        \| h^{(\ell-1)}_{\txt,\pa^{(L-\ell+1)}(v)}-\hnetwork^{(\ell-1)}_{\txt,v}\|_\infty
        & \leq 2m_\txt^{(\ell)} \max_{u\in \cV^{(L)}}\| q^{(\ell)}_{\txt,\pa^{(L-\ell)}(u)}-\qnetwork^{(\ell)}_{\txt,u}\|_\infty +2\delta
        \\ &\textstyle\leq \delta \times 2m_\txt^{(\ell)}\prod_{k=\ell+1}^L(2m_\txt^{(k)}+3)\big)+2\delta
\\ & \textstyle\leq  \delta \times (2m_\txt^{(\ell)}+2)\prod_{k=\ell+1}^L(2m_\txt^{(k)}+3),
    \end{align}
    where we used \Cref{lemma:Appendix-CLIP-Lipschitz-normalization} and $\|\bdelta_{\txt,v}^{(\ell-1)}\|_\infty \leq \delta$ for the first inequality, and \eqref{eq:Appendix-CLIP-ErrorPropagation-13} for the second inequality.
    Therefore, by induction, we obtained \eqref{eq:Appendix-CLIP-ErrorPropagation-1} for all $\ell=L-1,\dots,0$ and \eqref{eq:Appendix-CLIP-ErrorPropagation-12}  for all $\ell=L,\dots,1$. 

    
   Finally, we bound $|\scorefun_{\NN}-\scorefun_{\MP}|$ to prove \eqref{eq:Appendix-CLIP-ErrorPropagation-9}.
   By using \Cref{lemma:Appendix-CLIP-Lipschitz-logsumsoftmax}, and the bound $\| h^{(0)}_{\txt,\root}-\hnetwork^{(0)}_{\txt,\root}\|_\infty\leq \prod_{1\leq \ell\leq L}(2m_{\txt}^{(\ell)}+3)$ and $\| h^{(0)}_{\image,\root}-\hnetwork^{(0)}_{\image,\root}\|_\infty\leq \prod_{1\leq \ell\leq L}(2m_{\image}^{(\ell)}+3)$, we have that 
    \begin{align}
     |\scorefun_{\NN}-\scorefun_{\MP}|
     & \textstyle  = \big|f^{(0)}(\softmax(h^{(0)}_{\txt,\root}),\softmax(h^{(0)}_{\image,\root})) - f^{(0)}(\softmax(\hnetwork^{(0)}_{\txt,\root^{(L)}}),\softmax(\hnetwork^{(0)}_{\image,\root^{(L)}}))\big| 
      \\ & \leq \|h^{(0)}_{\txt,\root}-\hnetwork^{(0)}_{\txt,\root^{(L)}}\|_\infty+ \|h^{(0)}_{\image,\root}-\hnetwork^{(0)}_{\image,\root^{(L)}}\|_\infty 
       \\ &\textstyle \leq \delta \times 
     \big[\prod_{1\leq \ell\leq L}(2m_{\image}^{(\ell)}+3) + \prod_{1\leq \ell\leq L}(2m_{\txt}^{(\ell)}+3)\big]
       \label{eq:Appendix-CLIP-ErrorPropagation-6}.
    \end{align}
 \end{proof}

\subsection{Properties of the message passing algorithm}\label{subsection:Appendix-CLIP-Auxiliary}
As auxiliary lemmas, we state  boundedness of $h^{(\ell)}_v,$ $ q^{(\ell)}_v,$ $ \P[s|\bx],$ and $\truesimscore$. 
\Cref{lemma:Boundedness-1} omits the subscripts ``$\txt$'' and ``$\image$'' in this subsection because both text and image parts have similar bounds.
\begin{lemma}\label{lemma:Boundedness-1}
Consider the message passing algorithm in \eqref{eq:Appendix-CLIP-ErrorPropagation-15} and \eqref{eq:Appendix-CLIP-ErrorPropagation-14}. Under \Cref{ass:bounded_transition}, we have that
\begin{align}
\begin{aligned}
  & 
  \|f_{ \iota}^{(\ell)}(h)\|_{\infty}
 \leq \log S\transbd \quad (2\leq \ell \leq L), &&
 \|f_{ \iota}^{(1)}(h)\|_{\infty}
 \leq \log S\transbd^{1+\frac{1}{m^{(1)}}},
\end{aligned}
\end{align}
for $h\in \R^S$ with $\max_s h_s = 0$, and that
\begin{align}
\begin{aligned}
 &\|h_{v}^{(\ell)}\|_{\infty}\leq 2m^{(\ell+1)}\log S\transbd\quad (1\leq \ell \leq L-1),&&\|h_{v}^{(0)}\|_{\infty}\leq 2m^{(1)}\log S\transbd^{1+\frac{1}{m^{(1)}}},
 \end{aligned}
\end{align}  
for the variables $h_{v}^{(\ell)}$ of the message passing algorithm.
Furthermore, the conditional probability $\P[s|\bx]$ is bounded as
\begin{align}
    \frac{1}{\transbd^{2\mmaxone}\state}\leq \P[s|\bx] \leq 1.
\end{align}
\end{lemma}
\begin{proof}
    Because of the update \eqref{eq:Appendix-CLIP-ErrorPropagation-15}, one dimension of $h^{(\ell)}_v$ is zero, and the others are zero or negative.
    Therefore, for $2\leq \ell \leq L$, we have that
    \begin{align}
   \textstyle ( q_{v}^{(\ell)})_s
       =(f_{ \iota}^{(\ell)}(h))_s = \log \sum_{a\in [S]}\psi_{\iota}^{(\ell)}(s,a)e^{h_a}
       \leq \log S\transbd,
    \end{align}
    and $ ( q_{v}^{(\ell)})_s\geq - \log S\transbd$ holds in the same way.
    Also, for $\ell=1$, we have
    \begin{align}
   \textstyle ( q_{\root}^{(1)})_s
       =(f_{ \iota}^{(1)}(h))_s = \log \sum_{a\in [S]}\P[s]^{\frac{1}{m^{(1)}}}\psi_{\iota}^{(1)}(s,a)e^{h_a}
       \leq \log S\transbd^{1+\frac{1}{m^{(1)}}},
    \end{align}
    and $ ( q_{\root}^{(0)})_s\geq - \log S\transbd^{1+\frac{1}{m^{(1)}}}$ holds in the same way.

    Also, by using the above bounds on $q_{u}^{(\ell+1)}=f_{\iota(v)}^{(\ell+1)}(q_{u})$, we have
    \begin{align}
       \|h^{(\ell)}_{v}\|_\infty \textstyle = 2\|\sum_{u\in \cC(v)}q_{u}^{(\ell+1)}\|_\infty\leq 2|\cC(v)|\max_{u\in \cC(v)} \|q_{u}^{(\ell+1)}\|_\infty\leq 
       \begin{cases}
           2m^{(\ell+1)}\log S\transbd  & (\ell\geq 1),
           \\
           2m^{(1)}\log S\transbd^{1+\frac{1}{m^{(1)}}}& (\ell=0).
       \end{cases}
    \end{align}
    Here applying $\normalize$ only changes the bound by a factor of at most two.

Finally, we consider the lower bound on $\P[s|\bx]$.
   By Assumption~\ref{ass:bounded_transition},
   we see that $\min_{s\in[\state]}\P[s]\geq 1/(\state\transbd)$ and $\P[\bx|s]/\P[\bx|s']\in[\transbd^{-2\mmaxone},\transbd^{2\mmaxone}]$ for any $s,s'\in[\state]$. As a consequence,
   \begin{align}
   \P[s|\bx]=\frac{\P[\bx|s]\P[s]}{\sum_{s'\in[\state]}\P[\bx|s']\P[s']}
   \geq \P[s]\cdot \min_{s'\in[\state]}\frac{\P[\bx|s]}{\P[\bx|s']}\geq \frac{1}{\transbd^{2\mmaxone}\state}.\label{eq:poster_prob_bd1}
   \end{align}
\end{proof}
\begin{lemma}\label{lemma:BoundScore}
    Under \Cref{ass:bounded_transition}, the optimal similarity score function (adjusted up to constant shift) $\truesimscore(\xi,\xt)=\log \frac{\mathbb{P}[\xi,\xt]}{\mathbb{P}[\xi]\mathbb{P}[\xt]}$ is upper and lower bounded as 
    \begin{align}
    -2\mmaxone\log \transbd\leq \truesimscore(\xi,\xt)\leq 2\mmaxone\log \transbd.        
    \end{align}
\end{lemma}
\begin{proof}
Note that 
\begin{align}
\exp(\truesimscore(\xi,\xt))
&=\frac{\P[\xi,\xt]}{\P[\xi]\P[\xt]}
=
\frac{\P[\xi|\xt]}{\P[\xi]}
\overset{(i)}{=}
\frac{\sum_s\P[\xi|s]\P[s|\xt]}{\sum_s\P[\xi|s]\P[s]},
\end{align}
where step~(i) uses the conditional independence of $\xi,\xt$ given $\root=s$. Since 
\begin{align}
 \frac{\P[s|\xt]}{\P[s]} = 
 \frac{\P[\xt|s]}{\sum_{s'\in[\state]}\P[\xt|s']\P[s']}
 \in\Big[\min_{s,s'\in[\state]}\frac{\P[\xt|s]}{\P[\xt|s']},\max_{s,s'\in[\state]}\frac{\P[\xt|s]}{\P[\xt|s']}\Big]
\end{align}
by Bayes' formula, it follows from Assumption~\ref{ass:bounded_transition} that ${\P[\xt|s]}/{\P[\xt|s']}\in[\transbd^{-2\mmaxone},\transbd^{2\mmaxone}]$. Putting pieces together yields the desired result.
\end{proof}

\section{Proof of~\Cref{theorem:CDM-Generalization-two-step}}\label{sec:pf_theorem:CDM-Generalization-two-step}
\subsection{Overview}
To predict $\xi$ given $\xt$ and $\bz_t$, 
the message passing algorithm is the algorithm for computing the Bayes optimal denoiser. 
We describe the algorithm in \Cref{subsubsection:CDM-Intro-MP}, and discuss how to
 implement the message passing algorithm using transformers in \Cref{subsubsection:CDM-Intro-NN}.
 This section mainly focuses on the image part and sometimes we omit the subscript ``$\image$'' from, e.g., $m^{(\ell)}_\image$ and $d_\image$. 
\subsubsection{Belief propagation and message passing algorithms}
\label{subsubsection:CDM-Intro-MP}
The message passing algorithm for the conditional denoising problem consists of the text part and the image part.
The text part is the same as the procedure \eqref{eq:Appendix-CLIP-ErrorPropagation-15} and \eqref{eq:Appendix-CLIP-ErrorPropagation-14} in contrastive learning, which computes $h_{\txt,\root}^{(0)}=(\log \P[s|\xt])_{s\in [S]}\in \R^S$ from $h^{(L)}_{\txt,v}=x_{\txt,v}\ (v\in \cV_\txt^{(L)})$. 
The image part is divided into two processes: downsampling and upsampling.
The image part first conduct the downsampling process to compute $h_{\root}^{(0)}$ from $\bz$. 
Then, combining this $h_{\root}^{(0)}$ with the output of the text part $h_{\txt,\root}^{(0)}$, the upsampling process of the image computes $b_{v}^{(L)}$ for each node $v\in \cV^{(L)}_\image$, so that $\softmax(b_{v}^{(L)})$ is exactly equal to $(\P[x_{\image,v}=s|\bz_t,\xt])_{s\in [S]}$.
Intuitively, the downsampling process aggregates the information from the leaves to the root, while the upsampling constructs estimation of leaf nodes from the root to the leaves. 
Outputting the weighted average of $s$ with respect to $\softmax(b_{\uparrow,v}^{(L)})$ yields the Bayes optimal denoiser $(\truebbmcdm(\bz_t, \xt))_v$ of $\xi$.
We formally define the procedures for the image part in the following. 

\paragraph{Downsampling.}
The downsampling process of the image part aggregates information from the leaves to the root of the tree.
It starts with $h^{(L)}_{v}=\normalize((-t(s-z_{t,v}/t)^2/2)_{s\in [S]})\in \R^S\ (v\in \cV_\image^{(L)})$, and computes $(q^{(\ell)}_{v})_{v\in \cV_\image^{(\ell)}}$ and $(h^{(\ell)}_{v})_{v\in \cV_\image^{(\ell)}}$ in the decreasing order of $\ell$.
   \begin{align}
    \begin{aligned}
    \textstyle q_{v}^{(\ell)} &\textstyle= f_{\downarrow,\iota(v)}^{(\ell)}(h^{(\ell)}_{v}) \in \R^S,
    && v\in \cV^{(\ell)}_\image,\ \ell = L,\dots,1,
   \\ 
    \textstyle h^{(\ell-1)}_{v} &\textstyle = \normalize \big(\sum_{u\in \cC(v)}q_{u}^{(\ell)}\big) \in \R^S,
     &&v\in \cV^{(\ell-1)}_\image,\  \ell = L,\dots,1.
       \end{aligned}\label{eq:Appendix-CDM-ErrorPropagation-3}
    \end{align}
    Computation of $q_{v}^{(\ell)}$ and $h^{(\ell-1)}_{v}$ from $h^{(\ell)}_{v}$ is called the $\ell$-th step of the downsampling process (of the image part). 
    Here, $f_{\downarrow,\iota}^{(\ell)}$ are defined as
    \begin{align}
    \begin{aligned}
    & \textstyle   (f_{\downarrow,\iota}^{(\ell)}(h))_s = \log \sum_{a\in [S]}\psi_{\image, \iota}^{(\ell)}(s,a)e^{h_a}, 
        && h\in \R^S,\ s\in [S],\ \ell = L,\dots,1,
    \end{aligned}\label{eq:Appendix-CDM-ErrorPropagation-4}
    \end{align}
    This is also the same as \eqref{eq:Appendix-CLIP-ErrorPropagation-15} and \eqref{eq:Appendix-CLIP-ErrorPropagation-14} in contrastive learning, except that $\P[s]^{\frac{1}{m^{(1)}}}$ is not needed for $\ell=1$.
    
\paragraph{Upsampling.}
It starts with combining the information of the text part and image downsampling. 
Then, the algorithm computes the prediction of $\xi$ from the root to the leaves.
The update is written as
   \begin{align}
    \begin{aligned}
    \textstyle \bar{b}_{\root}^{(0)} &\textstyle= 
    h_{\root}^{(0)}
    +h_{\txt,\root}^{(0)}\in \R^S,
    \\
    \textstyle \bar{b}_{v}^{(\ell)} &\textstyle= f_{\uparrow,\iota(v)}^{(\ell)}(\normalize(\bar{b}^{(\ell-1)}_{\pa(v)}-q^{(\ell)}_{v}))
    +h^{(\ell)}_{v}\in \R^S,
    && v\in \cV^{(\ell)}_\image,\ \ell = 1,\dots,L,
     \\ \textstyle
     (\truebbmcdm&(\bz_t, \xt))_v
     \textstyle= \sum_{s\in [S]} s\cdot \softmax(\bar{b}_{v}^{(L)})_s,
     &&v\in \cV^{(L)}_\image,\ s\in [\staten]
       \end{aligned}\label{eq:Appendix-CDM-ErrorPropagation-5}
    \end{align}
    where 
    \begin{align}
    \begin{aligned}
  & \textstyle   (f_{\uparrow,\iota}^{(\ell)}(h))_s = \log \sum_{a\in [S]}\psi_{\image, \iota}^{(\ell)}(a,s)e^{h_a}, 
        && h\in \R^S,\ s\in [S],\ \ell = 1,\dots,L.
    \end{aligned}\label{eq:Appendix-CDM-ErrorPropagation-6}
    \end{align}

    The update in \eqref{eq:Appendix-CDM-ErrorPropagation-5} is equivalently written as (note that $\bar{b}_{\pa(v)}^{(\ell-1)}-q_{\image,v}^{(\ell)}=b_{v}^{(\ell)}$ holds)
    \begin{align}
    \begin{aligned}
     \!\!\!\!\!\!\!\!   \textstyle b_{v}^{(1)} &\textstyle= 
    \normalize(h_{\root}^{(0)}
    +h_{\txt,\root}^{(0)}-q^{(1)}_{\image,v})\in \R^S, &&v\in \cV^{(1)}_\image,
    \\
    \!\!\!\!\!\!\!\! \textstyle b_{v}^{(\ell+1)} &\textstyle= \normalize(f_{\image,\uparrow,\iota(\pa(v))}^{(\ell)}(b_{\pa(v)}^{(\ell)})
    +h^{(\ell)}_{\pa(v)}-q^{(\ell+1)}_{v})\in \R^S,
    && v\in \cV^{(\ell+1)}_\image,\ \ell = 1,\dots,L-1,
    \\
     \!\!\!\!\!\!\!\!\textstyle b_{v}^{(L+1)} &\textstyle= \normalize(f_{\uparrow,\iota(v)}^{(L)}(b_{v}^{(L)})
    +h^{(L)}_{v})\in \R^S,
    && v\in \cV^{(L)}_\image,
   \\ 
    \!\!\!\!\!\!\!\!
     \textstyle
     (\truebbmcdm &(\bz_t, \xt))_v
     \textstyle= \sum_{s\in [S]} s\cdot \softmax(b_{v}^{(L+1)})_s,
     &&v\in \cV^{(L)}_\image,\ s\in [\staten]
       \end{aligned}\label{eq:Appendix-CDM-ErrorPropagation-7}
    \end{align}
    Computation of $b_{v}^{(\ell)}$ is called the $\ell$-th step of the upsampling process (of the image part). 
    We will approximate \eqref{eq:Appendix-CDM-ErrorPropagation-7} instead of \eqref{eq:Appendix-CDM-ErrorPropagation-5}, because we want to avoid complication about $\normalize$.
    Specifically, while our transformer block consists of the feed forward, self-attention, and ``$\normalize$'', 
    applying subtraction ($\bar{b}^{(\ell-1)}_{\pa(v)}-q^{(\ell)}_{v}$), ``$\normalize$'', and nonlinear transformation $f_{\downarrow,\iota}^{(\ell)}$ 
    cannot be done in one block.
    
The correctness of the message passing algorithm is formally stated as follows.
Because of this, taking the weighted average of $s$ with respect to $\softmax(\bar{b}_{\image,v}^{(L)})$ yields the Bayes optimal prediction of $\xi$. 
    \begin{lemma}[MP is the optimal denoising algorithm]\label{lem:MP_denoise_optimal}
When applying the message passing algorithm introduced in \eqref{eq:Appendix-CDM-ErrorPropagation-5} and \eqref{eq:Appendix-CDM-ErrorPropagation-6}, it holds that $\softmax(b_{v}^{(L)})_s = \P[x_{\image,v}=s| \bz_t,\xt]$ for all $v\in \cV_\image^{(L)}$.
    \end{lemma}
    \begin{proof}
    Regarding the joint generative hierarchical model as a single tree, the message passing algorithm for this case is directly adopted from (MP-DNS) of \cite{mei2024u}. 
    \end{proof}

\subsubsection{Approximation with transformer networks}\label{subsubsection:CDM-Intro-NN}
We approximate the message passing algorithm with transformer networks.
We denote a transformer approximation of $h^{(0)}_{\txt,\root}=(\log\P[s|\xt])_{s\in [S]}$ by $\hnetwork_{\txt,d_\txt}^{(0)}\in \R^S$. 
This can be obtained by $\NN_\txt^{\bW_{\txt}}$ constructed in contrastive learning, or a transformation of $\what\Et(\xt)$ in the
two-stage training. Specifically, we let

\begin{align}
\hnetwork_{\txt,d_\txt}^{(0)}=\log(\readtxt(\what\Et(\xt))), ~~\text{ where }  \readtxt(z)\defn \proj_{[\exp(-\readbdcdmenc),\exp(\readbdcdmenc)]}(z)\label{eq:cdm_log_trun_text_def}
\end{align}
and $\readbdcdmenc\defn 4\mmaxone\log(\transbd)+\log\state$. 
We let \begin{align}\delta_{\txt}
\defn
\|\hnetwork_{\txt,r}^{(0)}
-
\hnetwork_{\txt,d}^{(0)}\|_\infty
\label{eq:def_delta_txt}
\end{align} denote the approximation error of $\hnetwork_{\txt,d_\txt}^{(0)}$. 
We will see how the final approximation error depends on $\delta_\txt$ in later sections.
We will use the numbering of nodes defined in \Cref{definition:Numbering}. 

Let $\hnetwork^{(L)}_v=h^{(L)}_{v}=\normalize((-t(x-z_{t,v}/t)^2/2)_{x\in [S]})\in \R^S$ for all $v\in \cV_\image^{(L)}$. 
After the positional encoding $\encode_{\cdm}$, we obtain the initial matrix $\hmatrix^{(L)}$ such that
\begin{align}
    \hmatrix^{(L)} = 
    \encode_{\cdm}(\bz,\what\Et(\xt)) =
    \begin{bmatrix}
            &&\boldzero&
            \\
            \hnetwork^{(L)}_1 & \hnetwork^{(L)}_2 & \cdots & \hnetwork^{(L)}_{d}
            \\
            \hnetwork_{\txt,d_\txt}^{(0)}&  \hnetwork_{\txt,d_\txt}^{(0)}&\cdots & \hnetwork_{\txt,d_\txt}^{(0)}
            \\
            &&\pomatrix&
    \end{bmatrix}\in \R^{(d_\feature+d_\positional)\times d},
\end{align}
where $\pomatrix\in \R^{d_\positional\times d}$ is a matrix that encodes the positions of the nodes, and the output of the text model $ \hnetwork_{\txt,d_\txt}^{(0)}$ is concatenated with every pixel.
Here the dimensions are defined as $d_\feature=(3L+3)S$ and $d_\positional=2L$.

The text network has $(2L+1)$ transformer blocks, and the structure of each block is the same as contrastive learning. 
The first $L$ blocks approximate downsampling, and each block is called the $\ell(=L,\dots,1)$-th block of downsampling using the decreasing order.
The latter $(L+1)$-blocks approximate upsampling, and each block is called the $\ell(=0,\dots,L)$-th block of upsampling using the increasing order.

First we consider downsampling. 
Starting from $\hmatrix^{(L)}$, we iteratively construct $\hmatrix^{(\ell)}\in \R^{(d_\feature+d_\positional)\times d}$  and $\qmatrix^{(\ell)}\in \R^{(d_\feature+d_\positional)\times d}$:
\begin{align}
  \hmatrix^{(\ell)} = 
        \begin{bmatrix}
        &&\boldzero&
        \\
            \hnetwork^{(\ell)}_1 & \hnetwork^{(\ell)}_2 & 
             \cdots
            &
            \hnetwork^{(\ell)}_d
            \\
            \vdots & \vdots & 
             \ddots
            &
            \vdots
            \\
            \qnetwork^{(L)}_1 & \qnetwork^{(L)}_2 & 
             \cdots
            &
            \qnetwork^{(L)}_d
            \\
            \hnetwork^{(L)}_1 & \hnetwork^{(L)}_2 & 
             \cdots
            &
            \hnetwork^{(L)}_d
            \\
            \hnetwork_{\txt,d_\txt}^{(0)}&  \hnetwork_{\txt,d_\txt}^{(0)}&\cdots & \hnetwork_{\txt,d_\txt}^{(0)}
            \\
            &&\pomatrix&
        \end{bmatrix},
        \quad
        \qmatrix^{(\ell)} = 
        \begin{bmatrix}
        &&\boldzero&
        \\
            \qnetwork^{(\ell)}_1 & \qnetwork^{(\ell)}_2 & 
             \cdots
            &
            \qnetwork^{(\ell)}_d
            \\
            \vdots & \vdots & 
             \ddots
            &
            \vdots
            \\
            \qnetwork^{(L)}_1 & \qnetwork^{(L)}_2 & 
             \cdots
            &
            \qnetwork^{(L)}_d
            \\
            \hnetwork^{(L)}_1 & \hnetwork^{(L)}_2 & 
             \cdots
            &
            \hnetwork^{(L)}_d
            \\
            \hnetwork_{\txt,d_\txt}^{(0)}&  \hnetwork_{\txt,d_\txt}^{(0)}&\cdots & \hnetwork_{\txt,d_\txt}^{(0)}
            \\
            &&\pomatrix&
        \end{bmatrix}.
\end{align}
Here $\hnetwork^{(\ell)}_v\ (\ell=L,L-1,\dots,0)$ and $\qnetwork^{(\ell)}_v\ (\ell=L,L-1,\dots,1)$ are $S$-dimensional real-valued vectors.
Except that their column dimension is different, $\hmatrix^{(\ell)}$ and $\qmatrix^{(\ell)}$ are the same as \Cref{subsubsection:CLIP-Overview-Transformer}. 

In the $\ell$-th block of downsampling, the feed forward layer $\FF^{(\ell)}_\downarrow$, a fully-connected ReLU network, receives $\hmatrix^{(\ell)}$ and outputs $\qmatrix^{(\ell)}$ by computing $\qnetwork^{(\ell)}_v$ from $\hnetwork^{(\ell)}_v$:
    \begin{align}
        \qmatrix^{(\ell)} = 
        \underbrace{\hmatrix^{(\ell)}}_{\text{skip connection}}
        + 
        \feedforward^{(\ell)}_\downarrow (\hmatrix^{(\ell)})
        =
        \hmatrix^{(\ell)}
        +
        \begin{bmatrix}
        \multicolumn{4}{l}{ \boldzero\ (\in \R^{((2\ell+L)S) \times d})} \\ 
       \qnetwork^{(\ell)}_1 & \qnetwork^{(\ell)}_2 &  \cdots & \qnetwork^{(\ell)}_d \\
        \multicolumn{4}{l}{ \boldzero \ (\in \R^{(d_\positional+(2L-2\ell+2)S) \times d})} 
        \end{bmatrix}
        ,
    \end{align}
    Then, the self-attention layer $\Attn^{(\ell)}$ uses $\qmatrix^{(\ell)}$ to construct $\hmatrix^{(\ell-1)}$ as
    \begin{align}
        \hmatrix^{(\ell-1)} = 
        \normalize\Big(\underbrace{\qmatrix^{(\ell)}}_{\text{skip connection}}
        +\Attn^{(\ell)}(\qmatrix^{(\ell)})
        \Big)
        =
        \normalize\Bigg(
        \qmatrix^{(\ell)}
        + 
        \begin{bmatrix}
            \boldzero & (\in \R^{((2\ell+L-1)S) \times d})\quad \ \ \ 
            \\
            {\star} &(\in \R^{S \times d})\quad \quad \quad \quad  \quad \quad
            \\
            \boldzero &(\in \R^{(d_\positional+(2L-2\ell+3)S) \times d})
        \end{bmatrix}
        \Bigg)
        .
    \end{align} 
    Here $\star$ means $[\hnetwork_1^{(\ell-1)}\ \hnetwork_2^{(\ell-1)}\ \cdots\  \hnetwork_d^{(\ell-1)}]$ before normalization. 

    We then consider upsampling. 
    After we obtain $\hmatrix^{(0)}$, we  iteratively compute $\bbmatrix^{(\ell)}\ (\ell=1,\dots,L+1)$:
    \begin{align}
           \bbmatrix^{(\ell)} = 
        \begin{bmatrix}
        &&\boldzero&\\
            \bnetwork^{(\ell)}_{1} & \bnetwork^{(\ell)}_{2} & 
             \cdots
            &
            \bnetwork^{(\ell)}_d
           \\
            \vdots & \vdots & 
             \ddots
            &
            \vdots
        \\
            \bnetwork^{(1)}_{1} & \bnetwork^{(1)}_{2} & 
             \cdots
            &
            \bnetwork^{(1)}_d
        \\
            \hnetwork^{(0)}_{1} & \hnetwork^{(0)}_{2} & 
             \cdots
            &
            \hnetwork^{(0)}_d
        \\
            \qnetwork^{(1)}_1 & \qnetwork^{(1)}_2 & 
             \cdots
            &
            \qnetwork^{(1)}_d
            \\
            \vdots & \vdots & 
             \ddots
            &
            \vdots
            \\
            \qnetwork^{(L)}_1 & \qnetwork^{(L)}_2 & 
             \cdots
            &
            \qnetwork^{(L)}_d
            \\
            \hnetwork^{(L)}_1 & \hnetwork^{(L)}_2 & 
             \cdots
            &
            \hnetwork^{(L)}_d
            \\
            \hnetwork_{\txt,d_\txt}^{(0)}&  \hnetwork_{\txt,d_\txt}^{(0)}&\cdots & \hnetwork_{\txt,d_\txt}^{(0)}
            \\
            &&\pomatrix&
        \end{bmatrix}.
    \end{align}
    Here, $\bnetwork^{(\ell)}_v\ ( v\in \cV_\image^{(L)}, \ell=1,\dots,L+1)$ are $S$-dimensional real-valued vectors.
    The $\ell$-th block of downsampling computes $\bbmatrix^{(\ell+1)}$ using a feed forward network $\feedforward^{(\ell)}_\uparrow$ with normalization: 
 \begin{align}
        \bbmatrix^{(\ell+1)} = 
         \normalize\Big(\underbrace{\bbmatrix^{(\ell)}}_{\text{skip connection}}
        + 
        \feedforward^{(\ell)}_\uparrow (\bbmatrix^{(\ell)})\Big)
        =
        \bbmatrix^{(\ell)}
        +
        \begin{bmatrix}
        \multicolumn{4}{l}{ \boldzero \ (\in \R^{((L-\ell)S) \times d})} \\ 
       \bnetwork^{(\ell+1)}_1 & \bnetwork^{(\ell+1)}_2 &  \cdots & \bnetwork^{(\ell+1)}_d \\
        \multicolumn{4}{l}{\boldzero \ (\in \R^{(d_\positional+(2L+\ell+2)S) \times d})} 
        \end{bmatrix}
        . 
        \label{eq:CDM-Bmatrix-1}
    \end{align}
   For $\ell=0$, replace $\bbmatrix^{(0)}$ by $\hmatrix^{(0)}$. 
    For upsampling, we do not need the self-attention layer.
    It is simply ignored by just setting $\bW_V=0$ (and remember that we still have the skip connection).

Finally, we will obtain $\bnetwork^{(L+1)}_v$, that approximates $b^{(L+1)}_v$.
In the readout layer $\read_{\cdm}$, we compute the prediction of $\xi$ based on $\bnetwork_v^{(L+1)}$.

In the following, our goal is to iteratively show that, for all $v\in \cV^{(L)}_\image$, 
\begin{align}
    &\hnetwork_v^{(\ell)}\approx h_{\pa^{(L-\ell)}(v)}^{(\ell)},
    \quad \qnetwork_v^{(\ell)}\approx q_{\pa^{(L-\ell)}(v)}^{(\ell)}, \quad 
    \bnetwork_v^{(\ell)}\approx b_{\pa^{(L-\ell)}(v)}^{(\ell)}
    \quad
    \text{for all $v\in \cV_\image^{(L)}$},
\end{align}
($\ell=L,\dots,0$ for $\hnetwork_v^{(\ell)}$, $\ell=L,\dots,1$ for $\qnetwork_v^{(\ell)}$, and $\ell=1,\dots,L$ for $\bnetwork_v^{(\ell)}$), and
\begin{align}
    \bnetwork_v^{(L+1)}\approx b_v^{(L+1)}.
\end{align}

We will now formally define each component of the pipeline.
\paragraph{Encoding $\encode_{\cdm}$.}
The positional encoding is the same as the one for contrastive learning  \eqref{eq:positional_encoding}.
The $v$th column of $\pomatrix$, $\pnetwork_v$, is written as
\begin{align}
& \!\!\!\! \pnetwork_v = 
  \\ & \!\!\!\!\textstyle\Big[\sin \big( \frac{2\pi\iota(v)}{m^{(L)}}\big)\  \cos \big(\frac{2\pi\iota(v)}{m^{(L)}}\big)\ \sin \big( \frac{2\pi\iota(\pa(v))}{m^{(L-1)}}\big)\  \cos \big( \frac{2\pi\iota(\pa(v))}{m^{(L-1)}}\big)\  \cdots  \  \sin \big(\frac{2\pi\iota(\pa^{(L-1)}(v))}{m^{(1)}}\big)\  \cos \big( \frac{2\pi\iota(\pa^{(L-1)}(v))}{m^{(1)}}\big)\Big]^\top\!\!.
  \label{eq:positional_encoding-2}
\end{align}

For two-stage training, where $\what\Et(x)$ approximates $\Et_{,\star}(x)=\P[s|\xt]$, 
we define $\hnetwork^{(0)}_{\txt,d_\txt}$ as $\hnetwork^{(0)}_{\txt,d_\txt}=(\log\readtxt( \what\Et(x))_s)_{s\in [S]}$.

\paragraph{Downsampling: position-wise feed forward block.}
%
Similarly to the contrastive learning, the feed forward layer at the $\ell$-th block yields
\begin{align}
   \qnetwork^{(\ell)}_{v} 
   =\hnetwork^{(\ell)}_{v} + \fnetwork_{\downarrow,\iota(\pa^{(L-\ell)}(v))}^{(\ell)}(\hnetwork^{(\ell)}_{v}), \quad v\in \cV^{(L)}_\image.
\end{align}
Thus, when $\hnetwork^{(\ell)}_{v}\approx h^{(\ell)}_v$ for $v\in \cV^{(\ell)}_\image$ and 
$ \fnetwork_{\iota}^{(\ell)} \approx f_{\iota}^{(\ell)}$, we have $\qnetwork^{(\ell)}_{v} \approx q^{(\ell)}_v$ for $v\in \cV^{(\ell)}_\image$. 

Following the notation in \Cref{definition:FullyConnected}, we state the following approximation error guarantee. 
\begin{lemma}[Approximation error of feed forward layer, downsampling]\label{lemma:Approximation-FF-CDM}
Fix $\ell\in [L]$ and $\delta>0$.
Assume that $\transbd^{-1} \leq \psi^{(\ell)}_{\iota}(s,a) \leq \transbd$ for all $s,a\in [\staten]$.
Then, there exists an $\NN\in \mathcal{F}(\depth,\widthvector,B)$ such that
\begin{align}
    \|\NN([h;\pnetwork_v]) - f^{(\ell)}_{\downarrow,\iota(\pa^{(L-\ell)}(v))}(h)\|_\infty \leq \delta, \quad v\in \cV^{(L)}_\image,
\end{align}
for all $h\in \R^S$ with $\max_s h_s = 0$.
The network parameters $\depth,\widthvector$ and $B$ are bounded as follows:
\begin{align}
    \depth \lesssim (\log\log(S\transbd/\delta))\log(S\transbd/\delta),\
    \|\widthvector\|_\infty  \lesssim m^{(\ell)}S(\log(S\transbd/\delta))^3+L,
    \
    B  \lesssim 2S(\transbd^2+\log(S\transbd/\delta)) + (m^{(\ell)})^2.
\end{align}
\end{lemma}
This is the same as \Cref{lemma:Approximation-FF}, except that $(f_{\downarrow,\iota}^{(\ell)}(h))_s = \log \sum_{a\in [S]}\psi_{\image, \iota}^{(\ell)}(s,a)e^{h_a}$ for $\ell=L$ and $1$.
It is easy to see that the proof of \Cref{lemma:Appendix-CLIP-Existence-log-sum-exp} covers these cases, and thus we do not repeat the proof.  

\paragraph{Downsampling: self-attention block.}
The self-attention layer $\AT^{(\ell)}$ of the $\ell$-th block $(\ell=L,L-1,\dots,1)$ yields
\begin{align}
\textstyle
   \hnetwork_v^{(\ell-1)} 
   = \normalize\Bigg(
   \underset{\iota(\pa^{(L-\ell')}(u))=\iota(\pa^{(L-\ell')}(v))\ (\ell'\ne \ell)}{\sum}
   \qnetwork^{(\ell)}_{u} +\bdelta^{(\ell-1)}_v\Bigg).
   \label{eq:CDM-self-attention-1}
\end{align}
Here $\normalize(x)_s=x_s - \max x_{s'}$.
Please refer to \Cref{subsubsection:CLIP-Overview-Transformer} for interpretation of the summation. 
We can see that, when $\qnetwork^{(\ell)}_v \approx q^{(\ell)}_{\pa^{(L-\ell)}(v)}$ and $\|\bdelta^{(\ell-1)}_v\|_\infty \ll 1$ for $v\in \cV^{(L)}_\image$, we have $\hnetwork^{(\ell-1)}_v \approx h^{(\ell-1)}_{\pa^{(L-\ell)}(v)}$ for $v\in \cV^{(L)}_\image$. 

Following the notation in \Cref{definition:Self-Attn}, we have the following approximation error guarantee.
\begin{lemma}[Approximation error of self-attention layer]\label{lemma:Self-attention-CDM}
For $\ell\in [L]$, there exists $\Attn\in \cA(D,B)$ with $D=d_\feature+d_\positional$ and $B\lesssim \log(d\delta^{-1}) + m^{(\ell)}$ such that
    \begin{align}
   \Attn(\qmatrix^{(\ell)})=
    \begin{bmatrix}
            \boldzero&  (\in \R^{(2\ell+L+1)S})\quad \ \ 
            \\
            \underset{\iota(\pa^{(L-\ell')}(u))=\iota(\pa^{(L-\ell')}(v))\ (\ell'\ne \ell)}{\sum}
            \qnetwork^{(\ell)}_{u} + \bdelta^{(\ell-1)}_v & (\in \R^{S})\ \quad \quad \quad \quad\quad 
            \\
            \boldzero &(\in \R^{d_\positional+(2L-2\ell+1)S})
        \end{bmatrix}_{v\in \cV^{(L)}}, 
    \end{align} 
    where $\bdelta_{v}^{(\ell-1)}\in \R^S$ satisfies $\|\bdelta_{v}^{(\ell-1)}\|_\infty \leq \delta   \max_{v'}\|\qnetwork^{(\ell)}_{v'}\|_\infty$.
\end{lemma}
Because the only difference from \Cref{lemma:Self-attention-CLIP} is the dimension of zeros, we do not repeat the proof. 
See \Cref{subsection:Selt-attention} for the proof of \Cref{lemma:Self-attention-CLIP}.

\paragraph{Upsampling: position-wise feed forward block.}
The upsampling is implemented with $(L+1)$-transformer blocks indexed in increasing order $\ell=0,\dots,L$, where the $\ell$-th block of upsampling consists of a feed forward layer $\FF_\uparrow^{(\ell)}$ with skip connection and normalization.
The $\ell$-th block of upsampling computes $\bnetwork^{(\ell+1)}_v$ from $\hnetwork^{(\ell)}_v$, $\qnetwork^{(\ell+1)}_v$, and $\bnetwork^{(\ell)}_v$.
\begin{align}
     \!\!\!\!\!\!\!\!   \textstyle \bnetwork_{v}^{(1)} &\textstyle= 
    \normalize(\hnetwork_{v}^{(0)}
    +\hnetwork_{\txt,d_\txt}^{(0)}-\qnetwork^{(1)}_{v})\in \R^S, &&v\in \cV^{(L)}_\image,
    \label{eq:Appendix-CDM-UP-Network-1}
    \\
    \!\!\!\!\!\!\!\! \textstyle \bnetwork_{v}^{(\ell+1)} &\textstyle= \normalize(\fnetwork_{\uparrow,\iota(\pa^{(L-\ell)}(v))}^{(\ell)}(\bnetwork_{v}^{(\ell)})
    +\hnetwork^{(\ell)}_{v}-\qnetwork^{(\ell+1)}_{v})\in \R^S,
    && v\in \cV^{(L)}_\image,\ \ell = 1,\dots,L-1,
    \label{eq:Appendix-CDM-UP-Network-2}
    \\
     \!\!\!\!\!\!\!\!\textstyle \bnetwork_{v}^{(L+1)} &\textstyle= \normalize(\fnetwork_{\uparrow,\iota(v)}^{(L)}(\bnetwork_{v}^{(L)})
    +\hnetwork^{(L)}_{v})\in \R^S,
    && v\in \cV^{(L)}_\image.
    \label{eq:Appendix-CDM-UP-Network-3}
    \end{align}
For each update, we can track the correspondence with the message passing algorithm.
Specifically, for \eqref{eq:Appendix-CDM-UP-Network-1}, when $\hnetwork_{v}^{(0)}\approx h_{\root}^{(0)}$, $\qnetwork^{(1)}_{v}\approx q^{(1)}_{\pa^{(L-1)}(v)}$, and $\hnetwork_{\txt,d_\txt}^{(0)}\approx h_{\txt, \root}^{(0)}$ for $v\in \cV_\image^{(L)}$, we have $\bnetwork_{v}^{(0)}\approx b^{(1)}_{\pa^{(L-1)}(v)}$ for $v\in \cV_\image^{(L)}$.
Similar discussion holds for \eqref{eq:Appendix-CDM-UP-Network-2} and \eqref{eq:Appendix-CDM-UP-Network-3} as well. 

We will show the following approximation error guarantee of $f^{(\ell)}_{\uparrow,\iota}$. 
Since it is easy to concatenate zeros to the first and last layer matrices and adjust the input and output dimensions, below we present the network $\NN$ as a function between relevant dimensions for simple presentation.
\begin{lemma}[Approximation error of feed forward layer, upsampling]\label{lemma:Approximation-FF-CDM-UP}
Fix $\ell\in \{0,\dots,L\}$ and $\delta>0$.
Assume that $\transbd^{-1} \leq \psi^{(\ell)}_{\iota}(s,a) \leq \transbd$ for all $s,a\in [\staten]$.
Then, there exist $\NN_1,\NN_2,\NN_3\in \mathcal{F}(\depth,\widthvector,B)$ such that
\begin{align}
&\NN_1([h;h';q]) = h+h'-q, &&
\\
&\|\NN_2([b;h;q;\pnetwork_v]) - (f^{(\ell)}_{\downarrow,\iota(\pa^{(L-\ell)}(v))}(b)+h-q)\|_\infty \leq \delta, && v\in \cV^{(L)}_\image,\ \ell=1,\dots,L-1
\\
&\|\NN_3([b;h;\pnetwork_v]) - (f^{(L)}_{\downarrow,\iota(v)}(b)+h)\|_\infty \leq \delta, && v\in \cV^{(L)}_\image,\ell=L,
\end{align}
for all $h,h',q,b\in \R^S$ with $\max_s b_s = 0$.
For all of these networks, the parameters $\depth,\widthvector$ and $B$ are bounded as follows:
\begin{align}
    \depth \lesssim (\log\log(S\transbd/\delta))\log(S\transbd/\delta),\
    \|\widthvector\|_\infty  \lesssim m^{(\ell)}S(\log(S\transbd/\delta))^3+L,
    \
    B  \lesssim 2S(\transbd^2+\log(S\transbd/\delta)) + (m^{(\ell)})^2.
\end{align}
\end{lemma}
The first network $\NN_1$ is just a linear mapping, represented as $[\bI_S\  \bI_S\ \bI_S]$. 
The proof for $\NN_2$ and $\NN_3$ is mostly the same as \Cref{lemma:Approximation-FF}.
The only difference is to add $h-q$ (or $h$) after computing $f^{(\ell)}_{\downarrow,\iota(\pa^{(L-\ell)}(v))}(b)$, which is easily done with one additional layer.
Therefore, we omit the proof of this lemma. 
See \Cref{subsection:FF} for the proof of \Cref{lemma:Approximation-FF}.

\paragraph{Normalization.}
In the attention network, since column vectors of $ \hmatrix^{(\ell)} $, $ \qmatrix^{(\ell)} $, and $ \bbmatrix^{(\ell)}$ are a collection of multiple $ \hnetwork^{(\ell)}_v $, $ \qnetwork^{(\ell)}_v $, and $\bnetwork^{(\ell)}_v $, 
we adopt a slightly different definition of ``$\normalize$'' for these column vectors, from the one for $S$-dimensional vectors. 
Let $\xnetwork =[\bnetwork^{(L+1)}\ \cdots \ \bnetwork^{(1)}\ \hnetwork^{(0)}\ \qnetwork^{(1)}\ \hnetwork^{(1)}\ \dots\ \qnetwork^{(L)}\ \hnetwork^{(L)}\ \hnetwork\ \pnetwork]\in \R^{d_\feature+d_\positional}$, where $\hnetwork, \hnetwork^{(\ell)} (\ell=L,\dots,0), \qnetwork^{(\ell)} (\ell=L,\dots,1), \bnetwork^{(\ell)}\ (\ell=1,\dots,L+1)$ are $S$-dimensional real-valued vectors and $\pnetwork\in \R^{d_\positional}$.
We define $\normalize$ as
\begin{align}
    \normalize(\xnetwork)
    = 
    \begin{bmatrix}
        \hnetwork^{(0)} \\ \qnetwork^{(1)}- \ones_S\max_{s\in S}\qnetwork^{(1)}_s \\ \hnetwork^{(1)} \\ \qnetwork^{(2)} - \ones_S\max_{s\in S} \qnetwork^{(2)}_s \\ \vdots \\ \qnetwork^{(L)} - \ones_S\max_{s\in S} \qnetwork^{(L)}\\ \hnetwork^{(L)}
        \\ 
        \hnetwork
        \\ \pnetwork
    \end{bmatrix}\in \R^{d_\feature+d_\positional},\quad \ones_S = \begin{bmatrix}
        1 \\ 1 \\ \vdots \\ 1
    \end{bmatrix}\in \R^S.\label{eq:clip_normalize_def-CDM}
\end{align}
For a matrix with its column dimension $d_\feature+d_\positional$, it is applied in a column-wise manner.

\paragraph{Readout layer $\read_\cdm$.}
In the readout layer, we output the prediction 
$\Mt$ 
of $\xi$ from $\bnetwork_v^{(L+1)}$ as
\begin{align}
    \textstyle 
     {\sf M}_{t,v}=\read_{\cdm}(\bbmatrix^{(L+1)})=\sum_{s\in [S]} s \cdot \softmax(\bnetwork_{v}^{(L+1)})_s,\ 
     v\in \cV^{(L)}_\image.\label{eq:readout_cdm_def}
\end{align}


\paragraph{The whole pipeline.}
Putting it all together, the neural network approximate the message passing algorithm (for the image part) in the following way.
The downsampling process is approximated as
 \begin{align}
    \begin{aligned}
    \textstyle \qnetwork_{v}^{(\ell)} &\textstyle= \fnetwork_{\uparrow,\iota(\pa^{(L-\ell)}(v))}^{(\ell)}(\hnetwork^{(\ell)}_{v}) \in \R^S,
    && v\in \cV^{(L)}_\image,\ \ell = L,\dots,1,
   \\ 
    \textstyle  \hnetwork_v^{(\ell-1)} 
  & =  \textstyle  \normalize\Bigg(
  \underset{\iota(\pa^{(L-\ell')}(u))=\iota(\pa^{(L-\ell')}(v))\ (\ell'\ne \ell)}{\sum}
  \qnetwork^{(\ell)}_{u} +\bdelta^{(\ell-1)}_v\Bigg) \in \R^S,
     &&v\in \cV^{(L)}_\image,\ \ell=L,\dots,1.
       \end{aligned}\label{eq:Appendix-CDM-ErrorPropagation-109}
    \end{align}
    Let $\hnetwork_{\txt,d_\txt}^{(0)}\approx h_{\txt,\root}^{(0)}$. 
    The upsampling process is approximated as
   \begin{align}
   \begin{aligned}
        \textstyle \bnetwork_{\image,v}^{(1)} &\textstyle= 
    \normalize(\hnetwork_{v}^{(0)}
    +\hnetwork_{\txt,d_\txt}^{(0)}-\qnetwork^{(1)}_{v})\in \R^S, &&v\in \cV^{(L)}_\image, 
    \\
    \textstyle \bnetwork_{v}^{(\ell+1)} &\textstyle=\normalize(\fnetwork_{\uparrow,\iota(\pa^{(L-\ell)}(v))}^{(\ell)}(\bnetwork_{v}^{(\ell)})
    +\hnetwork^{(\ell)}_{v}-\qnetwork^{(\ell+1)}_{v})\in \R^S,
    && v\in \cV^{(L)}_\image,\ \ell = 1,\dots,L-1
    \\
    \textstyle \bnetwork_{v}^{(L+1)} &\textstyle= \normalize(\fnetwork_{\uparrow,\iota(v)}^{(L)}(\bnetwork_{v}^{(L)})
    +\hnetwork^{(L)}_{v})\in \R^S,
    && v\in \cV^{(L)}_\image,
   \\ 
    \textstyle {\sf M}_{t,v} 
    =&\textstyle \sum_{s\in [S]} s \cdot \softmax(\bnetwork_{v}^{(L+1)}) \in \R^S,
     &&v\in \cV^{(L)}_\image.
       \end{aligned}\label{eq:Appendix-CDM-ErrorPropagation-10}
    \end{align}

For two step training, the hypothesis class to which a tuple $(\TF_{\cdm},\Adapter)$ belongs is defined as follows, formally restating (\ref{eqn:CDM_param_space}).
    For joint training, the parameter space $\Parspacecdmone$ is defined in \Cref{sec:joint_train_cdm}. 
    \begin{definition}[Eq.~(\ref{eqn:CDM_param_space}), restated]
    We say the collection of the parameters of  $(\TF_{\cdm},\Adapter)$ belongs to $\Parspacecdm$ if the following holds:
    The image transformer network $\TF_{\cdm}$ has $L$ blocks of feed forward (\Cref{definition:FullyConnected}), self-attention (\Cref{definition:Self-Attn}), and normalization.
    In each block, its feed forward $\FF$ and self-attention $\Attn$ satisfy 
    \begin{align}
        &\FF\in \cF (\depth,\widthvector=(D,*,\cdots,*,D),B), \text{ with } \|\widthvector\|_\infty \leq D',
        \quad \Attn\in \cA (D,B).
    \end{align}
    Furthermore, the adapter satisfies 
    \begin{align}
    \adaweighta\in \R^{S\times M},\ \adaweightb \in \R^{M\times S}, \
        \lops{\adaweighta}\leq B,\ \lops{\adaweightb} \leq B.
    \end{align}
    \end{definition}

The rest of this section is organized as follows. 
\Cref{subsection:Appendix-CDM-Main} proves \Cref{theorem:CDM-Generalization-two-step}, using \Cref{lemma:Approximation-FF-CDM,lemma:Self-attention-CDM,lemma:Approximation-FF-CDM-UP}, as well as the bound on the propagation of the intermediate errors \Cref{lemma:Appendix-CDM-errorpropagation}. 
\Cref{subsection:CDM-ErrorPropagation} proves the error propagation lemma (\Cref{lemma:Appendix-CDM-errorpropagation}).


\subsection{Proof of~\Cref{theorem:CDM-Generalization-two-step}}
\label{subsection:Appendix-CDM-Main}
Define 
\begin{align}\orisk^\star_{\cdm,t}\defn
\E_{(\xi, \xt, \bz_t)}
\Big[ \big\| \xi-\truebbmcdm(\bz_t, \xt)  \big\|_2^2 \Big],
\end{align}
where $\truebbmcdm(\bz_t,\xt) = \E_{(\xi,\xt,\bz_t)\sim\truemucdm}[\xi|\bz_t,\xt]$.
    Similar to the proof of~\Cref{theorem:CLIP-Generalization}, we have the following decomposition:
\begin{align}
&\quad
\E_{(\xi, \xt, \bz_t)}
\Big[ \big\| \truebbmcdm(\bz_t, \xt) - \Mt^{\what\Par}(\bz_t, \Et(\xt)) \big\|_2^2 \Big]
\\
&=
\risk_{\cdm,t}(\Mt^{\what\Par},\Et)
-
\orisk^\star_{\cdm,t}\\
&=\underbrace{\inf_{\Par\in\Parspacecdm}\risk_{\cdm,t}(\Mt^{\Par},\Et)-\orisk^\star_{\cdm,t}}_{\text{approximation error}}
\,\,+\,\,
\underbrace{\risk_{\cdm,t}(\Mt^{\what\Par},\Et)
-
\inf_{\Par\in\Parspacecdm}\risk_{\cdm,t}(\Mt^{\Par},\Et)}_{\text{generalization error}}.
\end{align}

We claim the follow bounds on the approximation and generalization error which we will prove momentarily.
\begin{itemize}
    \item [(a).] If we choose $J = \bigOtil(L)$, $D = \bigO(\state \layer)$, $D' =\bigOtil( \mmax\state \layer^{3})$, and $\bound = \bigOtil(\Binvbd+(\state\layer+\mmaxb^2)\sqrt{\diminadapt})$, then the approximation error
    \begin{align}
\!\!\!\!\inf_{\Par\in\Parspacecdm}\risk_{\cdm,t}(\Mt^{\Par},\Et)-\orisk^\star_{\cdm,t}
\leq
\dimimage\cdot
\bigOtil \Bigg( \sqrt{\frac{ (\state \layer^{8}\mmaxb^2+\diminadapt)\state^5\layer^3}{n}} 
+
\state^{7}\Binvbd^2\Big(\suffmeasure(\scorefun)+\frac{1}{\diminadapt}\Big)
\Bigg).\label{eq:pf_cdm_approximation_bd}
    \end{align}
    \item[(b).] Under the same choice of model class $\Parspacecdm$, the generalization error 
    \begin{align}
   \risk_{\cdm,t}(\Mt^{\what\Par},\Et)
-
\inf_{\Par\in\Parspacecdm}\risk_{\cdm,t}(\Mt^{\Par},\Et)
\leq
\bigOtil \Bigg( \dimimage\cdot\sqrt{\frac{ (\state \layer^{8}\mmaxb^2+\diminadapt)\state^5\layer^3}{n}} \Bigg)
    \end{align}
    with probability at least $1-1/n.$
\end{itemize}
Combining the claims yields Theorem~\ref{theorem:CDM-Generalization-two-step}. \\
\paragraph{(a) Approximation error.}
   Take some $\delta'>0$ which will be defined later.
    For the feed forward layers, we use  \Cref{lemma:Approximation-FF-CDM,lemma:Approximation-FF-CDM-UP} with $\delta=\delta'\ll 1$.
    For the self-attention layers, we use \Cref{lemma:Self-attention-CDM} with $\delta=\frac{\delta'}{\max_{v}\|\qnetwork^{(\ell)}_{v}\|_\infty}$.
    Following the argument in the proof of \Cref{theorem:CLIP-Generalization},  $\qnetwork^{(\ell)}_{v}=f^{(\ell)}_{\iota(\pa^{(L-\ell)}(v))}(\hnetwork^{(\ell)}_{v})$ is bounded by $3(1\lor\log S\transbd)$, and $\delta$ in \Cref{lemma:Self-attention-CDM} is bounded by $\frac{\delta'}{3(1\lor\log S\transbd)}$. 

    The error from each operation is then bounded by $\delta'$ in the $\|\cdot\|_\infty$-norm.
    Now we can apply \Cref{lemma:Appendix-CDM-errorpropagation} to obtain that
    %
       \begin{align}
      \|\Mt(\bz_t,\Et(\xt))-\truebbmcdm(\bz_t,\xt)\|_2 &  \textstyle 
      \leq  
      d_\image^\frac12 8^{L+1}S^2\delta \times \prod_{1\leq k\leq L}(2m_{\image}^{(k)}+3)+ d_\image^\frac12 S^2\delta_\txt
      \\ & \leq d_\image^\frac32 40^{L+1} S^2 \delta + d_\image^\frac12 S^2\delta_\txt
     \label{eq:ApproxError-5}
    \end{align}

        We choose 
    \begin{align}
        \delta'=\frac{1}{40^{L+1}d_\image\state^2}\Big(\frac{ (\state \layer^{8}\mmaxb^2+\diminadapt)\state^5\layer^3}{n}\Big)^{1/4}
    \end{align}
    with $\mmax=\max\{\max_{k}m^{(k)}_\txt, \max_{k}m^{(k)}_\image\}$.
    Moreover, from Proposition~\ref{prop:representation_equivalence}, the definition of $\delta_\txt$ in Eq.~\eqref{eq:def_delta_txt} and Lemma~\ref{lemma:Boundedness-1}, it can be verified that there exists some $\Adapter(\cdot)$ in Eq.~\eqref{eq:adapter_def_cdm} such that, $\lops{\adaweighta}\leq\polyshortprime\Binvbd,\lops{\adaweightb}\leq \polyshortprime(\state\layer+\mmax^2)\sqrt{\diminadapt}$, and
    \begin{align}
        \E_{\xt}\delta_\txt^2
        &\leq \E_{\xt}\|\log\readtxt(\what\Et(\xt))-\log
        \trueEt(\xt)\|_2^2
        \leq
        \polyshort\state^2\cdot
\E_{\xt}\|\what\Et(\xt)
-
        \trueEt(\xt)\|_2^2  \\
        &\leq
        \polyshort \state^{2} \cdot \Binvbd^2 \cdot \tslinvLip^2 \cdot \truedimclip\cdot ( \suffmeasure(\scorefun) + \diminadapt^{-1} )
        \leq 
         \polyshort \state^{3} \cdot \Binvbd^2 \cdot ( \suffmeasure(\scorefun) + \diminadapt^{-1} )
    \end{align}
    for some $\polyshort,\polyshortprime>0$ depending polynomially on $\transbd^{\mmaxone}$,
    where the last line follows since
    $\truedimclip=\state$, and $\tslinvLip\leq c\transbd^{2\mmaxone}$ by Lemma~\ref{lemma:BoundScore} and the fact that $\truescorelink^{-1}=\exp(\cdot)$.
    Putting pieces together, 
    according to \Cref{lemma:Approximation-FF} and \Cref{lemma:Self-attention-CLIP}, 
    we now know that there exists some parameter $\Par\in\Parspaceclip$ such that bound~\eqref{eq:pf_cdm_approximation_bd} is satisfied and
    \begin{align}
        D&\leq d_\feature+d_\positional = 3\state\layer+2\layer=\bigO(\state\layer),
        \\
        J &\lesssim (\log\log(\state\transbd/\delta'))\log(\state\transbd/\delta')=\bigOtil(\layer),
        \\
       D'=\linf{\bj} &\lesssim \mmax \state(\log(\state\transbd/\delta'))^3+d_\feature+d_\positional = \bigOtil(\mmaxb\state\layer^3),
        \\
        \bound &\lesssim \state(\transbd^2+\log(\state\transbd/\delta'))+\mmax^2+\log\frac{\dim\log(\state\transbd)}{\delta'}+(\Binvbd+(\state\layer+\mmax^2)\sqrt{\diminadapt}) \\
        &= \bigOtil(\Binvbd+(\state\layer+\mmaxb^2)\sqrt{\diminadapt}).
    \end{align}

\paragraph{(b) generalization error.}
Since $\Mt^\EstPar$ is the minimizer of $\what\risk_{\cdm,t}(\Mt^\Par,\Et)$ defined in Eq.~\eqref{eq:cdm-empirical-risk-minimization}, we have
\begin{align}
    \risk_{\cdm,t}(\Mt^\EstPar,\Et)-\inf_{\Par\in\Parspacecdm}\risk_{\cdm,t}(\Mt^\Par,\Et)
    &\leq
    2\sup_{\Par\in\Parspacecdm}|\what\risk_{\cdm,t}(\Mt^\Par,\Et)
    -
    \risk_{\cdm,t}(\Mt^\Par,\Et)|.\label{eq:erm_bound_cdm}
\end{align}

 Next, we verify the conditions for Lemma~\ref{lm:generalization_lemma} and then  apply the lemma to derive an upper bound for the R.H.S. of Eq.~\eqref{eq:erm_bound_cdm}. 

    In Lemma~\ref{lm:generalization_lemma}, take $\Theta=\Parspacecdm,~\rho(\Par,\Par')=\nrmp{\Par-\Par'}$, $z_i=(\xi^{(i)},\xt^{(i)},\bz_{t}^{(i)})$, and
   \begin{align}
   f(z_i;\Par)
   =
  \frac{1}{\dimimage} \big\| \xi^{(i)} - \Mt^\Par(\bz_t^{(i)}, \Et(\xt^{(i)})) \big\|_2^2.
 \end{align}
   \\
   
\myunder{Verification of condition~(a) in Lemma~\ref{lm:generalization_lemma}.}
We note that the set $\Parspacecdm$ with metric $\rho(\Par,\Par')=\nrmp{\Par-\Par'}$ has a diameter $\bound_\rho\defn2\bound$. Furthermore, the  dimension of $\Parspacecdm$ is bounded by $d_\rho\defn (\numfflayer+3)(2\layer+1)(D+D'+1)^2+\state+2\state\diminadapt=\bigOtil(\state^2\layer^8\mmaxb^2+2\state\diminadapt)$. Thus, by Example 5.8 in~\cite{wainwright_2019}, we have
$\log\cN(\Delta;\Parspacecdm,\nrmp{})\leq d_\rho\log(1+2 r/\Delta)\leq d_\rho\log(2 A_\rho r/\Delta)$ for $\Delta\in(0,2r]$ with $A_\rho=2$.\\

\myunder{Verification of condition~(b) in Lemma~\ref{lm:generalization_lemma}.}
Since $f(z_i;\Par)$ is $4\dimimage\state^{2}$-bounded by the construction of $\Mt^\Par$ and the fact that $\linf{\xi}\leq\state$, it follows that $f(z_i;\Par)-\E[f(z_i;\Par)]$ is $\sigma=c\state^{2}$-sub-Gaussian for all $\Par\in\Parspacecdm$  for some numerical constant $c>0$.\\

\myunder{Verification of condition~(c) in Lemma~\ref{lm:generalization_lemma}.}
By Lemma~\ref{lm:cdm_lip} and the boundedness condition, we have
\begin{align}
  &  |f(z_i;\Par) - f(z_i;\Par')|
   \\  &\leq 
   \frac{1}{\dimimage} |\< \Mt^{\Par'}(\bz_t^{(i)}, \Et(\xt^{(i)})- \Mt^\Par(\bz_t^{(i)}, \Et(\xt^{(i)})),
   2\xi^{(i)}-\Mt^{\Par'}(\bz_t^{(i)}, \Et(\xt^{(i)})- \Mt^\Par(\bz_t^{(i)}, \Et(\xt^{(i)}))
    \>|\\
   &\leq
   \frac{4\state}{\sqrt{\dimimage}} \| \Mt^{\Par'}(\bz_t^{(i)}, \Et(\xt^{(i)})- \Mt^\Par(\bz_t^{(i)}, \Et(\xt^{(i)}))\|_2
  \\
    & \leq \bound_f \nrmp{\Par-\Par'},~~~~~~~~\text{where}~~ \bound_f\defn ((c\bound)^{18\numfflayer\layer}\state^9\readbd^3\log^3\mmax)^{2\layer+2}\exp(2\readbd),
\end{align}where $\readbd=4\mmaxone\log\transbd.$ 
Therefore, we may choose $\sigma'=\bound_f$ and condition (c) is hence satisfied.

Now, invoking Lemma~\ref{lm:generalization_lemma} and plugging in the values of $d_\rho,\sigma,\sigma',A_\rho,\bound_\rho$, we find
\begin{align}
 \sup_{\Par\in\Parspacecdm}|\what\risk_{\cdm,t}(\Mt^\Par,\Et)
    -
    \risk_{\cdm,t}(\Mt^\Par,\Et)|
    &\leq \dimimage\cdot c\sigma \sqrt{\frac{d_\rho \log \left(2A_\rho \left(1 + B_\rho \sigma' / \sigma\right)\right) + \log(1 / \eta)}{n}}\\
    &\leq
    \bigOtil \Bigg( \dimimage\state^2\sqrt{\frac{ (\state \layer^{8}\mmaxb^2+\diminadapt)\state\layer^3+\log(1/\eta)}{n}} \Bigg)
\end{align} with probability at least $1-\eta.$ Setting $\eta=1/n$ completes the proof.

\subsection{Joint training of denoising function and text representation }\label{sec:joint_train_cdm}
In this section,
we analyze the sample complexity of jointly learning the conditional denoising models (CDMs) and the text representation within the JGHM framework. Following the setup of Section~\ref{sec:CDM_GHM}, 
suppose we are given a dataset of iid samples $\{(\bz_t^{(i)}, \xi^{(i)}, \xt^{(i)})\}_{i \in [n]} \sim_{iid} \truemucdm$.

The conditional denoiser is modeled as
\[
\Mt^\Par(\bz_t, \Et(\xt)) = \read_{\cdm} \circ \TF_{\cdm} \circ \encode_{\cdm}(\bz_t, \Et(\xt)), 
\] 
where $\Et(\xt)=\NN^{\bW_\txt}_\txt(\xt)$ as defined in Section~\ref{sec:clip_JGHM}, and the remaining components are the same as defined in Section~\ref{sec:CDM_GHM}, except that in the embedding $\encode_\cdm$, we let
\begin{align}
\hnetwork_{\txt,d}^{(0)}=\widetilde\readtxt(\Et(\xt)), ~~\text{ where }  \widetilde\readtxt(z)\defn \proj_{[-\readbdcdmenc,\readbdcdmenc]}(z), 
\end{align}
in contrast to Eq.~\eqref{eq:cdm_log_trun_text_def}. 
During  pre-training, we optimize the parameter $\Par = \bW_\cdm$, while hoding $\read_{\cdm}$ $\encode_{\cdm}$ as fixed. 
 More specifically, we find the model via empirical risk minimization 
\begin{align}\label{eq:cdm-empirical-risk-minimization-one-step}
\textstyle \what \Par = \underset{\Par \in \Parspacecdmone }{\argmin} \Big\{ \what\risk_{\cdm, t}(\Mt^\Par, \Et) \defn 
\frac{1}{n}\sum_{i=1}^n
\big\| \xi^{(i)} - \Mt^\Par(\bz_t^{(i)}, \Et(\xt^{(i)})) \big\|_2^2 \Big\}, 
\end{align}
where the parameter space is defined as
\begin{align}\label{eqn:CDM_param_space_one_step}
&~\Parspacecdmone \defn \Big\{ \bW_{\cdm},\bW_\txt \text{ as defined in Eq.~(\ref{eqn:W_matrix_CLIP_encoder})};
\\
&~~~~~~~~~~\nrmps{\btheta} \defn  \max_{i \in [\depth+1], \ell \in [2\layer+1]}\{ \| \ffnweight_{i, \cdm}^{(\ell)} \|_{\op}, \| {\querymatrix}_{,\cdm}^{(\ell)}\|_{\op}, \| {\keymatrix}_{,\cdm}^{(\ell)} \|_{\op}, \| {\valuematrix}_{,\cdm}^{(\ell)} \|_{\op} \}  \\
&~~~~~~~~~~~~~~~~~~~\vee
\max_{i \in [\depth+1], \ell \in [\layer]}\{ \| \ffnweight_{i, \txt}^{(\ell)} \|_{\op}, \| {\querymatrix}_{,\txt}^{(\ell)}\|_{\op}, \| {\keymatrix}_{,\txt}^{(\ell)} \|_{\op}, \| {\valuematrix}_{,\txt}^{(\ell)} \|_{\op} \}  \le B \Big\}.
\end{align}


Similar to Theorem~\ref{theorem:CDM-Generalization-two-step}, we have the following result
\begin{theorem}[Estimation error of conditional denoising function, joint training]\label{theorem:CDM-Generalization-one-step}
Suppose that \Cref{assumption:factorization} and \Cref{ass:bounded_transition} hold. For simplicity, assume $t=1$. Let $\Parspacecdmone$ be the set defined in Eq.~(\ref{eqn:CDM_param_space_one_step}), where $J = \bigOtil(\layer)$, $D = \bigO(\state \layer)$, $D' =\bigOtil(\mmaintextmax \state \layer^3)$, and $\bound = \bigOtil(\state\layer+\mmaintextmax^2)$. Let  $\EstPar$ be the empirical risk minimizer defined in Eq.~(\ref{eq:cdm-empirical-risk-minimization-one-step}). Then, with probability at least $1 - 1/n$, we have
\begin{align}
\E_{(\xi, \xt, \bz_t)}
\Big[\frac{1}{\dimimage} \big\| \truebbmcdm(\bz_t, \xt) - \Mt^{\EstPar}(\bz_t, \Et(\xt)) \big\|_2^2 \Big]
\le \bigOtil \Bigg( \sqrt{\frac{\state^6 \layer^{11}\mmaintextmax^2}{n}} 
\Bigg),\label{eqn:excess_risk_bound_CDM_joint}
\end{align}
where $\bigOtil$ hides polynomial factors in $(\log(\mmaintextmax \state \layer n), (\transbd)^{\mmaintextone})$.
\end{theorem}
\begin{proof}[Proof of Theorem~\ref{theorem:CDM-Generalization-one-step}]
    The proof follows from the same arugment as the proof of 
    Theorem~\ref{theorem:CDM-Generalization-two-step}, thus we only highlight the differences here.
    Similar to the proof of  Theorem~\ref{theorem:CDM-Generalization-two-step},
we claim that
\begin{itemize}
    \item [(a).] If we choose $J = \bigOtil(L)$, $D = \bigO(\state \layer)$, $D' =\bigOtil( \mmax\state \layer^3)$, and $\bound = \bigOtil(\state\layer+\mmaxb^2)$, then the approximation error
    \begin{align}
\inf_{\Par\in\Parspacecdmone}\risk_{\cdm,t}(\Mt^{\Par},\Et)-\orisk^\star_{\cdm,t}
\leq
\dimimage\cdot
\bigOtil \Bigg( \sqrt{\frac{ \state^6 \layer^{11}\mmaxb^2}{n}} 
\Bigg).\label{eq:pf_cdm_approximation_bd_one_step}
    \end{align}
    \item[(b).] Under the same choice of model class $\Parspacecdmone$, the generalization error 
    \begin{align}
   \risk_{\cdm,t}(\Mt^{\what\Par},\Et)
-
\inf_{\Par\in\Parspacecdmone}\risk_{\cdm,t}(\Mt^{\Par},\Et)
\leq
\bigOtil \Bigg( \dimimage\cdot\sqrt{\frac{ \state^6 \layer^{11}\mmaxb^2}{n}} \Bigg)
    \end{align}
    with probability at least $1-1/n.$
\end{itemize}
Combining the claims yields Theorem~\ref{theorem:CDM-Generalization-one-step}. \\

\paragraph{(a) approximation error}
By using the same parameter choise as that in Theorem~\ref{theorem:CLIP-Generalization}, we have
\begin{align}
    \delta_\txt \leq \delta' \prod_{1\leq k\leq L}(2m^{(k)}_\txt+3)
    \leq 5^{L}d_\txt
\end{align}
according to Eq.~\eqref{eq:Appendix-CLIP-ErrorPropagation-1}. 
As a result, we may choose $\delta'=\frac{\sqrt{S^2 L^{11} \mmaxb^2}}{5^{L}d_\txt \sqrt{n}}$ and obtain Eq.~\eqref{eq:pf_cdm_approximation_bd_one_step}.

\paragraph{(b) generalization error}
Likewise, we verify the conditions for Lemma~\ref{lm:generalization_lemma}.
 We take $\Theta=\Parspacecdmone$, $\rho(\Par,\Par')=\nrmp{\Par-\Par'}$, $z_i=(\xi^{(i)},\xt^{(i)},\bz_{t}^{(i)})$, and
   \begin{align}
   f(z_i;\Par)
   =
   \frac{1}{\dimimage} \big\| \xi^{(i)} - \Mt^\Par(\bz_t^{(i)}, \Et(\xt^{(i)})) \big\|_2^2.
 \end{align}
 Similarly, it can be verified that condition~(a) in Lemma~\ref{lm:generalization_lemma} is satisfied with $A_\rho =2,\bound_\rho=2\bound$, and number of parameters $d_\rho=\bigOtil(\state^2\layer^8\mmaxb^2);$
 $f(z_i;\Par)-\E[f(z_i;\Par)]$ is $\sigma=c\state^{2}$-sub-Gaussian for all $\Par\in\Parspacecdmone$; and 
 similar to Lemma~\ref{lm:cdm_lip},
 it can be verified that
\begin{align}
    |f(z_i;\Par) - f(z_i;\Par')|
    & \leq \bound_f \nrmp{\Par-\Par'},~~~~~~~~\text{where}~~ \bound_f\defn ((c\bound)^{18\numfflayer\layer}\state^9\readbd^3\log^3\mmax)^{3\layer+3}\exp(2\readbd).
\end{align}
Finally, invoking Lemma~\ref{lm:generalization_lemma} and plugging in the values of $d_\rho,\sigma,\sigma',A_\rho,\bound_\rho$ yields the desired bound.

\end{proof}

\subsection{Evaluation of error propagation}\label{subsection:CDM-ErrorPropagation}
Similarly to \Cref{lemma:Appendix-CLIP-errorpropagation} in \Cref{subsection:ErrorPropagation}, we evaluate the propagation of the errors.
We denote the estimation error of $h_{\txt,\root}^{(0)}$ by $\delta_\txt$, so that \Cref{lemma:Appendix-CDM-errorpropagation} can be used for both simultaneous training and two-stage training.
For simultaneous training of the image and text models, the error propagation lemma for contrastive learning (\Cref{lemma:Appendix-CLIP-errorpropagation}) can be used to bound $\delta_\txt$.
\begin{lemma}[Evaluation of error propagation]\label{lemma:Appendix-CDM-errorpropagation}
    Assume we have functions 
     $\fnetwork_{\downarrow, \iota}^{(\ell)}, \fnetwork_{\uparrow, \iota}^{(\ell)}\ (1\leq \ell \leq L,\iota\in [m_\txt^{(\ell)}])$ such that
    \begin{align}
    \begin{aligned}
        \|f_{\downarrow,\iota}^{(\ell)}(h)-\fnetwork_{\downarrow,\iota}^{(\ell)}(h)\|_\infty
        &\leq \delta,\quad \forall h\in \R^S\text{ such that $\max_{s\in S}h_s= 0$, }\ell\in [L]   , 
        \\
        \|f_{\uparrow,\iota}^{(\ell)}(h)-\fnetwork_{\uparrow,\iota}^{(\ell)}(h)\|_\infty
        &\leq \delta,\quad \forall h\in \R^S\text{ such that $\max_{s\in S}h_s= 0$, }\ell\in [L]   , 
    \end{aligned}
    \end{align}
    and  $\|\bdelta^{(\ell)}_{v}\|_\infty\leq \delta$ holds for all $\ell=L-1,\dots,0$ and $v\in \cV^{(L)}_\image$. 
    Moreover, we assume that $\|h_{\txt,\root}^{(0)}-\hnetwork_{\txt,d_\txt}^{(0)}\|_\infty \leq \delta_\txt$. 

    Consider the approximated update introduced in \eqref{eq:Appendix-CDM-ErrorPropagation-109} and \eqref{eq:Appendix-CDM-ErrorPropagation-10}. 
    Then, we have the following bound on the error propagation:
    \begin{align}
   & \textstyle 
   \max_{v\in \cV^{(L)}_\image}
   \|\hnetwork_{\downarrow,v}^{(\ell)}-h_{\downarrow,\iota(\pa^{(L-\ell)}(v))}^{(\ell)}\|_\infty
      \textstyle \leq \delta \times \prod_{\ell+1\leq k\leq L}(2m_{\image}^{(k)}+3),
     \label{eq:Appendix-CDM-ErrorPropagation-1}
    \\
   &  \textstyle 
\max_{v\in \cV^{(L)}_\image}
\|\qnetwork_{\downarrow,v}^{(\ell)}-q_{\downarrow,\iota(\pa^{(L-\ell)}(v))}^{(\ell)}\|_\infty
      \textstyle \leq \delta \times \prod_{\ell+1\leq k\leq L}(2m_{\image}^{(k)}+3),
    \label{eq:Appendix-CDM-ErrorPropagation-12}
    \\
   & \textstyle 
\max_{v\in \cV^{(\ell)}_\image}
\| \bnetwork_{v}^{(\ell)} - b_{\pa^{(L-\ell)}(v)}^{(\ell)}\|_\infty
       \textstyle \leq 8^{\ell}\delta \times \prod_{1\leq k\leq L}(2m_{\image}^{(k)}+3)+\delta_\txt,
    \label{eq:Appendix-CDM-ErrorPropagation-13}
    \\
&\textstyle \max_{v\in \cV^{(L)}_\image}
\|\bnetwork_{v}^{(L+1)} - b_{v}^{(L+1)}\|_\infty
       \textstyle \leq 8^{L+1}\delta \times \prod_{1\leq k\leq L}(2m_{\image}^{(k)}+3)+\delta_\txt
        \label{eq:Appendix-CDM-ErrorPropagation-14}
    .
    \end{align}
    Furthermore, we have
    \begin{align}
      \textstyle \max_{v\in \cV^{(L)}_\image} 
      \|{\sf M}_{t,v}-(\truebbmcdm (\bz_t, \xt))_v\|_\infty
      &  \textstyle \leq 8^{L+1}S^2\delta \times \prod_{1\leq k\leq L}(2m_{\image}^{(k)}+3)+S^2\delta_\txt.
     \label{eq:Appendix-CDM-ErrorPropagation-9}
    \end{align}
\end{lemma}
\begin{proof}
    The bounds \eqref{eq:Appendix-CDM-ErrorPropagation-1} and \eqref{eq:Appendix-CDM-ErrorPropagation-12} are the same as \eqref{eq:Appendix-CLIP-ErrorPropagation-1} and \eqref{eq:Appendix-CLIP-ErrorPropagation-12} of \Cref{lemma:Appendix-CLIP-errorpropagation}.
    Thus let us focus on the upsampling process of the image model. 
    We will prove \eqref{eq:Appendix-CDM-ErrorPropagation-13} using the induction, using \eqref{eq:Appendix-CDM-ErrorPropagation-1} and \eqref{eq:Appendix-CDM-ErrorPropagation-12}. 
    Until the final part, let us assume $\delta_\txt=0$.

    First, we prove that \eqref{eq:Appendix-CDM-ErrorPropagation-13} holds for $\ell=1$.
    For $v\in \cV^{(L)}_\image$, we have
    \begin{align}
        \|\bnetwork_{v}^{(1)}
        -
        b_{\pa^{(L-1)}(v)}^{(1)}\|_\infty
        &\textstyle\leq  
        2\|\hnetwork_{v}^{(0)}+\hnetwork_{\txt,d_\txt}^{(0)}-\qnetwork^{(1)}_{v}
        - (h_{\root}^{(0)}+h_{\txt,\root}^{(0)}- q^{(1)}_{\pa^{(L-1)}(v)})\|_\infty
        \\ & \textstyle \leq  
    2\|\hnetwork_{v}^{(0)}
    -h_{\root}^{(0)}\|_\infty
    +2\|\hnetwork_{\txt,d_\txt}^{(0)}-h_{\root}^{(0)}\|_\infty
    + 2\|\qnetwork^{(1)}_{v} - q^{(1)}_{\pa^{(L-1)}(v)}\|_\infty 
    \\ & \textstyle\leq 4\delta \times \prod_{1\leq k\leq L}(2m_{\image}^{(k)}+3), 
    \end{align}
    where we used \Cref{lemma:Appendix-CLIP-Lipschitz-normalization} for the first inequality and \eqref{eq:Appendix-CDM-ErrorPropagation-1} and \eqref{eq:Appendix-CDM-ErrorPropagation-12} for the last inequality.

    Then, let us assume that \eqref{eq:Appendix-CDM-ErrorPropagation-13} holds for $\ell$ and prove it for $\ell+1$ ($\ell=1,\dots,L-1$). 
    For $v\in \cV^{(L)}_\image$, we have
    \begin{align}
       & \|\bnetwork_{v}^{(\ell+1)}
        -
        b_{\pa^{(L-\ell-1)}(v)}^{(\ell+1)}\|_\infty
         \\ & \leq 
         2\|\fnetwork_{\uparrow,\iota(\pa^{(L-\ell)}(v))}^{(\ell)}(\bnetwork_{v}^{(\ell)})
    +\hnetwork^{(\ell)}_{v}-\qnetwork^{(\ell+1)}_{v}
    -
    (f_{\uparrow,\iota(\pa^{(L-\ell)}(v))}^{(\ell)}(b_{\pa^{(L-\ell)}(v)}^{(\ell)})
    +h^{(\ell)}_{\pa^{(L-\ell)}(v)}-q^{(\ell+1)}_{\pa^{(L-\ell-1)}(v)})\|_\infty
    \\ & \leq 
        2\|\fnetwork_{\uparrow,\iota(\pa^{(L-\ell)}(v))}^{(\ell)}(\bnetwork_{v}^{(\ell)})
        -f_{\uparrow,\iota(\pa^{(L-\ell)}(v))}^{(\ell)}(\bnetwork_{v}^{(\ell)})\|_\infty
        \\ & \quad +
        2\|f_{\uparrow,\iota(\pa^{(L-\ell)}(v))}^{(\ell)}(\bnetwork_{v}^{(\ell)})-f_{\uparrow,\iota(\pa^{(L-\ell)}(v))}^{(\ell)}(b_{\pa^{(L-\ell)}(v)}^{(\ell)})\|_\infty
    \\ & \quad +2\|\hnetwork^{(\ell)}_{v}-h_{\pa^{(L-\ell)}(v)}^{(\ell)}\|_\infty + 2\|\qnetwork^{(\ell+1)}_{v}-q_{\pa^{(L-\ell-1)}(v)}^{(\ell+1)}\|_\infty
      \\ & \leq 2\delta + 2\|\bnetwork_{v}^{(\ell)}-b_{\pa^{(L-\ell)}(v)}^{(\ell)}\|_\infty 
      + 2\|\hnetwork^{(\ell)}_{v}-h_{\pa^{(L-\ell)}(v)}^{(\ell)}\|_\infty + 2\|\qnetwork^{(\ell+1)}_{v}-q_{\pa^{(L-\ell-1)}(v)}^{(\ell+1)}\|_\infty
      \\ &\textstyle\leq 2\delta + 2 \times 8^\ell \delta \times \prod_{1\leq k\leq L}(2m_{\image}^{(k)}+3)
    + 4\delta \times \prod_{1\leq k\leq L}(2m_{\image}^{(k)}+3)
      \\ &\textstyle \leq  8^{(\ell+1)}\delta \times \prod_{1\leq k\leq L}(2m_{\image}^{(k)}+3)
    \end{align}
    where we used \Cref{lemma:Appendix-CLIP-Lipschitz-logsumexp} for the first inequality and \Cref{lemma:Appendix-CLIP-Lipschitz-normalization} for the third inequality. 
    Now \eqref{eq:Appendix-CDM-ErrorPropagation-13} is confirmed for $\ell+1$. 
    In the same way, \eqref{eq:Appendix-CDM-ErrorPropagation-14} is proved. Note that $\softmax$ is $1$-Lipschitz with respect to the $\|\cdot\|_{\infty}$ norm. 
    Thus, the bound \eqref{eq:Appendix-CDM-ErrorPropagation-9} directly follows from \eqref{eq:Appendix-CDM-ErrorPropagation-14}.

    Finally, we consider how the error from the text model $\delta_\txt$ propagates. 
    For this, we only need to bound how the message passing algorithm changes, because the difference between the message passing algorithm and its neural network approximation is already bounded. 
    According to \Cref{lemma:Appendix-CDM-errorpropagation^2}, that is bounded by $S^2\delta_\txt$. Now we have obtained the assertion. 
 \end{proof}

 \begin{lemma}\label{lemma:Appendix-CDM-errorpropagation^2}
    Suppose that we run the upsampling process of the message passing algorithm \eqref{eq:Appendix-CDM-ErrorPropagation-5} by changing
     $h_{\txt,\root}^{(0)}$ to ${h'}_{\txt,\root}^{(0)}$ with $\|{h'}_{\txt,\root}^{(0)}-h_{\txt,\root}^{(0)}\|_\infty\leq \delta_\txt$ while all the others are the same. 
     Then, the deviation of the new optimal prediction $\boldsymbol{m}_{\star,t,v}'$  from $\boldsymbol{m}_{\star,t,v}$ is bounded by
     \begin{align}
         \|\boldsymbol{m}_{\star,t,v}'-\boldsymbol{m}_{\star,t,v}\|_\infty \leq S^2\delta_\txt
     \end{align}
     for all $v\in \cV_\image^{(L)}$. 
 \end{lemma}
 \begin{proof}
 Because of $\softmax$ in the final part of Eq.~\eqref{eq:Appendix-CDM-ErrorPropagation-5}, we do not need to consider $\normalize$ in the message passing algorithm. Thus, let us consider how the change in $h_{\txt,\root}^{(0)}$ propagates in the following pipeline.
    \begin{align}
    \begin{aligned}
    \textstyle \bar{b}_{\image,\root}^{(0)} &\textstyle= 
    h_{\image,\root}^{(0)}
    +h_{\txt,\root}^{(0)}\in \R^S,
    \\
    \textstyle \bar{b}_{\image,v}^{(\ell)} &\textstyle= f_{\image,\uparrow,\iota(v)}^{(\ell)}(\bar{b}^{(\ell-1)}_{\image,\pa(v)}-q^{(\ell)}_{\image,v})
    +h^{(\ell)}_{\image,v}\in \R^S,
    && v\in \cV^{(\ell)}_\image,\ \ell = 1,\dots,L
   \\ 
    \textstyle 
    \boldsymbol{m}_{\star,t,v}
    =&\textstyle \sum_{s\in [S]} s \cdot \softmax(\bar{b}_{\image,v}^{(L)}) \in \R^S,
     &&v\in \cV^{(L)}_\image,
       \end{aligned}
    \end{align}
    According to \Cref{lemma:Appendix-CLIP-Lipschitz-logsumexp}, we know that the change of $\bar{b}_{\image,v}^{(\ell)}$ evaluated by $\|\cdot\|_\infty$-norm is bounded by the change of $\bar{b}_{\image,v}^{(\ell-1)}$.
    Thus, the change of $\bar{b}_{\image,v}^{(L+1)}$ is at most $\delta_\txt$ in the $\|\cdot\|_\infty$-norm. 
    Moreover, $\softmax$ is $1$-Lipschitz with respect to the $\|\cdot\|_\infty$-norm, and $s$ is bounded by $S$, which yields that the estimation of each leaf variable changes at most $S^2\delta_\txt$. 
 \end{proof}

\section{Proof of Theorem~\ref{thm:vlm_two_steps}}\label{sec:proof-vlm_two_steps}
\subsection{Overview}
This section solves the problem of estimating the posterior probability of the next word $\truemu(x_{\txt,i+1}|\xi,x_{\txt,1},$ $\dots,x_{\txt,i})$ for every $i=1,\dots,d_\txt-1$ in parallel. 
In this overview section, \Cref{subsubsection:NWP-Intro-BP} first introduces the belief propagation algorithm, which exactly calculates $\truemu(x_{\txt,i+1}|\xi,x_{\txt,1},\dots,x_{\txt,i})$ for each fixed $i$.
We then discuss how to parallelize the belief propagation algorithm into the message passing algorithm in \Cref{subsubsection:NWP-Intro-MP}, and finally explain how to implement the message passing algorithm with transformer networks in \Cref{subsubsection:NWP-Intro-NN}.

In the following, we identify nodes and integers by following \Cref{definition:Numbering}. 
When we refer to $v+1$, it means the leaf node $u$ such that $u=v+1$ in the interger representation of \Cref{definition:Numbering}.
We remark that, although the next node $v+1$ is not defined for $v=d_\txt$, we do not separate the case of $v=d_\txt$ when we use the notation $v+1$ in the following, because how to deal with the case of $v=d_\txt$ does not affect the prediction of $x_{\txt,2},\dots,x_{\txt,d_\txt}$.
Also, the discussion mainly focuses on the text processing, and therefore we sometimes omit the subscript ``$\txt$''. 

\subsubsection{Belief propagation algorithm}\label{subsubsection:NWP-Intro-BP}
To predict an unobserved leaf node of the text, the belief propagation algorithm can exactly calculates the posterior probability. 
Suppose that we have $\P[s|\xi]$ (, which can be approximated either by the transformer network $\NN_\image^{\bW_\image}$ in the contrastive learning or by the text embedding in the two-stage training). 
Given this, the belief propagation consists of the downsampling
\begin{align}\label{eqn:BP_nextword_downsample_text}
\begin{aligned}
       \message^{(L)}_{\downarrow,v}(x^{(L)}_{\txt,v}) =&~ \indicator[x^{(L)}_{\txt,v} = x_{\txt,v}]\ (v\leq i), \ \frac{1}{\staten}\ (\text{otherwise}), && v\in \cV^{(L)}_\txt,\\ 
    \message^{(\ell)}_{\downarrow, v}(x^{(\ell)}_{\txt,v})\ \propto\ &~ \textstyle  \sum_{x^{(\ell+1)}_{\txt,\cC(v)}} \prod_{v' \in \cC(v)} \Big( \psi_{\txt,\iota(v')}^{(\ell + 1)}(x_{\txt,v}^{(\ell)}, x^{(\ell + 1)}_{\txt,v'}) \message^{(\ell+1)}_{\downarrow,v'}(x^{(\ell + 1)}_{\txt,v'})\Big), && v\in \cV_\txt^{(\ell)},\ \ell = L-1, \ldots, 1.
\end{aligned}
\end{align}
and the upsampling
\begin{align}
\begin{aligned}
\message_{\uparrow, \root}^{(0)}(x^{(0)}_{\txt,\root}) =&~ 
\P[x^{(0)}_{\txt,\root}|\xi] \\ 
\message_{\uparrow,v}^{(\ell)}(x^{(\ell)}_{\txt,v})\ \propto\ &~ \textstyle \sum_{x^{(\ell-1)}_{\txt,\pa(v)}, x^{(\ell)}_{\txt,\cN(v)}}\psi^{(\ell)}(x^{(\ell-1)}_{\txt,\pa(v)}, x^{(\ell)}_{\txt,\cC(\pa(v))})\message_{\uparrow,\pa(v)}^{(\ell-1)}(x^{(\ell-1)}_{\txt,\pa(v)})\prod_{v' \in \cN(v)} \message_{\downarrow,v'}^{(\ell)}(x^{(\ell)}_{\txt,v'}),&&\\ & &&\hspace{-50mm} v\in \pa^{(L-\ell)}(i+1),\ \ell = 1,\ldots, L. 
\end{aligned}
\label{eqn:BP_nextword_upsample}
\end{align}
These beliefs $\message$ are normalized so that $\sum_{s}\message_s=1$.
The correctness of this algorithm is formally stated as follows.
\begin{lemma}[BP calculates the posterior probability of the next word exactly]\label{lem:BP_NextWord_exact}
When applying the belief propagation algorithm shown in \eqref{eqn:BP_nextword_downsample_text} and \eqref{eqn:BP_nextword_upsample}, it holds that $\message_{\uparrow,n}^{(L)}(x_{\txt,i+1}) = \truemu(x_{\txt,i+1}|\xi,x_{\txt,1},\dots,x_{\txt,i})$. 
\end{lemma}
\begin{proof}
    Referring to classical results \cite{pearl2022reverend,wainwright2008graphical, mezard2009information}, when we replace $\message_{\uparrow,\root}^{(0)}(x^{(0)}_{\root}) \ \propto \
    \P[x^{(0)}_{\root}|\xi]$ by the unconditioned $\message_{\uparrow, \root}^{(0)}(x^{(0)}_{\root})\ \propto\ 
    \P[x^{(0)}_{\root}]$ in \eqref{eqn:BP_nextword_upsample}, it holds that $\message_{\uparrow,i+1}^{(L)}(x_{\txt,i+1}) = \truemu(x_{\txt,i+1}|x_{\txt,1},\dots,x_{\txt,i})$.
    It is obvious to see that using $\message_{\uparrow, \root}^{(0)}(x^{(0)}_{\root}) = 
\P[x^{(0)}_{\root}|\xi] $ corresponds to conditioning on $\xi$ and that $\message_{\uparrow,i+1}^{(L)}(x_{\txt,i+1}) = \truemu(x_{\txt,i+1}|\xi,x_{\txt,1},\dots,x_{\txt,i})$.
\end{proof}

\subsubsection{Parallelization with message passing algorithm}\label{subsubsection:NWP-Intro-MP}
We then parallelize the belief propagation algorithm for different $i$ with the message passing algorithm.
We achieve this by grouping common variables across different $i$ into a single variables.
Remind that, for $v\in \cV_\txt^{(\ell)}$, $v^{(\ell')}$ means the node $u\in \cV_\image^{(\ell')}$ such that its corresponding integer is the same as that of $v$ (\Cref{definition:Numbering}).
Thus, for $u\in \cV_\txt^{(L)}$, $u^{(\ell)}\in \cN(\pa^{(L-\ell)}(v))$ means that $u$ has the corresponding node in the $\ell$-th level and that the corresponding node is a neighbor of $\pa^{(L-\ell)}(v)$. 

\paragraph{Downsampling.}
Starting with $h^{(L)}_{v}=x_{\txt,v}\ (v\in \cV^{(L)}_\txt)$, the downsampling process is defined as
\begin{align}
    \begin{aligned}
    \textstyle q_{v}^{(\ell)} &\textstyle= f_{\downarrow,\iota(\pa^{(L-\ell)}(v))}^{(\ell)}(h^{(\ell)}_{v}) \in \R^S,
    && v\in \cV^{(L)}_\txt,\ \ell = L,\dots,1,
   \\ 
    \textstyle h^{(\ell-1)}_{v} &\textstyle = \normalize \big(\sum_{\substack{{u}^{(\ell)}\in \cN(\pa^{(L-\ell)}(v))\\\text{ or }u=v}}\indicator [u\leq v] q_{{v'}}^{(\ell)}\big) \in \R^S,
     && v\in \cV^{(L)}_\txt,\ \ell = L,\dots,1,
       \end{aligned}\label{eq:Appendix-NextWord-MP-1}
    \end{align}
    where
    \begin{align}
    \begin{aligned}
     (f_{\downarrow, \iota}^{(L)}(x))_s &  \textstyle  = \log \psi_{\txt, \iota}^{(L)}(s,x),
        &&  x\in [\staten],\ s\in [\staten],
        \\
  (f_{\downarrow, \iota}^{(\ell)}(h))_s &   \textstyle  = \log \sum_{s\in [\staten]}\psi_{\txt, \iota}^{(\ell)}(s,a)e^{h_a}, 
        && h\in \R^{\staten},\ s\in [\staten], \ \ell = L-1,\dots,1.
    \end{aligned}\label{eq:Appendix-NextWord-MP-2}
    \end{align}
\paragraph{Upsampling.}
    The upsampling process is defined as
    \begin{align}
    \begin{aligned}
        \bar{b}^{(0)}_{v} & \textstyle  =h^{(0)}_{v} + (\log \P[s|\xi])_{s\in [\staten]}\in \R^S,\\
        \bar{b}^{(\ell)}_{v} & \textstyle = 
        \begin{cases}
            f^{(\ell)}_{\uparrow,\iota(\pa^{(L-\ell)}(v+1))}(\normalize(\bar{b}^{(\ell-1)}_{v}-q^{(\ell)}_{v}))+h^{(\ell)}_{v},
            & \text{(if $\pa^{(L-\ell)}(v) = \pa^{(L-\ell)}(v+1)$)}
            \\
        f^{(\ell)}_{\uparrow, \iota(\pa^{(L-\ell)}(v+1))}(\normalize(\bar{b}^{(\ell-1)}_{v})),
            & \text{(otherwise)}
        \end{cases} 
    \end{aligned}
    \\ v\in \cV^{(L)}_\txt,\ \ell = 1,2,\dots,L,
    \label{eq:Appendix-NextWord-MP-3}
    \end{align}
    where
    \begin{align}
    \label{eq:Appendix-NextWord-MP-4}
        (f_{\uparrow, \iota}^{(\ell)}(h))_s &   \textstyle  = \log \sum_{a\in [\staten]}\psi_{\txt, \iota}^{(\ell)}(a,s)e^{h_a}, 
        && \quad \quad \ \ \ h\in \R^{\staten},\ s\in [\staten], \ \ell = 1,2,\dots,L.
    \end{align}
    Then, it holds that $\softmax(\bar{b}^{(L)}_{i})_s = \mu_{\star}(x_{\txt,i+1}|\xi,x_{\txt,1},\dots,x_{\txt,i})$ for all $i=1,\dots,d-1$ and $s\in [\staten]$.

    \begin{proposition}[MP calculates the posterior probability of the next word in parallel]\label{proposition:MP-NWP}
        When applying the message passing algorithm defined in \eqref{eq:Appendix-NextWord-MP-1}, \eqref{eq:Appendix-NextWord-MP-2}, \eqref{eq:Appendix-NextWord-MP-3}, and \eqref{eq:Appendix-NextWord-MP-4}, for all $i=1,\dots,d-1$ and $s\in [S]$, it holds that $\softmax(\bar{b}^{(L)}_{i})_s = \truemu(x_{\txt,i}=s|\xi,x_{\txt,1},\dots,x_{\txt,i})$. 
    \end{proposition}
    
The original message passing algorithm has several issues when it comes to implementing it with transformer networks. 
First, in the downsampling process \eqref{eq:Appendix-NextWord-MP-1}, the number of $v'$ in the summation is not uniform across nodes in the same level.
Previously in contrastive learning and conditional diffusion model, we took average with self-attention and then applied the number of elements in the average (e.g., $m^{(\ell)}$) to compute summation. This cannot be directly adopted this time.
In addition, the upsampling process \eqref{eq:Appendix-NextWord-MP-3} uses subtraction ($\bar{b}^{(\ell-1)}_{v}-q^{(\ell)}_{v}$), ``$\normalize$'', and nonlinear transformation $f^{(\ell)}_{\uparrow,\iota}$ in this order, which is not implemented in one transformer block. 

Therefore, we prepare the following alternative version of the message passing algorithm to be implemented by a transformer network. 
The correspondence with the original version is easily confirmed, where 
$\bar{b}_{,v}^{(\ell-1)}-q_{\image,v}^{(\ell)}=b_{v}^{(\ell)}$ (if $\pa^{(L-\ell)}(v)=\pa^{(L-\ell)}(v+1)$) and $\bar{b}_{v}^{(\ell-1)}=b_{v}^{(\ell)}$ (otherwise).

\paragraph{Downsampling (alternative).} 
Define $a_v^{(\ell)} = \iota(\pa^{(L-\ell)}(v))+ \indicator[v^{(\ell)}\in \cV_\txt^{(\ell)}]$, where $v^{(\ell)}\in \cV_\txt^{(\ell)}$ means $v$ is the rightmost children of one of $u\in \cV_\txt^{(\ell)}$. 
Starting with $h^{(L)}_{v}=x_{\txt,v}\ (v\in \cV^{(L)}_\txt)$, the downsampling process is equivalently written as, for $v\in \cV^{(L)}_\txt$, 
\begin{align}
    \begin{aligned}
    \textstyle q_{v}^{(\ell)} &\textstyle= f_{\downarrow,\iota(\pa^{(L-\ell)}(v))}^{(\ell)}(h^{(\ell)}_{v}) \in \R^S,
    && v\in \cV^{(L)}_\txt,\ \ell = L,\dots,1
   \\ 
    \textstyle g^{(\ell)}_{v} &\textstyle = 
    \frac{1}{a_v^{(\ell)}}\sum_{{v'}^{(L-\ell)}\in \cC(\pa^{(L-\ell+1)}(v))}
    \indicator [v'\leq v] q_{{v'}}^{(\ell)} \in \R^S,
     && v\in \cV^{(L)}_\txt,\ \ell = L,\dots,1,
\\ 
    \textstyle h^{(\ell-1)}_{v} &\textstyle = 
    \normalize \big(a_v^{(\ell)} g^{(\ell)}_{v} +q_v^{(\ell)} - \indicator[v^{(\ell)}\in \cV_\txt^{(\ell)}]q_v^{(\ell)}\big) \in \R^S,
     && v\in \cV^{(L)}_\txt,\ \ell = L,\dots,1.
       \end{aligned}\label{eq:Appendix-NextWord-MP-10}
    \end{align}
    Computation of $q_{v}^{(\ell)}$, $g^{(\ell)}_{v}$, and $h^{(\ell-1)}_{v}$ from $h^{(\ell)}_{v}$ is called the $\ell$-th step of the downsampling process (of the text part). 
\paragraph{Upsampling (alternative).} 
    The upsampling \eqref{eq:Appendix-NextWord-MP-3} is equivalently written as, for $v\in \cV^{(L)}_\txt$, 
    \begin{align}
    \begin{aligned}
        b^{(1)}_{v} & \textstyle  = \normalize( h^{(0)}_{v} + (\log \P[s|\xi])_{s\in [S]} -q_{v}^{(1)})\in \R^S,\\
        b^{(\ell+1)}_{v} & \textstyle = 
        \begin{cases}
            \normalize( f^{(\ell)}_{\uparrow, \iota(\pa^{L-\ell}(v+1))}(b^{(\ell)}_{v})+  h^{(\ell)}_{v} - q^{(\ell+1)}_{v}),
            & \text{(if $\pa^{(L-\ell-1)}(v) = \pa^{(L-\ell-1)}(v+1)$)}
            \\
           \normalize( f^{(\ell)}_{\uparrow, \iota(\pa^{L-\ell}(v+1))}(b^{(\ell)}_{v})+ h^{(\ell)}_{v}) ,
            & 
            \begin{pmatrix}
            \text{if $\pa^{(L-\ell)}(v) = \pa^{(L-\ell)}(v+1)$}\\
            \text{but $\pa^{(L-\ell-1)}(v) \ne \pa^{(L-\ell-1)}(v+1)$}
            \end{pmatrix}  
            \\
            \normalize( f^{(\ell)}_{\uparrow, \iota(\pa^{(L-\ell)}(v+1))}(b^{(\ell)}_{v})),
            & \text{(otherwise)}
        \end{cases} \\ &\quad\quad\quad\quad\quad\quad\quad\quad\quad\quad\quad\quad\quad\quad\quad\quad\quad\quad\quad\quad\quad\quad\quad\quad\quad\quad\quad\quad\quad\quad\quad\quad\quad\quad \ell = 1,2,\dots,L.
    \end{aligned}
   \label{eq:Appendix-NextWord-MP-5}
    \end{align}
    so that $ \bar{b}^{(L)}_{v} = b^{(L+1)}_v$.
    Computation of $b_{\image,v}^{(\ell)}$ is called the $\ell$-th step of the upsampling process (of the text part). 

\subsubsection{Approximation with transformer networks}\label{subsubsection:NWP-Intro-NN}

We approximate the message passing algorithm with transformer networks. 
We denote a transformer approximation of $h^{(0)}_{\image,\root}=(\log \P[s|\xi])_{s\in [S]}$ by $\hnetwork_{\image,d_\image}^{(0)}\in \R^S$.
This can be obtained by $\NN_\image^{\bW_\image}$ constructed in contrastive learning, or we can assume this as a given variable in the two-stage training.
From now we focus on the text model. 
We use the numbering of nodes defined in \Cref{definition:Numbering}.

Let $\hnetwork^{(L)}_v=x_{\txt,v}\in [S]$ for all $v\in \cV_\txt^{(L)}$. 
After the encoding $\encode_\vlm$, we obtain a matrix $\hmatrix^{(L)}$ such that
\begin{align}
    \hmatrix^{(L)} = 
\encode_\vlm(\xt,\what\Ei(\xi))
    =
 \begin{bmatrix}
           &&&\boldzero&
            \\
         \hnetwork^{(L)}_0  &\hnetwork^{(L)}_1 & \hnetwork^{(L)}_2 & \cdots & \hnetwork^{(L)}_{d}
            \\
   \hnetwork_{\image,d_\image}^{(0)} &\hnetwork_{\image,d_\image}^{(0)}&  \hnetwork_{\image,d_\image}^{(0)}&\cdots & \hnetwork_{\image,d_\image}^{(0)}
      \\      &&&\pomatrix&
   \end{bmatrix}\in \R^{(d_\feature+d_\positional)\times (d+1)}.
\end{align}
The shape of $\hmatrix^{(L)}$ is $(d_\feature+d_\positional)\times (d+1)$, where
$d_\feature=(4L+2)\staten+1$ and $d_\positional=2L+2$. 
As previously, $\pomatrix\in \R^{d_\positional\times (d+1)}$ is a matrix that encodes the positions of the nodes in $d_\positional$-dimensional space, and $d_\feature$ is the dimensions for the intermediate variables.
The output of the image model $\hnetwork_{\image,d_\image}^{(0)}$ is concatenated with every token.
We added the leftmost column, which is treated as the variables corresponding to the token position $0$, and let $\hnetwork^{(L)}_0=0$.

The text network has $(2L+1)$ transformer blocks, and each transformer block consists of feed forward layer $\FF_{1}$, masked self-attention $\MAttn$ instead of the previous $\Attn$, feed forward layer $\FF_{2}$, and normalization. 
Using two feed forward layers in a single block is for the sake of clarity in the proof, and  it can simply be split into two separate blocks with one feed forward layer, if this is to be avoided.
The first $L$ blocks approximate downsampling, and each block is called the $\ell(=L,\dots,1)$-th block of downsampling using the decreasing order.
The latter $(L+1)$-blocks approximate upsampling, and each block is called the $\ell(=0,\dots,L)$-th block of upsampling using the increasing order.

First consider downsampling. 
Starting from $\hmatrix^{(L)}$, we will construct matrices $\hmatrix^{(\ell)}\ (\ell=L,\cdots,0),\ \qmatrix^{(\ell)}\ (\ell=L,\cdots,1),\ \gmatrix^{(\ell)}\ (\ell=L,\cdots,1)$ of shape $(d_\feature+d_\positional)\times (d+1)$, defined as
\begin{align}
  \hmatrix^{(\ell)} &= 
        \begin{bmatrix}
        &&\boldzero&
        \\
          \hnetwork^{(\ell)}_0 &  \hnetwork^{(\ell)}_1 & \
             \cdots
            &
            \hnetwork^{(\ell)}_d
            \\
         \vdots &   \vdots & 
             \ddots
            &
            \vdots
            \\
          \gnetwork^{(L)}_0 &  \gnetwork^{(L)}_1 & 
             \cdots
            &
            \gnetwork^{(L)}_d
            \\
          \qnetwork^{(L)}_0 &  \qnetwork^{(L)}_1 & 
             \cdots
            &
            \qnetwork^{(L)}_d
            \\
           \hnetwork^{(L)}_0 &  \hnetwork^{(L)}_1 & 
             \cdots
            &
            \hnetwork^{(L)}_d
            \\
        \hnetwork_{\image,d_\image}^{(0)}&  \hnetwork_{\image,d_\image}^{(0)}&\cdots & \hnetwork_{\image,d_\image}^{(0)}
            \\
            &&\pomatrix&
        \end{bmatrix},
        \
        \qmatrix^{(\ell)} = 
        \begin{bmatrix}
        &&\boldzero&
        \\
            \qnetwork^{(\ell)}_0 & \qnetwork^{(\ell)}_1 & 
             \cdots
            &
            \qnetwork^{(\ell)}_d
            \\
            \vdots & \vdots & 
             \ddots
            &
            \vdots
             \\
          \gnetwork^{(L)}_0 &  \gnetwork^{(L)}_1 & 
             \cdots
            &
            \gnetwork^{(L)}_d
            \\
            \qnetwork^{(L)}_0 & \qnetwork^{(L)}_1 & 
             \cdots
            &
            \qnetwork^{(L)}_d
            \\
            \hnetwork^{(L)}_0 & \hnetwork^{(L)}_1 & 
             \cdots
            &
            \hnetwork^{(L)}_d
            \\
         \hnetwork_{\image,d_\image}^{(0)}&  \hnetwork_{\image,d_\image}^{(0)}&\cdots & \hnetwork_{\image,d_\image}^{(0)}
            \\
            &&\pomatrix&
        \end{bmatrix}
        ,\\
        \gmatrix^{(\ell)} &= 
        \begin{bmatrix}
        &&\boldzero&
        \\
            \gnetwork^{(\ell)}_0 & \gnetwork^{(\ell)}_1 & 
             \cdots
            &
            \gnetwork^{(\ell)}_d
            \\
            \vdots & \vdots & 
             \ddots
            &
            \vdots
             \\
          \gnetwork^{(L)}_0 &  \gnetwork^{(L)}_1 & 
             \cdots
            &
            \gnetwork^{(L)}_d
            \\
            \qnetwork^{(L)}_0 & \qnetwork^{(L)}_1 & 
             \cdots
            &
            \qnetwork^{(L)}_d
            \\
            \hnetwork^{(L)}_0 & \hnetwork^{(L)}_1 & 
             \cdots
            &
            \hnetwork^{(L)}_d
            \\
         \hnetwork_{\image,d_\image}^{(0)}&  \hnetwork_{\image,d_\image}^{(0)}&\cdots & \hnetwork_{\image,d_\image}^{(0)}
            \\
            &&\pomatrix&
        \end{bmatrix}
        .
\end{align}
Here, $\hnetwork^{(\ell)},\qnetwork^{(\ell)}$ and $ \gnetwork^{(\ell)}$ are $\staten$-dimensional vectors except that $\hnetwork^{(L)}\in [\staten]$.

In the $\ell$-th block of downsampling, the feed forward layer $\FF^{(\ell)}_{\downarrow,1}$, a fully-connected ReLU network, receives $\hmatrix^{(\ell)}$ and outputs $\qmatrix^{(\ell)}$ by computing $\qnetwork^{(\ell)}_v$ from $\hnetwork^{(\ell)}_v$:
    \begin{align}
        \qmatrix^{(\ell)} = 
        \underbrace{\hmatrix^{(\ell)}}_{\text{skip connection}}
        + 
        \feedforward^{(\ell)}_{\downarrow,1} (\hmatrix^{(\ell)})
        =
        \hmatrix^{(\ell)}
        +
        \begin{bmatrix}
        \multicolumn{4}{l}{ \boldzero\ (\in \R^{((3\ell+L)S) \times (d+1)})} \\ 
       \qnetwork^{(\ell)}_0 & \qnetwork^{(\ell)}_1 &  \cdots & \qnetwork^{(\ell)}_d \\
        \multicolumn{4}{l}{ \boldzero \ (\in \R^{
        (d_{\positional}+(3L-3\ell+1)S+1) \times (d+1)} )}
        \end{bmatrix}
        ,
    \end{align}
    Then, the masked self-attention block $\MAttn^{(\ell)}$ constructs $\gmatrix^{(\ell-1)}$ by computing $\gnetwork^{(\ell)}_v$ from $\qnetwork_v^{(\ell)}$ as
    \begin{align}
        \gmatrix^{(\ell-1)} = 
        \underbrace{\qmatrix^{(\ell)}}_{\text{skip connection}}
        +\MAttn^{(\ell)}(\qmatrix^{(\ell)})
        =
        \qmatrix^{(\ell)}
        + 
        \begin{bmatrix}
        \multicolumn{4}{l}{ \boldzero\ (\in \R^{((3\ell+L-1)S) \times (d+1)})} \\ 
       \gnetwork^{(\ell)}_0 & \gnetwork^{(\ell)}_1 &  \cdots & \gnetwork^{(\ell)}_d \\
        \multicolumn{4}{l}{ \boldzero \ (\in \R^{
        (d_{\positional}+(3L-3\ell+2)S+1) \times (d+1)} )}
        \end{bmatrix}
        .
    \end{align} 
    Finally, the second feed forward layer $\FF_{\downarrow,2}^{(\ell)}$ constructs $\hmatrix^{(\ell-1)}_v$, using $\gnetwork_v^{(\ell)}$ and $\qnetwork_v^{(\ell)}$.
        \begin{align}
        \hmatrix^{(\ell-1)} = 
        \normalize\Big(\underbrace{\gmatrix^{(\ell)}}_{\text{skip connection}}
        +\FF_{\downarrow,2}^{(\ell)}(\gmatrix^{(\ell)})
        \Big)
        =
        \normalize\Bigg(
        \gmatrix^{(\ell)}
        + 
        \begin{bmatrix}
            \boldzero & (\in \R^{((3\ell+L-2)S) \times (d+1)})\quad \ \ \ 
            \\
            {\star} &(\in \R^{S \times d})\quad \quad \quad \quad  \quad \quad
            \\
            \boldzero &(\in \R^{(d_\positional+(3L-3\ell+3)S+1) \times (d+1)})
        \end{bmatrix}
        \Bigg)
        .
    \end{align} 
    Here $\star$ means $[\hnetwork_0^{(\ell-1)}\ \hnetwork_1^{(\ell-1)}\ \cdots\  \hnetwork_d^{(\ell-1)}]$ before normalization. 

We then consider upsampling. 
After we obtain $\hmatrix^{(0)}$, we iteratively compute $\bbmatrix^{(\ell)}\ (\ell=1,\dots,L+1)$:
\begin{align}
    \bbmatrix^{(\ell)} &= 
        \begin{bmatrix}
        &&\boldzero&\\
            \bnetwork^{(\ell)}_{0} & \bnetwork^{(\ell)}_{1} & 
             \cdots
            &
            \bnetwork^{(\ell)}_d
           \\
            \vdots & \vdots & 
             \ddots
            &
            \vdots
        \\
            \bnetwork^{(1)}_{0} & \bnetwork^{(1)}_{1} & 
             \cdots
            &
            \bnetwork^{(1)}_d
        \\
            \hnetwork^{(0)}_{0} & \hnetwork^{(0)}_{1} & 
             \cdots
            &
            \hnetwork^{(0)}_d
        \\
            \qnetwork^{(1)}_0 & \qnetwork^{(1)}_1 & 
             \cdots
            &
            \qnetwork^{(1)}_d
            \\
            \vdots & \vdots & 
             \ddots
            &
            \vdots
            \\
            \qnetwork^{(L)}_0 & \qnetwork^{(L)}_1 & 
             \cdots
            &
            \qnetwork^{(L)}_d
            \\
            \hnetwork^{(L)}_0 & \hnetwork^{(L)}_1 & 
             \cdots
            &
            \hnetwork^{(L)}_d
            \\
         \hnetwork_{\image,d_\image}^{(0)}&  \hnetwork_{\image,d_\image}^{(0)}&\cdots & \hnetwork_{\image,d_\image}^{(0)}
            \\
            &&\pomatrix&
        \end{bmatrix}.
\end{align}
Here, $\bnetwork^{(\ell)}$ are $\staten$-dimensional real-valued vectors.
The $\ell$-th block of downsampling computes $\bbmatrix^{(\ell+1)}$ using a feed forward network $\feedforward^{(\ell)}_\uparrow$ with normalization:
 \begin{align}
        \bbmatrix^{(\ell+1)} = 
         \normalize\Big(\underbrace{\bbmatrix^{(\ell)}}_{\text{skip connection}}
        + 
        \feedforward^{(\ell)}_\uparrow (\bbmatrix^{(\ell)})\Big)
        =
        \bbmatrix^{(\ell)}
        +
        \begin{bmatrix}
        \multicolumn{4}{l}{ \boldzero \ (\in \R^{((L-\ell)S) \times (d+1)})} \\ 
       \bnetwork^{(\ell+1)}_0 & \bnetwork^{(\ell+1)}_1 &  \cdots & \bnetwork^{(\ell+1)}_d \\
        \multicolumn{4}{l}{\boldzero \ (\in \R^{(d_\positional+(3L+\ell+1)\staten+1) \times (d+1)} }
        \end{bmatrix}
        . 
    \end{align}
       For $\ell=0$, replace $\bbmatrix^{(0)}$ by $\hmatrix^{(0)}$. 
   This is the same as \eqref{eq:CDM-Bmatrix-1} (except for difference in the column dimension).
   We do not need the self-attention layer and second feed forward layer, and we can ignore them by simply setting the weight matrices to zeros.

Finally, we obtain $\bnetwork^{(L+1)}_{v}$ for all $v=1,\dots,d-1$.
The readout layer $\read_\vlm$ computes $\softmax(\bnetwork^{(L+1)}_{v})$, 
which approximates $\mu_{\star}(x_{\txt,v+1}|\xi,x_{\txt,1},\dots,x_{\txt,v})$, for all $v=1,\dots,d-1$.  

In the following, our goal is to iteratively show that
\begin{align}
    &\hnetwork_v^{(\ell)}\approx h_v^{(\ell)},
    \quad \qnetwork_v^{(\ell)}\approx q_v^{(\ell)}, \quad 
    \gnetwork_v^{(\ell)}\approx g_v^{(\ell)}, \quad 
    \bnetwork_v^{(\ell)}\approx b_v^{(\ell)},
\end{align}
 ($\ell=L,\dots,0$ for $\hnetwork_v^{(\ell)}$, $\ell=L,\dots,1$ for $\qnetwork_v^{(\ell)}$ and $\gnetwork_v^{(\ell)}$, and $\ell=1,\dots,L+1$ for $\bnetwork_v^{(\ell)}$)
for all $v\in \cV^{(L)}_\txt$. 
For $v=0$, we will iteratively fill the zero vectors for all $\hnetwork_0^{(\ell)}, \qnetwork_0^{(\ell)},\gnetwork^{(\ell)}_0,$ and $\bnetwork_0^{(\ell)}$.

We now formally define each component of the pipeline.
\paragraph{Encoding $\encode_\vlm$.}
Denote the $v$-th column of $\pomatrix$ by $\pnetwork_v$. 
For $v\in \cV^{(L)}_\txt$, We define $\pnetwork_v\in \R^{2L+2}$ as
\begin{align}
\textstyle
    &\pnetwork_v = 
    \\ &\textstyle\Big[0, 1, \sin \big( \frac{2\pi\iota(v)}{m^{(L)}}\big), \cos \big(\frac{2\pi\iota(v)}{m^{(L)}}\big),\sin \big( \frac{2\pi\iota(\pa(v))}{m^{(L-1)}}\big), \cos \big( \frac{2\pi\iota(\pa(v))}{m^{(L-1)}}\big), \cdots  , \sin \big(\frac{2\pi\iota(\pa^{L-1}(v))}{m^{(1)}}\big), \cos \big( \frac{2\pi\iota(\pa^{L-1}(v))}{m^{(1)}}\big)\Big]^\top.
    \label{eq:positional_encoding-3}
\end{align}
The difference from the contrastive learning \eqref{eq:positional_encoding} and conditional diffusion model \eqref{eq:positional_encoding-2} is that we added $0$ and $1$ to the first two dimensions. 
For $v=0$, we define
\begin{align}
    \pnetwork_0 = \Big[1, 0, \bzero \Big]^\top \in \R^{2L+2} 
\end{align}
so that the first two dimensions are orthogonal to \eqref{eq:positional_encoding-3}. 

For two-stage training, where $\what\Ei(\xi)$ approximates $h^{(0)}_{\image,\root}=(\log \P[s|\xi])_{s\in [S]}$, we define $\hnetwork^{(0)}_{\image,d_\image}$ as $\hnetwork^{(0)}_{\image,d_\image}=(\log \readimage(\what\Ei(\xi)_s))_{s\in [S]}$. 

\paragraph{Downsampling: position-wise feed forward layers.}
The first feed forward layer $\FF^{(\ell)}_{\downarrow,1}$ of the $\ell$-th block $(\ell=L,\dots,1)$ approximates each $f_{\downarrow,\iota(\pa^{(L-\ell)}(v))}^{(\ell)}$.
Therefore, the feed forward block at the $\ell$th layer yields
\begin{align}
    \begin{aligned}
    \textstyle \qnetwork_{v}^{(\ell)} &\textstyle= \fnetwork_{\downarrow,\iota(\pa^{(L-\ell)}(v))}^{(\ell)}(\hnetwork^{(\ell)}_{v}) \in \R^S,
    && v\in \cV^{(L)}_\txt,\ \ell = L,\dots,1.
       \end{aligned}
    \end{align}
When $\hnetwork^{(\ell)}_{v}\approx h_{\pa^{(L-\ell)}(v)}^{(\ell)}$ for $v\in \cV^{(L)}$ and $\fnetwork_{\iota}^{(\ell)}\approx f_\iota^{(\ell)}$, we have $\qnetwork^{(\ell)}_{v}\approx q_{\pa^{(L-\ell)}(v)}^{(\ell)}$. 
Following the notation in \Cref{definition:FullyConnected}, we state the following approximation error guarantee. 
\begin{lemma}[Approximation error of the first feed forward layer]\label{lemma:Approximation-FF-NWP-1}
Fix $\ell\in [L]$ and $\delta>0$.
Assume that $\transbd^{-1} \leq \psi^{(\ell)}_{\iota}(s,a) \leq \transbd$ for all $s,a\in [\staten]$.
When $\ell=1$, also assume that $\transbd^{-1}\leq \P[s]\leq \transbd$ for all $s$. 
Then, there exists an $\NN\in \mathcal{F}(\depth,\widthvector,B)$ such that
\begin{align}
    \|\NN([h;\pnetwork_v]) - f^{(\ell)}_{\iota(\pa^{(L-\ell)}(v))}(h)\|_\infty \leq \delta, \quad v\in \cV^{(L)},
\end{align}
for all $h\in \R^S$ with $\max_s h_s = 0$ ($\ell\leq L-1$) or $h\in [S]$ ($\ell=L$).
The network parameters $\depth,\widthvector$ and $B$ are bounded as follows:
\begin{align}
    \depth \lesssim (\log\log(S\transbd/\delta))\log(S\transbd/\delta),\
    \|\widthvector\|_\infty  \lesssim m^{(\ell)}S(\log(S\transbd/\delta))^3,
    \
    B  \lesssim 2S(\transbd^2+\log(S\transbd/\delta)) + (m^{(\ell)})^2.
\end{align}
\end{lemma}
The only difference from  \Cref{lemma:Approximation-FF} is the dimension of $\pnetwork_v$, and thus we omit the proof. 

We then consider the second feed forward layer $\FF_{\downarrow,2}^{(\ell)}$. 
The role of this layer is to compute
\begin{align}
    a_v^{(\ell)} \gnetwork^{(\ell)}_{v} +\qnetwork_v^{(\ell)} - \indicator[v^{(\ell)}\in \cV_\txt^{(\ell)}]\qnetwork_v^{(\ell)},
\end{align}
and the following lemma shows that this computation can be done exactly.
\begin{lemma}[Approximation error of the second feed forward layer]\label{lemma:Approximation-FF-NWP-2}
Fix $\ell\in [L]$. 
There exists an $\NN\in \mathcal{F}(\depth,\widthvector,B)$ such that
\begin{align}
    \NN([g,q;\pnetwork_v]) = a_v^{(\ell)}g + q- \indicator[v^{(\ell)}\in \cV_\txt^{(\ell)}]q, \quad v\in \cV^{(L)},
\end{align}
for all $g,q\in \R^S$ with $\|g\|_\infty , \|q\|_\infty\leq C$.
The network parameters $\depth,\widthvector$ and $B$ are bounded as follows:
\begin{align}
        J_1 \lesssim 1,\quad \|\widthvector_1\|_\infty \lesssim S+d_\positional, 
        \quad B_1\lesssim L+\max_{\ell+1\leq k\leq L}(m^{(k)})^{2}+m^{(\ell)}C.
    \end{align}
\end{lemma}
The proof of this lemma is found in \Cref{subsection:NWP-Position-wiseFeedForward}.
\paragraph{Downsampling: masked self-attention layer.}
To obtain $\gmatrix^{(\ell-1)}$ from $\qmatrix^{(\ell)}$, we use the causal mask and multi-head attention. 
Let $k$ be a sequence length.
The causal mask $\mask_k$ is defined as
\begin{align}
    \mask_k = 
       \begin{bmatrix}
0 & 0 & 0 & \cdots & 0 \\
-C & 0 & 0 & \cdots & 0 \\
-C & -C & 0 & \cdots & 0 \\
\vdots & \vdots & \vdots & \ddots & \vdots \\
-C & -C & -C & \cdots & 0
\end{bmatrix}
\in \R^{k\times k},
\end{align}
where $C$ is a sufficiently large constant (so that $(i,j)$-element of $\softmax(\mask_k + ((\bW_{K,m}\ \cdot)^\top(\bW_{Q,m} \ \cdot)))$ is ignored in the following for $i>j$.)
Then a masked self-attention layer is defined as
\begin{align}
   \MAttn(\cdot) = 
  (\bW_{V}\ \cdot)\ \softmax(\mask_k + ((\bW_{K}\ \cdot)^\top(\bW_{Q} \ \cdot))) ,
\end{align}
where $\mask_k$ is added in an element-wise manner and $\softmax$ is applied column-wise.

\begin{definition}[A class of masked multi-head self-attention blocks]\label{definition:Self-MAttn}
    We define a class of masked self-attention blocks with as
\begin{align}
      \bar{\cA}(D,B) =  &\Big\{
        \textstyle
    (\bW_{V}\ \cdot)\ \softmax(\mask_d + ((\bW_{K}\ \cdot)^\top(\bW_{Q} \ \cdot)))\ \Big|\\ & \textstyle 
      \bW_{K}, \bW_{Q}, \bW_{V}\in \R^{d\times d},\
       \max_{i,j}|(\bW_{K})_{i,j}|, \max_{i,j}|(\bW_{Q})_{i,j}|, \max_{i,j}|(\bW_{V})_{i,j}|
       \leq B\Big\}.
    \end{align}
\end{definition}
We will construct weight matrices so that 
the self-attention layer $\MAttn^{(\ell)}$ of the $\ell$-th block $(\ell=L,L-1,\dots,1)$ yields
\begin{align}
\textstyle
\gnetwork^{(\ell-1)}_{v} &\textstyle = \frac{1}{a_v^{(\ell)}}\sum_{{u}^{(\ell)}\in \cC(\pa^{(L-\ell+1)}(v))}\indicator [u\leq v] \qnetwork_{u}^{(\ell)}+\bdelta^{(\ell)}_v\in \R^S
   \label{eq:NWP-self-attention-1}
\end{align}
with $\|\bdelta^{(\ell)}_v\|_\infty\ll 1$. 

The approximation error guarantee is stated as follows. 
\begin{lemma}[Approximation error of self-attention layer]\label{lemma:Self-attention-NWP}
For $\ell\in [L]$, there exists $\MAttn\in \bar{\cA}(D,B)$ with $D=d_\feature+d_\positional$ and $B\lesssim \log(d\delta^{-1})+ m^{(\ell)}$ such that
    \begin{align}
    &\MAttn(\qmatrix^{(\ell)})_v
   \\ & =
   \begin{cases}
  \begin{bmatrix}
            \boldzero&  (\in \R^{(3\ell+L-1)S})\quad \quad\quad 
            \\
             \frac{1}{a_v^{(\ell)}}\Big(\qnetwork^{(\ell)}_{0}+\sum_{{u}^{(\ell)}\in \cC(\pa^{(L-\ell+1)}(v))}\indicator [u\leq v]\qnetwork^{(\ell)}_{u} \Big)+ \bdelta^{(\ell)}_v & (\in \R^{S})\ \quad \quad \quad \quad\quad \quad
            \\
            \boldzero &(\in \R^{d_\positional+(3L-3\ell+2)S+1})
        \end{bmatrix}, & (v\in \cV^{(L)}_\txt)
        \\
        \begin{bmatrix}
        \bzero &(\in \R^{(3\ell+L-1)S})\quad \quad\quad 
        \\
        \qnetwork^{(\ell)}_0 &(\in \R^{S})\ \quad \quad \quad \quad\quad  \quad 
        \\
        \bzero &(\in \R^{d_\positional+(3L-3\ell+2)S+1})
        \end{bmatrix}.
         & (v=0)
    \end{cases}
    \end{align} 
    where $\bdelta_{v}^{(\ell)}\in \R^S$ satisfies $\|\bdelta_{v}^{(\ell)}\|_\infty \leq \delta   \max_{v'}\|\qnetwork^{(\ell)}_{v'}\|_\infty$.

\end{lemma}
Because we can iteratively see that $\qnetwork^{(\ell)}_0=0$, 
the column corresponding to $v\in \cV^{(L)}_\txt$ is $\frac{1}{a_v^{(\ell)}}\sum_{{u}^{(\ell)}\in \cN(\pa^{(L-\ell)}(v))}$ $\indicator [u\leq v]\qnetwork^{(\ell)}_{u} + \bdelta^{(\ell)}_v$, and the column corresponding to $v=0$ is exactly $\bzero\in \R^D$. 
When $\qnetwork^{(\ell)}_v \approx q^{(\ell)}_v$ and $\|\bdelta_v^{(\ell)}\|_\infty \ll 1$, we have $\gnetwork^{(\ell)}_v \approx g^{(\ell)}_v$.
The proof of this lemma can be found in \Cref{subsection:Selt-attention-NWP}. 

\paragraph{Upsampling: position-wise feed forward block.}
The upsampling constructs estimation of leaf nodes starting from the root.
The $(L+1)$ blocks of attention blocks (feed forward layers) can implement this process.
The $\ell$-th block of upsampling computes $\bnetwork^{(\ell+1)}_v$ from $\hnetwork^{(\ell)}_v$, $\qnetwork^{(\ell+1)}_v$, and $\bnetwork^{(\ell)}_v$.
    \begin{align}
    \begin{aligned}
        \bnetwork^{(1)}_{v} & \textstyle  =  \normalize(\hnetwork^{(0)}_{v} + \hnetwork_{\image,d_\image}^{(0)} -\qnetwork_{v}^{(1)})\in \R^S,\\
        \bnetwork^{(\ell+1)}_{v} & \textstyle = 
        \begin{cases}
            \normalize (\fnetwork^{(\ell)}_{\uparrow, \iota(\pa^{L-\ell}(v+1))}(\bnetwork^{(\ell)}_{v})+  \hnetwork^{(\ell)}_{v} - \qnetwork^{(\ell+1)}_{v}),
            & \text{(if $\pa^{(L-\ell-1)}(v) = \pa^{(L-\ell-1)}(v+1)$)}
            \\
           \normalize( \fnetwork^{(\ell)}_{\uparrow, \iota(\pa^{L-\ell}(v+1))}(\bnetwork^{(\ell)}_{v})+ \hnetwork^{(\ell)}_{v}) ,
            & 
            \begin{pmatrix}
            \text{if $\pa^{(L-\ell)}(v) = \pa^{(L-\ell)}(v+1)$}\\
            \text{but $\pa^{(L-\ell-1)}(v) \ne \pa^{(L-\ell-1)}(v+1)$}
            \end{pmatrix}  
            \\
            \normalize( \fnetwork^{(\ell)}_{\uparrow, \iota(\pa^{(L-\ell)}(v+1))}(\bnetwork^{(\ell)}_{v})),
            & \text{(otherwise)}
        \end{cases} \\ &\quad\quad\quad\quad\quad\quad\quad\quad\quad\quad\quad\quad\quad\quad\quad\quad\quad\quad\quad\quad\quad\quad\quad\quad\quad\quad\quad\quad\quad\quad\quad\quad\quad\quad \ell = 1,2,\dots,L.
    \end{aligned}
   \label{eq:Appendix-NextWord-MP-15}
    \end{align}
For each update, we can track the correspondence with the message passing algorithm.
\begin{lemma}[Approximation error of feed forward layer, upsampling]\label{lemma:Approximation-FF-VLM-UP}
Fix $\ell\in \{0,\dots,L\}$ and $\delta>0$.
Assume that $\transbd^{-1} \leq \psi^{(\ell)}_{\iota}(s,a) \leq \transbd$ for all $s,a\in [\staten]$.
Then, there exist $\NN_1,\NN_2\in \mathcal{F}(\depth,\widthvector,B)$ such that
\begin{align}
&\NN_1([h;h';q]) = h+h'-q, 
\\
   & 
        \begin{cases}
         \|  \NN_2([b;h;q]) -  (\fnetwork^{(\ell)}_{\uparrow, \iota(\pa^{L-\ell}(v+1))}(\bnetwork^{(\ell)}_{v})+  \hnetwork^{(\ell)}_{v} - \qnetwork^{(\ell+1)}_{v})\|_\infty \leq \delta,
            & \text{(if $\pa^{(L-\ell-1)}(v) = \pa^{(L-\ell-1)}(v+1)$)}
            \\
 \|  \NN_2 ([b;h;q]) - (\fnetwork^{(\ell)}_{\uparrow, \iota(\pa^{L-\ell}(v+1))}(\bnetwork^{(\ell)}_{v})+ \hnetwork^{(\ell)}_{v}) \|_\infty \leq \delta,
            & 
            \begin{pmatrix}
            \text{if $\pa^{(L-\ell)}(v) = \pa^{(L-\ell)}(v+1)$}\\
            \text{but $\pa^{(L-\ell-1)}(v) \ne \pa^{(L-\ell-1)}(v+1)$}
            \end{pmatrix}  
            \\
        \|  \NN_2 ([b;h;q]) - (\fnetwork^{(\ell)}_{\uparrow, \iota(\pa^{(L-\ell)}(v+1))}(\bnetwork^{(\ell)}_{v}))\|_\infty \leq \delta,
            & \text{(otherwise)}
        \end{cases} \\ &\quad\quad\quad\quad\quad\quad\quad\quad\quad\quad\quad\quad\quad\quad\quad\quad\quad\quad\quad\quad\quad\quad\quad\quad\quad\quad\quad\quad\quad\quad\quad\quad\quad\quad \ell = 1,2,\dots,L.
\end{align}
for all $h,h',q,b\in \R^S$ with $\max_s b_s = 0$.
For all of these networks, the parameters $\depth,\widthvector$ and $B$ are bounded as follows:
\begin{align}
    \depth &\lesssim (\log\log(S\transbd/\delta))\log(S\transbd/\delta),\\
    \|\widthvector\|_\infty  &\lesssim m^{(\ell)}S(\log(S\transbd/\delta))^3+ d_\positional,
    \\
    B  &\lesssim S(\transbd^2+\log(S\transbd/\delta)) + \max_{\ell\leq k\leq L}(m^{(k)})^{2}+L+C.
\end{align}
\end{lemma}
The proof will be placed in \Cref{subsection:NWP-Position-wiseFeedForward}. 

\paragraph{Normalization.}
Since each column vector of $ \hmatrix^{(\ell)} $, $ \qmatrix^{(\ell)} $, and $ \bbmatrix^{(\ell)}$ is a collection of multiple $ \hnetwork^{(\ell)}_v $, $ \qnetwork^{(\ell)}_v $, and $\bnetwork^{(\ell)}_v$, we adopt a slightly different definition of $\normalize$ than that used in message passing.
Specifically, for $\xnetwork =[\bnetwork^{(L+1)}\ \dots\ \bnetwork^{(1)}\ \hnetwork^{(0)}\ \gnetwork^{(0)}\ \qnetwork^{(1)}\ \hnetwork^{(1)}\ \dots\ \gnetwork^{(L)}\ \qnetwork^{(L)}\ \hnetwork^{(L)}\ \hnetwork \ \pnetwork]\in \R^{d_\feature+d_\positional}$ with $\hnetwork^{(L)}\in [S], \hnetwork^{(\ell)}\in \R^S\ (\ell=L-1,\dots,0), \qnetwork^{(\ell)}, \gnetwork^{(\ell)},\bnetwork^{(\ell)},\in \R^S$, we define
we define
\begin{align}
    \normalize(\xnetwork)
    = 
    \begin{bmatrix}
    \bnetwork^{(L+1)}- \ones_S\max_{s\in S}\qnetwork^{(L+1)}_s\\
    \vdots
    \\
    \bnetwork^{(1)}- \ones_S\max_{s\in S}\qnetwork^{(1)}_s
    \\
        \hnetwork^{(0)} \\ \gnetwork^{(1)}- \ones_S\max_{s\in S}\gnetwork^{(1)}_s \\ 
        \qnetwork^{(1)} \\
        \hnetwork^{(1)} \\ \vdots \\ \gnetwork^{(L)} - \ones_S\max_{s\in S} \gnetwork^{(L)}\\ 
        \qnetwork^{(L)} \\
        \hnetwork^{(L)} \\ \hnetwork \\ \pnetwork
    \end{bmatrix}\in \R^{d_\feature+d_\positional},\quad \ones_S = \begin{bmatrix}
        1 \\ 1 \\ \vdots \\ 1
    \end{bmatrix}\in \R^S.
\end{align}
For a matrix in $\R^{(d_\feature+d_\positional)\times (d+1)}$, it is applied in a column-wise manner.
\paragraph{Readout layer $\read_\vlm$}
Finally, the readout layer $\read_\vlm$ extracts $\bnetwork_v^{(L+1)}$ and apply an element-wise projection onto $[-\readbdvlmenc,\readbdvlmenc]$ and $\softmax.$
\begin{align}
  [\softmax(\proj_{[-\readbdvlmenc,\readbdvlmenc]^S}(\bnetwork_1^{(L+1)}))\ \cdots\ \softmax(\proj_{[-\readbdvlmenc,\readbdvlmenc]^S}(\bnetwork_d^{(L+1)}))],
  \label{eq:bdvlmenc}
\end{align}
where $ \proj_{[-\readbdvlmenc,\readbdvlmenc]^S}(x)=\argmin_{y\in[-\readbdvlmenc,\readbdvlmenc]^S}|x-y|$. 
According to \Cref{lemma:Boundedness-VLM}, setting $\readbdvlmenc\defn2\log(\state\transbd)$ allows us to ignore the effect of this truncation.
\paragraph{The whole pipeline.}
Putting it all together, the neural network approximate the message passing algorithm in the following way.
The downsampling process is approximated as
 \begin{align}
    \begin{aligned}
    \textstyle \qnetwork_{v}^{(\ell)} &\textstyle= \fnetwork_{\downarrow,\iota(\pa^{(L-\ell)}(v))}^{(\ell)}(\hnetwork^{(\ell)}_{v}) \in \R^S,
    && v\in \cV^{(L)}_\txt,\ \ell = L,\dots,1
   \\ 
    \textstyle \gnetwork^{(\ell)}_{v} &\textstyle = 
    \frac{1}{a_v^{(\ell)}}\sum_{{v'}^{(L-\ell)}\in \cC(\pa^{(L-\ell+1)}(v))}
    \indicator [v'\leq v] \qnetwork_{{v'}}^{(\ell)} \in \R^S,
     && v\in \cV^{(L)}_\txt,\ \ell = L,\dots,1,
\\ 
    \textstyle \hnetwork^{(\ell-1)}_{v} &\textstyle = 
    \normalize \big(a_v^{(\ell)} \gnetwork^{(\ell)}_{v} +\qnetwork_v^{(\ell)} - \indicator[v\in \cV_\txt^{(\ell)}]\qnetwork_v^{(\ell)}\big) \in \R^S,
     && v\in \cV^{(L)}_\txt,\ \ell = L,\dots,1.
       \end{aligned}\label{eq:Appendix-NWP-ErrorPropagation-9}
    \end{align}
    Let $\hnetwork_{\txt,d}^{(0)}\approx h_{\txt,\root}^{(0)}=(\log \P[s|\xi])_{s\in [S]}$. 
    The upsampling process is approximated as
    \begin{align}
    \begin{aligned}
        \bnetwork^{(1)}_{v} & \textstyle  = \normalize( \hnetwork^{(0)}_{v} +  \hnetwork_{\txt,d}^{(0)}-\qnetwork_{v}^{(1)})\in \R^S,\\
        \bnetwork^{(\ell+1)}_{v} & \textstyle = 
        \begin{cases}
            \normalize( \fnetwork^{(\ell)}_{\uparrow, \iota(\pa^{L-\ell}(v+1))}(\bnetwork^{(\ell)}_{v})+  \hnetwork^{(\ell)}_{v} - \qnetwork^{(\ell+1)}_{v}),
            & \text{(if $\pa^{(L-\ell-1)}(v) = \pa^{(L-\ell-1)}(v+1)$)}
            \\
           \normalize( \fnetwork^{(\ell)}_{\uparrow, \iota(\pa^{L-\ell}(v+1))}(\bnetwork^{(\ell)}_{v})+ \hnetwork^{(\ell)}_{v}) ,
            & 
            \begin{pmatrix}
            \text{if $\pa^{(L-\ell)}(v) = \pa^{(L-\ell)}(v+1)$}\\
            \text{but $\pa^{(L-\ell-1)}(v) \ne \pa^{(L-\ell-1)}(v+1)$}
            \end{pmatrix}  
            \\
            \normalize( \fnetwork^{(\ell)}_{\uparrow, \iota(\pa^{(L-\ell)}(v+1))}(\bnetwork^{(\ell)}_{v})),
            & \text{(otherwise)}
        \end{cases} \\ &\quad\quad\quad\quad\quad\quad\quad\quad\quad\quad\quad\quad\quad\quad\quad\quad\quad\quad\quad\quad\quad\quad\quad\quad\quad\quad\quad\quad\quad\quad\quad\quad\quad\quad \ell = 1,2,\dots,L.
    \end{aligned}
   \label{eq:Appendix-NWP-ErrorPropagation-10}
    \end{align}

   We summarize the network architecture (which slightly differs from the previous one) and the hypothesis class of $(\NN_\txt^{\bW_{\txt}},\Adapter)$ as follows.
    We focus on two step training, and the definition of the parameter space $\Parspacevlmone$ for joint training is introduced in \Cref{sec:joint_train_cdm}. 
    \begin{definition}[Eq.~(\ref{eqn:VLM_param_space}), restated]\label{definition:VLM_param_space}
    The image transformer network $\TF_{\vlm}$ has $L$ blocks of feed forward (\Cref{definition:FullyConnected}) with skip connection, masked self-attention (\Cref{definition:Self-MAttn}) with skip connection, feed forward (\Cref{definition:FullyConnected}) with skip connection, and normalization in this order.
    We say the collection of the parameters of  $(\TF_{\vlm},\Adapter)$ belongs to $\Parspacevlm$ if the following holds:
    In each block of the text transformer $\TF_{\vlm}$, its two feed forward layers $\FF_1,\FF_2$ and self-attention $\MAttn$ satisfy 
    \begin{align}
        &\FF\in \cF (\depth,\widthvector=(D,*,\cdots,*,D),B), \text{ with } \|\widthvector\|_\infty \leq D',
        \quad \MAttn\in \bar{\cA} (D,B).
    \end{align}
    Furthermore, the adapter satisfies 
    \begin{align}
    \adaweighta\in \R^{S\times M},\ \adaweightb \in \R^{M\times S}, \
        \lops{\adaweighta}\leq B,\ \lops{\adaweightb} \leq B.
    \end{align}
    \end{definition}
    
The rest of this section is organized as follows. 
\Cref{subsection:vlm_two_steps} discusses the two step training and proves \Cref{thm:vlm_two_steps}, using \Cref{lemma:Approximation-FF-NWP-1,lemma:Approximation-FF-NWP-2,lemma:Self-attention-NWP,lemma:Approximation-FF-VLM-UP}, as well as the bound on the propagation of the intermediate errors \Cref{lemma:Appendix-NWP-errorpropagation}. 
\Cref{sec:joint_train_vlm} discusses the joint training using these lemmas as well.
\Cref{lemma:Approximation-FF-NWP-1,lemma:Approximation-FF-NWP-2,lemma:Approximation-FF-VLM-UP} are proved in \Cref{subsection:NWP-Position-wiseFeedForward}, and \Cref{lemma:Self-attention-NWP} is proved in \Cref{subsection:NWP-Position-wiseFeedForward}.
\Cref{subsection:Appendix-NWP-errorpropagation} gives the error propagation lemma (\Cref{lemma:Appendix-NWP-errorpropagation}). 
\Cref{subsection:MP-NWP} gives the proof of \Cref{proposition:MP-NWP}.
Finally, \Cref{subsection:Appendix-VLM-Auxiliary} gives useful property on the message passing algorithm. 

\subsection{Proof of~\Cref{thm:vlm_two_steps}}\label{subsection:vlm_two_steps}
Similar to the proof of Theorem~\ref{theorem:CDM-Generalization-two-step},  define
\begin{align}\orisk^\star_{\vlm}\defn
\E_{(\xi, \xt)\sim\trueprobnext}
\Big[ \sum_{j \in [\dimtxt]} - \log \trueprobnext(\xtsub{j}| \xtsub{1:j-1},\Ei(\xi)) \Big].
\end{align}
Then we have the decomposition
\begin{align}
\distance(\trueprobnext, \probnext^\EstPar)
&=\risk_{\vlm}(\probnext^\EstPar,\Ei)
-
\orisk^\star_{\vlm}
\\
&=\underbrace{\inf_{\Par\in\Parspacevlm}\risk_{\vlm}(\probnext^\Par,\Ei)-\orisk^\star_{\vlm}}_{\text{approximation error}}
\,\,+\,\,
\underbrace{\risk_{\vlm}(\probnext^\EstPar,\Ei)
-
\inf_{\Par\in\Parspacevlm}\risk_{\vlm}(\probnext^\Par,\Ei)}_{\text{generalization error}}.
\end{align}

We state the following bounds on the approximation and generalization error, with proofs to follow shortly.
\begin{itemize}
    \item [(a).] If we choose $\numfflayer = \bigOtil(\layer)$, $D = \bigO(\state \layer)$, $D' =\bigOtil( \mmax\state \layer^{3})$, and $\bound = \bigOtil(\Binvbd+(\state\layer+\mmaxb^2)\sqrt{\diminadapt})$, then the approximation error
    \begin{align}
\inf_{\Par\in\Parspacevlm}\risk_{\vlm}(\probnext^\Par,\Ei)-\orisk^\star_{\vlm}
\leq
\dim_\txt\cdot
\bigOtil \Bigg( \sqrt{\frac{ (\state \layer^{8}\mmaxb^2+\diminadapt)\state\layer^3}{n}} 
+
\sqrt{\state^{5} \cdot \Binvbd^2 \cdot \Big( \suffmeasure(\scorefun) + \frac{1}{\diminadapt}\Big )}
\Bigg).\label{eq:ApproxError-claim-vlm}
    \end{align}
    \item[(b).] Under the same choice of model class $\Parspacevlm$, the generalization error 
    \begin{align}
   \risk_{\vlm}(\probnext^\EstPar,\Ei)
-
\inf_{\Par\in\Parspacevlm}\risk_{\vlm}(\probnext^\Par,\Ei)
\leq
\bigOtil \Bigg( \dimtxt\cdot\sqrt{\frac{ (\state \layer^{8}\mmaxb^2+\diminadapt)\state\layer^3}{n}} \Bigg)
    \end{align}
    with probability at least $1-1/n.$
\end{itemize}
Combining the claims yields Theorem~\ref{thm:vlm_two_steps}. \\
\paragraph{(a) Approximation error.}
    Take some $\delta'>0$ which will be defined later.
    For the feed forward layers, we use  \Cref{lemma:Approximation-FF-NWP-1,lemma:Approximation-FF-NWP-2,lemma:Approximation-FF-VLM-UP} with $\delta=\delta'\ll 1$.
    For the self-attention layers at the $\ell$-th step of the downsampling, we use \Cref{lemma:Self-attention-NWP} with $\delta=\frac{\delta'}{\max_{v}a_v^{(\ell)}\|\qnetwork^{(\ell)}_{v}\|_\infty}$.
    Following the argument in the proof of \Cref{theorem:CLIP-Generalization},  $\qnetwork^{(\ell)}_{v}=f^{(\ell)}_{\iota(\pa^{(L-\ell)}(v))}(\hnetwork^{(\ell)}_{v})$ and $\gnetwork_v^{(\ell)}$ are bounded by $3(1\lor\log S\transbd)$ for each $\ell$.
    Thus $C$ in \Cref{lemma:Approximation-FF-NWP-2} 
    and $\delta$ in \Cref{lemma:Self-attention-CDM} are bounded by $3(1\lor\log S\transbd)$ and $\frac{\delta'}{3(m^{(\ell)}_\txt+1)(1\lor\log S\transbd)}\leq \frac{\delta'}{(\mmaxb+1)(1\lor\log S\transbd)}$, respectively. 

    Furthermore, \Cref{lemma:Boundedness-VLM,lemma:Appendix-NWP-errorpropagation} yield that
    \begin{align}
      &\max_{(\xi,\xt,i)}|\log\trueprobnext(\xtsub{i}| \xtsub{1:i-1}, \xi )
      -\log\probnext^\EstPar(\xtsub{i}| \xtsub{1:i-1},\Ei(\xi))| \\
      & \textstyle \leq  
       S\transbd\Big(8^{\layer+1}\delta' \prod_{1\leq k\leq L}(2m_{\txt}^{(k)}+5)+ \delta_\image\Big)
       \\
      & \textstyle \leq  40^{\layer+1}\dim_\txt S\transbd \delta' + S\transbd\delta_\txt.
     \label{eq:ApproxError-2}
    \end{align}
    We 
    choose
    \begin{align}
        \delta' = \frac{\sqrt{(\state \layer^{8}\mmaxb^2+\diminadapt)\layer^3}}{40^{\layer+1}\dim_\txt \transbd\sqrt{S n}}.
    \end{align}
    Moreover, similar to the proof of Theorem~\ref{theorem:CDM-Generalization-two-step},
     from Proposition~\ref{prop:representation_equivalence} and \begin{align}\delta_\image=\|\log(\readimage(\what\Ei(\xi))-(\log \P[s|\xi])_{s\in [S]}\|_\infty
     \end{align}
     and Lemma~\ref{lemma:Boundedness-1}, it can be verified that there exists some $\Adapter(\cdot)$ in Eq.~\eqref{eq:adapter_def_cdm} such that, $\lops{\adaweighta}\leq\polyshortprime\Binvbd,\lops{\adaweightb}\leq \polyshortprime(\state\layer+\mmax^2)\sqrt{\diminadapt}$, and
    \begin{align}
        \E_{\xi}\delta_\image^2
        &\leq \E_{\xi}\|\log\readimage(\what\Ei(\xi))-\log
        \trueEi(\xi)\|_2^2
        \leq
     \polyshort\state^2\cdot
\E_{\xi}\|\what\Ei(\xi)
-
        \trueEi(\xi)\|_2^2  \\
        &\leq
        \polyshort \state^{2} \cdot \Binvbd^2 \cdot \tslinvLip^2 \cdot \truedimclip\cdot ( \suffmeasure(\scorefun) + \diminadapt^{-1} )
        \leq 
         \polyshort \state^{3} \cdot \Binvbd^2 \cdot ( \suffmeasure(\scorefun) + \diminadapt^{-1} )
    \end{align}
    for some $\polyshort,\polyshortprime>0$ depending polynomially on $\transbd^{\mmaxone}$.
    Putting pieces together, 
    according to \Cref{lemma:Approximation-FF-NWP-1,lemma:Approximation-FF-NWP-2,lemma:Self-attention-NWP,lemma:Approximation-FF-VLM-UP}, 
    we find that there exist some
$\Par\in\Parspacevlm$ that
yields  Eq.~\eqref{eq:ApproxError-claim-vlm}, where
    \begin{align}
        D&\leq d_\feature+d_\positional = (4S+2)L+1=\bigO(\state\layer),
        \\
        J &\lesssim (\log\log(SK/\delta'))\log(SK/\delta')=\bigOtil(\layer),
        \\
       D'=\linf{\bj} &\lesssim \mmax S(\log(SK/\delta'))^3+d_\feature+d_\positional = \bigOtil(\mmaxb\state\layer^3),
        \\
        \bound &\lesssim S(\transbd^2+\log(S\transbd/\delta'))+\mmax^2\log (S\transbd)+L+\log\frac{d\log(S\transbd)}{\delta'}
        +
        (\Binvbd+(\state\layer+\mmaxb^2)\sqrt{\diminadapt})\\
        &= \bigOtil(\Binvbd+(\state\layer+\mmaxb^2)\sqrt{\diminadapt}).
    \end{align}
    Here we recall $\mmax=\max\{\max_{k}m^{(k)}_\txt, \max_{k}m^{(k)}_\image\}$.
  
\paragraph{(b) generalization error.}
Since $\probnext^\EstPar$ is the minimizer of $\what\risk_{\vlm,t}(\probnext^\Par,\Ei)$ defined in Eq.~\eqref{eqn:VLM_empirical_risk_minimization}, we have
\begin{align}
    \orisk_{\vlm}(\probnext^\EstPar,\Ei)-\inf_{\Par\in\Parspacevlm}\orisk_{\vlm,t}(\probnext^\Par,\Ei)
    &\leq
2\sup_{\Par\in\Parspacevlm}|\what\risk_{\vlm}(\probnext^\Par,\Ei)
    -
    \orisk_{\vlm,t}(\probnext^\Par,\Ei)|.\label{eq:erm_bound_vlm}
\end{align}
Similar to the proof of Theorem~\ref{theorem:CLIP-Generalization}~and~\ref{theorem:CDM-Generalization-two-step}, we verify the conditions for Lemma~\ref{lm:generalization_lemma} and then  apply the lemma to derive an upper bound for the R.H.S. of Eq.~\eqref{eq:erm_bound_vlm}. 

    In Lemma~\ref{lm:generalization_lemma}, take $\Theta=\Parspacevlm,~\rho(\Par,\Par')=\nrmp{\Par-\Par'}$, $z_i=(\xi^{(i)},\xt^{(i)})$, and
   \begin{align}
   f(z_i;\Par)
   =
   -\frac{1}{\dimtxt}\sum_{j \in [\dimtxt]}  \log \probnext^\Par(\xtsub{j}| \xtsub{1:j-1},\Ei(\xi)) 
  .
 \end{align}
   \\
   
\myunder{Verification of condition~(a) in Lemma~\ref{lm:generalization_lemma}.}
We note that the set $\Parspacevlm$ with metric $\rho(\Par,\Par')=\nrmp{\Par-\Par'}$ has a diameter $\bound_\rho\defn2\bound$. Furthermore, the  dimension of $\Parspacevlm$ is bounded by $d_\rho\defn (2\numfflayer+3)(2\layer+1)(D+D'+1)^2+\state+2\state\diminadapt=\bigOtil(\state^2\layer^8\mmaxb^2+2\state\diminadapt)$. Thus, by Example 5.8 in~\cite{wainwright_2019}, we have
$\log\cN(\Delta;\Parspacevlm,\nrmp{})\leq d_\rho\log(1+2 r/\Delta)\leq d_\rho\log(2 A_\rho r/\Delta)$ for $\Delta\in(0,2r]$ with $A_\rho=2$.\\

\myunder{Verification of condition~(b) in Lemma~\ref{lm:generalization_lemma}.}
Since $f(z_i;\Par)$ is $\readbdvlmenc$-bounded by the definition of $\readvlm$ in Eq.~\eqref{eq:bdvlmenc}, it follows that $f(z_i;\Par)-\E[f(z_i;\Par)]$ is $\sigma=c\readbdvlmenc$-sub-Gaussian for all $\Par\in\Parspacevlm$  for some numerical constant $c>0$.\\

\myunder{Verification of condition~(c) in Lemma~\ref{lm:generalization_lemma}.}
By Lemma~\ref{lm:cdm_lip} and the boundedness condition, we have
\begin{align}
    |f(z_i;\Par) - f(z_i;\Par')|
    &\leq 
   \frac{1}{\dimtxt} \sum_{j \in [\dimtxt]}
    |   \log \probnext^\Par(\xtsub{j}| \xtsub{1:j-1},\Ei(\xi)) 
    -
     \log \probnext^{\Par'}(\xtsub{j}| \xtsub{1:j-1},\Ei(\xi)) |
   \\
    & \leq \bound_f \nrmp{\Par-\Par'},~~~~~~~~\text{where}~~ \bound_f\defn ((c\bound)^{18\numfflayer\layer}\state^4\readbd^3)^{4\layer+3}\transbd^{2\mmaxone},
\end{align}where $\readbd=4\mmaxone\log\transbd.$
Therefore, we may choose $\sigma'=\bound_f$ and condition (c) is hence satisfied.

Now, invoking Lemma~\ref{lm:generalization_lemma} and plugging in the values of $d_\rho,\sigma,\sigma',A_\rho,\bound_\rho$, we find
\begin{align}
 \sup_{\Par\in\Parspacecdm}|\what\risk_{\cdm,t}(\Mt^\Par,\Et)
    -
    \orisk_{\cdm,t}(\Mt^\Par,\Et)|
    &\leq \dimtxt\cdot c \sigma \sqrt{\frac{d_\rho \log \left(2A_\rho \left(1 + B_\rho \sigma' / \sigma\right)\right) + \log(1 / \eta)}{n}}\\
    &\leq
    \bigOtil \Bigg( \dimtxt\cdot\sqrt{\frac{ (\state \layer^{8}\mmaxb^2+\diminadapt)\state\layer^3+\log(1/\eta)}{n}} \Bigg)
\end{align} with probability at least $1-\eta.$ Setting $\eta=1/n$ completes the proof.

\subsection{Joint training of the vision-language model and the image representation}
\label{sec:joint_train_vlm}
Similar to Appendix~\ref{sec:joint_train_cdm}, in this section,
we consider jointly learning the vision-language models (VLMs) and the image representation within the JGHM framework. Following the setup of Section~\ref{sec:VLM_GHM}, 
suppose we are given a dataset of iid samples $\{( \xi^{(i)}, \xt^{(i)})\}_{i \in [n]} \simiid \truemu$.

The next word predictors 
\[
\probnext^\Par(\,\cdot \,| \xtsub{1:j-1},\Ei(\xi)) = \read_{\vlm} \circ \TF_{\vlm} \circ \encode_{\vlm}(\xtsub{1:j-1}, \Adapter(\Ei(\xi))),
\]  for $i\in[\dimtxt]$,
where $\Ei(\xt)=\NN^{\bW_\image}_\image(\xi)$ as defined in Section~\ref{sec:clip_JGHM}, and the remaining components are the same as defined in Section~\ref{sec:VLM_GHM}, except that in the embedding $\encode_\vlm$, we let
\begin{align}
\hnetwork_{\image,d}^{(0)}=\widetilde\readimage(\Ei(\xi)), ~~\text{ where }  \widetilde\readimage(z)\defn \proj_{[-\readbdvlmenc,\readbdvlmenc]}(z), 
\end{align}
in contrast to Eq.~\eqref{eq:bdvlmenc}.
We solve the empirical risk minimization 
\begin{align}\label{eq:vlm-empirical-risk-minimization-one-step}
\EstPar = \arg\min_{\Par \in \Parspacevlmone} \Big\{ \what\risk_{\vlm}(\probnext^\Par, \Ei) \defn \frac{1}{n}\sum_{i=1}^n
\Big[ \sum_{j \in [\dimtxt]} - \log \probnext^\Par(\xtsub{j}| \xtsub{1:j-1},\Ei(\xi)) \Big] \Big\},
\end{align}
where the parameter space is defined as
\begin{align}\label{eqn:VLM_param_space_one_step}
&~\Parspacevlmone \defn \Big\{ \bW_{\vlm},\bW_\image \text{ as defined in Eq.~\eqref{eqn:W_matrix_CLIP_encoder}};
\\
&~~~~~~~~~~\nrmps{\Par} \defn  \max_{i \in [{ 2\numfflayer+2}], \ell \in [{2\layer+1 }]}\{ \| \ffnweight_{i, \vlm}^{(\ell)} \|_{\op}, \| {\querymatrix}_{,\vlm}^{(\ell)}\|_{\op}, \| {\keymatrix}_{,\vlm}^{(\ell)} \|_{\op}, \| {\valuematrix}_{,\vlm}^{(\ell)} \|_{\op} \} \\
&~~~~~~~~~~~~~~~~
\vee 
\max_{i \in [\numfflayer+1]], \ell \in [\layer]}\{ \| \ffnweight_{i, \image}^{(\ell)} \|_{\op}, \| {\querymatrix}_{,\image}^{(\ell)}\|_{\op}, \| {\keymatrix}_{,\image}^{(\ell)} \|_{\op}, \| {\valuematrix}_{,\image}^{(\ell)} \|_{\op} \}  \le \bound \Big\}
\end{align}

Similar to Theorem~\ref{thm:vlm_two_steps} and Theorem~\ref{theorem:CDM-Generalization-one-step}, we state the following result without providing a formal proof.
\begin{theorem}[Sampling error of the conditional next-token predictors, joint training]\label{thm:vlm_one_step}
Suppose that \Cref{assumption:factorization} and \Cref{ass:bounded_transition} hold. Let $\Parspacevlmone$ be the set defined in Eq.~(\ref{eqn:VLM_param_space_one_step}), where $J = \bigOtil(\layer)$, $D = \bigO(\state \layer)$, $D' =\bigOtil(\mmaintextmax \state \layer^3)$, and $\bound = \bigOtil(\state\layer+\mmaintextmax^2)$. Let  $\EstPar$ be the empirical risk minimizer defined in Eq.~(\ref{eq:vlm-empirical-risk-minimization-one-step}). Then, with probability at least $1 - 1/n$, we have
\begin{align}\label{eqn:excess_risk_bound_VLM_joint}
\distance(\trueprobnext, \probnext^\EstPar)
&\defn
\E_{ (\xi, \xt) \sim \P_{\image,\txt}}
\Big[ \sum_{i \in [\dimtxt]} \KL\Big( \trueprobnext(\xtsub{i}| \xtsub{1:i-1}, \xi ) \Big|\Big| \probnext^\EstPar(\xtsub{i}| \xtsub{1:i-1},\Ei(\xi)) \Big) \Big]\\
&
\le
\dim_\txt\cdot
\bigOtil \Bigg( \sqrt{\frac{ \state^2 \layer^{11}\mmaxb^2}{n}} 
\Bigg),
\end{align}
where $\bigOtil$ hides polynomial factors in $(\log(\mmaintextmax \state \layer n), (\transbd)^{\mmaintextone})$.
\end{theorem}

\subsection{Position-wise Feed Forward Layer (proof of Lemma~\ref{lemma:Approximation-FF-NWP-2}~and~\ref{lemma:Approximation-FF-VLM-UP})}
\label{subsection:NWP-Position-wiseFeedForward}
This section proves \Cref{lemma:Approximation-FF-NWP-2,lemma:Approximation-FF-VLM-UP} for approximation with feed forward networks.
\begin{proof}[Proof of \Cref{lemma:Approximation-FF-NWP-2}]
First we explain how to compute $\indicator[v^{(\ell)}\in \cV_\txt^{(\ell)}]q$. 
We consider the value of
\begin{align}
\textstyle
    \sum_{\ell=L}^{\ell+1}\pnetwork_{v,2(L-\ell+2)}
    =\sum_{\ell=L}^{\ell+1} \cos \big(\frac{2\pi \iota(\pa^{(L-\ell)}(v))}{m^{(\ell)}}\big).
    \label{eq:Approximation-FF-NWP-2-1},
\end{align}
which is implemented with one linear layer on $\pnetwork_v$. 
Eq.\eqref{eq:Approximation-FF-NWP-2-1} is equal to $L-\ell$ if $v^{(\ell)}\in \cV_\txt^{(\ell)}$, and otherwise at most $(L-\ell) - \big(1-\max_{\ell+1\leq k\leq L}\big(\frac{2\pi}{m^{(k)}}\big)\big)$. 
Thus, we apply \Cref{lem:ReLU_indicator} with $a=L-\ell$ and $\delta = 1-\max_{\ell+1\leq k\leq L}\big(\frac{2\pi}{m^{(k)}}\big) \gtrsim \min_{\ell+1\leq k\leq L}(m^{(k)})^{-2}$ to obtain the network that implements $\indicator[v^{(\ell)}\in \cV_\txt^{(\ell)}]$. 
Therefore, there exists a network that implements \begin{align}[0;q;1-\indicator[v^{(\ell)}\in \cV_\txt^{(\ell)}]=\indicator[v^{(\ell)}\notin \cV_\txt^{(\ell)}];\indicator[v^{(\ell)}\in \cV_\txt^{(\ell)}]]  \end{align}, where
$
        J \lesssim 1,\ \|\widthvector\|_\infty \lesssim S+d_\positional, 
        \ B\lesssim L+\max_{\ell+1\leq k\leq L}(m^{(k)})^{2}.
$ 
Once we obtain this vector, we follow the argument of \Cref{lemma:IdeitifyTheRank} to obtain the network $\NN_1$ that implements $\indicator[v^{(\ell)}\in \cV_\txt^{(\ell)}]q$, with 
    \begin{align}
        J_1 \lesssim 1,\quad \|\widthvector_1\|_\infty \lesssim S+d_\positional, 
        \quad B_1\lesssim L+\max_{\ell+1\leq k\leq L}(m^{(k)})^{2}+C.
    \end{align}

Next we consider how to compute $a_v^{(\ell)}g$.
Note that $a_v^{(\ell)}g=\iota(\pa^{(L-\ell)}(v))g+ \indicator[v^{(\ell)}\in \cV_\txt^{(\ell)}]g$.
$\indicator[v^{(\ell)}\in \cV_\txt^{(\ell)}]g$ is implemented similarly to $\NN_1$.
The first part $\iota(\pa^{(L-\ell)}(v))g$ is obtained by replacing each $\NN_1,\cdots,\NN_{m^{(\ell)}}$ by $\iota(\pa^{(L-\ell)}(v))g$ in \Cref{lemma:IdeitifyTheRank}. 
Concatenating these two networks, we obtain $\NN_2$ that implements $a_v^{(\ell)}g$, where 
    \begin{align}
        J_2 \lesssim 1,\quad \|\widthvector_2\|_\infty \lesssim S+d_\positional, 
        \quad B_2\lesssim L+\max_{\ell+1\leq k\leq L}(m^{(k)})^{2}+m^{(\ell)}C.
    \end{align}

Finally, by concatenating $-\NN_1$, $\NN_2$, and  (the identify function for) $q$, we get the desired network. 
\end{proof}

\begin{proof}[Proof of \Cref{lemma:Approximation-FF-VLM-UP}]
$\NN_1$ is just a linear function, and thus we focus on $\NN_2$. 

First, we explain how to implement $\fnetwork^{(\ell)}_{\uparrow, \iota(\pa^{L-\ell}(v+1))}$.
Approximation of each $f^{(\ell)}_{\uparrow, \iota}$ follows from \Cref{lemma:Appendix-CLIP-Existence-log-sum-exp}, which is denoted by $\fnetwork^{(\ell)}_{\uparrow, \iota}=\NN_{3,\iota}\ (\iota=1,\dots,m^{(\ell)})$.
The size of these networks is bounded by 
    \begin{align}
   \depth \lesssim (\log\log(\staten\transbd/\delta))\log(\staten\transbd/\delta),\quad
    \|\widthvector\|_\infty  \lesssim \staten(\log(\staten\transbd/\delta))^3,
    \quad
    B  \lesssim 2\staten(\transbd^2+\log(\staten\transbd/\delta)).
\end{align}
Note that
\begin{align}
    \iota(\pa^{L-\ell}(v+1))
    &=
    \begin{cases}
        \iota(\pa^{L-\ell}(v)) + 1 & (\text{if $v^{(\ell)}\in \cV_\txt^{(L)}$  and $\iota(\pa^{L-\ell}(v))< m^{(\ell)}$})
    \\
        1
       &
       (\text{if $v^{(\ell)}\in \cV_\txt^{(L)}$  and $\iota(\pa^{L-\ell}(v))= m^{(\ell)}$})
    \\
        \iota(\pa^{L-\ell}(v)) & (\text{otherwise})
    \end{cases}
    \\
    &=
    \begin{cases}
        1 & (\text{if $\iota(\pa^{L-\ell}(v)) + \indicator[v^{(\ell)}\in \cV_\txt^{(\ell)}]=m^{(\ell)}+1$})
    \\
        \iota(\pa^{L-\ell}(v)) + \indicator[v^{(\ell)}\in \cV_\txt^{(\ell)}]
       &
       (\text{otherwise})
    \end{cases}.
\end{align}
Thus, for $1\leq i \leq m^{(\ell)}$, 
\begin{align}
   & \indicator[\iota(\pa^{L-\ell}(v+1))=i]  
  \\ &  =\begin{cases}
        \indicator[\iota(\pa^{L-\ell}(v)) + \indicator[v^{(\ell)}\in \cV_\txt^{(\ell)}]=i]
        & (\text{if $i\ne 1$})
        \\
        \indicator[\iota(\pa^{L-\ell}(v)) + \indicator[v^{(\ell)}\in \cV_\txt^{(\ell)}]=1]
        +
        \indicator[\iota(\pa^{L-\ell}(v)) + \indicator[v^{(\ell)}\in \cV_\txt^{(\ell)}]=m^{(\ell)}+1]
        & (\text{if $i= 1$})
    \end{cases}.
\end{align}
Using this fact, consider how to implement $\indicator[\iota(\pa^{L-\ell}(v)) + \indicator[v^{(\ell)}\in \cV_\txt^{(\ell)}]=i]\ (1\leq i \leq m^{(\ell)}+1)$. 
$\indicator[v^{(\ell)}\in \cV_\txt^{(L)}]$ is implemented in the proof of \Cref{lemma:Approximation-FF-NWP-2}, and $\iota(\pa^{L-\ell}(v))$ is implemented by using $\iota(\pa^{L-\ell}(v))=\sum_{i=1}^{m^{(\ell)}}i\indicator[i=\iota(\pa^{L-\ell}(v))]$ and \Cref{lem:ReLU_indicator}.
Once we obtain $\indicator[v^{(\ell)}\in \cV_\txt^{(\ell)}]+\iota(\pa^{L-\ell}(v))$, we apply \Cref{lem:ReLU_indicator} again with $\delta=1$.
Therefore, for $1\leq i \leq m^{(\ell)}$, there exists a network that implements $\indicator[\iota(\pa^{L-\ell}(v+1))=i]$, where
\begin{align}
    \depth \lesssim 1,\
    \|\widthvector\|_\infty  \lesssim m^{(\ell)}+S+d_\positional,
    \
    B  \lesssim L + \max_{\ell\leq k\leq L}(m^{(k)})^{2}.
\end{align}

We parallelize these indicators and $\NN_{3,i}$ to obtain the vector 
\begin{align}
    [\NN_{3,1};\NN_{3,2};\dots;\NN_{3,m^{(\ell)}};\NN_{3,1}; (\indicator[\iota(\pa^{L-\ell}(v+1))=i])_{i=1}^{m^{(\ell)}}].
    \label{eq:Lemma42-targetvector}
\end{align}
Once we obtain this vector we can follow the proof of \Cref{lemma:IdeitifyTheRank}.
Therefore, $\fnetwork^{(\ell)}_{\uparrow, \iota(\pa^{L-\ell}(v+1))}$ is implemented by a network
\begin{align}
    \depth &\lesssim (\log\log(S\transbd/\delta))\log(S\transbd/\delta),\\
    \|\widthvector\|_\infty  &\lesssim m^{(\ell)}S(\log(S\transbd/\delta))^3+ d_\positional,
    \\
    B  &\lesssim S(\transbd^2+\log(S\transbd/\delta)) + \max_{\ell\leq k\leq L}(m^{(k)})^{2}+L.
\end{align}

Next, we consider how to distinguish the three cases--(i) $\pa^{(L-\ell-1)}(v) = \pa^{(L-\ell-1)}(v+1)$, (ii) $\pa^{(L-\ell)}(v) = \pa^{(L-\ell)}(v+1)$ but $\pa^{(L-\ell-1)}(v) \ne \pa^{(L-\ell-1)}(v+1)$, and (iii) otherwise. 
Consider the values
\begin{align}
\textstyle
    \sum_{\ell=L}^{\ell+2}\pnetwork_{v,2(L-\ell+2)}=\sum_{\ell=L}^{\ell+2}\cos\Big(\frac{2\pi \iota(\pa^{(L-\ell)}(v))}{m^{(\ell)}}\Big),
    \label{eq:PA-L+1}
\end{align}
and
\begin{align}
\textstyle
    \sum_{\ell=L}^{\ell+1}\pnetwork_{v,2(L-\ell+2)}=\sum_{\ell=L}^{\ell+1}\cos\Big(\frac{2\pi \iota(\pa^{(L-\ell)}(v))}{m^{(\ell)}}\Big).
    \label{eq:PA-L}
\end{align}
\eqref{eq:PA-L+1} is equal to $L-\ell-2$ iff $\pa^{(L-\ell-1)}(v)=\pa^{(L-\ell-1)}(v+1)$, and \eqref{eq:PA-L} is equal to $L-\ell-1$ iff $\pa^{(L-\ell)}(v)=\pa^{(L-\ell)}(v+1)$. 
Therefore, the vector of the indicator functions $[\indicator[\eqref{eq:PA-L+1}=L-\ell-2], \indicator[\eqref{eq:PA-L+1}=L-\ell-2]-\indicator[\eqref{eq:PA-L}=L-\ell-1], \indicator[\eqref{eq:PA-L}=L-\ell-1]]$ correspond to the vector of the three cases $[\indicator[\text{(i)}], \indicator[\text{(ii)}], \indicator[\text{(iii)}]]$.
It is easy to implement $\indicator[\eqref{eq:PA-L+1}=L-\ell-2]$ and $\indicator[\eqref{eq:PA-L}=L-\ell-1]$ by following \eqref{eq:Approximation-FF-NWP-2-1}.
Now, we can determine whether to add $h$ and $q$, and the rest of the proof is the same as \Cref{lemma:IdeitifyTheRank}. 

Putting it all together, we obtain the desired network.
\end{proof}

\subsection{Self-attention layer (proof of \Cref{lemma:Self-attention-NWP})}
\label{subsection:Selt-attention-NWP}
    Define the auxiliary key and the query matrices  $\bWauxiliary^{(\ell)}_{K},\bWauxiliary^{(\ell)}_{Q}\in \R^{d_\positional \times d_\positional}$ as
    $
        \bWauxiliary^{(\ell)}_{K} = \bI_{d_\positional},
    $ 
    and
    \begin{align}
        (\bWauxiliary^{(\ell)}_{Q})_{i,j} = 
        \begin{cases}
            \alpha & \text{if $j=2$ and $i=4,6,\dots,2(L-\ell+1)$, or $2(L-\ell+2)+1\leq i=j\leq 2(L+1)$},
            \\
            (L-1)\alpha & \text{$(i,j)=(1,2)$},
            \\
            0 & \text{otherwise}.
        \end{cases}
    \end{align}

    Then, the value of QK matrix is 
    \begin{align}
    &\textstyle
    ((\bWauxiliary_{K}^{(\ell)}\qmatrix^{(\ell)})^\top
    (\bWauxiliary_{Q}^{(\ell)}\qmatrix^{(\ell)}))_{u,v}
   \\ &= 
    \textstyle
        \alpha\sum_{\ell'< \ell} \cos\big(\frac{2\pi \iota(\pa^{(L-\ell')}(v))}{m^{(\ell')}}\big)
    \\ &\quad  \textstyle   +
        \alpha\sum_{\ell'>\ell}
        \big[\sin\big(\frac{2\pi \iota(\pa^{(L-\ell')}(v))}{m^{(\ell')}}\big)\sin\big(\frac{2\pi \iota(\pa^{(L-\ell')}(u))}{m^{(\ell')}}\big)
        +
        \cos\big(\frac{2\pi \iota(\pa^{(L-\ell')}(u))}{m^{(\ell')}}\big)\cos\big(\frac{2\pi \iota(\pa^{(L-\ell')}(v))}{m^{(\ell')}}\big)\big]
     \\ &=  
     \textstyle
      \alpha \sum_{\ell'< \ell} \cos\big(\frac{2\pi \iota(\pa^{(L-\ell')}(v))}{m^{(\ell')}}\big)
      +\alpha\sum_{\ell'> \ell}\cos\big(\frac{2\pi \iota(\pa^{(L-\ell')}(v))-\iota(\pa^{(L-\ell')}(u))}{m^{(\ell')}}\big)
      .
       \label{eq:NWP-QK-matrix-aux-1}
    \end{align}
    For $v=0$, the attention mask ensures that the output is always $\qnetwork_0^{(\ell)}$, and thus let us focus on $v\in \cV_{\txt}^{(L)}$.
   In \eqref{eq:NWP-QK-matrix-aux-1}, the maximum value is $(L-1)\alpha$, which is achieved when 
    $u\in\cV^{(L)}$ and $u^{(\ell)} = \cC(\pa^{(L-\ell+1)}(v))$, or 
    $u=0$.
    Otherwise, $((\bWauxiliary_K^{(\ell)}\qmatrix^{(\ell)})^\top
    (\bWauxiliary_Q^{(\ell)}\qmatrix^{(\ell)}))_{u,v}$ is smaller than $\alpha(L-1)$ by
    $1-\max_{\ell'}\cos\big(\frac{2\pi}{m^{(\ell')}}\big) \gtrsim  \min_{\ell'}(m^{(\ell')})^{-2}$. 
   
   Therefore, by following the argument of \Cref{lemma:QK-matrix} and taking $\alpha \simeq \log (d/\delta)$, 
    we have
    \begin{align}
    \|(\softmax(\mask_{d_\positional} + (\bWauxiliary^{(\ell)}_{K,1}\bP)^\top
        (\bWauxiliary^{(\ell)}_{ Q,1}\bP)) - \softmax(\bA^{(\ell)} ))_{u,v}\|_\infty \leq \delta,\quad u,v\in \cV^{(L)},
        \label{eq:Softmax-Identity-2}
    \end{align}
    where $\bA^{(\ell)}\in \R^{(d+1)\times (d+1)}$ is a matrix such that $\bA^{(\ell)}_{u,v}=\frac{1}{a_v^{(\ell)}}$ if
    ``$u,v\in\cV^{(L)}$ and $u^{(\ell)} \in  \cC(\pa^{(L-\ell)}(v))$'',  $\bA^{(\ell)}_{u,v}=1$ for $u,v=0$, and $\bA^{(\ell)}_{u,v}=0$ otherwise.
    There is no approximation error in the column corresponding to $v=0$ because the mask excludes the dependency on all the other variables. 

    By following the proof of \Cref{lemma:Self-attention-CLIP}, we obtain matrices $\bW^{(\ell)}_{K},\bW^{(\ell)}_{Q},\bW^{(\ell)}_{V}\in \R^{D \times D}$ 
    with $\|\bW^{(\ell)}_{K}\|, \|\bW^{(\ell)}_{Q}\|,$ $ \|\bW^{(\ell)}_{V}\|\lesssim \log (d/\delta)$
    such that
    \begin{align}
    &\big[(\bW_{V}^{(\ell)}\qmatrix^{(\ell)})\softmax\big(\mask +(\bW_{K}^{(\ell)}\qmatrix^{(\ell)})^\top
    (\bW_{Q}^{(\ell)}\qmatrix^{(\ell)})\big)\big]_v
     \\ & =
   \begin{cases}
  \begin{bmatrix}
            \boldzero&  (\in \R^{(3\ell+L-1)S})\quad \quad\quad 
            \\
             \frac{1}{a_v^{(\ell)}}\Big(\qnetwork^{(\ell)}_{0}+\sum_{{u}^{(\ell)}\in \cC(\pa^{(L-\ell+1)}(v))}\indicator [u\leq v]\qnetwork^{(\ell)}_{u} \Big)+ \bdelta^{(\ell)}_v & (\in \R^{S})\ \quad \quad \quad \quad\quad \quad
            \\
            \boldzero &(\in \R^{d_\positional+(3L-3\ell+2)S+1})
        \end{bmatrix}, & (v\in \cV^{(L)}_\txt)
        \\
        \begin{bmatrix}
        \bzero &(\in \R^{(3\ell+L-1)S})\quad \quad\quad 
        \\
        \qnetwork^{(\ell)}_0 &(\in \R^{S})\ \quad \quad \quad \quad\quad  \quad 
        \\
        \bzero &(\in \R^{d_\positional+(3L-3\ell+2)S+1})
        \end{bmatrix}.
         & (v=0)
    \end{cases}
    \end{align} 
    where $\|\bdelta^{(\ell)}_{v}\|_\infty \leq \delta\max_{v'}\|\qnetwork_{v'}^{(\ell)}\|_\infty$. 

\subsection{Evaluation of error propagation}
\label{subsection:Appendix-NWP-errorpropagation}
\begin{lemma}[Evaluation of error propagation]\label{lemma:Appendix-NWP-errorpropagation}
    Assume we have functions 
     $\fnetwork_{\downarrow, \iota}^{(\ell)}, \fnetwork_{\uparrow, \iota}^{(\ell)}\ (1\leq \ell \leq L,\iota\in [m_\txt^{(\ell)}])$ such that
    \begin{align}
    \begin{aligned}
        \|f_{\downarrow,\iota}^{(\ell)}(h)-\fnetwork_{\downarrow,\iota}^{(\ell)}(h)\|_\infty
        &\leq \delta,\quad \forall h\in \R^S\text{ such that $\max_{s\in S}h_s= 0$, }\ell\in [L]   , 
        \\
        \|f_{\uparrow,\iota}^{(\ell)}(h)-\fnetwork_{\uparrow,\iota}^{(\ell)}(h)\|_\infty
        &\leq \delta,\quad \forall h\in \R^S\text{ such that $\max_{s\in S}h_s= 0$, }\ell\in [L]   , 
    \end{aligned}
     \label{eq:Appendix-NWP-ErrorPropagation-3}
    \end{align}
    and  $a_v^{(\ell)}\|\bdelta^{(\ell)}_{v}\|_\infty\leq \delta$ holds for all $\ell=L,\dots,1$ and $v\in \cV^{(L)}_\txt$. 
    Moreover, we assume that $\|h_{\image,\root}^{(0)}-\hnetwork_{\image,d}^{(0)}\|_\infty \leq \delta_\image$. 

    Consider the approximated update introduced in \eqref{eq:Appendix-NWP-ErrorPropagation-9} and \eqref{eq:Appendix-NWP-ErrorPropagation-10}. 
    Then, we have the following bound on the error propagation:
    \begin{align}
   & \textstyle \max_{v\in \cV^{(L)}_\txt}\|h_{v}^{(\ell)}-\hnetwork_{v}^{(\ell)}\|_\infty
      \textstyle \leq \delta \times (2m_{\txt}^{(\ell+1)}+4)\prod_{\ell+2\leq k\leq L}(2m_{\txt}^{(k)}+5),
     \label{eq:Appendix-NWP-ErrorPropagation-1}
    \\
   &  \textstyle 
     \max_{v\in \cV^{(L)}_\txt}\|q_{v}^{(\ell)}-\qnetwork_{v}^{(\ell)}\|_\infty
      \textstyle \leq \delta \times \prod_{\ell+1\leq k\leq L}(2m_{\txt}^{(k)}+5),
    \label{eq:Appendix-NWP-ErrorPropagation-12}
    \\
&\textstyle \max_{v\in \cV^{(L)}_\txt}\|
        b_{v}^{(L+1)} - \bnetwork_{v}^{(L+1)}\|_\infty
       \textstyle \leq 8^{L+1}\delta \prod_{1\leq k\leq L}(2m_{\txt}^{(k)}+5)+\delta_\txt
        \label{eq:Appendix-NWP-ErrorPropagation-14}
    .
    \end{align}
    Furthermore, we have
    \begin{align}
     & \textstyle  \|\softmax(\bnetwork^{(L+1)}_v)_s-\truemu(x_{\txt,v+1}=s|\xi,x_{\txt,1},\dots,x_{\txt,v})\|_\infty
     \\ &  \textstyle \leq 8^{L+1}\delta \prod_{1\leq k\leq L}(2m_{\txt}^{(k)}+5)+\delta_\image, && s\in [\staten],\ v=1,2,\dots,d-1.
     \label{eq:Appendix-NWP-ErrorPropagation-15}
    \end{align}
\end{lemma}
\begin{proof}
    The error from the image model is evaluated by \Cref{lemma:Appendix-CDM-errorpropagation^2}.
    Thus in the following we will assume $\delta_\image = 0$.

    First, we prove \eqref{eq:Appendix-NWP-ErrorPropagation-1} and \eqref{eq:Appendix-NWP-ErrorPropagation-12}. 
    Because $\hnetwork_{v}^{(L)}=h_{v}^{(L)}$, \eqref{eq:Appendix-NWP-ErrorPropagation-3} implies that
    \begin{align}
        \|q_{v}^{(L)}-\qnetwork_{v}^{(L)}\|_\infty \leq \delta.
    \end{align}
    By \Cref{lemma:Appendix-CLIP-Lipschitz-normalization} and $a_v^{(L)}\|\bdelta_{v}^{(L)}\|_\infty \leq \delta$, we have
    \begin{align}
        \|h_{v}^{(L-1)}-\hnetwork_{v}^{(L-1)}\|_\infty
        \leq
        2(m_\txt^{(L)}+1) \max_{u\in \cV_\txt}\| f^{(L)}_{\downarrow,\iota(u)}(x_{\txt,u}^{(L)})-\fnetwork^{(L)}_{\downarrow,\iota(u)}(x_{\txt,u}^{(L)})\|_\infty + 2\delta \leq (2m_\txt^{(L)}+4)\delta.
    \end{align}
    This confirms \eqref{eq:Appendix-NWP-ErrorPropagation-1} for $\ell=L-1$.

    Suppose that \eqref{eq:Appendix-NWP-ErrorPropagation-1} holds for some $\ell(\leq L-1)$ and prove \eqref{eq:Appendix-NWP-ErrorPropagation-12} for $\ell$ and
    \eqref{eq:Appendix-NWP-ErrorPropagation-1} for $\ell-1$.
    For \eqref{eq:Appendix-NWP-ErrorPropagation-12},
    \begin{align}
       & \| q^{(\ell)}_{v}-\qnetwork^{(\ell)}_{v}\|_\infty
       \\ & =\max_{v\in \cV_\txt}\| f^{(\ell)}_{\downarrow,\iota(\pa^{(L-\ell)}(v))}(h_{v}^{(\ell)})-\fnetwork^{(\ell)}_{\downarrow,\iota(\pa^{(L-\ell)}(v))}(\hnetwork_{v}^{(\ell)})\|_\infty 
        \\ &\leq 
        \max_{v\in \cV_\txt}\| f^{(\ell)}_{\downarrow,\iota(\pa^{(L-\ell)}(v))}(h_{v}^{(\ell)})-\fnetwork^{(\ell)}_{\downarrow,\iota(\pa^{(L-\ell)}(v))}(\hnetwork_{v}^{(\ell)})\|_\infty 
        +
        \| \fnetwork^{(\ell)}_{\downarrow,\iota(\pa^{(L-\ell)}(v))}(\hnetwork_{v}^{(\ell)})-\fnetwork^{(\ell)}_{\downarrow,\iota(\pa^{(L-\ell)}(v))}(\hnetwork_{v}^{(\ell)})\|_\infty 
\\ & \leq \textstyle
        \max_{v\in \cV_\txt}\|h_{v}^{(\ell)}-\hnetwork_{v}^{(\ell)}\|_\infty
        +
        \delta
 \\ &\textstyle\leq \delta + \delta \times (2m_{\txt}^{(\ell+1)}+4)\prod_{\ell+2\leq k\leq L}(2m_{\txt}^{(k)}+5)
 \\ & \textstyle \leq \delta \times \prod_{\ell+1\leq k\leq L}(2m_{\txt}^{(k)}+5),
 \label{eq:Appendix-NWP-ErrorPropagation-13}
    \end{align}
where we used \Cref{lemma:Appendix-CLIP-Lipschitz-logsumexp} for the second inequality.
Also, 
    \begin{align}
        \| h^{(\ell-1)}_{v}-\hnetwork^{(\ell-1)}_{v}\|_\infty
        & \leq 2(m_\txt^{(\ell)}+1) \max_{v\in \cV_\txt}\| q^{(\ell)}_{v}-\qnetwork^{(\ell)}_{v}\|_\infty +2\delta
        \\ &\textstyle\leq \delta \times 2(m_\txt^{(\ell)}+1)\prod_{k=\ell+1}^L(2m_\txt^{(k)}+5)\big)+2\delta
\\ & \textstyle\leq  \delta \times (2m_\txt^{(\ell)}+4)\prod_{k=\ell+1}^L(2m_\txt^{(k)}+5),
    \end{align}
    where we used \Cref{lemma:Appendix-CLIP-Lipschitz-normalization} and $a_v^{(\ell)}\|\bdelta_{\txt,v}^{(\ell)}\|_\infty \leq \delta$ for the first inequality, and \eqref{eq:Appendix-NWP-ErrorPropagation-13} for the second inequality.
    Therefore, by induction, we obtained \eqref{eq:Appendix-NWP-ErrorPropagation-1} for all $\ell=L,\dots,0$ and \eqref{eq:Appendix-NWP-ErrorPropagation-12} for all $\ell=L,\dots,1$.

    \eqref{eq:Appendix-NWP-ErrorPropagation-14} is derived by following \Cref{lemma:Appendix-CDM-errorpropagation}.
    Finally, from the Lipschitzness of softmax, we obtain \eqref{eq:Appendix-NWP-ErrorPropagation-15}.
\end{proof}

\subsection{Proof of \Cref{proposition:MP-NWP}}\label{subsection:MP-NWP}
    Fix $i\ (1\leq i\leq d-1)$. We show that $\softmax(\bar{b}^{(L)}_{i})_s=\message_{\uparrow,i+1}^{(L)}(s)$ for all $s\in [\staten]$. (Remember \Cref{lem:BP_NextWord_exact}, which states that $\message_{\uparrow,i}^{(L)}(x_{\txt,i+1})_s = \truemu(x_{\txt,i+1}=s|\xi,x_{\txt,1},\dots,x_{\txt,i})$.)
    Without loss of generality, for all $\ell$ and $\iota$, we assume that $\sum_{a\in [S]}\psi^{(\ell)}_{\txt,\iota}(s,a)$ is constant for all $s\in [S]$.

    We first consider the downsampling. We will verify that, for $v\in \cV^{(\ell)}_\txt$, 
    \begin{align}
        \message_{\downarrow,v}^{(\ell)}(s)=
        \begin{cases}
            \softmax(h_{v^{(L)}}^{(\ell)})_s & (\text{if $v\leq i$}), \\
            \softmax(h_{i}^{(\ell)})_s & (\text{if $v = \pa^{(L-\ell)}(i)$}), \\
            \frac{1}{S} & (\text{otherwise}),
        \end{cases}
        \label{eq:BP=MP-1}
    \end{align}
    for $\ell=L-1,\dots,0$ by induction. 
    
    We first verify that \eqref{eq:BP=MP-1} holds for $\ell=L-1$.
    For $v\in \cV^{(L-1)}_\txt$, if $v\leq i$, all children of $v$ are observed, and we have that
    \begin{align}
        \message_{\downarrow,v}^{(L-1)}(s) 
       \ \propto\
        \textstyle
        \prod_{v' \in \cC(v)} \psi_{\txt,\iota(v')}^{(L)}(s, x_{\txt,v'})
    =
        \prod_{v' \in \cN(v^{(L)})\cup \{v^{(L)}\}} \big( \psi_{\txt,\iota(v')}^{(L)}(s, x_{\txt,v'}) \big)^{\indicator[v'\leq v]}\ \propto\  \softmax(h_{v^{(L)}}^{(L-1)})_s,
    \end{align}
    else if $\pa(i)=v$, we have
    \begin{align}
        \message_{\downarrow,v}^{(L-1)}(s) 
        &\ \propto\ \textstyle
        \sum_{x^{(L)}_{\txt,\cC(v)}} \prod_{v' \in \cC(v)} \Big( \psi_{\txt,\iota(v')}^{(L)}(s, x^{(L)}_{\txt,v'}) \message^{(L)}_{\downarrow, v'}(x^{(L)}_{\txt,v'})\Big)
      \\ &=
        \textstyle
        \prod_{v' \in \cC(v)} \Big(\indicator [v'\leq i]\psi_{\txt,\iota(v')}^{(L)}(s, x_{\txt,v'})  + \indicator [v'> i]\frac{1}{S}\Big)
       \\ &\ \propto\ \textstyle
        \prod_{v' \in \cN(i)\cup \{i\}} \Big( \psi_{\txt,\iota(v')}^{(L)}(s, x_{\txt,v'}) \Big)^{\indicator[v'\leq i]}
          \\ &\ \propto\ \textstyle
        \softmax(h_{i}^{(L-1)})_s,
    \end{align}
    and else, when $\pa(i)< v$, none of the leaf nodes under $v$ is observed and we have 
    \begin{align}
         \message_{\downarrow,v}^{(L-1)}(s) 
            &\ \propto\ \textstyle  \sum_{x^{(L)}_{\txt,\cC(v)}} \prod_{v' \in \cC(v)} \Big( \psi_{\txt,\iota(v')}^{(L)}(s, x^{(L)}_{\txt,v'}) \message^{(L)}_{\downarrow, v'}(x^{(L)}_{\txt,v'})\Big)
        = \textstyle  \sum_{x^{(L)}_{\txt,\cC(v)}} \prod_{v' \in \cC(v)} \Big( \psi_{\txt,\iota(v')}^{(L)}(s, x^{(L)}_{\txt,v'}) \frac{1}{S}\Big) \ \propto\ \frac1S.
    \end{align}
    
   Then, assuming that \eqref{eq:BP=MP-1} holds for some $\ell\in [L-1]$, we will prove \eqref{eq:BP=MP-1} for $\ell-1$. 
   If $v\leq i$, 
   because all $v' \in \cC(v)$ satisfy $v'\leq i$ and thus $\message^{(\ell)}_{\downarrow, v'}(x^{(\ell)}_{\txt,v'}) \ \propto\  \exp\big(\big(h_{v'}^{(\ell)}\big)_{x^{(\ell)}_{\txt,v'}}\big)$, 
   we have that
    \begin{align}
        \message_{\downarrow,v}^{(\ell-1)}(s) 
        &\ \propto\ \textstyle
        \sum_{x^{(\ell)}_{\txt,\cC(v)}} \prod_{v' \in \cC(v)} \Big( \psi_{\txt,\iota(v')}^{(\ell)}(s, x^{(\ell)}_{\txt,v'}) \message^{(\ell)}_{\downarrow, v'}(x^{(\ell)}_{\txt,v'})\Big)
      \\ &= \textstyle
       \prod_{v' \in \cC(v)} \Big( \sum_{x^{(\ell)}_{\txt,v'}} \psi_{\txt,\iota(v')}^{(\ell)}(s, x^{(\ell)}_{\txt,v'}) \message^{(\ell)}_{\downarrow, v'}(x^{(\ell)}_{\txt,v'})\Big)
       \\ &\ \propto\ \textstyle
       \prod_{v' \in \cC(v)} \Big( \sum_{x^{(\ell)}_{\txt,\cC(v)}} \psi_{\txt,\iota(v')}^{(\ell)}(s, x^{(\ell)}_{\txt,v'}) \exp\big(\big(h_{v'^{(L)}}^{(\ell)}\big)_{x^{(\ell)}_{\txt,v'}}\big)\Big)
     \\ &\ \propto\ \textstyle
        \softmax\big(\prod_{v' \in \cC(v)}q_{v'^{(L)}}^{(\ell)}\big)_s,
    \\ &= \textstyle
        \softmax\big(\prod_{\substack{{v'}^{(L-\ell)}\in \cN(\pa^{(L-\ell)}(v^{(L)}))\\ \text{or }v'=v}}\indicator[v'\leq v]q_{v'}^{(\ell)}\big)_s,
          \\ &= \textstyle
        \softmax(h_{v^{(L)}}^{(\ell-1)})_s,
    \end{align}
    else if $v=\pa^{(L-\ell+1)}(i)$, we have that
     \begin{align}
       & \message_{\downarrow,v}^{(\ell-1)}(s) 
      \\  &\ \propto\ \textstyle
        \sum_{x^{(\ell)}_{\txt,\cC(v)}} \prod_{v' \in \cC(v)} \Big( \psi_{\txt,\iota(v')}^{(\ell)}(s, x^{(\ell)}_{\txt,v'}) \message^{(\ell)}_{\downarrow, v'}(x^{(\ell)}_{\txt,v'})\Big)
      \\ &=\textstyle
       \prod_{v' \in \cC(v)} \Big(\indicator[v'\leq i] \sum_{x^{(\ell)}_{\txt,v'}} \psi_{\txt,\iota(v')}^{(\ell)}(s, x^{(\ell)}_{\txt,v'}) \message^{(\ell)}_{\downarrow, v'}(x^{(\ell)}_{\txt,v'})+ \indicator[v'> i] \sum_{x^{(\ell)}_{\txt,v'}} \psi_{\txt,\iota(v')}^{(\ell)}(s, x^{(\ell)}_{\txt,v'}) \message^{(\ell)}_{\downarrow, v'}(x^{(\ell)}_{\txt,v'})\Big)
 \\ &\ \propto\ \textstyle
       \prod_{v' \in \cC(v)} \Big(\indicator[v'\leq i] \sum_{x^{(\ell)}_{\txt,v'}} \psi_{\txt,\iota(v')}^{(\ell)}(s, x^{(\ell)}_{\txt,v'}) \message^{(\ell)}_{\downarrow, v'}(x^{(\ell)}_{\txt,v'})+ \indicator[v'> i] \sum_{x^{(\ell)}_{\txt,v'}} \psi_{\txt,\iota(v')}^{(\ell)}(s, x^{(\ell)}_{\txt,v'}) \frac1S\Big)
       \\ &\ \propto\ \textstyle
       \prod_{v' \in \cC(v)} \Big(\indicator[v'\leq i] \sum_{x^{(\ell)}_{\txt,\cC(v)}} \psi_{\txt,\iota(v')}^{(\ell)}(s, x^{(\ell)}_{\txt,v'}) \exp\big(\big(h_{v'^{(L)}}^{(\ell)}\big)_{x^{(\ell)}_{\txt,v'}}\big)\Big)
       \\ &= \textstyle
       \prod_{\substack{{v'}^{(L-\ell)}\in \cN(\pa^{(L-\ell)}(v^{(L)}))\\ \text{or }v'=v}} (\indicator[v'\leq i] q^{(\ell)}_{v'})
          \\ &= \textstyle
        \softmax(h_{i}^{(\ell-1)})_s,
    \end{align}
    and else, when $\pa^{(L-\ell+1)}(i)<v$, none of the leaf nodes under $v$ is observed and we have
    \begin{align}
        \message_{\downarrow,v}^{(\ell-1)}(s) 
            &\ \propto\ \textstyle  \sum_{x^{(\ell)}_{\txt,\cC(v)}} \prod_{v' \in \cC(v)} \Big( \psi_{\txt,\iota(v')}^{(\ell)}(s, x^{(\ell)}_{\txt,v'}) \message^{(\ell)}_{\downarrow, v'}(x^{(\ell)}_{\txt,v'})\Big)
        = \textstyle  \sum_{x^{(\ell)}_{\txt,\cC(v)}} \prod_{v' \in \cC(v)} \Big( \psi_{\txt,\iota(v')}^{(\ell)}(s, x^{(\ell)}_{\txt,v'}) \frac{1}{S}\Big) \ \propto\ \frac1S.
    \end{align}
    Therefore, we have obtained \eqref{eq:BP=MP-1} for $\ell-1$, and the induction proves that \eqref{eq:BP=MP-1} holds for all $\ell=L-1,\dots,0$. 
    
    We next consider the upsampling. 
    Let $\ell_{\star}$ be the largest $\ell$ such that $\pa^{(L-\ell_{\star})}(i) = \pa^{(L-\ell_{\star})}(i+1)$ holds.
    We will verify that, for $\ell=0,1,\dots,L$ and $v=\pa^{(L-\ell)}(i+1)$, 
    \begin{align}
        \message^{(\ell)}_{\uparrow,v}(s) = 
        \begin{cases}
            \softmax(\bar{b}^{(\ell)}_{\uparrow,i}-h^{(\ell)}_{\downarrow,i})_s & (\ell =0,1,\dots, \ell_{\star}),
            \\
            \softmax(\bar{b}^{(\ell)}_{\downarrow,i})_s & (\ell = \ell_{\star}+1,\dots,L),
        \end{cases}
        \label{eq:BP=MP-2}
    \end{align}
    by induction. 

    Checking \eqref{eq:BP=MP-2} for $\ell=0$ is done by just comparing the definitions of $\message^{(0)}_{\uparrow,\root}$ and $\bar{b}^{(0)}_{\uparrow,v}$. 

    Suppose that \eqref{eq:BP=MP-2} holds for some $\ell$ and $v= \pa^{(L-\ell)}(i+1)$.
    We will prove that \eqref{eq:BP=MP-2} holds for $\ell+1$ and $v= \pa^{(L-\ell-1)}(i+1)$.
    If $\ell+1 \leq \ell_{\star}$, we have that $v= \pa^{(L-\ell-1)}(i)=\pa^{(L-\ell-1)}(i+1)$, and that
    \begin{align}
       & \message^{(\ell+1)}_{\uparrow,v}(x^{(\ell+1)}_{\txt,v})
        \\ & \propto\ \textstyle
        \sum_{x^{(\ell)}_{\txt,\pa(v)}, x^{(\ell+1)}_{\txt,\cN(v)}}\psi^{(\ell+1)}_{\txt,\iota(v)}(x^{(\ell)}_{\txt,\pa(v)}, x^{(\ell+1)}_{\txt,\cC(\pa(v))})\message_{\uparrow, \pa(v)}^{(\ell)}(x^{(\ell)}_{\txt,\pa(v)})\prod_{v' \in \cN(v)} \message_{\downarrow,v'}^{(\ell+1)}(x^{(\ell+1)}_{\txt,v'})
        \\ &  \propto\ \textstyle
        \sum_{x^{(\ell)}_{\txt,\pa(v)}, x^{(\ell+1)}_{\txt,\cN(v)}}\psi^{(\ell+1)}_{\txt,\iota(v)}(x^{(\ell)}_{\txt,\pa(v)}, x^{(\ell+1)}_{\txt,\cC(\pa(v))})\message_{\uparrow, \pa(v)}^{(\ell)}(x^{(\ell)}_{\txt,\pa(v)})\prod_{v' \in \cN(v)} \indicator[v'\leq i]\message_{\downarrow,v'}^{(\ell+1)}(x^{(\ell+1)}_{\txt,v'})
        \\ & \textstyle =
        \sum_{x^{(\ell)}_{\txt,\pa(v)}}
        \Big(
        \psi^{(\ell+1)}_{\txt,\iota(v)}(x^{(\ell)}_{\txt,\pa(v)}, x^{(\ell+1)}_{v})\message_{\uparrow, \pa(v)}^{(\ell)}(x^{(\ell)}_{\txt,\pa(v)})
        \\ & \textstyle \quad \quad \quad \quad \quad \quad \prod_{v' \in \cN(v)} \Big(\sum_{x^{(\ell+1)}_{\txt,v'}}\psi^{(\ell+1)}_{\txt,\iota(v')}(x^{(\ell)}_{\txt,\pa(v)}, x^{(\ell+1)}_{\txt,v'})\exp(h^{(\ell+1)}_{v'^{(L)}}(x^{(\ell+1)}_{\txt,v'}))\Big)^{\indicator[v'\leq i]}\Big)
        \\ & \textstyle =
        \sum_{x^{(\ell)}_{\txt,\pa(v)}}
        \Big(
        \psi^{(\ell+1)}_{\txt,\iota(v)}(x^{(\ell)}_{\txt,\pa(v)}, x^{(\ell+1)}_{\txt,v})\message_{\uparrow, \pa(v)}^{(\ell)}(x^{(\ell)}_{\txt,\pa(v)})
        \exp\Big(\sum_{v' \in \cN(v)} \indicator[v'\leq i]q^{(\ell+1)}_{v'^{(L)}}(x^{(\ell)}_{\txt,\pa(v)})\Big)\Big)
        \label{eq:BP=MP-3} 
        \\ & \textstyle \ \propto \
        \sum_{x^{(\ell)}_{\txt,\pa(v)}}
        \Big(\psi^{(\ell+1)}_{\txt,\iota(v)}(x^{(\ell)}_{\txt,\pa(v)}, x^{(\ell+1)}_{\txt,v})\exp\Big(\bar{b}^{(\ell)}_{i}(x^{(\ell)}_{\txt,\pa(v)})-h^{(\ell)}_{i}(x^{(\ell)}_{\txt,\pa(v)})+h^{(\ell)}_{i}(x^{(\ell)}_{\txt,\pa(v)})-q^{(\ell+1)}_{i}(x^{(\ell)}_{\txt,\pa(v)})\Big)\Big)
        \\ &\label{eq:BP=MP-4} 
        \\ & \textstyle =
        \sum_{x^{(\ell)}_{\txt,\pa(v)}}
        \Big(\psi^{(\ell+1)}_{\txt,\iota(v)}(x^{(\ell)}_{\txt,\pa(v)}, x^{(\ell+1)}_{\txt,v})\exp\Big(\bar{b}^{(\ell)}_{i}(x^{(\ell)}_{\txt,\pa(v)})-q^{(\ell+1)}_{i}(x^{(\ell)}_{\txt,\pa(v)})\Big)\Big)
          \\ & \textstyle \ \propto \ \exp\Big(f^{(\ell+1)}_{\uparrow,\iota(v)}\Big(\bar{b}^{(\ell)}_{\uparrow,i}-q^{(\ell+1)}_{\downarrow,i}\Big)\Big)_{x^{(\ell+1)}_{\txt,v}}
          \\ & \textstyle \ \propto \
          \softmax\Big(\bar{b}^{(\ell+1)}_{i}-h^{(\ell+1)}_{i}\Big)_{x^{(\ell+1)}_{\txt,v}}.
    \end{align}
    In \eqref{eq:BP=MP-4}, because $\pa^{(L-\ell-1)}(i)$ and $v$ means the same child node of $\pa(v)$, the condition ``${v'}^{(L-\ell-1)}\in \cN(v)$'' is equivalent to ``${v'}^{(L-\ell-1)}\in \cN(\pa^{(L-\ell-1)}(i))$'' and does not overlap with ``$v'=i$''. Therefore, in \eqref{eq:BP=MP-4}, we used that $\sum_{v' \in \cN(v)} \indicator[v'\leq i]q^{(\ell+1)}_{v'^{(L)}}(x^{(\ell)}_{\txt,\pa(v)})=\sum_{\substack{{v'}^{(L-\ell-1)}\in \cN(\pa^{(L-\ell-1)}(i))\\ \text{or }v'=i}}\indicator[v'\leq i]q^{(\ell+1)}_{{v'}}-q^{(\ell+1)}_{i}(x^{(\ell)}_{\txt,\pa(v)})=h^{(\ell)}_{i}(x^{(\ell)}_{\txt,\pa(v)})-q^{(\ell+1)}_{i}(x^{(\ell)}_{\txt,\pa(v)})$. 
    Else if $\ell= \ell_{\star}$, we have that
    \begin{align}
       & \message^{(\ell+1)}_{\uparrow,v}(x^{(\ell+1)}_{\txt,v})
 \ \propto\ \eqref{eq:BP=MP-3} \\ &= \textstyle 
        \sum_{x^{(\ell)}_{\txt,\pa(v)}}
        \Big(\psi^{(\ell+1)}_{\txt,\iota(v)}(x^{(\ell)}_{\txt,\pa(v)}, x^{(\ell+1)}_{\txt,v})\exp\Big(\bar{b}^{(\ell)}_{i}(x^{(\ell)}_{\txt,\pa(v)})-h^{(\ell)}_{i}(x^{(\ell)}_{\txt,\pa(v)})+h^{(\ell)}_{i}(x^{(\ell)}_{\txt,\pa(v)})\Big)\Big)
        \label{eq:BP=MP-5} 
        \\ & \textstyle =
        \sum_{x^{(\ell)}_{\txt,\pa(v)}}
        \Big(\psi^{(\ell+1)}_{\txt,\iota(v)}(x^{(\ell)}_{\txt,\pa(v)}, x^{(\ell+1)}_{\txt,v})\exp\Big(\bar{b}^{(\ell)}_{i}(x^{(\ell)}_{\txt,\pa(v)})\Big)\Big)
     \\ & \textstyle \ \propto \ \exp\Big(f^{(\ell+1)}_{\uparrow,\iota(v)}\Big(\bar{b}^{(\ell)}_{i}\Big)\Big)_{x^{(\ell+1)}_{\txt,v}}
      \\ & \textstyle \ \propto \ \softmax\Big(\bar{b}^{(\ell+1)}_{i}\Big)_{x^{(\ell+1)}_{\txt,v}}.
    \end{align}
    In \eqref{eq:BP=MP-5}, because $\pa^{(L-\ell-1)}(i)$ and $v$ are different child nodes of $\pa(v)$, and  
    ``${v'}^{(L-\ell-1)}\in \cN(\pa^{(L-\ell-1)}(i)) $ or $v'=i$'' is equivalent to ${v'}^{(L-\ell-1)}\in \cN(v) $, we used that
    \begin{align}\sum_{v' \in \cN(v)} \indicator[v'\leq i]q^{(\ell+1)}_{v'^{(L)}}(x^{(\ell)}_{\txt,\pa(v)})=\sum_{\substack{{v'}^{(L-\ell-1)}\in \cN(\pa^{(L-\ell-1)}(i))\\ \text{or }v'=i}}\indicator[v'\leq i]q^{(\ell+1)}_{{v'}}=h^{(\ell)}_{i}(x^{(\ell)}_{\txt,\pa(v)})
    \end{align}
    (ignoring $\normalize$).
    Else, when $\ell> \ell_{\star}$, $v=\pa^{(L-\ell-1)}(i+1)$ does not have observed leaf nodes as its descendants, and
    \begin{align}
          \message^{(\ell+1)}_{\uparrow,v}(x^{(\ell+1)}_{\txt,v})
 &\ \propto\ \eqref{eq:BP=MP-3} 
          \textstyle =
        \sum_{x^{(\ell)}_{\txt,\pa(v)}}
        \Big(\psi^{(\ell+1)}_{\txt,\iota(v)}(x^{(\ell)}_{\txt,\pa(v)}, x^{(\ell+1)}_{\txt,v})\exp\Big(\bar{b}^{(\ell)}_{i}(x^{(\ell)}_{\txt,\pa(v)})\Big)\Big)
      \textstyle \\ & \ \propto \ \exp\Big(f^{(\ell+1)}_{\uparrow,\iota(v)}\Big(\bar{b}^{(\ell)}_{i}\Big)\Big)_{x^{(\ell+1)}_{\txt,v}} \ \propto \ \softmax\Big(\bar{b}^{(\ell+1)}_{i}\Big)_{x^{(\ell+1)}_{\txt,v}}.
    \end{align}
    
    Now, by induction, we have \eqref{eq:BP=MP-2} for all $\ell=0,1,\dots,L$ and $v\in \pa^{(L-\ell)}(i+1)$.

    It always holds that $\ell_{\star}<L$. Therefore, we obtain that $\message^{(L)}_{\uparrow,v} =  \softmax(\bar{b}^{(L)}_{i})$, which finishes the proof.

   \subsection{Bound on the posterior probability}\label{subsection:Appendix-VLM-Auxiliary}
As an auxiliary lemma, we state the boundedness of $\bar{b}^{(\ell)}_v.$ 
\begin{lemma}\label{lemma:Boundedness-VLM}
Under Assumption~\ref{assumption:factorization}~and~\ref{ass:bounded_transition}, we have that
\begin{align}
       \frac{1}{S\transbd} \leq \truemu(x_{\txt,v+1}=s|\xi,x_{\txt,1},\dots,x_{\txt,v})
       \leq 1,
    \end{align}
    for all $v=1,\dots,d-1$. 
\end{lemma}
\begin{proof}
    Consider the message passing algorithm in \eqref{eq:Appendix-NextWord-MP-1} and \eqref{eq:Appendix-NextWord-MP-3}. 
    For $\ell=L$, $\pa^{(L-L)}(v)\ne \pa^{(L-L)}(v+1)$ always holds.
    Because $\bar{b}_{v}^{(L)}=f_{\uparrow, \iota(v+1)}^{(L)}(\normalize(\bar{b}_{v+1}^{(L-1)}))$ and
    \begin{align}
         (f_{\uparrow, \iota}^{(\ell)}(h))_s &   \textstyle  = \log \sum_{a\in [\staten]}\psi_{\txt, \iota}^{(\ell)}(a,s)e^{h_a},\quad (\psi_{\txt, \iota}^{(\ell)}(a,s)\geq \transbd^{-1})
    \end{align}
    $\bar{b}_{v}^{(L)}$ is bounded by $-\log \transbd \leq (\bar{b}_{v}^{(L)})_s$, and $\softmax(\bar{b}_{v}^{(L)})_s=\truemu(x_{\txt,v+1}=s|\xi,x_{\txt,1},\dots,x_{\txt,v})$ (this equivalence is proven in \Cref{proposition:MP-NWP}) is bounded by 
    \begin{align}
       \textstyle \frac{1}{S\transbd} \leq \truemu(x_{\txt,v+1}=s|\xi,x_{\txt,1},\dots,x_{\txt,v})
       \leq 1,
    \end{align}
    for all $v=1,\dots,d-1$. 
\end{proof}

\section{Auxiliary lemmas}

\subsection{Lipschitzness of transformers}
In this section, we establish the Lipschitz continuity of the transformers in their parameters. Let $\|\tokenmatrix\|_{2,\infty}\defn \max_{i\in N}\|\tokenmatrix_i\|_{2}$  denote the column-wise $(2,\infty)$-norm for any matrix $\tokenmatrix\in\R^{M\times N}$. For any $\radius>0$, we let $\trunspace_{\radius}=\{\tokenmatrix:\|\tokenmatrix\|_{2,\infty}\leq\radius\}$
be the ball of radius $\radius$ in $\|\cdot\|_{2,\infty}$. W.l.o.g., we assume the radius $\radius\geq1$.
\begin{lemma}[Lipschitzness of the feedforward layer]\label{lm:ffn_lip}
For a $\numfflayer+1$-layer feedforward (FF) network parameterized by $\Par_{\ff}=(\ffnweight_1\in\R^{\hiddendim\times(\embeddim+1)},\ffnweight_{\numfflayer+1}\in\R^{\hiddendim\times (\hiddendim+1)},\{\ffnweight_j\in\R^{\hiddendim\times(\hiddendim+1)}\}_{2\leq j\leq \numfflayer})$, we introduce the norm
(as in Eq.~\ref{eqn:CLIP_param_space})
\begin{align}
    \nrmp{\Par_\ff}= \max_{j\in[\numfflayer+1]}\lops{\ffnweight_j}
.
\end{align}
 Define the parameter space
\begin{align}
  \ParSpace_{\ff,\bound} \defn\{\Par_\ff:\nrmp{\Par_\ff}\leq \bound\}.
\end{align}
Then for $\tokenmatrix\in\trunspace_{\radius},\Par_\ff\in\ParSpace_{\ff,\bound}$, the function $(\Par_\ff,\tokenmatrix)\mapsto\FF_{\Par_{\ff}}(\tokenmatrix)+\tokenmatrix$ is $(\numfflayer+1)\bound^\numfflayer\radius$-Lipschitz w.r.t. $\Par_{\ff}$ in $\nrmp{\cdot}$ and  $1+\bound^{\numfflayer+1}$-Lipschitz w.r.t. $\tokenmatrix$ in $\|\cdot\|_{2,\infty}$.
\end{lemma} 

\begin{proof}[Proof of \Cref{lm:ffn_lip}]
    By definition, for the $i$-th column $\tokenmatrix_i$ of the matrix $\tokenmatrix\in\R^{D\times N}$, we have\footnote{We incorporate the intercept term into the token matrix to simplify the notation.}
    \begin{align}
    \FF_{\Par_\ff}(\tokenmatrix_i)
    =
    \ffnweight_{\numfflayer+1} \cdot \ReLU(\ffnweight_{\numfflayer} \cdots \ReLU(\ffnweight_{1} \cdot \tokenmatrix_i))
   .
    \end{align}
    Therefore, for $\Par'_{\ff}=(\ffnweight_{1:\numfflayer+1}')\in\ParSpace_{\ff,\bound}$
    \begin{align}
      &~\quad\| \FF_{\Par_\ff}(\tokenmatrix)+\tokenmatrix- \FF_{\Par'_\ff}(\tokenmatrix)-\tokenmatrix\|_{2,\infty}\\
      & =\max_{i}
     \|
     \ffnweight_{\numfflayer+1} \cdot \ReLU(\ffnweight_{\numfflayer} \cdots \ReLU(\ffnweight_{1} \cdot \tokenmatrix_i))
      -
      \ffnweight'_{\numfflayer+1} \cdot \ReLU(\ffnweight'_{\numfflayer} \cdots \ReLU(\ffnweight'_{1} \cdot \tokenmatrix_i))
      \|_2 
      \\
      &\leq
      \sum_{j=1}^{\numfflayer+1}
      \max_{i}
     \|
       \ffnweight'_{\numfflayer+1} \cdots\ffnweight'_{j+1} \ReLU(\ffnweight_{j} \cdots \ReLU(\ffnweight_{1} \cdot \tokenmatrix_i))
      -
    \ffnweight'_{\numfflayer+1} \cdots\ffnweight'_{j+1} \ReLU(\ffnweight'_{j} \cdots \ReLU(\ffnweight_{1} \cdot \tokenmatrix_i))
      \|_2 \\
      &\overset{(i)}{\leq}
      \sum_{j=1}^{\numfflayer+1}
      \max_i
      \lops{\ffnweight'_{J+1}\cdots \ffnweight'_{j+1}}\cdot
      \lops{\ffnweight_{j}-\ffnweight_j'}\cdot\|\ReLU(\ffnweight_{j-1} \cdots \ReLU(\ffnweight_{1} \cdot \tokenmatrix_i))\|_2
    \\
 &\overset{(ii)}{\leq}
 \bound^\numfflayer\radius\cdot (\sum_{j=1}^{\numfflayer+1} \lops{\ffnweight_j-\ffnweight_j'})
 \leq (\numfflayer+1)\bound^\numfflayer\radius\cdot \nrmp{\Par_{\ff}-\Par_{\ff'}}
 ,
    \end{align}
where steps~(i)~and~(ii) use the fact that $\|\ReLU(\bx)-\ReLU(\by)\|_2\leq \|\bx-\by\|_2$.
Similarly,
for any matrices $\tokenmatrix,\tokenmatrix'$
    \begin{align}
      &~\quad\| \FF_{\Par_\ff}(\tokenmatrix)+\tokenmatrix- \FF_{\Par_\ff}(\tokenmatrix')-\tokenmatrix'\|_{2,\infty}\\
      & =\max_{i}
     \|\tokenmatrix_i+
     \ffnweight_{\numfflayer+1} \cdot \ReLU(\ffnweight_{\numfflayer} \cdots \ReLU(\ffnweight_{1} \cdot \tokenmatrix_i))
      -\tokenmatrix'_i
      -
   \ffnweight_{\numfflayer+1} \cdot \ReLU(\ffnweight_{\numfflayer} \cdots \ReLU(\ffnweight_{1} \cdot \tokenmatrix'_i))
      \|_2 
      \\
      &\leq
      \max_i\|\tokenmatrix_i-
        \tokenmatrix'_i\|_2
        +
      \prod_{j=1}^{\numfflayer+1}\lops{\ffnweight_{j}}
        \cdot\max_i\|\tokenmatrix_i-
        \tokenmatrix'_i\|_2
     \leq   
       (1+ \bound^{\numfflayer+1})\cdot \|\tokenmatrix-
        \tokenmatrix'\|_{2,\infty},
        \end{align}
        where the last line uses $\|\ReLU(\bx)-\ReLU(\by)\|_2\leq \|\bx-\by\|_2$.
\end{proof}

\begin{lemma}[Lipschitzness of the attention layer]\label{lm:attention_lip}
For a single attention layer $\Attn_{\Par_{\attn}}(\cdot)$ parameterized by $\Par_{\attn}=(\querymatrix,\keymatrix,\valuematrix)$, we introduce the norm
\begin{align}
    \nrmp{\Par_\attn}= \max\{\lops{\querymatrix}
    ,\lops{\keymatrix}
    ,\lops{\valuematrix}\}
    ,
\end{align}
where $\querymatrix,\keymatrix,\valuematrix\in\R^{D\times D}$  are the query, key, value matrices. 
 Define the parameter space
\begin{align}
  \ParSpace_{\attn,\bound} \defn\{\Par_\attn:\nrmp{\Par_\attn}\leq \bound\}.
\end{align}
Then for $\tokenmatrix\in\trunspace_{\radius},\Par_\attn\in\ParSpace_{\attn,\bound}$, the function $(\Par_\attn,\tokenmatrix)\mapsto\Attn_{\Par_{\attn}}(\tokenmatrix)$ is $\radius(1+4e\bound^2\radius^2)$-Lipschitz w.r.t. $\Par_{\attn}$ in $\nrmp{\cdot}$ and  $  1+\bound(1+4e\bound^2\radius^2)$-Lipschitz w.r.t. $\tokenmatrix$ in $\|\cdot\|_{2,\infty}$.
\end{lemma}

\begin{proof}[Proof of \Cref{lm:attention_lip}]
Adopt the shorthand $\softmaxshort$ for the softmax activation.    By definition, for any input $\tokenmatrix\in\R^{\hiddendim\times\horizon}$,  the output of attention $\Attn_{\Par_{\attn}}(\tokenmatrix)$ is given by
    \begin{align}
        \outtokenmatrix_i\defn[\Attn_{\Par_{\attn}}(\tokenmatrix)]_{i}+\tokenmatrix_i &= 
        \sum_{j=1}^\horizon
        \softmaxshort(\<\querymatrix\tokenmatrix_i,\keymatrix\tokenmatrix_j\>)\cdot\valuematrix\tokenmatrix_j+\tokenmatrix_i,~~~~\text{for }i\in[\horizon].
    \end{align}
Similarly, for $\Par'_{\attn}=(\querymatrix',\keymatrix',\valuematrix')$, the output is given by
\begin{align}
     \outtokenmatrix'_i\defn[\Attn_{\Par'_{\attn}}(\tokenmatrix)]_{i}+\tokenmatrix_i &= 
        \sum_{j=1}^\horizon
        \softmaxshort(\<\querymatrix'\tokenmatrix_i,\keymatrix'\tokenmatrix_j\>)\cdot\valuematrix'\tokenmatrix_j+\tokenmatrix_i,~~~~\text{for }i\in[\horizon].
\end{align}
Note that
$
   \|(\Attn_{\Par_{\attn}}(\tokenmatrix)+\tokenmatrix)-(\Attn_{\Par'_{\attn}}(\tokenmatrix)+\tokenmatrix)\|_{2,\infty} 
  = \max_{i\in[\horizon]}\| \outtokenmatrix_i- \outtokenmatrix'_i\|_2$. For any $i\in[\horizon]$, we have
\begin{align}
&\| \outtokenmatrix_i- \outtokenmatrix'_i\|_2
\\ &=
\|\sum_{j=1}^\horizon
        \softmaxshort(\<\querymatrix\tokenmatrix_i,\keymatrix\tokenmatrix_j\>)\cdot\valuematrix\tokenmatrix_j
        -
        \sum_{j=1}^\horizon
        \softmaxshort(\<\querymatrix'\tokenmatrix_i,\keymatrix'\tokenmatrix_j\>)\cdot\valuematrix'\tokenmatrix_j\|_2\\
        &\leq
         \sum_{j=1}^\horizon  
         \lops{\softmaxshort(\<\querymatrix\tokenmatrix_i,\keymatrix\tokenmatrix_j\>)\valuematrix- \softmaxshort(\<\querymatrix'\tokenmatrix_i,\keymatrix'\tokenmatrix_j\>)\valuematrix'}\cdot \|\tokenmatrix_j\|_2\\
         &\leq
         \radius \cdot \sum_{j=1}^\horizon \Big[ \lops{\softmaxshort(\<\querymatrix\tokenmatrix_i,\keymatrix\tokenmatrix_j\>)(\valuematrix-\valuematrix')} 
         +
        \lops{(\softmaxshort(\<\querymatrix\tokenmatrix_i,\keymatrix\tokenmatrix_j\>)-\softmaxshort(\<\querymatrix'\tokenmatrix_i,\keymatrix'\tokenmatrix_j\>))\valuematrix'}\Big]
        \\
        &\leq
        \termattn_{a1}+\termattn_{a2},
\end{align}
where 
\begin{align}
     \termattn_{a1}   &\defn
         \radius \cdot \sum_{j=1}^\horizon\softmaxshort(\<\querymatrix\tokenmatrix_i,\keymatrix\tokenmatrix_j\>)\lops{\valuematrix-\valuematrix'}
         \overset{(i)}{\leq}  \radius \cdot\lops{\valuematrix-\valuematrix'},~~~~~\label{eq:pf_attn_term1}
         \\
    \termattn_{a2}   &\defn 
    \radius \cdot      \sum_{j=1}^\horizon
      |(\softmaxshort(\<\querymatrix\tokenmatrix_i,\keymatrix\tokenmatrix_j\>)-\softmaxshort(\<\querymatrix'\tokenmatrix_i,\keymatrix'\tokenmatrix_j\>)|\cdot \lops{\valuematrix'}\Big],\\
      &\overset{(ii)}{\leq}  
      2e\bound\radius
      \cdot 
      \max_{j}|\<\querymatrix\tokenmatrix_i,\keymatrix\tokenmatrix_j\>-\<\querymatrix'\tokenmatrix_i,\keymatrix'\tokenmatrix_j\>|\\
       &\overset{(iii)}{\leq}  
      2e\bound^2\radius^3
       \cdot 
      (\lops{\querymatrix-\querymatrix'}+\lops{\keymatrix-\keymatrix'})
      \label{eq:pf_attn_term2}.
\end{align}
In the above equations, step~(i) uses the property of softmax activation that $\sum_{j=1}^\horizon\softmaxshort(\<\querymatrix\tokenmatrix_i,\keymatrix\tokenmatrix_j\>)=1$; step~(ii) follows from \Cref{lm:softmax_lip}; step~(iii) follows from a triangle inequality and the boundedness assumption on $\tokenmatrix,\querymatrix,\keymatrix$, namely,
\begin{align}
  \max_{j}|&\<\querymatrix\tokenmatrix_i,\keymatrix\tokenmatrix_j\>-\<\querymatrix'\tokenmatrix_i,\keymatrix'\tokenmatrix_j\>|\\ 
  &\leq 
   \max_{j}|\<\querymatrix\tokenmatrix_i,\keymatrix\tokenmatrix_j\>-\<\querymatrix'\tokenmatrix_i,\keymatrix\tokenmatrix_j\>|
   +
   |\<\querymatrix'\tokenmatrix_i,\keymatrix\tokenmatrix_j\>-\<\querymatrix'\tokenmatrix_i,\keymatrix'\tokenmatrix_j\>|\\
   &\leq
   \max_{j}\|\tokenmatrix_i\|_2\|\tokenmatrix_j\|_2\lops{\keymatrix}\cdot\lops{\querymatrix-\querymatrix'}+\|\tokenmatrix_i\|_2\|\tokenmatrix_j\|_2\lops{\querymatrix'}\cdot\lops{\keymatrix-\keymatrix'}\\
   &\leq \bound\radius^2 (\lops{\querymatrix-\querymatrix'}+\lops{\keymatrix-\keymatrix'}) .
\end{align}
Putting equation~\eqref{eq:pf_attn_term1}~and~\eqref{eq:pf_attn_term2} together yields the Lipschitz continuity w.r.t. $\Par$.

For token matrix $\tokenmatrix'\in\R^{D\times\horizon}$, let $\outtokenmatrixb'\defn\Attn_{\Par_{\attn}}(\tokenmatrix')+\tokenmatrix'$. Then
\begin{align}
    \outtokenmatrixb'_i\defn[\Attn_{\Par_{\attn}}(\tokenmatrix')]_{i}+\tokenmatrix'_i &= 
        \sum_{j=1}^\horizon
        \softmaxshort(\<\querymatrix\tokenmatrix'_i,\keymatrix\tokenmatrix'_j\>)\cdot\valuematrix\tokenmatrix'_j+\tokenmatrix'_i,~~~~\text{for }i\in[\horizon]. 
\end{align}
Similarly, we have 
$
   \|(\Attn_{\Par_{\attn}}(\tokenmatrix)+\tokenmatrix)-(\Attn_{\Par_{\attn}}(\tokenmatrix')+\tokenmatrix')\|_{2,\infty} 
  = \max_{i\in[\horizon]}\| \outtokenmatrix_i- \outtokenmatrixb'_i\|_2$. For any $i\in[\horizon]$,
\begin{align}
\| \outtokenmatrix_i- \outtokenmatrixb'_i\|_2
&=
\|\sum_{j=1}^\horizon
        \softmaxshort(\<\querymatrix\tokenmatrix_i,\keymatrix\tokenmatrix_j\>)\cdot\valuematrix\tokenmatrix_j+\tokenmatrix_i
        -
        \sum_{j=1}^\horizon
        \softmaxshort(\<\querymatrix\tokenmatrix'_i,\keymatrix\tokenmatrix'_j\>)\cdot\valuematrix\tokenmatrix'_j-\tokenmatrix'_i\|_2\\
        &\leq
        \|\tokenmatrix_i-\tokenmatrix'_i\|_2
         \sum_{j=1}^\horizon  
         \softmaxshort(\<\querymatrix\tokenmatrix_i,\keymatrix\tokenmatrix_j\>)\cdot \lops{\valuematrix}\|\tokenmatrix_j-\tokenmatrix'_j\|_2\\
         &\quad
         +
         \sum_{j=1}^\horizon  
         |\softmaxshort(\<\querymatrix\tokenmatrix_i,\keymatrix\tokenmatrix_j\>)- \softmaxshort(\<\querymatrix\tokenmatrix'_i,\keymatrix\tokenmatrix'_j\>)|\cdot \lops{\valuematrix}\|\tokenmatrix'_j\|_2 
         \\
         &\leq
        (1+\bound) \cdot\|\tokenmatrix-\tokenmatrix'\|_{2,\infty}
        +2 e\bound\radius\cdot \max_{j} |\<\querymatrix\tokenmatrix_i,\keymatrix\tokenmatrix_j\>- \<\querymatrix\tokenmatrix'_i,\keymatrix\tokenmatrix'_j\>|\\
        &\leq 
        (1+\bound(1+4e\bound^2\radius^2))\cdot \|\tokenmatrix-\tokenmatrix'\|_{2,\infty},
\end{align}
where the last line follows from 
\begin{align}
  \max_{j}| &\<\querymatrix\tokenmatrix_i,\keymatrix\tokenmatrix_j\>-\<\querymatrix\tokenmatrix'_i,\keymatrix\tokenmatrix'_j\>| \\ 
  &\leq 
   \max_{j}|\<\querymatrix\tokenmatrix_i,\keymatrix\tokenmatrix_j\>-\<\querymatrix\tokenmatrix'_i,\keymatrix\tokenmatrix_j\>|
   +
   |\<\querymatrix\tokenmatrix'_i,\keymatrix\tokenmatrix_j\>-\<\querymatrix\tokenmatrix'_i,\keymatrix\tokenmatrix'_j\>|\\
   &\leq
   \max_{j}\|\tokenmatrix_j\|_2\lops{\querymatrix}\lops{\keymatrix}\cdot{\|\tokenmatrix_i-\tokenmatrix'_i\|_2}+\|\tokenmatrix'_i\|_2\lops{\querymatrix}\lops{\keymatrix}\cdot\|\tokenmatrix_j-\tokenmatrix'_j\|_2\\
   &\leq 2\bound^2\radius\cdot \|\tokenmatrix-\tokenmatrix'\|_{2,\infty}.
\end{align}
\end{proof}


\begin{lemma}[Lipschitzness of the transformer layer]\label{lm:one_layer_tf_lip}
Consider the parameter space of transformer blocks
\begin{align}
    \ParSpace_{\tf,\bound} \defn\{\Par=(\Par_{\ff},\Par_{\attn}),~\nrmp{\Par}\leq \bound\},
\end{align}
where $\nrmp{\cdot}$ is defined in Eq.~\eqref{eqn:CLIP_param_space}.
 Let $\TF(\cdot):\R^{D\times\horizon}\mapsto\R^{D\times\horizon}$ denote the transformer consists of one attention layer (with normalization) and one $\numfflayer+1$-layer feedforward map, i.e.,
  \begin{align}
      \TF_{\Par_{\tf}}(\tokenmatrix) = \normalize(\Attn_{\Par_{\attn}}(\tokenmatrixb)+\tokenmatrixb),~~\text{where}~~
\tokenmatrixb=\FF_{\Par_{\ff}}(\tokenmatrix)+\tokenmatrix.
  \end{align}
Assume $\bound,\radius \geq1$.    Then for $\tokenmatrix\in\trunspace_{\radius},\Par_\attn\in\ParSpace_{\tf,\bound}$, the function $(\Par_\tf,\tokenmatrix)\mapsto\TF_{\Par_{\tf}}(\tokenmatrix)$ is $\bound_{\TF}(\radius)$-Lipschitz w.r.t. $\Par_{\attn}$ in $\nrmp{\cdot}$ and  $  \bound_{\TF}(\radius)$-Lipschitz w.r.t. $\tokenmatrix$ in $\|\cdot\|_{2,\infty}$, where $\bound_\TF(\radius)\defn (c\bound)^{3\numfflayer+6}\sqrt{\state}\radius^3$ for some numerical constant $c>0$.
\end{lemma}
\begin{proof}[Proof of \Cref{lm:one_layer_tf_lip}]
For any $\Par_{\tf},\Par'_{\tf}\in\ParSpace_{\tf,\bound}$,
let $\tokenmatrixb=\FF_{\Par_{\ff}}(\tokenmatrix)+\tokenmatrix$ and $\tokenmatrixb'=\FF_{\Par'_{\ff}}(\tokenmatrix)+\tokenmatrix$.
 We have
 \begin{align}
     \|\tokenmatrixb-\tokenmatrixb'\|_{2,\infty}
     &=
     \|\FF_{\Par_{\ff}}(\tokenmatrix)
     -\FF_{\Par'_{\ff}}(\tokenmatrix)
     \|_{2,\infty}  
     \leq 
     (\numfflayer+1)\bound^\numfflayer \radius\cdot  \nrmp{\Par_{\ff}-\Par'_{\ff}},
 \end{align}
 where the last step uses \Cref{lm:ffn_lip}. 
 Adopt the shorthand $\normalizeshort(\cdot)$ for $\normalize(\cdot)$.
 Moreover, by the definition of  $\FF(\cdot)$, it can be verified that
 $\|\tokenmatrixb \|_{2,\infty},\|\tokenmatrixb' \|_{2,\infty}\leq (\bound^{\numfflayer+1}+1)\radius$. Therefore,
\begin{align}
    \|\TF_{\Par_\tf}(\tokenmatrix)-\TF_{\Par'_\tf}(\tokenmatrix)\|_{2,\infty}
     &\leq
    \|\normalizeshort(\Attn_{\Par_\attn}(\tokenmatrixb)+\tokenmatrixb)-\normalizeshort(\Attn_{\Par'_\attn}(\tokenmatrixb')+\tokenmatrixb')\|_{2,\infty}
    \\
  &\leq
   2\sqrt{\state} \|\Attn_{\Par_\attn}(\tokenmatrixb)+\tokenmatrixb)-\Attn_{\Par'_\attn}(\tokenmatrixb')+\tokenmatrixb'\|_{2,\infty}
    \\  
    &\leq
  2\sqrt{\state}\cdot[ \|\Attn_{\Par_\attn}(\tokenmatrixb)-\Attn_{\Par'_\attn}(\tokenmatrixb)\|_{2,\infty}
    \\ 
    &\quad \quad +
     \|\Attn_{\Par'_\attn}(\tokenmatrixb)-\Attn_{\Par'_\attn}(\tokenmatrixb')\|_{2,\infty}+\|\tokenmatrixb-\tokenmatrixb'\|_{2,\infty}\,]
    \\
    &\leq
    2\sqrt{\state} \widebar{\radius}(1+4e\bound^2\widebar\radius^2)\cdot\nrmp{\Par_\attn-\Par'_\attn}
    +
    2\sqrt{\state} (2+\bound(1+4e\bound^2\widebar\radius^2))\cdot \|\tokenmatrixb-\tokenmatrixb'\|_{2,\infty},
\end{align}
where $\widebar\radius\defn (\bound^{\numfflayer+1}+1)\radius$ and the third inequality uses \Cref{lm:attention_lip}. Putting pieces together yields
\begin{align}
    \|\TF_{\Par_\tf}(\tokenmatrix)-\TF_{\Par'_\tf}(\tokenmatrix)\|_{2,\infty}\leq \bound_{\TF}(\radius) \cdot \nrmp{\Par_\tf-\Par'_\tf} .
\end{align}
Similarly, for any $\tokenmatrix,\tokenmatrix'\in\trunspace_{\radius}$ and $\Par_{\tf}\in\ParSpace_{\tf,\bound}$, let 
 $\outtokenmatrix=\FF_{\Par_{\ff}}(\tokenmatrix)+\tokenmatrix$ and $\outtokenmatrix'=\FF_{\Par_{\ff}}(\tokenmatrix')+\tokenmatrix'$. Then 
  \begin{align}
     \|\outtokenmatrix-\outtokenmatrix'\|_{2,\infty}
     &=
     \|\FF_{\Par_{\ff}}(\tokenmatrix)+\tokenmatrix
     -\FF_{\Par_{\ff}}(\tokenmatrix')-\tokenmatrix'
     \|_{2,\infty}  \\
     &\leq 
    (1+\bound^{\numfflayer+1})\cdot  \|\tokenmatrix-\tokenmatrix'\|_{2,\infty},
 \end{align}
 where the last step uses \Cref{lm:ffn_lip}. Moreover, basic algebra gives 
$\|\outtokenmatrix \|_{2,\infty},\|\outtokenmatrix' \|_{2,\infty}\leq (\bound^{\numfflayer+1}+1)\radius$. We thus have
\begin{align}
     \|\TF_{\Par_\tf}(\outtokenmatrix)-\TF_{\Par_\tf}(\outtokenmatrix')\|_{2,\infty}
     &\leq
    \|\normalizeshort(\Attn_{\Par_\attn}(\outtokenmatrix)+
    \outtokenmatrix)-\normalizeshort(\Attn_{\Par_\attn}(\outtokenmatrix')+\outtokenmatrix')\|_{2,\infty}
    \\
     &\leq
   2\sqrt{\state} \cdot[\|\Attn_{\Par_\attn}(\outtokenmatrix)-\Attn_{\Par_\attn}(\outtokenmatrix')\|_{2,\infty}
   +\|\outtokenmatrix-\outtokenmatrix'\|_{2,\infty}\,]
    \\
   &\leq  
    2\sqrt{\state} (2+\bound(1+4e\bound^2\widetilde\radius^2))\cdot \|\outtokenmatrix-\outtokenmatrix'\|_{2,\infty}\\
    &\leq  
   \bound_{\TF}(\radius)\cdot \|\tokenmatrix-\tokenmatrix'\|_{2,\infty} 
    ,  
\end{align}
where $\widetilde\radius\defn(\bound^{\numfflayer+1}+1)\radius$ and the third inequality uses \Cref{lm:attention_lip}.
\end{proof}

\begin{lemma}[Lipschitzness of the transformer]\label{lm:tf_lip}
Consider the space
\begin{align}
    \ParSpace_{\tf,\layer,\bound} \defn\{\Par=(\Par^{(1:\layer)}_{\ff},\Par^{(1:\layer)}_{\attn}),~\nrmp{\Par}\leq \bound\},
\end{align}
where $\nrmp{\cdot}$ is as defined in Eq.~\eqref{eqn:CLIP_param_space}.
 Let $\NN_\image^{\Par}(\cdot):[\state]^{\dimimage}\mapsto\R^{\statesize\times\horizon}$ denote the image network that consists of  $\layer$ transformer blocks in Lemma~\ref{lm:one_layer_tf_lip} and the embedding function $\encode(\cdot)$, i.e.,
  \begin{align}
      \NN_\image^{\Par}(\xi) = \TF_{\Par^{(1:\layer)}}(\encode(\xi)).
  \end{align}
    Then for $\Par\in\ParSpace_{\tf,\layer,\bound}$, the function $\xi\mapsto\NN_\image^{\Par}(\xi)$ is $\bound_\NN \defn((c\bound)^{18\numfflayer\layer}\state^4)^\layer$-Lipschitz w.r.t. $\Par$ in $\nrmp{\cdot}$ for some numerical constant $c>0$. Moreover, let $\tokenmatrix\defn\encode(\xi)$. Then 
    the function $\TF_{\Par^{(1:L)}}(\tokenmatrix)$ 
  is $\bound_\NN$-Lipschitz w.r.t. $\tokenmatrix$  in $\|\cdot\|_{2,\infty}$. Same results hold for the text network $\NN_\txt^\Par(\cdot).$
\end{lemma}

\begin{proof}[Proof of \Cref{lm:tf_lip}]
Let $\tokenmatrix = \encode(\xi)$ and $\radius=\radius_\layer = {\state}\sqrt\layer$. For $0\leq i\leq \layer-1$, define
 $\radius_i\defn (2\bound)^{i(\numfflayer+2)}\radius$.
 Then it can be verified by induction that for any $0\leq\ell\leq\layer$
 \begin{align}
 \|\TF_{\Par^{(\layer-\ell+1:\layer)}}(\tokenmatrix)\|_{2,\infty}\leq  \radius_{\layer-\ell}
 \end{align}
 for any $\Par\in\ParSpace_{\tf,\layer,\bound},\tokenmatrix\in\trunspace_\radius$ and $\ell\in[\layer]$.
 With this bound at hand,
    for any $\Par,\widetilde\Par\in\ParSpace_{\tf,\layer,\bound}$
    \begin{align}
   \|\NN_\image^\Par(\tokenmatrix)-\NN_\image^{\widetilde\Par}(\tokenmatrix)\|_{2,\infty}
        &=
       \|
      \TF_\Par(\tokenmatrix)-\TF_{\widetilde\Par}(\tokenmatrix)
       \|_{2,\infty}
       \\
               &\overset{(i)}{\leq}
        \sum_{\ell=1}^\layer 
        \|\TF_{\Par^{(1:\ell-1)}}(\TF_{\Par^{(\ell)}}(\TF_{\widetilde\Par^{(\ell+1:\layer)}}(\tokenmatrix)))
        -
        \TF_{\Par^{(1:\ell-1)}}(\TF_{\widetilde\Par^{(\ell)}}(\TF_{\widetilde\Par^{(\ell+1:\layer)}}(\tokenmatrix)))\|_{2,\infty}\\
        &\overset{(ii)}{\leq}
         \bound \sum_{\ell=1}^\layer 
      \prod_{j=1}^{\ell} \bound_{\TF}(\radius_{\ell})\cdot \nrmp{\Par^{(\ell)}-\widetilde\Par^{(\ell)}}\leq
\layer\bound\cdot  \prod_{j=1}^\layer \bound_{\TF}(\radius_{j})\cdot\nrmp{\Par-\widetilde\Par}
    \end{align}
where  step~(i) follows from a triangle inequality and step~(ii) uses~\Cref{lm:one_layer_tf_lip}. Plugging in the definition of $\bound_\TF(\cdot)$ yields the desired bound.

Similarly, for two embedding matrices $\tokenmatrix,\tokenmatrixb$, we have
\begin{align}
     \|\NN_\image^\Par(\tokenmatrix)-\NN_\image^{\Par}(\tokenmatrixb)\|_{2,\infty}
       &=
       \|
      \TF_\Par(\tokenmatrix)-\TF_{\Par}(\tokenmatrixb)
       \|_{2,\infty}
       \\
               &\overset{}{\leq}
               \prod_{\ell=1}^\layer \bound_{\TF}(\radius_{\ell})\cdot \|\tokenmatrix-\tokenmatrixb\|_{2,\infty}
      \leq
\bound_\NN\cdot  \|\tokenmatrix-\tokenmatrixb\|_{2,\infty}.
\end{align}
where the last line follows from Lemma~\ref{lm:one_layer_tf_lip} and the definition of $\bound_\NN.$

\end{proof}

\begin{lemma}[Properties of the score function]\label{lm:simscore_lip}
Consider the space
\begin{align}
    \ParSpace_{\simscore,\layer,\bound} \defn\{\Par=(\bW_\image,\bW_\txt,\linkweight),~\nrmp{\Par}\leq \bound\},
\end{align}
where $\nrmp{\cdot}$ is defined in Eq.~\eqref{eqn:CLIP_param_space}.
Let  the score function \begin{align} 
\simscore_\NN^\btheta(\xi, \xt) = \NNs^{\linkweight}\big(\softmax(\NN_\image^{\bW_{\image}}(\xi)),\softmax(\NN_\txt^{\bW_{\txt}}(\xt))\big).
\end{align} 
    Then for $\Par\in\ParSpace_{\tf,\layer,\bound}$, the function $(\xi,\xt)\mapsto\simscore_\NN^\btheta(\xi, \xt)$ is $\bound_\scorefun\defn((c\bound)^{18\numfflayer\layer}\state^4)^{\layer+1}$-Lipschitz w.r.t. $\Par$ in $\nrmp{\cdot}$ for all fixed $(\xi,\xt)\in\cXi\times\cXt$ for some numerical constant $c>0$. 
    Moreover, $\exp(\simscore_\NN^\btheta(\xi, \xt))\in[1/c_1,c_1]$ with $c_1=\exp({\readbd)}.$
\end{lemma}

\begin{proof}[Proof of Lemma~\ref{lm:simscore_lip}]
  Use $\softmaxshort(\cdot)$ as a shorthand notation for $\softmax(\cdot)$ and let
  \begin{align}\widetilde\NNs^\linkweight(\bx,\by) \defn \exp(\NNs^\linkweight(\bx,\by))=
 \readsimscore\big(\sum_{s=1}^\state\linkweight_s x_s
 y_s\big),
  \end{align}
  where we recall $\readsimscore(z)\defn \proj_{[\exp(-\readbd),\exp(\readbd)]}(z)$.
  For two parameters $\Par,\widetilde\Par\in\ParSpace_{\state,\layer,\bound}$, we have
    \begin{align}
       &\quad~|\exp(\simscore_\NN^\Par(\xi, \xt))-\exp(\simscore_\NN^{\widetilde\Par}(\xi, \xt))|\\
&=
|\widetilde\NNs^{\linkweight}\big(\softmaxshort(\NN_\image^{\bW_{\image}}(\xi)),\softmaxshort(\NN_\txt^{\bW_{\txt}}(\xt))\big)
       -
\big(\widetilde\NNs^{\widetilde\linkweight}\big(\softmaxshort(\NN_\image^{\widetilde\bW_{\image}}(\xi)),\softmaxshort(\NN_\txt^{\widetilde\bW_{\txt}}(\xt))\big) 
|\\
       &\leq 
      \Sterm_1+  \Sterm_2 +\Sterm_3,
      \end{align}
      where 
      \begin{align}
          \Sterm_1
          &\defn|
       \widetilde\NNs^{\linkweight}\big(\softmaxshort(\NN_\image^{\bW_{\image}}(\xi)),\softmaxshort(\NN_\txt^{\bW_{\txt}}(\xt))\big)
       -
        \widetilde\NNs^{\widetilde\linkweight}\big(\softmaxshort(\NN_\image^{\bW_{\image}}(\xi)),\softmaxshort(\NN_\txt^{\bW_{\txt}}(\xt))\big) 
        |\\
        &\leq
        \|\linkweight-\widetilde\linkweight\|_\infty\cdot\|\softmaxshort(\NN_\image^{\bW_{\image}}(\xi)\|_{1}\cdot\|\softmaxshort(\NN_\txt^{\bW_{\txt}}(\xt)\|_{1}\leq \nrmp{\Par-\widetilde\Par},\vspace{0.2em}
        \\
   \Sterm_2
          &\defn|
       \widetilde\NNs^{\widetilde\linkweight}\big(\softmaxshort(\NN_\image^{\bW_{\image}}(\xi)),\softmaxshort(\NN_\txt^{\bW_{\txt}}(\xt))\big)
       -
        \widetilde\NNs^{\widetilde\linkweight}\big(\softmaxshort(\NN_\image^{\widetilde\bW_{\image}}(\xi)),\softmaxshort(\NN_\txt^{\bW_{\txt}}(\xt))\big) 
        |
        \\
        &\leq
        \|\widetilde\linkweight\|_\infty\cdot
        \|\softmaxshort(\NN_\image^{\bW_{\image}}(\xi))
        -
        \softmaxshort(\NN_\image^{\widetilde\bW_{\image}}(\xi))
        \|_{1}
        \cdot\|\softmaxshort(\NN_\txt^{\bW_{\txt}}(\xt)\|_{\infty}\\
        &       
        \overset{(i)}{\leq}
       2e\bound\cdot
         \|\NN_\image^{\bW_{\image}}(\xi)
        -
       \NN_\image^{\widetilde\bW_{\image}}(\xi)
        \|_{2}\leq 2e\bound\cdot\bound_\NN\cdot\nrmp{\Par-\widetilde\Par}
       , \label{eq:simscore_lip_pf_2}
       \\
       \Sterm_3
          &\defn|
       \widetilde\NNs^{\widetilde\linkweight}\big(\softmaxshort(\NN_\image^{\widetilde\bW_{\image}}(\xi)),\softmaxshort(\NN_\txt^{\bW_{\txt}}(\xt))\big)
       -
        \widetilde\NNs^{\widetilde\linkweight}\big(\softmaxshort(\NN_\image^{\widetilde\bW_{\image}}(\xi)),\softmaxshort(\NN_\txt^{\widetilde\bW_{\txt}}(\xt))\big) 
        |\\
        &\leq
        \|\widetilde\linkweight\|_\infty\cdot
        \|\softmaxshort(\NN_\txt^{\bW_{\txt}}(\xt))
        -
        \softmaxshort(\NN_\txt^{\widetilde\bW_{\txt}}(\xt))
        \|_{1}
        \cdot\|\softmaxshort(\NN_\image^{\widetilde\bW_{\image}}(\xi)\|_{\infty}\\
        &
        \overset{(ii)}{\leq}
       2e\bound\cdot
         \|\NN_\txt^{\bW_{\txt}}(\xt)
        -
       \NN_\txt^{\widetilde\bW_{\txt}}(\xt)
        \|_{2}\leq 2e\bound\cdot\bound_\NN\cdot\nrmp{\Par-\widetilde\Par}
       ,  
      \end{align}
     where step~(i)~and~(ii) uses Lemma~\ref{lm:softmax_lip}.  Putting pieces together we find that $\exp(\simscore_\NN^\Par(\xi,\xt))$ is $(1+4e)\bound\bound_\NN$-Lipschitz continuous in $\Par$ in $\nrmp{\cdot}$.

The upper and lower bounds on $\exp(\scorefun_\NN^\Par(\xi,\xt))$ follows immediately from the definition of the readout function $\readsimscore(\cdot)$ in Eq.~\eqref{eq:readout_def}.

\end{proof}

\begin{lemma}[Lipschitzness of the CLIP representation]\label{lm:clip_repr_lip}
Consider the space
\begin{align}
    \ParSpace_{\clip,\layer,\bound} \defn\{\Par=(\adaweighta,\adaweightb,\bW_\txt),~\nrmp{\Par}\leq \bound\},
\end{align}
where $\nrmp{\cdot}$ is defined in Eq.~\eqref{eqn:CLIP_param_space}. Let $\softmaxshort(\cdot)$ denote the softmax function. 
Under the definition of $\Adapter(\cdot)$ in Eq.~\eqref{eq:adapter_def_cdm},
with slight abuse of notation, let the CLIP representation of the text data
\begin{align}
  \what \Et_{,\Par}(\xt) 
  =\Adapter(\NN_\txt^{\bW_\txt}(\xt))
=\adaweighta\softmaxshort(\log(\readsimscore(\adaweightb_{}\softmaxshort(\NN_\txt^{\bW_\txt}(\xt) )))~),
  \end{align}
  where $\readsimscore(\cdot)$ is the truncation function defined in Eq.~\eqref{eq:readout_def}.
    Then for $\Par\in\ParSpace_{\clip,\layer,\bound}$, the function $ \what\Et_{,\Par}(\xt)$ is 
$\bound_\Adapter\defn\exp(\readbd)((c\bound)^{18\numfflayer\layer}\state^4)^{\layer+1}$-Lipschitz w.r.t. $\Par$ in $\nrmp{\cdot}$ for all  $\xt\in\cXt$ for some numerical constant $c>0$, where $\readbd= 4\mmaxone\log\transbd^{}$.
Moreover,  $ \what\Et_{,\Par}:\R^{\dimtxt}\mapsto\R^{\state}$ is $1$-Lipschitz w.r.t. $\adaweighta$ and $  2e\bound\exp(\readbd)$-Lipschitz w.r.t. $\adaweightb$ in $\lops{\cdot}$. Same results hold for the adapter of the image representation.
\end{lemma}

\begin{proof}[Proof of Lemma~\ref{lm:clip_repr_lip}]
For two parameters $\Par,\widetilde\Par$, we have
\begin{align}
    \| \what \Et_{,\Par}(\xt)- \what \Et_{,\widetilde\Par}(\xt)\|_2
    &\leq
    \Sterm_1+\Sterm_2+\Sterm_3,
\end{align}
where 
\begin{align}
     \Sterm_1&\defn
   \Big\| (\adaweighta-\widetilde\adaweighta) \softmaxshort(\log(\readsimscore(\adaweightb_{}\softmaxshort(\NN_\txt^{\bW_\txt}(\xt) )))~)
    \Big\|_2,
    \\
      \Sterm_2&
      \defn
   \Big\| \widetilde\adaweighta \Big(\softmaxshort(\log(\readsimscore(\widetilde\adaweightb_{}\softmaxshort(\NN_\txt^{\bW_\txt}(\xt) )))~)
   -
\softmaxshort(\log(\readsimscore(\adaweightb_{}\softmaxshort(\NN_\txt^{\bW_\txt}(\xt) )))~)
\Big)
    \Big\|_2,\\ 
     \Sterm_3
     &
      \defn
  \Big\| \widetilde\adaweighta \Big(\softmaxshort(\log(\readsimscore(\widetilde\adaweightb_{}\softmaxshort(\NN_\txt^{\bW_\txt}(\xt) )))~)
   -
\softmaxshort(\log(\readsimscore(\widetilde\adaweightb_{}\softmaxshort(\NN_\txt^{\widetilde\bW_\txt}(\xt) )))~)
\Big) 
    \Big\|_2
    .
\end{align}

By properties of the softmax function, we have
  \begin{align}
      \Sterm_1
      &\leq 
      \lops{\adaweighta-\widetilde\adaweighta} \cdot\|\softmaxshort(\log(\readsimscore(\adaweightb_{}\softmaxshort(\NN_\txt^{\bW_\txt}(\xt) )))~)\|_2\\
      &\leq
        \lops{\adaweighta-\widetilde\adaweighta} \cdot\|\softmaxshort(\log(\readsimscore(\adaweightb_{}\softmaxshort(\NN_\txt^{\bW_\txt}(\xt) )))~)\|_1 =   \|\adaweighta-\widetilde\adaweighta\|_2, \text{~~~and~}\vspace{0.2em}
        \\
        \Sterm_2&
      \leq
      \lops{\widetilde\adaweighta }\cdot
   \Big\| \softmaxshort(\log(\readsimscore(\widetilde\adaweightb_{}\softmaxshort(\NN_\txt^{\bW_\txt}(\xt) )))~)
   -
\softmaxshort(\log(\readsimscore(\adaweightb_{}\softmaxshort(\NN_\txt^{\bW_\txt}(\xt) )))~)
    \Big\|_2\\
    &\overset{(i)}{\leq}
   2e \bound\cdot 
   \|
\log(\readsimscore(\adaweightb_{}\softmaxshort(\NN_\txt^{\bW_\txt}(\xt) )))
   -
\log(\readsimscore(\widetilde\adaweightb_{}\softmaxshort(\NN_\txt^{\bW_\txt}(\xt) )))
\|_\infty\\
&\overset{(ii)}{\leq}
2e\bound\exp(\readbd)\cdot \max_{j\in[\state]}\|\adaweightb_{,j:}-\widetilde\adaweightb_{,j:}\|_2,
  \end{align}
  where  step~(i) uses Lemma~\ref{lm:softmax_lip} and step~(ii) follows from the definition of $\readsimscore(\cdot)$. Similarly, we have
  \begin{align}
      \Sterm_3\leq
  2e\bound^2\exp(\readbd)\cdot\|
   \NN_\txt^{\bW_\txt}(\xt)-
   \NN_\txt^{\widetilde\bW_\txt}(\xt)
   \|_2 
   \leq    2e\bound^2\exp(\readbd)\cdot \bound_\NN\cdot\nrmp{\Par-\widetilde\Par}
  \end{align} by Lemma~\ref{lm:tf_lip}. Putting the bounds on $\Sterm_1,\Sterm_2,\Sterm_3$ together yields Lemma~\ref{lm:clip_repr_lip}.

\end{proof}

\begin{lemma}[Lipschitzness of the conditional diffusion model]\label{lm:cdm_lip}
Consider the space
\begin{align}
    \ParSpace_{\cdm,\layer,\bound} \defn\{\Par=(\adaweighta,\adaweightb,\bW_\cdm),~\nrmp{\Par}\leq \bound\},
\end{align}
where $\nrmp{\cdot}$ is defined in Eq.~\eqref{eq:cdm-empirical-risk-minimization}. 
Let
\begin{align}
 \Mt^\Par(\bz_t, \Et(\xt)) = \read_{\cdm}(\TF_{\cdm}(\encode_{\cdm}(\bz_t, \Adapter(\Et(\xt))))).
  \end{align}
    Then for $\Par\in\ParSpace_{\cdm,\layer,\bound}$, the function $  \Mt^\Par(\bz_t, \Et(\xt))$ is 
$\bound_\cdm$-Lipschitz w.r.t. $\Par$ in $\nrmp{\cdot}$, where
$\bound_\cdm\defn ((c\bound)^{18\numfflayer\layer}\state^9\readbd^3\log^3\mmax)^{2\layer+2}\exp(2\readbd)$ and $\readbd=4\mmaxone\log\transbd+\log\state$
for some numerical constant $c>0$.
\end{lemma}

\begin{proof}[Proof of Lemma~\ref{lm:cdm_lip}]
For any two parameters $\Par,\widetilde\Par\in\ParSpace_{\cdm,\layer,\bound} $, we have
\begin{align}
   &\qquad \| \Mt^\Par(\bz_t, \Et(\xt))- \Mt^{\widetilde\Par}(\bz_t, \Et(\xt))\|_2
   \\
    &\overset{(i)}{\leq}
    c\state\cdot
    \sqrt{\dimimage}\cdot\|\TF^{\bW_\cdm}_{\cdm}(\encode_{\cdm}(\bz_t, \Adapter^{\adaweighta,\adaweightb}(\Et(\xt))))
    \\ 
    &\quad\quad -
    \TF^{\widetilde\bW_\cdm}_{\cdm}(\encode_{\cdm}(\bz_t, \Adapter^{\widetilde\adaweighta,\widetilde\adaweightb}(\Et(\xt))))\|_{2,\infty}\\
    &\leq\Sterm_1+\Sterm_2
\end{align}
for some numerical constant $c>0,$
where step~(i) uses the definition of $\read_\cdm$ in Eq.~\eqref{eq:readout_cdm_def} and Lemma~\ref{lm:softmax_lip}, and
\begin{align}
    \Sterm_1
    &\defn\|\TF^{\bW_\cdm}_{\cdm}(\encode_{\cdm}(\bz_t, \Adapter^{\adaweighta,\adaweightb}(\Et(\xt))))
    -
    \TF^{\widetilde\bW_\cdm}_{\cdm}(\encode_{\cdm}(\bz_t, \Adapter^{\adaweighta,\adaweightb}(\Et(\xt))))\|_{2,\infty}\\
    &\overset{(ii)}{\leq}
    ((c\bound)^{18\numfflayer\layer}\state^9\readbd^3\log^3\mmax)^{2\layer+1}\nrmp{\bW_\cdm-\widetilde\bW_\cdm},\\
      \Sterm_2
    &\defn\|\TF^{\widetilde\bW_\cdm}_{\cdm}(\encode_{\cdm}(\bz_t, \Adapter^{\adaweighta,\adaweightb}(\Et(\xt))))
    -
    \TF^{\widetilde\bW_\cdm}_{\cdm}(\encode_{\cdm}(\bz_t, \Adapter^{\widetilde\adaweighta,\widetilde\adaweightb}(\Et(\xt))))\|_{2,\infty}\\
    &\overset{(iii)}{\leq}
    ((c\bound)^{18\numfflayer\layer}\state^9\readbd^3\log^3\mmax)^{2\layer+1}\exp(\readbd)\cdot \|
    \Adapter^{\adaweighta,\adaweightb}(\Et(\xt)))
    -
    \Adapter^{\widetilde\adaweighta,\widetilde\adaweightb}(\Et(\xt)))
    \|_{2,\infty}
    \\&\overset{(iv)}{\leq}
       ((c\bound)^{18\numfflayer\layer}\state^9\readbd^3\log^3\mmax)^{2\layer+2}\exp(2\readbd)
       \cdot \nrmp{\Par-\widetilde\Par}
\end{align} 
where step~(ii) uses Lemma~\ref{lm:tf_lip} and note that $\|\encode_\cdm(\bz_t,\Adapter(\Et(\xt)))\|_{2,\infty}\leq \radius\defn c\state^{2.5}\log\mmax{\layer}\readbd$ for some numerical constant $c>0,$ step~(iii) follows from Lemma~\ref{lm:tf_lip} and the definition of $\encode_\cdm$, and step~(iv) uses Lemma~\ref{lm:clip_repr_lip}. Putting pieces together yields Lemma~\ref{lm:cdm_lip}.

\end{proof}

\begin{lemma}[Lipschitzness of the vision-language model]\label{lm:vlm_lip}
Consider the space
\begin{align}
\ParSpace_{\vlm,\layer,\bound} \defn\{\Par=(\adaweighta,\adaweightb,\bW_\vlm),~\nrmp{\Par}\leq \bound\},
\end{align}
where $\nrmp{\cdot}$ is defined in Eq.~\eqref{eq:cdm-empirical-risk-minimization}. 
For any $j\in[\dimtxt]$, let
\begin{align}
 \log \probnext^\Par(\xtsub{j}| \xtsub{1:j-1},\Ei(\xi)) 
 =
   \log\circ \read_{\vlm}(\TF_{\vlm}(\encode_{\vlm}(\xtsub{1:j-1}, \Adapter(\Ei(\xi))))).
  \end{align}
    Then for $\Par\in\ParSpace_{\vlm,\layer,\bound}$, the function $ \log \probnext^\Par(\xtsub{j}| \xtsub{1:j-1},\Ei(\xi))   $ is 
$\bound_\vlm$-Lipschitz w.r.t. $\Par$ in $\nrmp{\cdot}$ for all  $(\xi,\xt)\in\cXi\times\cXt$, where
$\bound_\vlm\defn ((c\bound)^{18\numfflayer\layer}\state^4\readbd^3)^{4\layer+3}\exp(2\readbd)$ and $\readbd=4\mmaxone\log\transbd+\log\state$
for some numerical constant $c>0$.
\end{lemma}

\begin{proof}[Proof of Lemma~\ref{lm:vlm_lip}]
Similarly to the proof of Lemma~\ref{lm:cdm_lip}, 
for any two parameters $\Par,\widetilde\Par\in\ParSpace_{\vlm,\layer,\bound} $, we have
\begin{align}
   &\qquad|\log \probnext^\Par(\xtsub{j}| \xtsub{1:j-1},\Ei(\xi))
   -
   \log \probnext^{\widetilde\Par}(\xtsub{j}| \xtsub{1:j-1},\Ei(\xi))|
   \\
    &\overset{(i)}{\leq}
   \|\TF^{\bW_\vlm}_{\vlm}(\encode_{\vlm}(\xtsub{1:j-1}, \Adapter^{\adaweighta,\adaweightb}(\Ei(\xi))))
     \\ 
    &\quad \quad -
    \TF^{\widetilde\bW_\cdm}_{\cdm}(\encode_{\cdm}(\xtsub{1:j-1}, \Adapter^{\widetilde\adaweighta,\widetilde\adaweightb}(\Ei(\xi))))\|_{2,\infty}\\
    &\leq\Sterm_3+\Sterm_4
\end{align}
for some numerical constant $c>0,$
where step~(i) uses the definition of $\read_\vlm$ and Lemma~\ref{lm:log_softmax}, and
\begin{align}
    \Sterm_3
    &\defn\|\TF^{\bW_\vlm}_{\vlm}(\encode_{\vlm}(\xtsub{1:j-1}, \Adapter^{\adaweighta,\adaweightb}(\Ei(\xi))))
    \\ 
    & \quad\quad  -
    \TF^{\widetilde\bW_\vlm}_{\vlm}(\encode_{\vlm}(\xtsub{1:j-1}, \Adapter^{\adaweighta,\adaweightb}(\Ei(\xi))))\|_{2,\infty}\\
    &\overset{(ii)}{\leq}
    ((c\bound)^{18\numfflayer\layer}\state^4\readbd^3)^{4\layer+2}\nrmp{\bW_\vlm-\widetilde\bW_\vlm},\\
      \Sterm_4
    &\defn\|\TF^{\widetilde\bW_\cdm}_{\vlm}(\encode_{\vlm}(\xtsub{1:j-1}, \Adapter^{\adaweighta,\adaweightb}(\Ei(\xi))))
    \\ 
    &\quad\quad -
    \TF^{\widetilde\bW_\vlm}_{\vlm}(\encode_{\vlm}(\xtsub{1:j-1}, \Adapter^{\widetilde\adaweighta,\widetilde\adaweightb}(\Ei(\xi))))\|_{2,\infty}\\
    &\overset{(iii)}{\leq}
    ((c\bound)^{18\numfflayer\layer}\state^4\readbd^3)^{4\layer+2}\exp(\readbd)\cdot \|
    \Adapter^{\adaweighta,\adaweightb}(\Ei(\xi)))
    -
    \Adapter^{\widetilde\adaweighta,\widetilde\adaweightb}(\Ei(\xi)))
    \|_{2,\infty}
    \\&\overset{(iv)}{\leq}
       ((c\bound)^{18\numfflayer\layer}\state^4\readbd^3)^{4\layer+3}\exp(2\readbd)
       \cdot \nrmp{\Par-\widetilde\Par},
\end{align} 
where step~(ii) follows from a modified version of Lemma~\ref{lm:tf_lip} (note that one layer of transformer used in VLMs can be represented by two layers of transformer used in Lemma~\ref{lm:tf_lip}), and the fact that \begin{align}
\|\encode_\vlm(\xtsub{1:j-1},\Adapter(\Ei(\xi)))\|_{2,\infty}\leq \radius\defn c\state\readbd
\end{align} 
for some numerical constant $c>0,$ step~(iii) follows from Lemma~\ref{lm:tf_lip} and the definition of $\encode_\vlm$, and step~(iv) uses Lemma~\ref{lm:clip_repr_lip}. Combining the bounds yields Lemma~\ref{lm:vlm_lip}.

\end{proof}

\subsection{Lipschitzness of basic operations}\label{subsection:Lip-Basic-Functions}
\begin{lemma}[Lipschitzness of the normalization operator,  $\|\cdot\|_{\infty,\infty}$-norm]\label{lemma:Appendix-CLIP-Lipschitz-normalization}
    Let $\normalize(h)_s:=h_s-\max_{s'}h_{s'}$ for $h\in \R^S$. 
    For $h,h'\in\R^S$, 
    \begin{align}
    \label{eq:Appendix-CLIP-Lipschitz-Normalize-1}
        \|\normalize(h)-\normalize(h')\|_\infty\leq 2\|h-h'\|_\infty.
    \end{align}
    Therefore, for $q_{i},q'_{i}\in \R^S\ (i=1,2,\dots,m)$, we have
    \begin{align}
    \label{eq:Appendix-CLIP-Lipschitz-Normalize-2}
     \textstyle 
     \|\normalize(\sum_{i}q_{i})-\normalize(\sum_i q'_{i})\|_\infty\leq 2m\max_i\|q_{i}-q_{i}'\|_\infty.
    \end{align}
\end{lemma}
\begin{proof}
    Note that $|\max_s h_s - \max_s h'_s| \leq \|h-h'\|_\infty$.
    For each coordinate, we have
    \begin{align}
        \normalize(h)_s-\normalize(h')_s 
       = (h_s - \max_{s'} h_{s'}) - (h'_s - \max_{s'} h'_{s'})
       = h_s-h'_s - (\max_{s} h'_{s'}-\max_{s'} h'_{s'}).
    \end{align}
    Here $h_s-h'_s$ and $\max_{s} h'_{s'}-\max_{s'} h'_{s'}$ are bounded by $\|h-h'\|_\infty$, which yields \eqref{eq:Appendix-CLIP-Lipschitz-Normalize-1}. 

    \eqref{eq:Appendix-CLIP-Lipschitz-Normalize-2} follows in a straightforward way:
    \begin{align}
        \textstyle
        \|\normalize(\sum_{i}q_{i})-\normalize(\sum_i q'_{i})\|_\infty
        \leq 2\|\sum_{i}q_{i}-\sum_{i}q_{i}'\|_\infty \leq 2m \max_{i}\|q_{i}-q_{i}'\|_\infty.
    \end{align}
\end{proof}

\begin{lemma}[Lipschitzness of the normalized layer, $\|\cdot\|_{2,2}$-norm]\label{lm:normalized_lip}
For any $\tokenmatrix\in\R^{D\times\horizon}$, the normalized function $\normalize(\cdot)$ defined in Eq.~\eqref{eq:clip_normalize_def} is
 $2\sqrt{\state}$-Lipschitz w.r.t. $\tokenmatrix$ in $\|\cdot\|_{2,\infty}$.
\end{lemma}
\begin{proof}
The lemma follows by noting that
\begin{align}
\left\| 
\begin{bmatrix}
    \qnetwork^{(1)} - \ones_S \max_{s \in S} \qnetwork^{(1)}_s \\
    \qnetwork^{(2)} - \ones_S \max_{s \in S} \qnetwork^{(2)}_s \\
    \vdots \\
    \qnetwork^L - \ones_S \max_{s \in S} \qnetwork^L
\end{bmatrix}
\right\|_2 
\leq 
\left\| 
\begin{bmatrix}
    \qnetwork^{(1)}  \\
    \qnetwork^{(2)}  \\
    \vdots \\
    \qnetwork^{(L)} 
\end{bmatrix}
\right\|_2
+
\left\| 
\begin{bmatrix}
   \max_{s \in S} \qnetwork^{(1)}_s \\
   \max_{s \in S} \qnetwork^{(2)}_s  \\
    \vdots \\
   \max_{s \in S} \qnetwork^{(L)}_s
\end{bmatrix}
\right\|_2
\cdot \|\mathbf{1}_\state\|_2
\leq 
2\sqrt{\state}
\cdot\left\| 
\begin{bmatrix}
    \qnetwork^{(1)}  \\
    \qnetwork^{(2)}  \\
    \vdots \\
    \qnetwork^{(L)} 
\end{bmatrix}
\right\|_2.
\end{align}
\end{proof}

\begin{lemma}[Lipschitzness of the softmax activation]\label{lm:softmax_lip}
For any $\bu,\bv\in\R^{\horizon}$, we have
\begin{align}
    \Big\|\frac{e^{\bu}}{\|e^{\bu}\|_1}-\frac{e^{\bv}}{\|e^{\bv}\|_1}\Big\|_1
    \leq 2e\cdot \|\bu-\bv\|_\infty.
\end{align} 
\end{lemma}
\begin{proof}[Proof of \Cref{lm:softmax_lip}]
Write $\bu=(u_1,\ldots,u_\horizon)$ and $\bv=(v_1,\ldots,v_\horizon)$. 
By definition of $\ell_1$-norm, we have
\begin{align}
    \Big\|\frac{e^{\bu}}{\|e^{\bu}\|_1}-\frac{e^{\bv}}{\|e^{\bv}\|_1}\Big\|_1
    &=   \sum_{i=1}^\horizon\frac{|e^{u_i}\|e^{\bv}\|_1-e^{v_i}\|e^{\bu}\|_1|}{\|e^{\bu}\|_1\|e^{\bv}\|_1}
    \leq
    \sum_{i=1}^\horizon\frac{|e^{u_i}-e^{v_i}|\cdot \|e^{\bv}\|_1}{\|e^{\bu}\|_1\|e^{\bv}\|_1} +
     \sum_{i=1}^\horizon\frac{e^{v_i}\cdot|\|e^{\bv}\|_1-\|e^{\bu}\|_1|}{\|e^{\bu}\|_1\|e^{\bv}\|_1}\\
  &=
     \sum_{i=1}^\horizon\frac{|e^{u_i}-e^{v_i}|}{\|e^{\bu}\|_1}
      +
      \frac{|\|e^{\bv}\|_1-\|e^{\bu}\|_1|}{\|e^{\bu}\|_1}
     \overset{(i)}{\leq}
      2\sum_{i=1}^\horizon\frac{e^{u_i+|u_i-{v_i}|}\cdot|u_i-{v_i}|}{\|e^{\bu}\|_1}\\
      &\leq 2 e^{\|\bu-\bv\|_\infty}\cdot \|\bu-\bv\|_\infty,
\end{align} 
where step~(i) follows from a triangle inequality and the fact that $|e^x-e^y|\leq  e^{x+|x-y|}\cdot|x-y|$ for all $x,y\in\R$.
When $\|\bu-\bv\|_2\leq1$, it follows that
\begin{align}
      \Big\|\frac{e^{\bu}}{\|e^{\bu}\|_1}-\frac{e^{\bv}}{\|e^{\bv}\|_1}\Big\|_1
      \leq 2e\cdot\|\bu-\bv\|_\infty.
\end{align}When $\|\bu-\bv\|_{\infty}\geq 1$, since $
\|\frac{e^{\bu}}{\|e^{\bu}\|_1}\|_1
=
\|\frac{e^{\bv}}{\|e^{\bv}\|_1}\|_1=1,$ we have
\begin{align}
    \Big\|\frac{e^{\bu}}{\|e^{\bu}\|_1}-\frac{e^{\bv}}{\|e^{\bv}\|_1}\Big\|_1
      \leq
       \Big\|\frac{e^{\bu}}{\|e^{\bu}\|_1}\Big\|_1+\Big\|\frac{e^{\bv}}{\|e^{\bv}\|_1}\Big\|_1
    =
      2\leq 
      2e\cdot\|\bu-\bv\|_\infty.  
\end{align}
Combining the two cases completes the proof.
\end{proof}

\begin{lemma}[Lipschitzness of log-softmax]\label{lm:log_softmax}
    For any $\bu,\bv\in\R^N$, we have 
    \begin{align}
    \linf{\log \Big(\frac{e^\bu}{\lone{e^\bu}}\Big)-\log \Big(\frac{e^\bv}{\lone{e^\bv}}\Big)}\leq 2\linf{\bu-\bv}
    \end{align}
\end{lemma}
\begin{proof}[Proof of Lemma~\ref{lm:log_softmax}]
Let $\bw:=\bu-\bv$. Then
  \begin{align}
  &~~~~ \linf{\log \Big(\frac{e^\bu}{\lone{e^\bu}}\Big)-\log \Big(\frac{e^\bv}{\lone{e^\bv}}\Big)}
    \leq
    \linf{\bu-\bv}+|{\log \lone{e^\bu}-\log \lone{e^\bv}}|\\
    &\overset{(i)}{=}
    \linf{\bu-\bv}+\int_{0}^1 \<\frac{e^{\bv+t\bw}}{\lone{e^{\bv+t\bw}}},\bw\> dt
     \leq 
    \linf{\bu-\bv}+\int_{0}^1 \lone{\frac{e^{\bv+t\bw}}{\lone{e^{\bv+t\bw}}}}\cdot\linf{\bw} dt\\
    &=2\linf{\bu-\bv},
    \end{align} where step~(i) uses the Newton-Leibniz formula.
\end{proof}

\begin{lemma}[Lipschitzness of log-sum-exponential]\label{lemma:Appendix-CLIP-Lipschitz-logsumexp}
    For $h\in \R^S$ and $\Psi\in \R_+^{S\times S}$, define $f(h)\in \R^S$ by
    \begin{align}
      \textstyle  f(h)_s:= \log \sum_{s'\in [S]}\Psi_{ss'}e^{h_{s'}}.
    \end{align}
    Then, for $h,h'\in \R^S$, we have
    \begin{align}
         \|f(h)- f(h')\|_\infty \leq \|h-h'\|_\infty.
    \end{align}
\end{lemma}
\begin{proof}
    By differentiating $f_s$, we have
    \begin{align}
        [\nabla f(h)]_s = \Big( \frac{\Psi_{ss'}\exp(h_{s'})}{\sum_{s'' \in [S]} \Psi_{ss''} \exp(h_{s''})} \Big)_{s'},
    \end{align}
    which implies that $\|\nabla f(h)\|_1\leq 1$ for all $h$.
    Therefore, 
    \begin{align}
        \|f(h)- f(h')\|_\infty \leq \|\nabla f\|_1 \|h-h'\|_\infty \leq  \|h-h'\|_\infty.
    \end{align}
\end{proof}
\begin{lemma}[Lipschitzness of log-sum-softmax]\label{lemma:Appendix-CLIP-Lipschitz-logsumsoftmax}
    For $h,h'\in \R^S$ and $\Psi\in \R_+^{S}$, define $f(h,h)\in \R$ by
    \begin{align}
      \textstyle  f(h_1,h_2):= \log \sum_{s\in [S]}\Psi_{s}\softmax(h)_s\softmax(h')_s.
    \end{align}
    Then, for all $h_1,h_2\in \R^S$ and $h_1',h_2'\in \R^S$, we have
    \begin{align}
         \|f(h_1,h_2)- f(h_1',h_2')\|_\infty \leq \|h_1-h_1'\|_\infty+\|h_2-h_2'\|_\infty.
    \end{align}
\end{lemma}
\begin{proof}
    Let us first fix $h_2$.
    By differentiating $f$ by $h_1$, we have
    \begin{align}
        [\nabla_{h_1} f(h_1,h_2)]_s = \frac{\Phi_{s}[\softmax(h_1)_s-(\softmax(h_1)_s)^2]\softmax(h_2)_{s}}{\sum_{s'\in [S]}\Phi_{s'} \softmax(h_1)_{s'} \softmax(h_2)_{s'}}.
    \end{align}
    By following the argument of \Cref{lemma:Appendix-CLIP-Lipschitz-logsumexp}, we have
    \begin{align}
        \|f(h_1,h_2)- f(h_1',h_2)\|_\infty \leq \|h_1-h_1'\|_\infty.
    \end{align}
    In the same way, we have
    \begin{align}
        \|f(h_1',h_2)- f(h_1',h_2')\|_\infty \leq \|h_2-h_2'\|_\infty.
    \end{align}
    Adding these two bounds together, we obtain the assertion.
\end{proof}

\subsection{Properties of empirical processes}\label{sec:properties_of_empirical_processes}
\begin{lemma}[Proposition A.4 of~\cite{bai2024transformers}]\label{lm:generalization_lemma}
Let $\{X_\Par\}_{\Par \in \ParSpace}$ be a zero-mean random process defined as
\[
X_w = \frac{1}{n} \sum_{i=1}^n f(z_i; \Par) - \mathbb{E}_z[f(z; \Par)],
\]
where $z_1, \dots, z_n$ are i.i.d. samples from a distribution $\mathbb{P}_z$. Assume the following conditions hold:
\begin{itemize}
\item[(a)] The index set $\ParSpace$ is equipped with a metric $\rho$ and has a diameter $B_\rho$. Furthermore, there exists a constant $A_\rho$ such that for any subset $\ParSpace'$ of radius $r$ in $\ParSpace$, the covering number satisfies:
\[
\log \mathcal{N}(\Delta; \ParSpace', \rho) \leq d_\rho \log \frac{2 A_\rho r}{\Delta}, \quad \forall 0 < \Delta \leq 2r.
\]
\item[(b)] For any fixed $\Par \in \ParSpace$ and $z$ sampled from $\mathbb{P}_z$, the random variable $f(z; \Par) - \mathbb{E}_z[f(z; \Par)]$ is $\sigma$-sub-Gaussian. That is,
\[
\mathbb{E}\left[e^{\lambda (f(z; w) - \mathbb{E}_z[f(z; w)])}\right] \leq e^{\lambda^2 \sigma^2 / 2}, \quad \forall \lambda \in \mathbb{R}.
\]
\item[(c)] For any $\Par, \Par' \in \ParSpace$ and $z$ sampled from $\mathbb{P}_z$, the random variable $f(z; \Par) - f(z; \Par')$ is $\sigma' \rho(\Par, \Par')$-sub-Gaussian. That is,
\[
\mathbb{E}\left[e^{\lambda (f(z; \Par) - f(z; \Par'))}\right] \leq e^{\lambda^2 (\sigma')^2 \rho^2(\Par, \Par') / 2}, \quad \forall \lambda \in \mathbb{R}.
\]
\end{itemize}

\noindent Under these assumptions, with probability at least $1 - \eta$, we have
\[
\sup_{\Par \in \ParSpace} |X_\Par| \leq c \sigma \sqrt{\frac{d_\rho \log \left(2A_\rho \left(1 + B_\rho \sigma' / \sigma\right)\right) + \log(1 / \eta)}{n}},
\]
where $c>0$ is some numerical constant.
\end{lemma}

{
\begin{lemma}[The empirical InfoNCE loss]\label{lm:empirical_infonce_loss_bounded_diff_concen}
For any function $f:\cXi\times\cXt\mapsto \R$ such that $\|f\|_\infty\leq B$,
define the empirical InfoNCE loss as
 \begin{align}
  \what\risk_{\clip,K}(f) \defn - \frac{1}{K}\sum_{k=1}^{K} \log \frac{\exp(f(\xi_{,k}^{(i)}, \xt_{,k}^{(i)})  ) }{\sum_{j \in [K]} \exp(f(\xi_{,k}^{(i)}, \xt_{, j}^{(i)}))}
    - \frac{1}{K}\sum_{k=1}^{K}\log \frac{\exp(f(\xi_{,k}^{(i)}, \xt_{,k}^{(i)})  ) }{\sum_{j \in [K]} \exp(f(\xi_{, j}^{(i)}, \xt_{,k}^{(i)}))}.
 \end{align}
 Then,
 \begin{align}
     \what\risk_{\clip,K}(f) - \mathbb{E}[\what\risk_{\clip,K}(f)]
 \end{align}
 is $c\min\{B, e^{2B}/\sqrt{K}\}$-sub-Gaussian for some universal constant $c>0$. Moreover, for any $f_1,f_2:\cXi\times\cXt\mapsto \R$ such that $\|f_1\|_\infty,\|f_2\|_\infty\leq B$, we have
$\what\risk_{\clip,K}(f_1)-\mathbb{E}[\what\risk_{\clip,K}(f_1)]-(\what\risk_{\clip,K}(f_2)-\mathbb{E}[\what\risk_{\clip,K}(f_2)])$ is $4\|f_1-f_2\|_\infty$-sub-Gaussian.
\end{lemma}
\begin{proof}[Proof of Lemma~\ref{lm:empirical_infonce_loss_bounded_diff_concen}]
    By considering the case where the denominator’s exponent has content $\pm B$ and the numerator’s exponent has content $\mp B$, it is easy to see that the difference between the maximum and minimum values of $\widehat{\risk}_{\clip,K}(f)$ is bounded by $8B$. Therefore, $\what\risk_{\clip,K}(f) - \mathbb{E}[\widehat{\risk}_{\clip,K}(f)]$ is $4B$-sub-Gaussian.

    Next, we show that $\what\risk_{\clip,K}(f) - \mathbb{E}[\what\risk_{\clip,K}(f)]$ is $\frac{c e^{2B}}{\sqrt{K}}$-sub-Gaussian.
    By symmetry, it suffices to consider the behavior of the first term of $\what\risk_{\clip,K}(f)$:
    \begin{align}
        \frac{1}{K}\sum_{k=1}^{K} \log \frac{\exp(f(\xi_{,k}^{(i)}, \xt_{,k}^{(i)})  ) }{\sum_{j \in [K]} \exp(f(\xi_{,k}^{(i)}, \xt_{, j}^{(i)}))}.
        \label{eq:empirical_infonce_loss_bounded_diff_concen_image}
    \end{align}
    To apply the concentration properties of functions with bounded differences, we evaluate the deviation when one of $(\xi_{,k}^{(i)}, \xt_{,j}^{(i)})_{k \in [K]}$ is replaced.
    
    By replacing $\xi_{,l}^{(i)}$ with $\xib_{,l}^{(i)}$, the variation of the first term is as follows:
    \begin{align}
       & \frac{1}{K}\sum_{k\ne l} \log \frac{\exp(f(\xi_{,k}^{(i)}, \xt_{,k}^{(i)})  ) }{\sum_{j \in [K]} \exp(f(\xi_{,k}^{(i)}, \xt_{,j}^{(i)}))} + \frac{1}{K} \log \frac{\exp(f(\xib_{,l}^{(i)}, \xt_{,l}^{(i)})  ) }{\sum_{j \in [K]} \exp(f(\xib_{,l}^{(i)}, \xt_{,j}^{(i)}))}
     \\  & \quad -
        \frac{1}{K}\sum_{k=1}^{K} \log \frac{\exp(f(\xi_{,k}^{(i)}, \xt_{,k}^{(i)})  ) }{\sum_{j \in [K]} \exp(f(\xi_{,k}^{(i)}, \xt_{, j}^{(i)}))}
     \\ &   = \frac{1}{K} \log \frac{\exp(f(\xib_{,l}^{(i)}, \xt_{,l}^{(i)})  ) }{\sum_{j \in [K]} \exp(f(\xib_{,l}^{(i)}, \xt_{,j}^{(i)}))} - \frac{1}{K} \log \frac{\exp(f(\xi_{,l}^{(i)}, \xt_{,l}^{(i)})  ) }{\sum_{j \in [K]} \exp(f(\xi_{,l}^{(i)}, \xt_{,j}^{(i)}))} \label{eq:empirical_infonce_loss_bounded_diff_concen_image_diff}.
    \end{align}
    Using the boundedness of $f$, \eqref{eq:empirical_infonce_loss_bounded_diff_concen_image_diff} is upper and lower bounded by $\frac{2B}{K}$ and $-\frac{2B}{K}$, respectively. 

    By replacing $\xt_{, l}^{(i)}$ with $\xtb_{, l}^{(i)}$, the variation of the first term is as follows:
    \begin{align}
        & \frac{1}{K}\sum_{k\ne l} \log \frac{\exp(f(\xi_{,k}^{(i)}, \xt_{,k}^{(i)})  ) }{\sum_{j\ne l} \exp(f(\xi_{,k}^{(i)}, \xt_{,j}^{(i)})) + \exp(f(\xi_{,k}^{(i)}, \xtb_{,l}^{(i)}))} 
     \\  & \quad 
     + \frac{1}{K} \log \frac{\exp(f(\xi_{,l}^{(i)}, \xtb_{,l}^{(i)})  ) }{\sum_{j \ne l} \exp(f(\xi_{,k}^{(i)}, \xtb_{,j}^{(i)})) + \exp(f(\xi_{,k}^{(i)}, \xtb_{,l}^{(i)}))} - \frac{1}{K}\sum_{k=1}^{K} \log \frac{\exp(f(\xi_{,k}^{(i)}, \xt_{,k}^{(i)})  ) }{\sum_{j \in [K]} \exp(f(\xi_{,k}^{(i)}, \xt_{, j}^{(i)}))}
     \\  & = \log \frac{\sum_{j \in [K]} \exp(f(\xi_{,k}^{(i)}, \xt_{, j}^{(i)}))}{\sum_{j\ne l} \exp(f(\xi_{,k}^{(i)}, \xt_{,j}^{(i)})) + \exp(f(\xi_{,k}^{(i)}, \xtb_{,l}^{(i)}))} + \frac1K \log \frac{\exp(f(\xi_{,l}^{(i)}, \xtb_{,l}^{(i)})  ) }{\exp(f(\xi_{,l}^{(i)}, \xt_{,l}^{(i)})  ) }.
     \label{eq:empirical_infonce_loss_bounded_diff_concen_text_diff}
    \end{align}
    Using the boundedness of $f$ and the fact that $\log(1+x)\leq x$  for $x>-1$, it can be verified that \eqref{eq:empirical_infonce_loss_bounded_diff_concen_text_diff} is upper and lower bounded by $\frac{2e^{2B}}{K}$ and $-\frac{2e^{2B}}{K}$, respectively. 

    Combining the two bounds above, replacing one sample $(\xi_{,l}^{(i)}, \xt_{,l}^{(i)})$ with $(\xib_{,l}^{(i)}, \xtb_{,l}^{(i)})$ changes \eqref{eq:empirical_infonce_loss_bounded_diff_concen_image} at most $\frac{4e^{2B}+4B}{K}(\leq \frac{6e^{2B}}{K})$.
    By concentration properties of functions with bounded differences (e.g., Corollary~2.21 in~\cite{wainwright_2019}), the deviation of \eqref{eq:empirical_infonce_loss_bounded_diff_concen_image} from its mean is $\frac{3e^{2B}}{\sqrt{K}}$-sub-Gaussian.
    Therefore, $\what\risk_{\clip,K}(f) - \mathbb{E}[\what\risk_{\clip,K}(f)]$ is $\frac{6e^{2B}}{\sqrt{K}}$-sub-Gaussian.
    Now, the first assertion of this lemma holds with $c=6$.

    For the second assertion, Lemma~\ref{lm:log_softmax} implies that
    \begin{align}
         |\what\risk_{\clip,K}(f_1) -  \what\risk_{\clip,K}(f_2)| \leq 4\|f_1-f_2\|_\infty,
    \end{align}
    for a fixed $(\xi_{,k}^{(i)}, \xt_{,k}^{(i)})_{k\in [K]}$.
    This directly implies that $\what\risk_{\clip,K}(f_1)-\mathbb{E}[\what\risk_{\clip,K}(f_1)]-(\what\risk_{\clip,K}(f_2)-\mathbb{E}[\what\risk_{\clip,K}(f_2)])$ is $4\|f_1-f_2\|_\infty$-sub-Gaussian.

\end{proof}

\begin{remark}[Exponential dependency on $\overline{m}$]\label{remark:exponential_dependency_on_B}
    In the proof of \Cref{theorem:CLIP-Generalization} in Appendix~\ref{subsection:Proof-of-CLIP}, we have only used $cB$-sub-Gaussianity of $\what\risk_{\clip,K}(f) - \mathbb{E}[\what\risk_{\clip,K}(f)]$.
    By applying the fact that $\what\risk_{\clip,K}(f) - \mathbb{E}[\what\risk_{\clip,K}(f)]$ is $ce^{2B}/\sqrt{K}$-sub-Gaussian instead, it is easy to see that the rate becomes
    \begin{align}
    \bigOtil \Bigg( \sqrt{\frac{ \state^2 \layer^{11}e^{4\mmaxb}\mmaxb^2+\log(1/\eta)}{\numbatch K}} \Bigg).
\end{align}
    Here, the denominator involves $K$, and the convergence rate decreases with the power $-\tfrac{1}{2}$ of the total number of data, $\numbatch \times K$, rather than with the number of batches, $\numbatch$.
    In exchange for the better dependence on $K$, the bound exponentially depends on $\mmaxb$.
    Viewing $\mmaxb$ as a constant makes this permissible; nevertheless, for this reason, in \Cref{theorem:CLIP-Generalization} we adopt rates that are polynomial in all parameters.
    
    We believe that, in order to obtain the factor $1/\sqrt{K}$ in the sub-Gaussian parameter in Lemma~\ref{lm:empirical_infonce_loss_bounded_diff_concen}, an exponential dependence on $B$ is unavoidable.
    We will provide an example to justify this.
    To simplify the argument, consider the case when $f(\xi_{,k}^{(i)},\xt_{,j}^{(i)})$ only depends on $\xi_{,k}^{(i)}$ and we will write $f(\xi_{,k}^{(i)},\xt_{,j}^{(i)})=y_k$.
    Define the distribution of $y_k$ as
    \begin{align}
        y_k = 
        \begin{cases}
            B & (\text{w.p. }e^{-B/2})
            \\
            -\frac{B}{e^{B/2}-1} & (\text{w.p. }1-e^{-B/2})
        \end{cases}.
    \end{align}
    Then, the mean and variance of $e^{y_k}$ are given respectively as follows:
    \begin{align}
    \mathbb{E}[e^{y_k}]
   & = e^{B/2} + \left(1-e^{-B/2}\right)\exp\!\left(-\frac{B}{e^{B/2}-1}\right) = \Theta(e^{B/2}),
    \\
    \mathrm{Var}(e^{y_k})
   & = e^{3B/2} + \left(1-e^{-B/2}\right)\exp\!\left(-\tfrac{2B}{e^{B/2}-1}\right)
    - \left( e^{B/2} + \left(1-e^{-B/2}\right)\exp\!\left(-\tfrac{B}{e^{B/2}-1}\right)\right)^2 = \Theta(e^{3B/2}).
    \end{align}
    The difference in order between $\mathbb{E}[e^{y_k}]$ and $\sqrt{\mathrm{Var}(e^{y_k})}$ will become important later.

Then, $ \what\risk_{\clip,K}(f) $ is simplified as
\begin{align}
  \what\risk_{\clip,K}(f) &= - \frac{1}{K}\sum_{k=1}^{K} \log \frac{\exp(f(\xi_{,k}^{(i)}, \xt_{,k}^{(i)})  ) }{\sum_{j \in [K]} \exp(f(\xi_{,k}^{(i)}, \xt_{, j}^{(i)}))}
    - \frac{1}{K}\sum_{k=1}^{K}\log \frac{\exp(f(\xi_{,k}^{(i)}, \xt_{,k}^{(i)})  ) }{\sum_{j \in [K]} \exp(f(\xi_{, j}^{(i)}, \xt_{,k}^{(i)}))}
    \\ &= - \frac{1}{K}\sum_{k=1}^{K} \log \frac{e^{y_k}}{\sum_{j \in [K]} e^{y_k}}
    - \frac{1}{K}\sum_{k=1}^{K}\log \frac{e^{y_k}}{\sum_{j \in [K]} e^{y_j}}
  \\ &= 2\log K + \log \frac1K\sum_{j \in [K]}e^{y_j}
    - \frac{1}{K}\sum_{k=1}^{K}y_k.
    \label{eq:remark-exp-dep-on-B-simple-target}
 \end{align}
 When $K$ is sufficiently large so that $\frac1K\sum_{j \in [K]}e^{y_j}$ and $\frac{1}{K}\sum_{k=1}^{K}y_k$ concentrate around their expectations, \eqref{eq:remark-exp-dep-on-B-simple-target} is approximated as
 \begin{align}
 \what\risk_{\clip,K}(f) &= 2\log K  + \log \bigg(1+\frac{1}{K}\sum_{k=1}^{K}\frac{e^{y_k}-\mathbb{E}[e^{y_k}]}{\mathbb{E}[e^{y_k}]}\bigg)+ \log \mathbb{E}[e^{y_1}] + \frac{1}{K}\sum_{k=1}^{K}(y_k-\mathbb{E}[y_k]) + E[y_1]
    \\ &  \approx \frac{1}{K}\sum_{k=1}^{K}\bigg[\frac{e^{y_k}-\mathbb{E}[e^{y_k}]}{\mathbb{E}[e^{y_k}]}+y_k-\mathbb{E}[y_k]\bigg](+\text{const.}).
 \end{align}
 Remembering that $\mathbb{E}[e^{y_k}]=\Theta(e^{B/2})$ and $\mathrm{Var}(e^{y_k})=\Theta(e^{3B/2})$, 
 the variance of $\frac{e^{y_k}-\mathbb{E}[e^{y_k}]}{\mathbb{E}[e^{y_k}]}$ is $\Theta(e^{B/2})$, and thus $\what\risk_{\clip,K}(f) - \mathbb{E}[\what\risk_{\clip,K}(f) ] $ is of order $e^{B/4}/\sqrt{K}$. As a consequence, the sub-Gaussian parameter of $\what\risk_{\clip,K}(f) - \mathbb{E}[\what\risk_{\clip,K}(f) ] $ is at least exponentially dependent on $B.$
 \end{remark}
}

\section{Experimental details}\label{sec:simulations-detail}

This section provides details for the experimental results presented in \Cref{sec:simulations}, along with additional experiments.

\subsection{Experimental setup}

\paragraph{The JGHM data distribution.} \label{sec:simu-detail-jghm} 
We generate the dataset from the distribution of Joint Generative Hierarchical Model (JGHM) of \Cref{sec:GHMs}. The root distribution $\P(x_{\root})$ is taken to be uniform over $\state$ states. The transition functions 
\(\{\psi^{(\ell)}_{\square,\iota}\}_{\square \in \{\image,\txt\},\,\iota \in [\state],\,\ell \in [L]}\) 
are constructed as follows:
\begin{align}
&~[\psi^{(\ell)}_{\square,\iota}(s, s') ]_{s, s' \in [\state]} = (1 - \pflip) \times \Perm_{\square, \iota}^{(\ell)} + \pflip \times \softmax_{\row}(\bG_{\square, \iota}^{(\ell)}), ~~\square \in \{ \image, \txt\}, \iota \in [\state], \ell \in [L], \\
&~(\Perm_{\square, \iota}^{(\ell)}, \bG_{\square, \iota}^{(\ell)} ) \sim_{iid} (\Perm, \bG), ~~\Perm, \bG \in \R^{\state \times \state},  \Perm \text{ is a random permutation matrix}, \bG \text{ has iid Gaussian entries}. 
\end{align}
This formulation implies that for each parent-child pair $(x_v^{(\ell-1)}, x_{v'}^{(\ell)})$ where $x_v^{(\ell-1)} = s$, the child node $x_{v'}^{(\ell)}$ takes the value $\Perm_{\square, \iota}^{(\ell)}(s)$ (corresponding to the non-zero element in the $s$-th row of $\Perm_{\square, \iota}^{(\ell)}$) with probability $(1 - \pflip)$. 
With probability $\pflip$, the child node $x_{v'}^{(\ell)}$ follows a multinomial distribution parameterized by $\softmax_{\row}(\bG_{\square, \iota}^{(\ell)})_{s :}$. In our experiments, we maintain a fixed set of matrices $(\Perm_{\square, \iota}^{(\ell)}, \bG_{\square, \iota}^{(\ell)})$ by using a consistent random seed for generation.

The parameter $\pflip$ determines the conditional entropy of the leaf nodes $\bx_{\square}$ given the root node $x_{\root}$. When $\pflip = 0$, $\bx_{\square}$ is a deterministic function of $x_{\root}$ (given fixed matrices $(\Perm_{\square, \iota}^{(\ell)})_{\square, \iota, \ell}$). Conversely, when $\pflip = 1$, $\bx_{\square}$ given $x_{\root}$ exhibits high conditional entropy. Predicting $x_{\root}$ from $\bx_{\square}$ is relatively straightforward for small values of $\pflip$, but becomes increasingly challenging as $\pflip$ approaches $1$.

In our simulations, we set the depth $L=4$, the states $\state=\{1,\ldots,10\}$, and $\nchild_\image = \nchild_\txt=3$, and vary the transition randomness parameter $\pflip$ from $0.02$ to $0.4$ with increments of $0.02$. Note that in this case $\dimimage=\dimtxt=\dim=81$. At last, we set the number of pairs in a sample to be $K=4$.

\paragraph{Belief propagation.}  \label{sec:simu-detail-bp} Given the transition functions $\{ \psi^{(\ell)}_{\square,\iota} \}$, we can compute the true similarity score 
\[
\textstyle \truesimscore(\xi, \xt) = \log [ \sum_{x_\root \in [\state]} \P(x_\root | \xi) \P(x_\root | \xt) / \P(x_\root) ]
\]
by calculating the conditional probabilities $(\P(x_\root | \xi), \P(x_\root | \xt))$ via belief propagation. This enables us to obtain the global minimum of the CLIP risk $\min_{\simscore} \orisk_{\clip, K}(\simscore)$ as defined in Eq.~(\ref{eqn:CLIP_risk}) (\Cref{subsubsection:CLIP-Intro-BP}). Similarly, belief propagation algorithm can be applied to find the global minima of both the CDM risk $\min_{\Mt, \Et}\risk_{\cdm, t}(\Mt, \Et)$ from Eq.~(\ref{eqn:conditional_denoising_with_CLIP}) (\Cref{subsubsection:CDM-Intro-MP}) and the VLM risk $\min_{\probnext, \Ei} \risk_{\vlm}(\probnext, \Ei)$ from Eq.~(\ref{eqn:VLM_with_CLIP}) (\Cref{subsubsection:NWP-Intro-BP}).

\paragraph{Guided training.} \label{sec:simu-detail-guide} Here we detail the settings for the guided penalty. 

\vskip0.4em
\myunder{CLIP training.}
For CLIP training, belief propagation involves only downsampling. Let $\hmatrix^{(\ell-1)}_{\image}$ and $\hmatrix^{(\ell-1)}_{\txt}\in \real^{D\times d}$ represent the outputs of $\Attn^{(\ell)}_{\image}(\hmatrix^{(\ell)}_{\image})$ and $\Attn^{(\ell)}_{\txt}(\hmatrix^{(\ell)}_{\txt})$ respectively, for $\ell=L,\ldots, 1$. The inputs to these attention layers are $\hmatrix^{(L)}_{\image}$ and $\hmatrix^{(L)}_{\txt}$.  We define $\Hmessage^{(\ell)}_{\square}$ as the messages passed from the parent nodes in $\cV^{(\ell)}_{\square}$ to the child nodes in $\cV^{(\ell-1)}_{\square}$ (for $\square \in \{ \image, \txt\}$): 
\[
\Hmessage^{(\ell)}_{\square} = 
\begin{bmatrix}
    h^{(\ell)}_{\square, \pa^{(L-\ell)}(v)}
\end{bmatrix}_{v =1,\ldots, d} \in \real^{S\times d}.
\] 
The guided penalty $r^{(\ell)}_{\square}$ at each layer is given by:
\[
r^{(\ell)}_{\square} = \|  \hmatrix^{ (\ell)}_{\square,(L-\ell)s : (L+1 -\ell)s, :} - \Hmessage^{(\ell)}_{\square} \|_2^2,
\] 
where different rows $((L-\ell)s : (L+1 -\ell)s)$ in $\hmatrix^{(\ell)}_{\square}$ are used for different layers $\ell$ to align with $\Hmessage^{(\ell)}_{\square}$. The total guided penalty is computed as the weighted sum across all layers: 
\[
\textstyle r = \sigma\sum_{\ell = 0}^{L-1} \big( r^{(\ell)}_{\txt} + r^{(\ell)}_{\image} \big),
\]
where $\sigma$ is a hyperparameter controlling the penalty strength.

\vskip0.4em
\myunder{CDM training.} Note that for the CDMs, we have $2L+1$ layers. The belief propagation process is split into a downsampling phase and an upsampling phase. For the downsampling we have $\hmatrix^{(\ell-1)}_{\square} = \Attn^{(L+1+\ell)}_{\square} (\hmatrix^{(\ell)}_{\square})$ for $\ell = L+1, \ldots, 1$. For the upsampling, we have $\bbmatrix_{\square}^{(\ell)}  = \Attn^{(L+2- \ell)}(\bbmatrix_{\square}^{(\ell-1)}) $  for $\ell = 1, \ldots, L$. Hence, the input is $\hmatrix^{(L+1)}_{\square}$ and the output is $\bbmatrix_{\image}^{(L)}$. 
For down sampling, there are two kinds of messages $\Qmessage^{(\ell)}_{\image}$ and $ \Hmessage^{(\ell)}_{\image} $ i.e.,  
\[
\Hmessage^{(\ell)}_{\image} = 
\begin{bmatrix}
    h^{(\ell)}_{\image, \pa^{(L-\ell)}(v)}
\end{bmatrix}_{v =1,\ldots, d}, \; \Qmessage^{(\ell)}_{\image} = 
\begin{bmatrix}
    q^{(\ell)}_{\image, \pa^{(L-\ell)}(v)}
\end{bmatrix}_{v =1,\ldots, d} \in \real^{S\times d}.
\] 
Then the penalty for the image part in downsampling is defined as
\[
r_{\image, \downarrow} = \sum_{\ell=0}^{L} \big( \|  \hmatrix^{ (\ell)}_{\image,(L-\ell)s : (L+1 -\ell)s, :} - \Hmessage^{(\ell)}_{\image} \|_2^2 + \| \hmatrix^{ (\ell)}_{\image,(2L-\ell)s  : (2L+1 -\ell)s, :} - \Qmessage^{(\ell)}_{\image} \|_2^2 \big). 
\] 
Regarding the text part, there are two scenarios. If we do not use clip features, we follow a procedure similar to clip-guided training. In that case,
\[
\Hmessage^{(\ell)}_{\txt} = 
\begin{bmatrix}
    h^{(\ell+1)}_{\txt, \pa^{(L-\ell)}(v)}
\end{bmatrix}_{v =1,\ldots, d} \in \real^{S\times d},
\] 
and 
\[
    r_{\txt, \downarrow} = \sum_{\ell=1}^{L+1}  \|  \hmatrix^{ (\ell)}_{\txt,(L-\ell)s : (L+1 -\ell)s, :} - \Hmessage^{(\ell)}_{\txt} \|_2^2.
\]
Otherwise, if we do use the clip features, then $\hmatrix^{(\ell)}_{\txt} \in \real^{D\times 1}$ and we only ensure that the information is retained after the 
$L$-th layer:
\[
    r_{\txt, \downarrow} =  \|  \hmatrix^{ (L+1)}_{\txt, :s, :} - \hmatrix^{(1)}_{\txt, :s,:} \|_2^2. 
\]
Finally, there is an upsampling penalty only for the image part. An additional type of message, $\Bmessage^{(\ell)}_{\image}$ i.e.,
\[
\Bmessage^{(\ell)}_{\image} = 
\begin{bmatrix}
    b^{(\ell)}_{\image, \pa^{(L-\ell)}(v)}
\end{bmatrix}_{v =1,\ldots, d}.
\] 
Finally, the total penalty is defined as
\[
r_{\image, \uparrow} =  \sum_{\ell=0}^{L} \big( \|  \bbmatrix^{ (\ell)}_{\image,(L-\ell)s : (L+1 -\ell)s, :} - \Hmessage^{(\ell)}_{\image} \|_2^2 + \| \bbmatrix^{ (\ell)}_{\image,(2L-\ell)s  : (2L+1 -\ell)s, :} - \Qmessage^{(\ell)}_{\image} \|_2^2  + \| \bbmatrix^{ (\ell)}_{\image,(3L-\ell)s  : (3L+1 -\ell)s, :} - \Bmessage^{(\ell)}_{\image} \|_2^2\big). 
\] 
The total penalty is defined as $r = \sigma \big( r_{\image, \downarrow} + r_{\txt, \downarrow} + r_{\image, \uparrow} \big). $

\vskip0.4em
\myunder{VLM training.} VLM also involves downsampling and upsampling. The information structure is almost the same as that of the CDMs, with the primary distinction being the swapping of roles between image and text. We can define $r_{\txt, \downarrow}, r_{\txt, \uparrow}$ and $r_{\image, \downarrow}$ in a similar manner. The total penalty is defined as $r = \sigma \big( r_{\txt, \downarrow} + r_{\image, \downarrow} + r_{\txt, \uparrow} \big)$. 

\paragraph{Learning rates and penalties.}  \label{sec:simu-detail-lr} After doing a grid search for parameters, we choose the following combinations of learning rates and penalties.

\begin{table}[h!]
\centering
\begin{tabular}{|l|l|l|l|l|}
\hline
Task & Model          & max lr & min lr & penalty ($\sigma$) \\ \hline
CLIP & \ATF{}    & 3e-4   & 3e-7   &         \\ \hline
CLIP & \GTF{}      & 1e-3   & 1e-6   & 1e-3    \\ \hline
CLIP & \STF{}     & 3e-4   & 3e-7   &         \\ \hline
CDM  & \ATF{}    & 1e-3   & 1e-6   &         \\ \hline
CDM  & \GTF{}      & 1e-2   & 1e-5   & 1e-1    \\ \hline
CDM  & \STF{}     & 1e-3   & 1e-6   &         \\ \hline
CDM  & \JT{} & 1e-3   & 1e-6   &         \\ \hline
VLM  & \ATF{}    & 1e-3   & 1e-6   &         \\ \hline
VLM  & \STF{}     & 1e-3   & 1e-6   &         \\ \hline
VLM  & \GTF{}      & 1e-3   & 1e-6   & 1e-3    \\ \hline
VLM  & \JT{} & 3e-4   & 3e-7   &         \\ \hline
\end{tabular}
\caption{ Learning rates and penalties for different models.}
\label{tab:lrp}
\end{table}

\paragraph{Adam-W parameters.}  \label{sec:simu-detail-adam} We use the Adam-W optimizer~\cite{loshchilov2017decoupled} for all our models. The parameters \(\beta_1\) and \(\beta_2\) are set to \(0.9\) and \(0.999\), respectively. The weight decay is configured to \(0.01\), and the error term is set to \(10^{-8}\). Additionally, we apply norm clipping with a maximum \(\ell_2\) norm of \(1.0\). Finally, we employ a cosine annealing learning rate scheduler with the number of warm-up steps set to \(0\).

\paragraph{Network architecture.}   \label{sec:simu-detail-network} Now we introduce the details of network architectures.
\vskip0.4em

\myunder{CLIP architecture.} In CLIP training, we parameterize the similarity score function as 
\begin{align}\simscore^\btheta(\xi, \xt) = \langle \NN_{\image}^{\bW_\image}(\xi), \NN_{\txt}^{\bW_\txt}(\xt) \rangle
\end{align} 
using an inner-product link function and neural networks $(\NN_{\image}^{\bW_\image}, \NN_{\txt}^{\bW_\txt})$ as encoders. Each encoder neural network $\NN_{\square}^{\bW_{\square}}(\bx_\square) = \read(\TF(\encode(\bx_\square)))$ is composed of a trainable embedding function $\encode: \R^d \to \R^{D \times d}$, a trainable read-out function $\read: \R^{D \times d} \to \R^\state$, and a $(L+1)$-layer transformer $\TF: \R^{D \times d} \to \R^{D \times d}$ based on the architecture from \cite{vaswani2017attention}, modified with RMSNorm instead of LayerNorm, and a pre-norm instead of post-norm. Note that we choose $L=4$ and the hidden dim $D=128$. 
\vskip0.4em

\myunder{CDM architecture.} In joint CDM training, the conditional denoising function is parameterized as the following: $\Mt(\bz_t, \xt) = \read(\TF(\encode_{\image}(\bz_t), \encode_{\txt}(\xt)))$, where $\encode_{\image}: \R^d \to \R^{D \times d}$ and $\encode_{\txt}: \R^d \to \R^{D \times d}$ are trainable embedding functions, $\read: \R^{D \times 2d} \to \R^d$ is a trainable read-out function, and $\TF: \R^{D \times 2d} \to \R^{D \times 2d}$ is a $(2L+1)$-layer transformer. 

In cases of partial training with a fixed CLIP embedding $\what \Et(\xt)$, the conditional denoising function becomes $\Mt(\bz_t, \what \Et(\xt)) = \read(\TF(\encode_{\image}(\bz_t), \encode_{\txt}(\what \Et(\xt))))$, with a trainable embedding function $\encode_{\image}: \R^d \to \R^{D \times d}$, a fixed embedding function $\encode_{\txt}: \R^{\state} \to \R^{d}$, a trainable read-out function $\read: \R^{D \times (d+1)} \to \R^d$, and a transformer $\TF: \R^{D \times (d+1)} \to \R^{D \times (d+1)}$ consisting of $(2L+1)$ layers. Note that we set the hidden dim $D=256$ here.
\vskip0.4em

\myunder{VLM architecture.}
Similarly, in the joint training of VLMs, the conditional next-token probability is parameterized as $\probnext(x_{\txt, k} = \cdot | \xi) = \softmax(\read(\TF(\encode_{\txt}(x_{\txt, 1:k-1}), \encode_{\image}(\xi))))$, with trainable embedding functions $\encode_{\image}: \R^d \to \R^{D \times d}$ and $\encode_{\txt}: \R^{k-1} \to \R^{D \times (k-1)}$, a trainable read-out function $\read: \R^{D \times (d+k-1)} \to \R^\state$, and a $(2L+1)$-layer transformer $\TF: \R^{D \times (d+k-1)} \to \R^{D \times (d+k-1)}$. Note that we set the hidden dim $D=256$ here.

In cases of partial training with a fixed CLIP embedding $\what \Ei(\xi)$, the conditional next-token probability becomes the following: $\probnext(x_{\txt, k} = \cdot | \what \Ei(\xi)) = \softmax(\read(\TF(\encode_{\txt}(x_{\txt, 1:k-1
}), \encode_{\image}(\what \Ei(\xi)))))$, with a fixed embedding function 
$\encode_{\image}: \R^\state \to \R^{d}$, a trainable embedding function $\encode_{\txt}: \R^{k-1} \to \R^{D \times (k-1)}$, a trainable read-out function $\read: \R^{D \times (d + k - 1)} \to \R^\state$, and a $(2L+1)$-layer transformer $\TF: \R^{D \times (d + k - 1)} \to \R^{D \times (d + k - 1)}$.

\paragraph{ZSC settings.} For the ZSC, we choose the number of samples to be $M=250$ in \Cref{fig:zsc-ood,fig:zsc-train}. 

\paragraph{Computational resource.} All our experiments are performed on 8 Nvidia Tesla A100 GPUs (80GB memory) and 12 Nvidia Tesla V100 GPUs (16GB memory). The total GPU time is approximately 3000 GPU hours.

\subsection{Ablation studies}\label{sec:ablations}

{
\subsubsection{Further discussion of Assumption~\ref{ass:bounded_score}}\label{sec:ablation_bounded}

Many of our theoretical results depend on the boundedness conditions in \Cref{ass:bounded_score}, which assume both the  score function $\simscore(\xi,\xt)$ and the logarithm of the probability ratio $\log\frac{\P(\xi,\xt)}{\P(\xi)\P(\xt)}$ are bounded between $[-\log\boundscoreconst,\log\boundscoreconst]$
for some constant $\boundscoreconst>0$.
In practice~\cite{chen2020simple,radford2021learning}, the score function $\simscore(\xi,\xt)$ is chosen as  \begin{align}\simscore(\xi,\xt)=\langle \Ei(\xi), \Et(\xt)\rangle/\tau
    \end{align} for some temperature parameter $\tau>0$ and normalized representations $\Ei(\cdot)$ and $\Et(\cdot)$ such that $\|\Ei(\xi)\|_2\\=1$ and $\|\Et(\xt)\|_2=1$ for all $\xi\in\cXi,\xt\in\cXt$. In this case, the absolute score function $|\simscore(\xi,\xt)|$ is bounded by $1/\tau$. }

\begin{table}[h]
\centering
{
\begin{tabular}{lcc}
\toprule
\textbf{Setting} & \textbf{Val-Top1 (\%)} & \textbf{Val-Top5 (\%)} \\
\midrule
Trainable $\tau$ & $34.81 \pm 0.04$ & $17.61 \pm 0.07$ \\
Fix $\tau=0.07$   & $35.12 \pm 0.03$ & $18.21 \pm 0.11$ \\
Fix $\tau=0.1$    & $31.95 \pm 0.17$ & $15.13 \pm 0.15$ \\
\bottomrule
\end{tabular}
}
\caption{%
{
{ Top-1 and top-5 validation accuracies (\%) across different choices of $\tau$. The
ResNet-50 model was pretrained on the CC3M dataset (3M samples) for 32 epochs under various $\tau$ configurations. 
For the trainable $\tau$ case, $\tau$ was initialized at $0.07$ and constrained to be no smaller than $0.01$ during training, following the design of \cite{radford2021learning}. 
Each setting was evaluated over 3 random seeds, and results are reported as mean $\pm$ standard deviation.%
}}
}
\label{tab:zsc-exp}
\end{table}

{
To further examine the boundedness assumption of the score function $|\simscore(\xi,\xt)|$, we conducted a controlled experiment by pretraining ResNet-50 on the CC3M dataset (3M samples) \cite{sharma2018conceptual_captions}  with different $\tau$ configurations. In addition to the standard design from \cite{radford2021learning}, where $\tau$ is trainable (initialized at $0.07$ and clipped to be at least $0.01$), we trained models with fixed $\tau$ values of $0.07$ and $0.1$, which impose progressively tighter bounds on the score function. We then evaluated these models on the ImageNet-1k validation set \cite{deng2009imagenet} in a zero-shot classification setup.

As shown in Table~\ref{tab:zsc-exp}, fixing $\tau=0.07$ yields comparable performance to the standard trainable setting, while increasing $\tau$ to $0.1$ results in only a modest decrease in accuracy. These findings empirically support our assumption that the score function can be chosen to be bounded.

Nevertheless, the boundedness assumptions on the  probability ratio $\frac{\P(\xi,\xt)}{\P(\xi)\P(\xt)}$ may be relatively strong for certain real-world multimodal data, and it is challenging to estimate the supremum of the probability ratio in real-world multimodal distributions (e.g., image and text) as we only have limited samples from it.

 Technically, Assumption~\ref{ass:bounded_score} is mainly used for change-of-measure arguments in the proof of Proposition~\ref{prop:approximate_global_min_contrastive_loss}~and~\ref{prop:representation_equivalence}. 
 For example, $\boundscoreconst$ in Assumption~\ref{ass:bounded_score} provides an upper bound on the ratio 
 $\frac{\E_{(\xi,\xt)\sim\P_{\image,\txt}}[f(\xi,\xt)] }{\E_{(\xi,\xt)\sim\P_{\image}\times\P_{\txt}}[f(\xi,\xt)]}$ 
 for all functions $f\geq0$. However, for the restricted class of functions considered in the proof, 
 a smaller upper bound may be possible. We leave the further relaxation of Assumption~\ref{ass:bounded_score} to future work.

}

\subsubsection{More samples per category improves zero-shot performance}\label{sec:ablation_vary_m}
\Cref{fig:zsc-m} illustrates the risks associated with zero-shot learning as a function of the number of samples. The experimental setup for \Cref{fig:zsc-m} is nearly identical to that of \Cref{fig:zsc-train}, with the exception that we fix $\pflip=0.2$ and vary $M$. Here, $M$ ranges from $5$ to $250$.
We observe that a larger number of samples leads to more accurate predictions.

{

\Cref{fig:zsc-imagenet} evaluates the zero-shot classification performance of a ResNet-50 model pretrained on the CC12M dataset \cite{changpinyo2021cc12m}, tested on the ImageNet-1k validation set \cite{deng2009imagenet}. We adopt the standard 80 text templates introduced in OpenAI’s CLIP paper \cite{radford2021learning}. For each trial, we randomly permute the 80 templates and incrementally increase the number of templates following this order. This process is repeated 16 times, and we report the mean and standard deviation of cross-entropy loss, top-1 accuracy, and top-5 accuracy.

To validate our theoretical prediction that these performance bounds follow a rate of $C + O(1/M)$ (\Cref{prop:validity_zero_shot_classification}), where $C$ is a constant and $M$ is the number of templates, we also fit the curves in both synthetic dataset and the real dataset with functions of the form $f(M) = A + B/M$ (with $A$ and $B$ as parameters). 
The results in Figure~\ref{fig:zsc-fit} and Figure~\ref{fig:zsc-imagenet} show that the fitted curves (dashed) align closely with the empirical results (solid), with $R^2 > 0.98$, thereby supporting our theoretical findings.
}
\if\showfigure1
\begin{figure}[h!]
\centering
\includegraphics[width=0.5\linewidth]{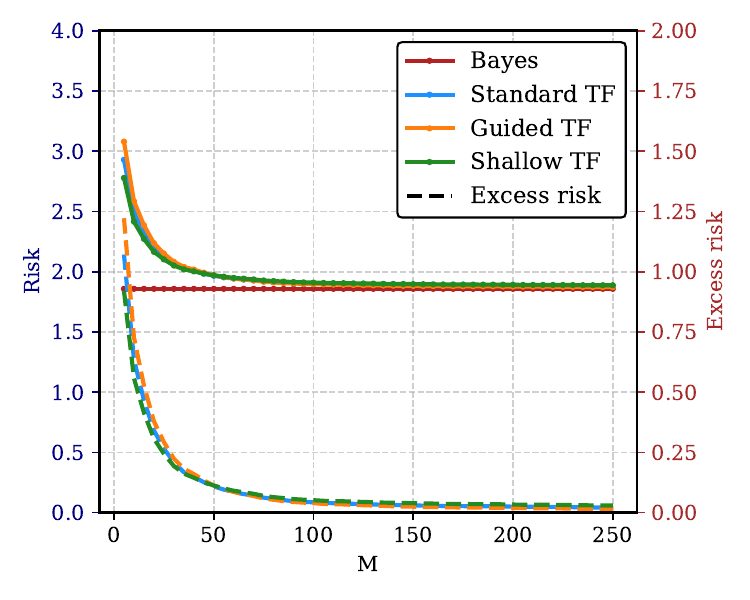}
\caption{
 Risks of zero-shot learning versus number of samples. The setup of \Cref{fig:zsc-m} is almost  the same as that of \Cref{fig:zsc-train}, except that we fix  $\pflip=0.2$ and vary $M$, where $M$ ranges from $5$ to $250$.
We can observe that larger numbers of samples lead to improved predictions.
}\label{fig:zsc-m}
\end{figure}
\fi

\if\showfigure1
\begin{figure}[h!]
\centering
\begin{minipage}{0.32\textwidth}
\includegraphics[width=\linewidth]{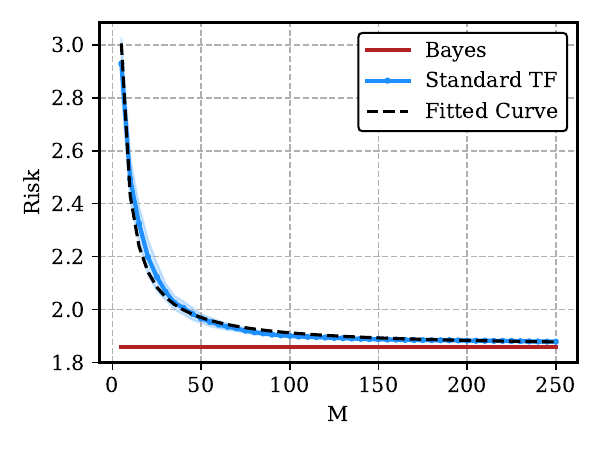}
\subcaption{ Standard TF}
\label{fig:zsc-ce}
\end{minipage}
\begin{minipage}{0.32\textwidth}
\includegraphics[width=\linewidth]{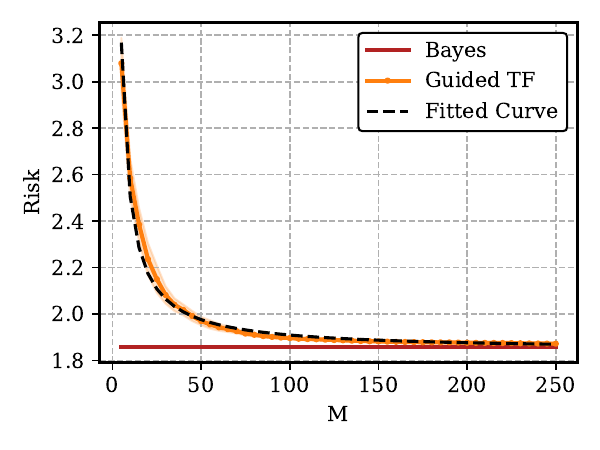}
\subcaption{ Guided TF}
\label{fig:zsc-acc1}
\end{minipage}
\begin{minipage}{0.32\textwidth}
\includegraphics[width=\linewidth]{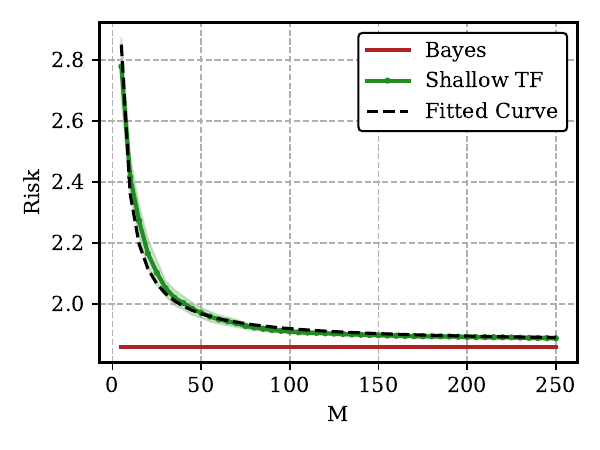}
\subcaption{ Shallow TF}
\label{fig:zsc-acc2}
\end{minipage}

\caption{{{Fitted curves for the risks in zero-shot learning. Each risk curve is fitted by the function $f(M) = A + B/M$, with parameters $A$ and $B$. The fitted results are shown as dark dashed lines, while the empirical risks are plotted as solid lines. All fitted curves achieve an $R^2$ value above 0.98.}
}}\label{fig:zsc-fit}
\end{figure}
\fi

\if\showfigure1
\begin{figure}[h!]
\centering
\begin{minipage}{0.32\textwidth}
\includegraphics[width=\linewidth]{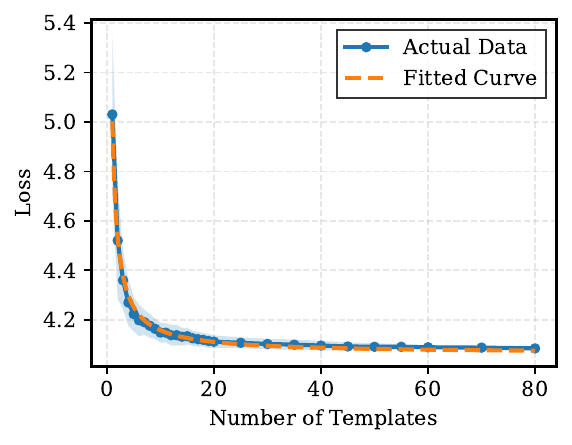}
\subcaption{ Cross Entropy Loss}
\label{fig:zsc-ce_image}
\end{minipage}
\begin{minipage}{0.32\textwidth}
\includegraphics[width=\linewidth]{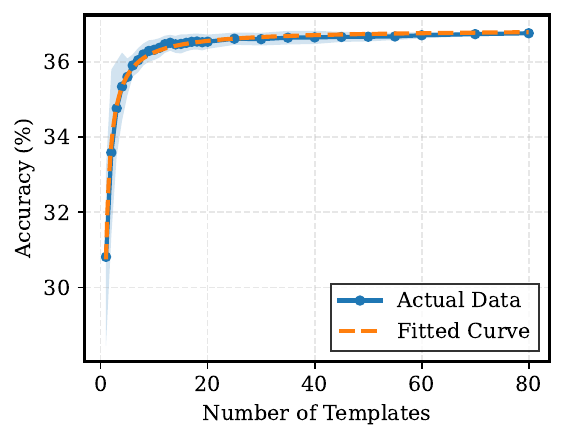}
\subcaption{ Top-1 Accuracy}
\label{fig:zsc-acc1_image}
\end{minipage}
\begin{minipage}{0.32\textwidth}
\includegraphics[width=\linewidth]{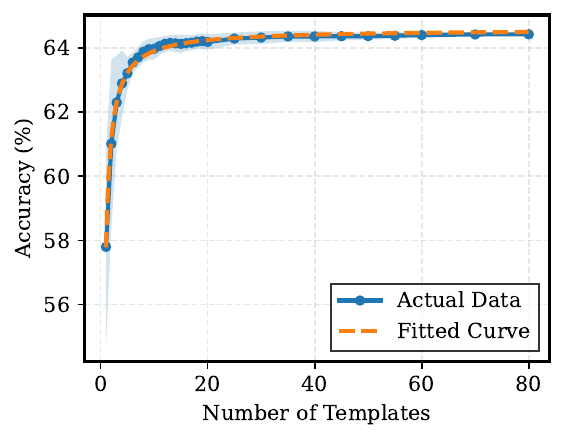}
\subcaption{ Top-5 Accuracy}
\label{fig:zsc-acc2_image}
\end{minipage}

\caption{{{  Zero-shot classification performance of a ResNet-50 model pretrained on CC12M, evaluated on the ImageNet-1k validation set.  We use the standard 80 text templates from OpenAI’s CLIP paper \cite{radford2021learning}. For each run, we sample a random permutation of these 80 templates and progressively increase the number of templates in that order. This procedure is repeated 16 times, and we report the mean and standard deviation of cross-entropy loss, top-1 accuracy, and top-5 accuracy. We also fit the results with functions of the form $f(M) = A + B/M$, where $A$ and $B$ are parameters. Fitted curves are shown as orange dashed lines, while empirical results are shown as blue solid lines. All fitted curves achieve an $R^2$ value above 0.98. }
}}\label{fig:zsc-imagenet}
\end{figure}
\fi

\if\showfigure1
\begin{figure}[t]
\centering
\begin{minipage}{0.38\textwidth}
\includegraphics[width=\linewidth]{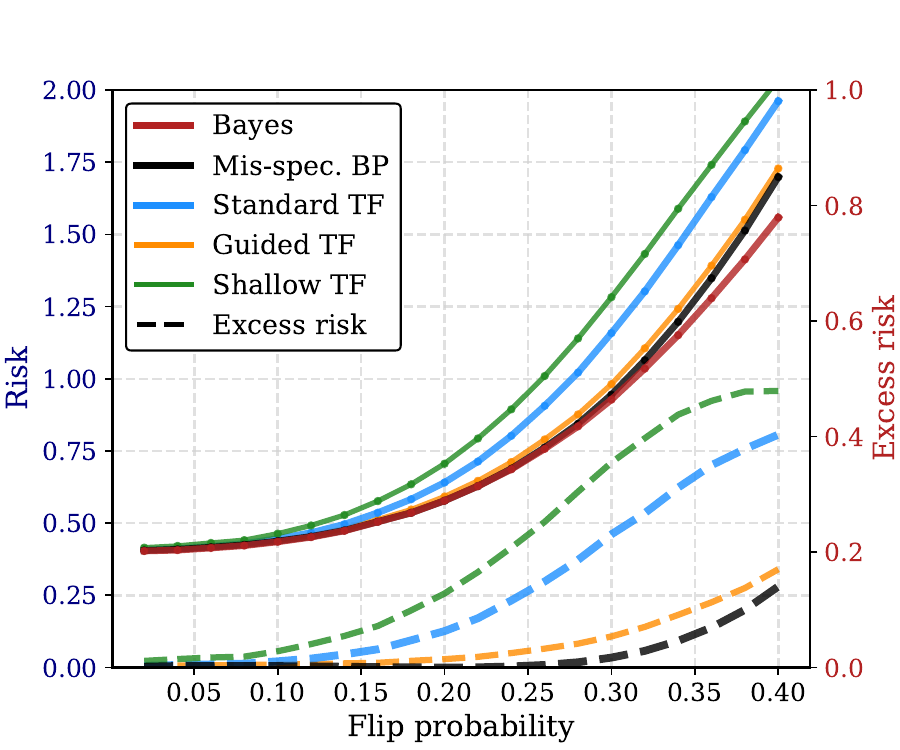}
\subcaption{ CLIP OOD risk}
\label{fig:clip-ood}
\end{minipage}
\begin{minipage}{0.38\textwidth}
\includegraphics[width=\linewidth]{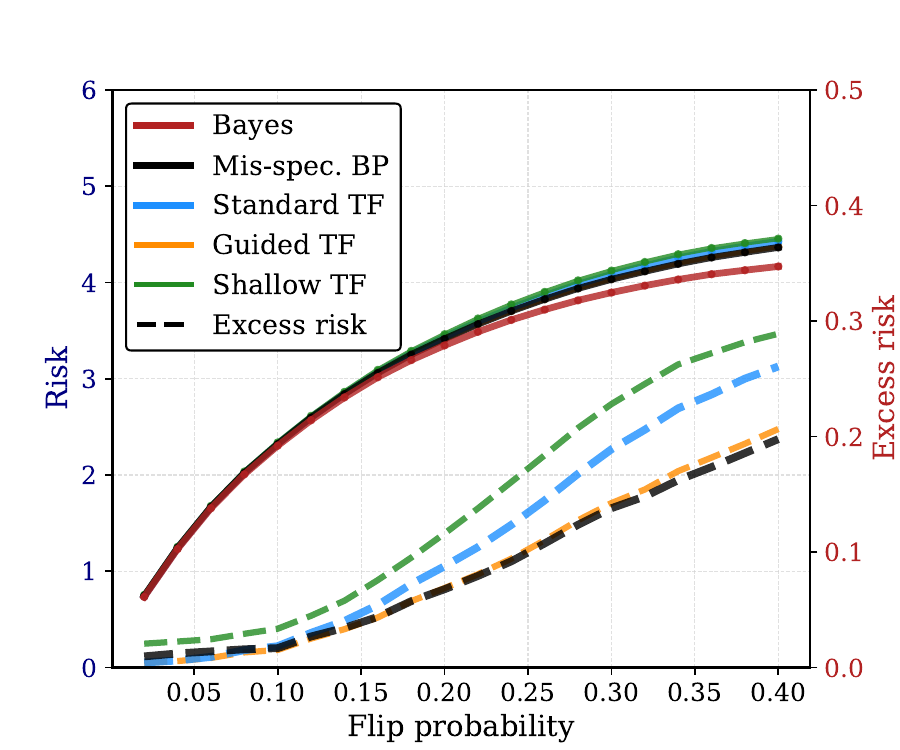}
\subcaption{ ZSC OOD risk}
\label{fig:zsc-ood}
\end{minipage}

\begin{minipage}{0.38\textwidth}
\includegraphics[width=\linewidth]{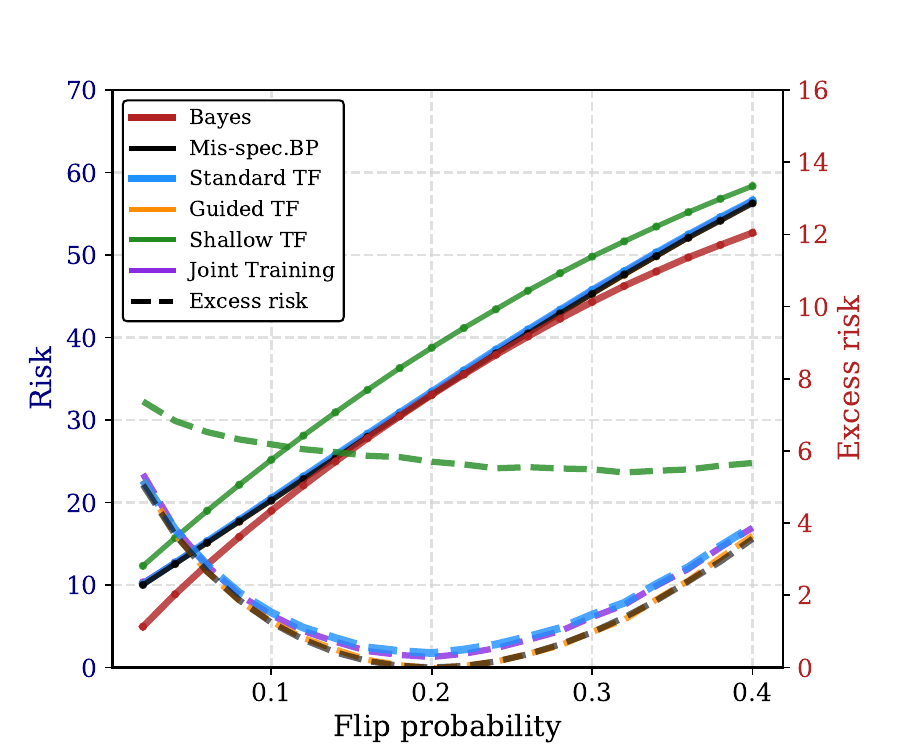}
\subcaption{ CDM OOD risk}
\label{fig:cdm-ood}
\end{minipage}
\begin{minipage}{0.38\textwidth}
\includegraphics[width=\linewidth]{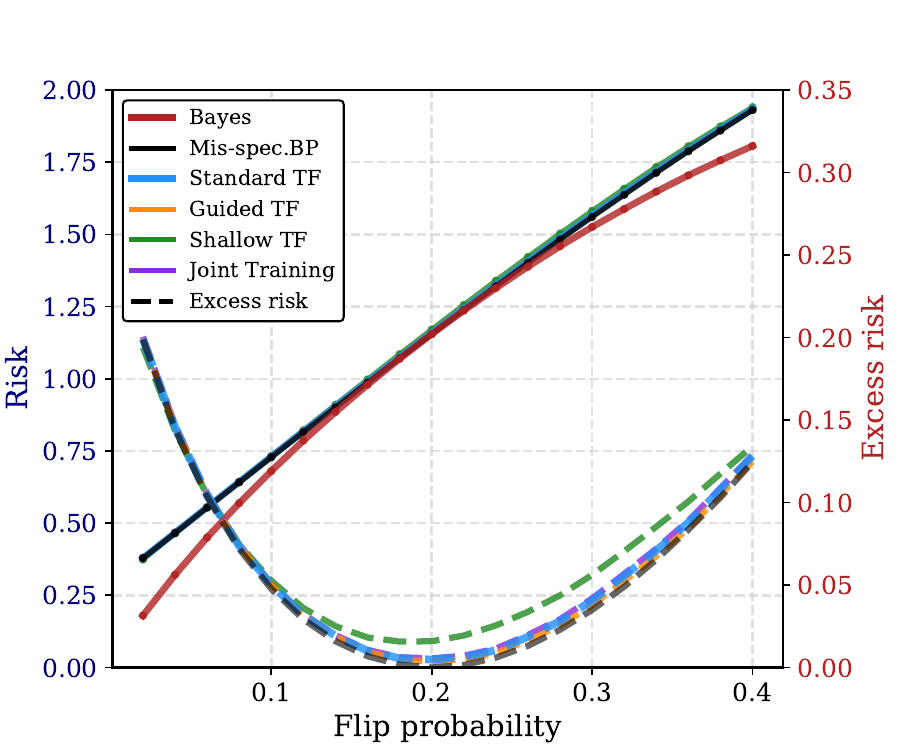}
\subcaption{ VLM OOD risk}
\label{fig:vlm-ood}
\end{minipage}
\caption{{ Out-of-distribution (OOD) risks (solid curves) and excess risks (dashed curves) as a function of the parameter $\pflip$ for CLIP training, ZSC, CDM, and VLM. Models are trained with a fixed $\pflip=0.2$. Across all setups, \GTF{} closely matches the performance of \MBP{}. In the CDM (\ref{fig:cdm-ood}) and VLM (\ref{fig:vlm-ood}) setups, \ATF{} performs similarly to \GTF{}, whereas in the CLIP training (\ref{fig:clip-ood}) and ZSC (\ref{fig:zsc-ood}) setups, \ATF{} shows a greater gap from \GTF{}. This suggests that standard-trained transformers may perform closer to the belief-propagation algorithm when the in-distribution risk is smaller. }}\label{fig:OOD}
\end{figure}
\fi

\subsubsection{Out-of-distribution test} \Cref{fig:OOD} shows the out-of-distribution (OOD) risk (solid curve) and excess risk (dashed curve) as functions of the parameter $\pflip$, for CLIP training (\Cref{fig:clip-ood}), ZSC (\Cref{fig:zsc-ood}), CDM (\Cref{fig:cdm-ood}), and VLM (\Cref{fig:vlm-ood}). In all these experiments, models are trained with a fixed $\pflip=0.2$, and their risks are evaluated under varying $\pflip$ values that are out-of-distribution. The following observations can be made:
\begin{itemize}[topsep=3pt]
\setlength\itemsep{3pt}
\item As expected, across all settings, guided training (\GTF{}) closely matches the performance of the misspecified BP algorithm (\MBP{}), while shallow transformers (\STF{}) perform much worse compared to \MBP{}.
\item In the CDM (\Cref{fig:cdm-ood}) and VLM (\Cref{fig:vlm-ood}) setups, \ATF{} performs similarly to \GTF{}, whereas in the CLIP training (\Cref{fig:clip-ood}) and ZSC (\Cref{fig:zsc-ood}) setups, \ATF{} shows a greater gap from \GTF{}. This suggests that standard-trained transformers may perform closer to the belief-propagation algorithm when the in-distribution risk is smaller.
\end{itemize}

\subsubsection{OOD tests with different $\pflip$ in image and text trees}

\Cref{fig:ood2} shows OOD risks and excessive risks of transformer architectures and belief propagation for VLMs and
CDMs.
The settings of \Cref{fig:cdm-ood2,fig:vlm-ood2} are nearly identical to those of \Cref{fig:cdm-ood,fig:vlm-ood}. The only difference is that we fix the text \(\pflip = 0.2\) while varying the image \(\pflip\) in \Cref{fig:cdm-ood2}, and conversely, we fix the image \(\pflip = 0.2\) while varying the text \(\pflip\) in \Cref{fig:vlm-ood2}. We observe that the trends are similar to those in \Cref{fig:cdm-ood,fig:vlm-ood}.

\if\showfigure1
\begin{figure}[h!]
\centering
\begin{minipage}{0.47\textwidth}
\includegraphics[width=\linewidth]{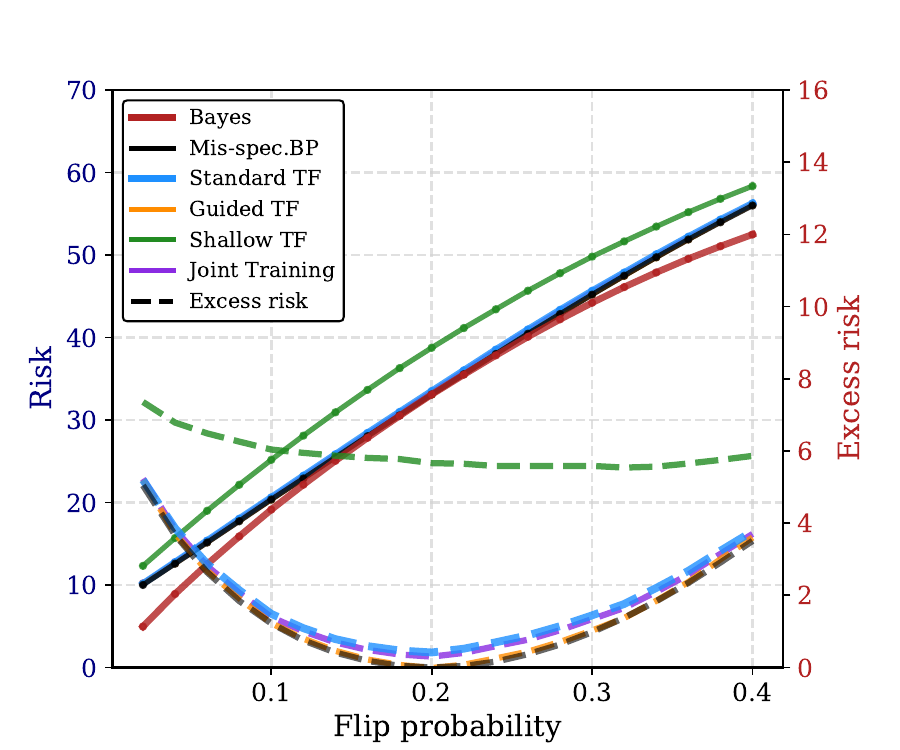}
\subcaption{{ CDM OOD risk w/ fixed text $\pflip$}}
\label{fig:cdm-ood2}
\end{minipage}
\hspace{-.2em}
\begin{minipage}{0.47\textwidth}
\includegraphics[width=\linewidth]{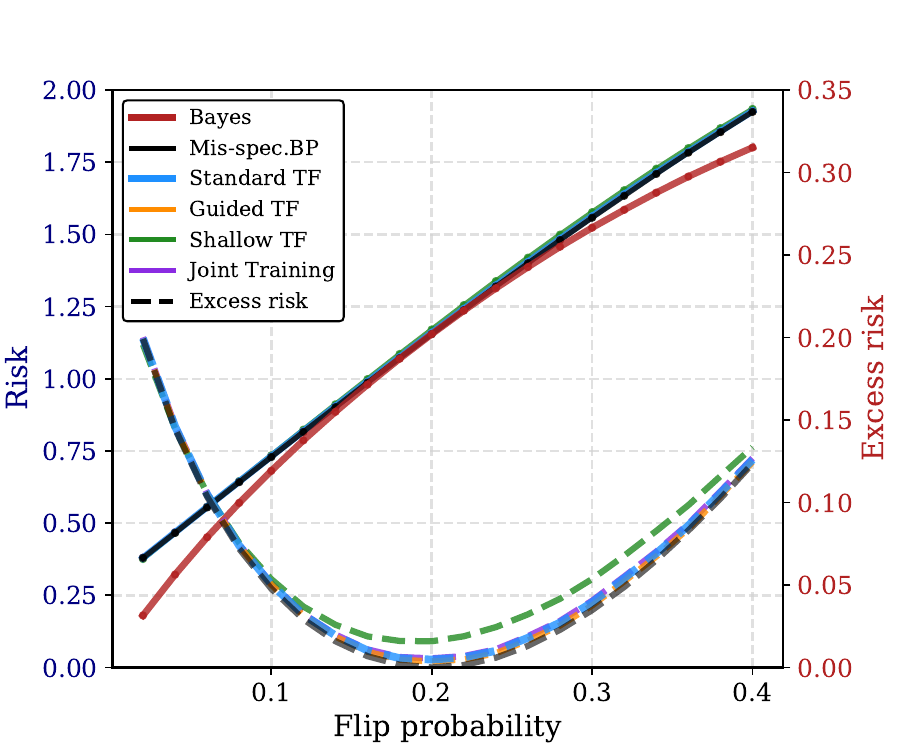}
\subcaption{ VLM OOD risk w/ fixed image $\pflip$}
\label{fig:vlm-ood2}
\end{minipage}
\caption{  OOD risks and excessive risks of various transformer architectures and belief propagation for VLMs and CDMs. These figures exhibit same trends as \Cref{fig:cdm-ood,fig:vlm-ood}. (a) fix the text $\pflip=0.2$ but vary the image $\pflip$. (b) fix the image $\pflip=0.2$ but vary the text $\pflip$. }\label{fig:ood2}
\end{figure}
\fi

\clearpage

\fi

\end{document}